\documentstyle[times,ACL]{illcdiss}
\newcommand{\FSM}{\mbox{$(Q, \Sigma, S, T, F)$}}
\newcommand{\SFSM}{\mbox{$(Q, \Sigma, S, T, F, P)$}}
\newcommand{\LA}{\langle}
\newcommand{\RA}{\rangle}
\newcommand{\tuple}[1]{\LA #1\RA}

\newcommand{\hspf}{} 
\newcommand{\hspft}{} 
\newcommand{\hspfth}{} 
\newcommand{\sumMPD}{{\bf Max}} 
\newcommand{\SizeOf}[1]{$|#1|$}
\newcommand{\TSGsize}{\SizeOf{{\cal C}}}
\newcommand{\lraN}{\rightarrow}
\newcommand{\lra}{$\rightarrow$}
\newcommand{\ITEM}[3]{$#1$\lra$#2\bullet #3$}
\newcommand{\ITEMN}[3]{#1\rightarrow #2\bullet #3}
\newcommand{\ITEMNM}[3]{\hspft #1\rightarrow #2\bullet #3}
\newcommand{\STACK}[2]{{#1}_{#2}^{}}

\newcommand{\defEqual}{\stackrel{def}{=}}
\newcommand{\DEFINE}[2]{\begin{description}\item[#1] {\it #2}\end{description}}
\newcommand{\TB}{\cal TB}

\newcommand{\CorpM}{{\cal C}}
\newcommand{\Corp}{${\cal C}$}
\newcommand{\VN}{$V_{N}$\/}
\newcommand{\VT}{$V_{T}$\/}
\newcommand{\TSG}{\mbox{(\VN,~\VT,~$S$,~\Corp)}} 
\newcommand{\STSG}{\mbox{(\VN, \VT, $S$, ${\cal C}$, $PT$)}} 
\newcommand{\Rules}{${\cal R}$\/}
\newcommand{\RulesN}{{\cal R}}

\newcommand{\CFG}{\mbox{(\VN,~\VT,~$S$,~\Rules)}}
\newcommand{\SCFG}{\mbox{(\VN,~\VT,~$S$,~\Rules,~P)}} 
 
\newcommand{\SETM}[2]{\{#1~|~#2\}}
\newcommand{\SET}[2]{\mbox{\{#1~$|$~#2\}}\/}
\newcommand{\FUN}[3]{#1:~#2~\lra~#3\/}

\newcommand{\RuleM}[2]{{#1 \lraN #2}}
\newcommand{\Rule}[2]{\mbox{#1~\lra~#2}\/}

\newcommand{\lmd}{\longrightarrow}
\newcommand{\lmdir}{$\stackrel{*}{{\rightarrow}}$}
\newcommand{\lmpdir}{$\stackrel{+}{{\rightarrow}}$}

\newcommand{\bIf}{{\bf if}~}

\newcommand{\bThen}{{\bf then}~}
\newcommand{\bElse}{{\bf else}~}

\newcommand{\bOf}{{\bf of}~}
\newcommand{\bCase}{{\bf case}~}

\newcommand{\AND}{{\footnotesize\bf and}}
\newcommand{\OR}{{\footnotesize\bf or}}
\newcommand{\bOtherwise}{{\bf otherwise}~}
\newcommand{\mul}{\times}
\newcommand{\multi}{\times}

\newcommand{\Example}[2]{\begin{#1} #2 \end{#1}} 
\newcommand{\EExample}[3]{\begin{#1} #2 \label{#3}\end{#1}} 
\newcommand{\entry}[2]{\mbox{$[#1, #2]$}}
\newcommand{\entryN}[2]{\mbox{[#1, #2]}}
\newcommand{\MEASM}{{\cal M}}
\newcommand{\MEASN}{$\cal M$}
\newcommand{\MEAS}{\MEASN~}
\newcommand{\Front}{$\underline{F}$}

\newcommand{\A}{$\cal A$}
\newcommand{\An}{{\cal A}}

\newcommand{\RBSTSG}{\mbox{(\VN, \VT, $S$, \Rules, \A, $Viable?$, $P$, $PF$)}\/} 
\newcommand{\COMSIZE}{}
\newcommand{\comment}[1]{{\COMSIZE\tt\bf /*~#1~*/}}

\newcommand{\bREPEAT}{{\bf Repeat}}
\newcommand{\bUNTIL}{{\bf until}}
\newcommand{\shftl}{\hspace*{-.4cm} }
\newcommand{\REALS}{${\cal R}$}
\newcommand{\TABL}[4]{
              \begin{tabular}{ll}
              #1 & #2 \\
              #3 & #4 \\
              \end{array}}

\newcommand{\TABS}[4]{
              \begin{tabular}{l}
              \mbox{#1} \\
              \mbox{#2} \\
              \mbox{#3} \\
              \mbox{#4} \\
              \end{tabular}}

\begin{document}
\pagestyle{plain}
\pagenumbering{roman}


{\pagestyle{empty}

%
%
\setcounter{page}{1}

\par\vskip 2.5cm
\begin{center}
\pagestyle{empty}
\leftline{\huge\sc Learning Efficient Disambiguation}
\par\vspace {1.7cm}
\leftline{\Large\sc Khalil Sima'an}
\end{center}
\vfill
%
\par\vspace {1cm}
%
\clearpage
\par\vskip 2cm
\begin{center}
ILLC Dissertation Series 1999-02  			
\par\vspace {4cm}
\illclogo{4cm}
\par\vspace {4cm}
\noindent%
For further information about ILLC-publications, please contact\\[2ex]
Institute for Logic, Language and Computation\\
Universiteit van Amsterdam\\
Plantage Muidergracht 24\\
1018 TV Amsterdam\\
phone: +31-20-5256090\\
fax: +31-20-5255101\\
e-mail: illc@wins.uva.nl
\end{center}
\clearpage
%
\begin{center}
\mbox{}
\vskip 2cm
{\LARGE\sc
Learning Efficient Disambiguation\\[24pt]
}
\vskip 2cm
{\Large\sc Leren Effici\"{e}nt te Desambigueren} \\
(met een samenvatting in het Nederlands)

\vfill
{\Large PROEFSCHRIFT}\\[24pt]
ter verkrijging van de graad van doctor \\
aan de Universiteit Utrecht \\
op gezag van de Rector Magnificus, prof.\ dr.\ H.O.\ Voorma \\
ingevolge het besluit van het College voor Promoties\\
in het openbaar te verdedigen\\
op woensdag 31 maart 1999\\
des ochtends te 10.30 uur. \\[24pt]
door \\[24pt]
{\Large\sc Khalil Sima'an} \\[24pt]
geboren op 12 september 1964 te Haifa
\vskip 2.5cm
\end{center}

\clearpage

\noindent
\begin{tabbing}
Promotoren:\= ~Prof. ir. S. P. J. Landsbergen, Universiteit Utrecht\\
           \> ~Prof. dr. ir. R. J. H. Scha, Universiteit van Amsterdam
\end{tabbing}
%
\vfill
\noindent
This work was partially funded by the Netherlands Organization for Scientific
Research (NWO) and the Foundation for Language Technology (STT, Utrecht 
University). It was facilitated by support from the Institute for Logic,
Language and Computation (ILLC) and from the Utrecht Institute of 
Linguistics (Uil-OTS). 
\vfill
\noindent
Learning efficient disambiguation / Khalil Sima'an. \\
Thesis, Utrecht University - With summary in Dutch\\
ISBN 90-73446-88-0\\
Subject headings: natural language processing/machine learning/probabilistic parsing.\\
\vskip 0.2cm
\noindent
Cover design by Yael Seggev\\
\noindent
\copyright\ ~by Khalil Sima'an.
\vskip 1cm
%
%

\clearpage
} 


\include{./D-ILLC/guide_dedication}
\tableofcontents
\acknowledgments
\vspace*{-1.5cm}
{\large\bf L}ate in an evening in November~1993, I received a bizarre phone-call concerning
a research position in a project on parsing natural language. 
I was told that the project is about resolving ambiguity, that it is for two years only 
(stressing that a PhD is not the goal) and that it pays better than being a PhD-student 
(an ``immoral" approach :-)). It sounded like adventure because I had already met Remko Scha 
a couple of times the year before, when I was writing my Master's thesis on ambiguity. 
During one of these times I asked Remko ``how do you people in natural language processing  
get rid of ambiguity from a natural language grammar", Remko answered tersely  
``we are not interested in making natural language grammars unambiguous". 
As a computer scientist I was puzzled; I felt that Computer Science is 
a ``safer" place to be than those ``ambiguous linguistic environments". In an interview
for the job I also met Rens Bod and Steven Krauwer, who was the intended project leader.
The week before the interview I had read the papers on DOP.
Because I was told that there were no polynomial-time parsing algorithms for DOP, I sat 
down and designed such an algorithm. During the interview I explained some of 
the details of the algorithm, Remko and Steven were interested in seeing this written 
down first, Rens was surprised and did not believe it was possible. Despite of that, I was 
hired to develop a parser for DOP in a two year project called CLASK.
Meanwhile, Remko and his group were involved in a national project (``OVIS") of the 
Netherlands organization for Scientific Research (NWO). The results of CLASK constituted my
``visa" for joining ``OVIS" for one year. After that, Remko and I decided that it is time to 
concentrate on writing a thesis; NWO and the Foundation for Language and Speech (STT) decided 
to support this proposal. 

This thesis exists thanks to various project proposals submitted together with Remko Scha. 
Without the support of Remko Scha (ILLC), Steven Krauwer (STT), Jan Landsbergen (OTS) and 
Alice Dijkstra (NWO), this thesis would have remained virtual. 
Our proposals would not have become projects without additional support from Loe Boves,
Martin Everaart, Gertjan van Noord, Eric Reuland, and the STT-board.

I am grateful to my promoters for the involvement and the supervision. They listened, discussed,
read, commented and corrected always with so much patience. I am especially indebted to 
Christer Samuelsson and Remko Bonnema who read and commented on earlier versions 
of all chapters; in particular, Christer detected and suggested corrections to a serious 
error in the original paper that led to chapter~\ref{CHComplexity}. I thank also Ameen Abu-Hanna,
Yaser Yacoob and Yoad Winter for reading and commenting on earlier versions.
Apart from the aforementioned people, this thesis benefited from discussions with Erik Aarts,
Rens Bod, Boris Cormons, Walter Daelemans, Antal van Den Bosch, Aravind K.\ Joshi, 
Ron Kaplan, Mark-Jan Nederhof, Renee Pohlmann, B.\ Srinivas, Jorn Veenstra and Jacoob Zavrel. 
And I thank the dissertation-committee members for their effort: Walter Daelemans, Jan van Eijck,
Michael Moortgat, Anton Nijholt and Christer Samuelsson. 

I am grateful to Remko Bonnema for allowing me to use the software tools that he
developed beside and around my parser. I thank both the Priority Programme of the Netherlands 
organization for Scientific Research (NWO) and the Alfa Informatica at the University of 
Amsterdam for providing the word-graphs, the OVIS tree-bank and the hardware for conducting
the experiments. I thank SRI-Cambridge (UK), especially David Carter, Steve Pullman 
and Manny Rayner, for supplying the ATIS tree-bank for the experiments. 
I thank the support-team of STT and OTS: Brigitte Burger, Leslie Dijkstra, 
Sibylla Nijhof, Annette Nijstad and Margriet Paalvast.
And I am grateful to Yael Seggev for designing the cover of this thesis.

Although this thesis was due about half a year ago, its existence now might still come 
somewhat as a surprise. Shortly after I officially started preparing for writing it, about 
eighteen months ago, I was hit by health troubles.  
During these hard times, when it seemed that there was only one possible outcome for 
any dice I would cast$\ldots$ 
I was surrounded by so many caring, supportive and loving people. 
Among these people I would like to name here my friends \'{A}adel, Ameen, 
Jan, Jelena, Louis, Neeltje, Patricia, Saeed, Sjoerd, Sophie, Wessel, Yael, Yaser 
and Yoad. I re-mention Ameen (intentionally~!) who supported me in all ways from the 
first moment I arrived in Amsterdam about ten years ago (after ``ruining" my soul
-~in a joint complicity together with Yaser~- on the beach of Haifa for so many years). 
I also mention the dear families Rhebergen and van-Nee for so much care and support:
Marian, Peter, Marije, Snoopy~(Miaao), 2$\times$Didi, Tineke, Theo, Sitske, 
${\cal TT}$\ and Mw.\ van Nee-van Lonkhu\"{y}zen. Marian and Peter are acknowledged despite
of reminding me so often that I am merely a ``gastarbeider" and asking me to empty my 
pockets from stones every time I enter their house (this is called ``Dutch 
hospitality" ;-)).\ I thank also Hanna and Sumayya Abu-Hanna for the long 
lasting friendship and support.

If I was able to write this thesis, it is due to the time that my parents,
Mariam and Butros, spent on educating me when I was a child. It was a hostile environment
around them, an environment that denied from them their youth, beloved ones, home, roots
and belongings. Still they were able to put me on the track that lead to
this thesis$\ldots$ this thesis is actually theirs. My brothers Camil and Nabil, and my sister 
Camilia have always been so loving, supportive and caring. They are the dearest.
I wish we could be more often together. Camil is acknowledged again for ``revenging" 
from Peter on behalf of myself.

Finally, the amoora Didi. She evokes the waves that keep the waters that surround
me so fresh. During the cold and dark times I found the warmest shelter within her smiles 
and tears. She showed me far places where only some travelers go, and as it seems now, 
she intends to show me new places where especially anthropologists would want to go$\ldots$
{\sc Adios~!}


\cleardoublepage
\pagestyle{headings}
\pagenumbering{arabic}

\newcommand{\SETWIDTH}{\setlength\textwidth{13cm}}
\newcommand{\RESETWIDTH}{\setlength\textwidth{15cm}}
\newtheorem{FirstExamp}{Example}[chapter]
\newtheorem{SecondExamp}[FirstExamp]{Example}
\chapter{Introduction}
\label{CHIntro}
{
\section{A brief summary}
Many contemporary performance models of natural language parsing resolve ambiguity
by acquiring probabilistic grammars from tree-banks that 
represent language use in limited domains. Among these performance models,
the Data Oriented Parsing (DOP) model represents a memory-based approach to language modeling.
The DOP model casts a {\em whole tree-bank}, which is assumed to represent the language 
experience of an adult in some domain, into a probabilistic grammar called Stochastic 
Tree-Substitution Grammar (STSG). 

A remarkable fact about contemporary performance models is their entrenched 
inefficiency. Despite of their ability to learn from tree-banks, these models do not 
account for two appealing properties of human language processing: 
firstly, {\em that more frequent utterances are processed more efficiently}, and secondly, 
{\em that utterances in specific contexts, typical for limited domains of language use,
are usually less ambiguous than they are in general contexts}.
This thesis defends the proposition that the absence of mechanisms that represent these 
and similar properties is a major source for the inefficiency of performance models.
Besides this source of inefficiency of performance models in general, the 
DOP model in particular suffers from other inveterate sources of inefficiency: the huge STSGs 
that it acquires and the complexity of disambiguation by means of STSGs.

This thesis studies solutions to the inefficiency of performance models in general and 
the DOP model in particular. The principal idea for removing these sources of inefficiency
is to incorporate ``efficiency properties" of human behavior in limited domains of language 
use, such as the properties stated above, into existing performance models. Efficiency 
properties can be observed through {\em the statistical biases of the linguistic phenomena}\/
that are found in tree-banks that represent limited domains of human language use.
These properties can be incorporated into a performance
model through the combination of two methods of learning from a domain-specific tree-bank:
an {\em off-line method}\/ that constrains the recognition-power and the ambiguity of 
the linguistic annotation of the tree-bank such that it {\em specializes}\/ it for the domain, 
and an {\em on-line performance model}\/ that acquires less ambiguous and more efficient 
probabilistic grammars from that less redundant tree-bank.
With this idea as departure point, this thesis studies both on-line and off-line learning 
of ambiguity resolution in the context of the DOP model. To this end
\begin{itemize}
 \item it presents a framework for specializing performance models, especially the DOP model, and
       broad-coverage grammars to limited domains by ambiguity reduction. Ambiguity-reduction 
       specialization takes place off-line by using a tree-bank that is representative of
       a limited domain of language use.
 \item it presents deterministic polynomial-time algorithms for parsing and disambiguation under 
       the DOP model for various tasks such as sentence disambiguation and word-graph 
       (speech-recognizer's output) disambiguation.
       Crucially, these algorithms have time complexity linear in STSG size.
       It is noteworthy that prior to the first publication of these algorithms, 
       parsing and disambiguation under the DOP model took place solely by means
       of inefficient {\em non-deterministic exponential-time}\/ algorithms.
 \item it provides proofs that some actual problems of probabilistic disambiguation under 
       Stochastic Context-Free Grammars and Stochastic Tree-Substitution Grammars are NP-Complete.
       The most remarkable among these problems is the problem of computing
       the most probable sentence from a word-graph under a Stochastic Context-Free 
       Grammar (SCFG). 
 \item it reports on an extensive empirical study of the DOP model and the specialization 
       algorithms on two independent domains that feature two languages and two different tasks. 
\end{itemize}
The rest of this chapter presents a brief general introduction to this thesis.
Section~\ref{SecBroadC} caters especially to readers who are not familiar with the
relevant developments and debates in the field of Computational Linguistics
in general and in Corpus-based Linguistics in particular. It pinpoints the present
research in the general direction of Computational Linguistics by describing the 
course of arguments that lead to the evolution of what currently are known as
performance models of language. 
Keywords in this section are: linguistic grammars, parsing, competence models,
ambiguity, overgeneration, probabilistic grammars, corpus-based models, learning and
performance models. Section~\ref{SecDOP} provides
a short introduction to Data Oriented Parsing. Section~\ref{SecEfficiency} discusses
shortly the principal subject of this thesis: efficiency of performance models.
Section~\ref{SecProbStatement} states the problems that this thesis deals with, the 
hypotheses that it defends, and its contributions.
Finally section~\ref{SecHisOver} 
provides an overview of the other chapters.
}
%
\section{Ambiguity and performance models}
\label{SecBroadC}
Humans interact in speech and in writing and they are usually able to understand the messages 
that they exchange. The main problem that keeps the researchers busy in the field of Natural 
Language Processing (NLP) is how to model this linguistic capacity.
Besides the ``elevated" scientific interest and curiosity, the research in NLP is
also driven and often even financed by ``humble" economical interest. Needless to say,
it is very attractive to automate tasks in which language plays a central role, e.g. 
systems that you can command through speech, systems that can have a dialogue with 
humans in order to provide them with information and services (e.g.~time-table information
and ticket reservation), systems that translate the European Commission's lengthy reports 
simultaneously into fifteen or more languages. In fact, some impoverished systems have already
found their way to the market, speech-recognition systems that understand some words and 
even sentences. However, it is not an exaggeration to say that the field of NLP is a baby 
that has just discovered that it can stand~up.

A major assumption that underlies the research in NLP is that  human languages 
share a common internal structure.  This assumption is essential for NLP because it implies 
that it is possible to capture the many and diverse languages in one single model: the model
of human languages. 
What this model should look like and what methods it should be based on is still a subject 
of debate in NLP research. However, most NLP researchers agree on the need for a 
divide-and-conquer modeling strategy: the model of understanding a spoken/written message 
is divided into a sequence of modules, each dealing with a subtask of linguistic 
understanding. Examples of these modules are speech-recognition 
(constructing words from speech signals), morphological analysis (exposing the structure of 
words), part-of-speech tagging (categorizing words), syntactic analysis (exposing the structure
of sentences) and semantic analysis (assigning meanings to sentences). 
By dividing the complex task into smaller subtasks (with suitable 
interfaces between the corresponding modules), NLP researchers hope that it will be 
``easier" to {\em understand and model}\/ each of the subtasks separately. 

In this thesis we are mostly interested in syntactic analysis, also called {\em parsing}.
Syntactic analysis is concerned with discovering the internal structures of sentences
in order to facilitate the construction of representations of their meaning.
The syntax of a sentence is a kind of skeleton that supports its ``semantic flesh".
Usually, syntactic analysis is not a goal in itself but rather a kind of fore-play which 
prepares for semantic analysis. 
%
\subsection{Competence and performance models}
In the past four decades of {\em computational linguistic}\/ research, the major concern has 
been to develop models that characterize {\em what sentences are grammatical and how the meanings 
of these sentences are constructed from basic units}; these basic units can be seen as the bits 
and pieces of the syntactic skeleton and the corresponding muscles of the semantic flesh. 
In this, computational linguistics describes a language as a set of sentences, a set of analyses 
and a correspondence between these two sets.
Usually, this triple is described by a grammar that shows how sentences are constructed
from smaller phrases, e.g. verb-phrases and noun-phrases, that are in turn constructed from
yet smaller phrases or words. A grammar, thus, allows decomposing or {\em parsing}\/
a sentence into its basic units; the process of parsing a sentence using a grammar 
results in analyses (we also say that the grammar~- or a {\em parser}\/ that is based 
on it~- {\em assigns}\/ analyses to the sentence).
Computational linguistics aims at developing the types of grammars that seem most suited 
for describing natural languages. For this difficult task, the computational linguists had 
to make an inevitable assumption as to what kind of language use interests them: computational
linguists assume that the subject matter of their studies is ``idealized" language. The 
computational linguists refer to the grammars that they develop as {\em competence}\/ models 
of language~\cite{Chomsky65}, as opposed to {\em performance}\/ models of language, i.e. models 
of ``non-idealized" linguistic behavior of humans. 
\subsection{Overgeneration and undergeneration}
The focus of computational linguistics on developing grammatical descriptions resulted in lack of
attention for modeling {\em the input-output behavior of the human linguistic system}. 
In applications that involve natural language, grammar engineers try in the first place to
develop grammars that bridge the gap between these grammatical descriptions and actual language 
use.   Despite of the immense efforts, these grammars suffer from 
many problems among which two are most severe: overgeneration and undergeneration. 
Overgeneration, also known\footnote{
Some linguists use the term {\em overgeneration}\/ differently from the term {\em ambiguity}.
This is because these linguists view a language as {\em a predefined and fixed}\/ set 
of utterances. When the grammar recognizes sequences of words that are not in the language, 
the grammar overgenerates; when the grammar assigns to an utterance more than one analysis,
the grammar is ambiguous. Thus, in this line of reasoning a grammar can be ambiguous but not 
overgenerating. In our experience-based approach, we view a language as a probability
distribution determined by experience rather than being an a priori fixed set. Moreover,
we view parsing as assigning a {\em single analysis}\/ to every sentence. In our view,
an ambiguous grammar, that does not offer any means to discriminate between the various 
analyses that it assigns to the same sentence, is an overgenerating grammar. And because
we do not believe in clear-cut grammaticality judgments but in a continuum of
``grammaticality levels", overgeneration coincides with ambiguity.
} as {\em ambiguity}, is the phenomenon that a grammar tends
to assign too many analyses to a sentence, most of which are not perceived by humans.
Undergeneration, on the other hand, is the phenomenon that a grammar does not assign
to a sentence the analyses that are perceived by humans (often the grammar does not assign
any analyses at all). In this thesis we will focus on the problem of overgeneration.
Next we provide two examples of overgeneration. The first exemplifies this problem 
in natural language and explains why linguistic grammars do not aim at solving it. And the 
second example exemplifies why this problem is so severe in linguistic grammars.
\Example{FirstExamp}
{Consider the sentence ``John found Mary a nice woman". It has two different interpretations:
 {\em John considered Mary to be a nice woman}\/ or  {\em John found a nice woman for Mary}. 
 Most people perceive only one interpretation (often the first). Nevertheless, it is
 essential that a grammar of English be able to analyze this sentence both ways. Only within
 enough context and with access to knowledge resources beyond language (world-knowledge), 
 the choice of the right analysis might become clear. Therefore, linguistic competence grammars
 do not try to select the correct analysis but assign (at least) both analyses to the
 sentence. \\
}

\begin{figure}[hbt]
\epsfxsize=13cm
\center{
\epsfbox{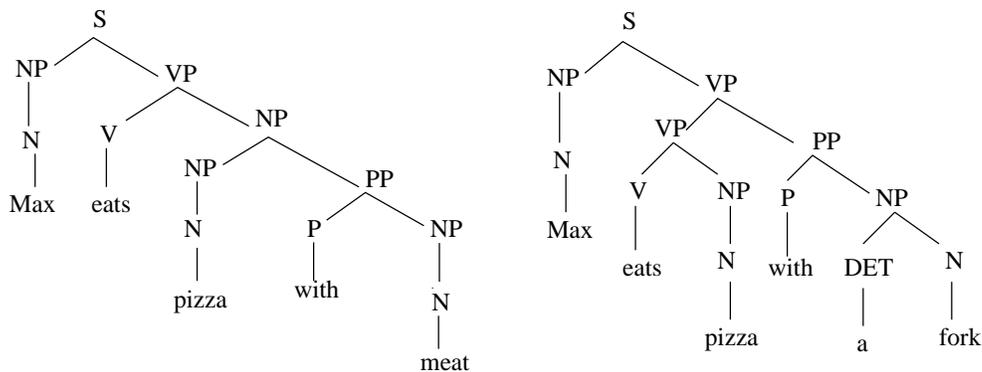}
}
\caption{Syntactic analyses}
\label{FigAmbExamples}
\end{figure}

Linguistic grammars assign to the preceding example sentence many other analyses
beside these two. Most analyses are not perceived by humans and are a byproduct of
the complex grammar rules. An empirical study of this problem~\cite{MartinChurchPatil} 
shows that actual linguistic grammars assign hundreds but even thousands of analyses to the
very same sentence. The following example shows how this can happen.\\

\Example{SecondExamp}
{Consider the two sentences ``Max eats pizza with a fork" and ``Max eats pizza with meat". 
 The meanings of both sentences are clear to the reader and the correct syntactic analyses 
 of both sentences are found in figure~\ref{FigAmbExamples}. 
 A competence grammar of English should contain the correct analysis for each of the 
 two sentences. 
 However, this means that such a grammar assigns to each of these sentences at least
 two analyses: the correct analysis and another analysis that is derived by combining rules
 that originate from the analysis of the other sentence. \\
}

A grammar that undergenerates or overgenerates is not very useful in practice.
A system based on such a grammar tends to be brittle and inaccurate. Sentences that are
assigned no analyses by the grammar are not understood by the system, and sentences that
are assigned too many analyses confuse the system. Therefore, for building a system that
involves a  serious linguistic task, a grammar engineer must find a way to weed out the 
wrong analyses from the grammar but keep the correct ones in it. This is important because 
if the wrong analysis is assigned to a sentence it might result in the wrong meaning.
Consider what happens if Max eats the fork and the pizza; this is exactly what happens 
if the analysis at the right-side of figure~\ref{FigAmbExamples} is properly adjusted and 
assigned to the sentence at the left-side of that figure.

It is by no means easy to get rid of overgeneration and undergeneration from a grammar for 
a serious portion of a natural language. Firstly, it is very hard for a human to keep track 
of the complex interactions between the many rules of a grammar. And secondly, hacking the 
grammar to get rid of overgeneration usually results in extreme undergeneration (and vice versa).
As a matter of fact, there are no serious natural language grammars out there that do not 
suffer from overgeneration as well as undergeneration.
\subsection{The probabilistic-linguistic approach}
A decade and a half ago a different approach to the problem of ambiguity in linguistic 
grammars revived due to its success in speech-recognition: the probabilistic approach. 
The slogan of this approach is: {\em allow linguistic grammars to overgenerate but resolve 
the ambiguities by assigning probabilities to the different analyses of a sentence}.
Assigning probabilities to the analyses of a sentence enables  selecting one
analysis: the analysis with the highest probability. In fact, according to the probabilistic 
approach it is possible to assign probabilities that minimize the chance of committing 
errors in selecting an analysis\footnote{
A probabilistic model can minimize the chance of committing error to the extent that it is
true to the task at hand (see~\cite{Mitchell97} on Bayesian modeling). 
If for example the probabilistic grammar is very shallow, 
e.g. sentences are assigned only the sentential-category without any internal structure, it  
can minimize the error in assigning a sentential-category to a sequence of words. The lower
bound on the error-rate depends on how important are the hierarchical linguistic structures  
to the task at hand. A good example of this scenario are the n-gram
models of part-of-speech tagging that seem to have a lower-bound on error-rate (roughly 3-5\%).
}. Typically, assigning probabilities to the analyses is achieved by 
attaching probabilities to the rules of the linguistic grammar, resulting in probabilistic 
or stochastic linguistic grammars 
e.g.~\cite{Fujisaki84,ViterbiCFG,Jelinek,Resnik,Schabes92,SchabesWaters93}

In one view on the probabilistic approach, the probabilities assigned to a linguistic 
grammar are considered means for approximating phenomena that the grammar is not aimed at 
modeling, e.g. broad-context (discourse) dependencies and world-knowledge. The probabilities, 
thus, constitute averages of many variables that influence 
the choice of the correct analysis of a sentence. Although this view is valid, it
might not do the probabilistic approach full justice. Many researchers believe that the 
probabilistic approach to language has psychological relevance. 
A relevant psychological observation here 
is that {\em humans tend to register frequencies and differences between frequencies}.
Since probabilistic models employ relative frequencies to estimate probabilities, they can 
be seen as implementing this observation. In any case, since we are here less 
interested in psychological studies, we will refrain from discussing them and we
refer the interested reader to the discussions provided in~\cite{Scha90,Scha92,RENSDES}
and the references that they cite.
\subsection{The corpus-based approach}
Although the probabilistic approach offers linguistic grammars a way out of the 
ambiguity maze, it does not offer a solution to the problem of developing performance
models of language. Linguistic grammars model idealized language use. Extending
them with probabilities does not make them suitable for modeling linguistic input-output 
behavior of humans. Many constructions will be missing from the linguistic grammar and
have to be engineered by humans. Moreover, there is a related fundamental question with
regard to whether probabilities should be attached to {\em linguistic competence grammar}\/ 
rules in the first place.
Generally speaking, probabilities are more meaningful when they are attached to dependencies
and relations that are more significant for the task they are employed for. 
In language modeling, often relations between words, between phrasal-categories and, for some 
sentences, even between whole constituents are most significant in determining the correct 
analyses of sentences. Therefore, attaching probabilities to the rules of a linguistic 
competence grammar is equivalent to attaching them only to a small portion of the linguistic 
relations that are significant to syntactic and semantic analysis.


A major shift in developing performance models took place less than a decade ago.
It can be characterized as a shift from a {\em top-down}\/ approach to a {\em bottom-up}\/
approach to modeling. Rather than considering the linguistically developed competence models
as central to performance models, the bottom-up approach puts collections of real-world data,
{\em corpora}, at the center of its activity. With the corpus of data at the center, 
this approach, the {\em corpus-based}\/ approach, aims at {\em acquiring}\/ or {\em learning}\/ 
from the data a suitable performance model. Crucially, the data is collected in such a way 
that it is representative of language use in some domain.
The data may be {\em annotated}\/ with linguistic information, or it may
be ``raw". If the data are annotated, the annotation is usually based on linguistic knowledge.
This offers  a  kind of ``back-door" where linguistic knowledge enters a performance model.
Learning from annotated data is analogous to tutoring (supervising) a novice in some task 
by showing him examples and their solutions. If the data are ``raw", this is similar 
to learning from scratch without the help of a supervisor (i.e. unsupervised learning).

By acquiring performance models from corpora, the corpus-based approach offers the
hope that the acquired performance models will be better suited for processing new similar data.  
Another attractive feature, market-wise, of automatic learning of performance models 
is the independence from manual labor, which is not always available and not always 
consistent.

Currently, there are different corpus-based methods in linguistics.
Some of these methods have been successful in acquiring useful performance models for
some (relatively simple) linguistic tasks, e.g. part-of-speech tagging of free text (PoSTagging).
Examples\footnote{There are also linguists that acquire their grammars and models 
from corpora manually, e.g. the Constraint Grammar (CG) approach~\cite{ENGCG}.
This can be considered a linguistically-oriented corpus-based approach 
to acquiring linguistic models. Currently, the CG~approach outperforms purely statistical
methods in the task of PoSTagging~\cite{SamuelssonVoitilainen}. 
See~\cite{Samuelsson98} for a discussion of the role of 
linguistics in statistical learning, and for a summary of the CG approach, where it (sensibly)
meets and where it differs from other corpus-based approaches.}
of successful corpus-based methods for PoSTagging are Hidden 
Markov Models (HMMs)\  \cite{BahlMercer76,Church88}, Transformation-Based Error 
Driven Learning~\cite{BrillPoSTagging,BrillThesis}, Instance-Based 
Learning~\cite{TilburgPoSTagging} and Maximum-Entropy Modeling~\cite{Ratnaparkhi96}.
There are a few more linguistic tasks for which the corpus-based approach currently offers 
suitable solutions, e.g. morphological analysis, phonological analysis, chunking or 
recognition of noun-phrases. However, in general, these tasks constitute the less complex 
subtasks of sentence processing. For the more complex tasks, e.g. syntactic and semantic analysis, none of the contemporary corpus-based approaches can claim similar success yet.

Most contemporary corpus-based methods of syntactic analysis are 
probabilistic, e.g.
\cite{SchabesIO,Magerman94,RENSDES,Charniak96,Collins96,Collins97,Ratnaparkhi97,Srinivas97,GoodmanDES}; 
they acquire both their grammars and the probabilities 
of the grammars from {\em tree-banks}, i.e. corpora that contain syntactically analyzed 
sentences. Some of these approaches achieve a parsing accuracy that constitutes an improvement
even on well-developed linguistic grammars and parsers~\cite{Magerman94}. 
Nevertheless, none of these young methods has demonstrated yet that it is the most suitable
for applications of natural language disambiguation.
In this situation, the field of Computational Linguistics has become an
arena where young methods compete to prove their vitality and strength. 
Computational Linguistics has become a multidisciplinary field where tools from many fields,
e.g. Linguistics, Logic, Probability Theory, Information Theory, Computer Science and Machine 
Learning, are imported, sharpened and customized for linguistic processing.
\section{Ambiguity and Data Oriented Parsing}
\label{SecDOP}
Data Oriented Parsing (DOP)~\cite{Scha90,Scha92} is a performance model of language that 
is based on the observation that {\em adults tend to process linguistic input on the basis
of their memory of past linguistic experience}. The linguistic experience of an adult is 
represented by a tree-bank of analyzed sentences. To process new input, DOP relies on some 
psychological observations that boil down to the statement that 
``frequencies of analyses that are perceived in the past influence the choice of the 
analysis of the current input" (see~\cite{Scha90,Scha92,RENSDES}). Thus, DOP does not only
memorize the past analyses but also their frequencies. To analyze a new sentence
DOP exploits similarities between the sentence at hand and all sentences that occurred in 
the past. The frequencies of past analyses allow a quantification of the notion 
of similarity in DOP.

DOP processes a new sentence by first recalling all relevant subanalyses of analyses from
the tree-bank. The notion of a subanalysis is taken here in the broadest sense: any part
of an analysis that does not violate the atomicity of grammar rules that constitute
that analysis. Then these subanalyses are assembled, in the same fashion as grammar
rules, into analyses for the current sentence. The process of assembling the subanalyses is
governed by their {\em relative frequencies}\/ in the tree-bank. In this process,
the relative frequencies are treated as probabilities and Probability Theory is brought 
into action for computing probabilities of assembled subanalyses and analyses. The most
suitable analysis for the current sentence is selected from among the many 
assembled analyses as the one with the largest probability.

A first implementation of this model is worked out in~\cite{RENSDES}. Bod implements
Scha's DOP model as a probabilistic grammar that is projected from a tree-bank.
Crucially, as Scha prescribes, this probabilistic grammar consists of all subanalyses
of the analyses in the tree-bank together with their relative frequencies.
And in conformity with earlier work on generative probabilistic grammars, the relative
frequencies of subanalyses are conditioned on their root-node category.\\

The fact that DOP employs a memory of past experience relates DOP to a Machine 
Learning tradition with many names (and subtle differences): Memory-Based Learning, 
Analogy-Based Reasoning, Instance-Based Learning and Similarity-Based 
Learning (see~\cite{StanfillWaltz,Aha,AamodtPlaza,DaelemansMBL}). 
At least as interesting, however, is the fact that DOP extends the Memory-Based 
approach in two important ways: the analysis of new input does not rely on a flat 
representation of the past-analyses but on a hierarchical~- possibly recursive~- 
linguistic structure of the analyses,
and the analogy function is a probability function\footnote{
DOP acquires its probabilities using Maximum-Likelihood Estimation (MLE), which
is a restricted form of Bayesian-Learning (see section~\ref{SecIndLearning}).
} (with the relative frequency as interpretation of the notion of probability).
\section{Efficiency problems of performance models}
\label{SecEfficiency}
A remarkable fact about current performance models of natural language parsing, corpus-based
or not, is their entrenched inefficiency of processing. Since performance models are meant 
for real-world language use, their natural habitat is the world of applications. In general,
the application of any performance model to real tasks is governed by various limitations 
on space, time and data. Models that do not provide efficient solutions that fit into 
the available space and time are impractical and will not scale up to larger applications.
And methods that require huge amounts of data are economically crippled. 
Efficiency is indeed an important factor in applying natural language parsing models 
in practice.

One of the most inefficient performance models of language is the DOP model. Due to its 
enormous probabilistic grammars, the DOP model suffers from extremely high time and space costs. 
Despite of the fact that corpus-based performance models such as DOP
learn accurate disambiguation from tree-banks, none of them accounts for efficiency
aspects of human language processing.
Efficiency in human behavior in general and in linguistic behavior in particular, is 
a hallmark of intelligence.
Models that resort to exhaustive search of a huge space of possibilities usually miss 
crucial aspects of the observable behavior of the human linguistic system.

Currently, many applications that involve natural language are aimed 
at {\em limited language use}. The language use in these applications is limited 
to an extent that is determined by {\em system design}\/ (e.g. restricted dialogue) and/or
by {\em the domain of application}, e.g. travel-information and ticket-reservation systems, 
or computer-manual translators. 
Upon studying existing performance models of natural language, one finds that 
the actual time and space consumptions of these models depend {\em solely}\/ on characteristic 
measures of the probabilistic grammar and the {\em individual}\/ input utterances,
e.g. utterance length, grammar size and ambiguity; the time and space consumptions of these
models are {\em not affected by biases that are typical for limited domains of language use}.
In fact, it is clear that current performance models have overlooked some attractive 
{\em efficiency properties}\/ that are attributed to the linguistic behavior of humans, 
especially in relation to limited domains of language use. Language use in limited domains 
shows much less variation than language use in less limited 
domains \cite{WinogradFlores,SamuelssonThesis}, and humans tend to process frequent input 
more efficiently \cite{Scha90}. These are properties of human behavior on {\em samples}\/ 
(i.e. representative collections) of utterances and analyses in limited domains, rather than 
on individual utterances. Performance models currently do not exploit such properties to 
improve their efficiency.

Earlier work has acknowledged the importance of some efficiency properties  
of samples of utterances and analyses in limited domains.
In~\cite{Rayner88,SamuelssonRayner91,Samuelsson94,Neumann94,Srinivas97} such properties
are exploited in order to improve the efficiency of parsing by {\em broad-coverage linguistic 
grammars}\/ that can be considered competence models of natural language. Except for
\cite{Samuelsson94}, these efforts exploit efficiency properties by {\em precompiling
examples}\/ using a {\em pure form}\/ of Explanation-Based Learning 
(EBL)\ \cite{DeJong81,DeJong+Mooney86,MitchellEtAl86,Harmelen+Bundy} (see section~\ref{SecEBL}). 
Samuelsson~\cite{Samuelsson94} was the first to observe that these are properties of 
{\em samples}\/ (rather than individual analyses) and can be exploited for extending 
EBL with statistical reasoning.
Encouraged and inspired by these efforts, this thesis addresses efficiency problems of 
performance models in general and the DOP model in particular by focusing on efficiency 
properties of language use in limited domains.
\section{Problem statement, hypotheses and contributions}
\label{SecProbStatement}
This section states the problems that this thesis addresses, sketches the solutions that 
it provides and summarizes its contributions.
\subsection{Problem statement}
This thesis focuses on efficiency aspects and complexity problems of contemporary performance 
models in general and the DOP model in particular. It studies and provides solutions for two
related problems. The first problem concerns acquiring and applying these 
performance models under actual limitations on the available data, space and time.
This problem is most urgent in the DOP model and its various 
instantiations~\cite{RENS92,RENS95,Charniak96,SekineGrishman95,BonnemaEtAl97,BodKaplanSchaI}.
And the second problem is the independence of the actual time and space complexities of 
disambiguation algorithms under current performance models {\em from the domain of language 
use}. This is a consequence of the fact that current performance models do not exploit
general efficiency properties of language use in limited domains.
Next, each of these problems is elaborated.
%
\paragraph{Problem 1:}
   A base-line research agenda for any performance model of parsing and disambiguation 
   consists of two elements: algorithms that are efficient enough to enable 
   {\em reliable}\/ empirical experimentation, and a thorough understanding of the computational
   complexity of problems of parsing and disambiguation. The DOP model suffers from the lack 
   of both. Next we elaborate on each of these two subjects.
    \begin{description}
      \item [Algorithms:]
           The lack of efficient algorithms for DOP and similar models can be attributed
            to two different time and space complexity issues:
        \begin{description}
          \item [Exponential-time:] The DOP disambiguation algorithms developed prior to this
            work\footnote{The first versions of the present work were published 
            in~1994~\cite{MyNEMLAP94}.} (Monte-Carlo parsing~\cite{RENSMonteCarlo}) are 
            {\em non-deterministic exponential time}\footnote{
            Although Bod~\cite{RENSDES} claims that his algorithm is non-deterministic 
         polynomial-time, \cite{GoodmanDES} shows that Monte-Carlo parsing is exponential-time.}.  
         Due to their inefficiency, these methods prevented reliable empirical experimentation 
         (based on the cross-validation technique) with the DOP model in the past~\cite{GoodmanDES}.
         In real-world applications, parsing and disambiguation cannot be based on these 
         methods because they do not scale up to actual applications.
      \item [Grammar size:] 
        The DOP model employs very large probabilistic grammars. \linebreak
        Therefore, two problems arise 
        in acquiring and employing them in practice. Firstly, from a certain point on, the size 
        of a probabilistic grammar becomes a major factor in determining the efficiency of 
        disambiguation. For the 
        actual DOP probabilistic grammars this is indeed the main factor that determines their 
        actual time- and space-consumption. And secondly, the larger the probabilistic grammar,
        the more probability parameters it has. The more parameters a model has, the more data 
        is necessary for acquiring good relative frequencies as estimates of these parameters.
        This is the problem of {\em data-sparseness}. Essentially this boils down to the 
        economical observation that constructing large enough tree-banks is expensive.
      \end{description}
      \item [Complexity:] Prior to this work\footnote{The first publication
         of our complexity results is~\cite{MyCOLING96}.} there existed 
         no studies of the computational time- and space-complexities of actual problems 
         of disambiguation under the DOP model and similar probabilistic models.
    \end{description}
\paragraph{Problem 2:}
      An important observation about limited domains is that humans tend to express 
      themselves in the same way most of the time~\cite{WinogradFlores,SamuelssonThesis}. 
      The direct implication of this observation is that in such domains humans tend to 
      employ only part of their linguistic capacity. Existing performance models that are 
      acquired from tree-banks, annotated in terms of broad-coverage (i.e. domain independent)
      grammars, are not equipped to account for this. Therefore, parsing and disambiguation 
      algorithms for these models have time and space consumptions that are {\em independent 
      of the properties of samples of sentences and analyses from these limited domains}.

      Two of these properties are of interest here. Firstly, the frequencies of utterances  
      in limited domains usually constitute a non-uniform distribution; in this respect,
      humans are able to anticipate on more frequent input in  limited domains in order to 
      process it more efficiently~\cite{Scha90}. And secondly, based on the observations
      of~\cite{WinogradFlores}, domain specific language use is usually less ambiguous 
      than it is in domain-independent competence models and performance models that are 
      based on them.

      The observation of~\cite{WinogradFlores} constitutes the main motivation 
      behind the various efforts at acquiring linguistic competence grammars that are 
      {\em specialized}\/ for limited 
      domains~\cite{Rayner88,SamuelssonRayner91,Samuelsson94,Srinivas97,Neumann94}.
      The task that these efforts address\footnote{
        Some of these efforts employed probabilistic disambiguation after parsing (spanning
        the parse-space). However, their specialization methods concentrated only on the 
        parsing part of the system.
      } is how to specialize {\em linguistic broad-coverage 
      grammars}\/ (rather than full performance models that use probabilistic grammars)
      to specific domains. Their main goal is to acquire a specialized grammar with a 
      {\em limited but sufficient coverage}\/ (i.e. sentence recognition power). 
      However, for current performance models this does not {\em directly}\/ address two
      important issues. Firstly, that domain specific language use is usually {\em much less 
      ambiguous than the general case}. And secondly, that current {\em performance models are 
      probabilistic corpus-based}\/ rather than pure linguistic.
%
\subsection{Fundamental hypotheses}
The main hypothesis of this thesis is that more frequent input in limited domains 
can be processed more efficiently if language use in limited domains is modeled
as unambiguously as possible. This is stated here as a requirement on performance models 
that are acquired from tree-banks:
\begin{verse}
{\sl domain specific language should be modeled as unambiguously as possible 
     by a specialized performance model}. 
\end{verse}
Within an Information Theoretic interpretation of this requirement, the property that 
more frequent input is usually processed more efficiently becomes a derivative;
in order to model domain specific language use as unambiguously as possible,
frequency must play a central role. When frequent input is modeled as unambiguously as 
possible, it is usually processed faster and it requires less space. 

By addressing the second problem, the first problem is also addressed partially, especially
the grammar-size issue. When a probabilistic grammar is broad-coverage, it includes many 
probabilistic relations that are highly improbable in the domain. By specializing the 
probabilistic grammar to the domain and removing ambiguities that are not domain specific, 
many of these relations are also removed from the probabilistic grammar. This results in 
smaller probabilistic grammars.

Based on this hypothesis, this thesis defends the idea that the main solution to these 
problems lies in employing two {\em complementary and interdependent}\/ systems for
ambiguity resolution in natural language parsing and disambiguation:
\begin{enumerate}
\item An {\em off-line partial-disambiguation} system based on {\em grammar 
      specialization\linebreak 
      through ambiguity reduction}. This system is acquired through the automatic learning 
      of a less ambiguous grammar from a tree-bank representing a specific domain. 
\item An {\em on-line full-disambiguation} system represented by the DOP model. 
\end{enumerate}
These two manners of disambiguation are {\em complementary}: on-line disambiguation is applied
only where it is {\em impossible to disambiguate off-line without causing undergeneration}.
And, crucially, they are {\em interdependent}\/ since a specialized less ambiguous grammar, 
acquired off-line, can serve for {\em specialized}\/ re-annotation of the tree-bank;
a DOP model that is obtained from this {\em specialized}\/ tree-bank is called a
{\em specialized DOP (SDOP) model}. 
\subsection{Contributions}
This thesis develops a new off-line disambiguation framework for the specialization
of performance models and broad-coverage grammars, dubbed the Ambiguity-Reduction Specialization
(ARS) framework. Based on the fundamental hypothesis stated above, the framework 
focuses specialization on how to reduce ambiguity without loss of accuracy 
and coverage; it formally casts the task of {\em specialization}\/ as a 
{\em constrained-optimization learning problem}\/ based on Information Theoretic formulae.
The ARS framework provides general guidelines for specializing the DOP model 
and other probabilistic models. It is implemented in algorithms for acquiring 
specialized grammars from tree-banks, algorithms for acquiring SDOP models, and 
novel parsing and disambiguation algorithms that combine the specialized grammar 
with the original grammar and the SDOP model with the original DOP model. 

For on-line disambiguation, this thesis contributes efficient deterministic 
polynomial-time and space algorithms. Important for the DOP model is that these algorithms 
have time- and space-complexities that are linear in grammar-size, and that they are equipped
with effective heuristics that control the size of DOP grammars. These algorithms constitute
a considerable improvement in time- and space-consumption (a reduction of two orders of
magnitude) on earlier non-deterministic algorithms.
Besides these algorithms, the thesis provides a study of the computational complexity of 
probabilistic disambiguation under the DOP model and some related probabilistic grammars. 
The study contains proofs that some of these problems belong to the class of NP-Complete 
problems, i.e. they are intractable (as long as the NP-Complete problems are considered 
intractable). 

The algorithms that the thesis contributes are implemented as computer programs in two 
systems: the Data-Oriented Parsing and Disambiguation System (DOPDIS) and the Data-Oriented 
Ambiguity Reduction System (DOARS). Using these systems, the thesis also contributes 
an empirical study of the various algorithms on two independent domains\footnote{The Dutch
railway time-table inquiry domain (OVIS) and the American (DARPA) air travel inquiry 
domain (ATIS).} 
and on two related tasks (sentence-understanding and speech-understanding). It is noteworthy 
that these experiments are currently among the first and certainly the most extensive 
that test the DOP model on large tree-banks using cross-validation testing. 
\section{Thesis overview}
\label{SecHisOver}
The structure of this thesis reflects the shift in the focus of my personal interest 
from developing and optimizing parsing algorithms to developing 
algorithms that learn how to parse efficiently in order to cope with problems that are 
considered not feasible in current performance models of natural language.
Next I describe briefly what each chapter is about.
\begin{description}
\item [Chapter~\ref{CHDOPinML}] provides the reader
 with the terminology, notation and background that is necessary to the
 following chapters. It also provides a more elaborate overview of this thesis in
 the light of that background knowledge. The chapter mainly contains a brief description
 of probabilistic grammars, the DOP model, and some relevant paradigms of Machine 
 Learning (Bayesian Learning and Explanation-Based Learning).
\item [Chapter~\ref{CHComplexity}] presents proofs that some actual problems of probabilistic 
  disambiguation under models that are similar to the DOP model are NP-Complete. Among these
  problems: computing the most-probable parse for a sentence (or a word-graph) under Stochastic 
  Tree-Substitution Grammars (STSGs), and computing the most-probable sentence from a word-graph
  under Stochastic Context-Free Grammars (and STSGs).
\item [Chapter~\ref{CHARS}] presents the Ambiguity Reduction Specialization (ARS) framework 
  and algorithms that are based on it for specializing DOP and broad-coverage grammars.
  It also presents parsing and disambiguation algorithms that benefit from specialization.
  Some of these algorithms are general and apply to broad-coverage grammars, but others
  are specific to the DOP model.
\item [Chapter~\ref{CHOptAlg4DOP}] presents efficient parsing and disambiguation algorithms
  for DOP. Apart from parsing and disambiguation of sentences, these algorithms
  are also adapted for parsing and disambiguation of speech-recognizer output in the form 
  of word-graphs (or word-lattices). These algorithms underly the DOPDIS system.
\item [Chapter~\ref{CHARSImpExp}]
  presents the implementation details of the current learning, parsing and disambiguation
  algorithms of chapter~\ref{CHARS} and exhibits an empirical study of the DOP model
  and the Specialized DOP models on two domains that represent two languages (Dutch and English)
  and two tasks (sentence-understanding and speech-understanding).
\item [Chapter~\ref{CHDiscussion}] discusses the results and contributions of 
   this thesis.
\end{description}
%
%
%
%

\begin{figure}[htb]
\begin{center}
\mbox{
\epsfxsize=130mm
\epsfbox{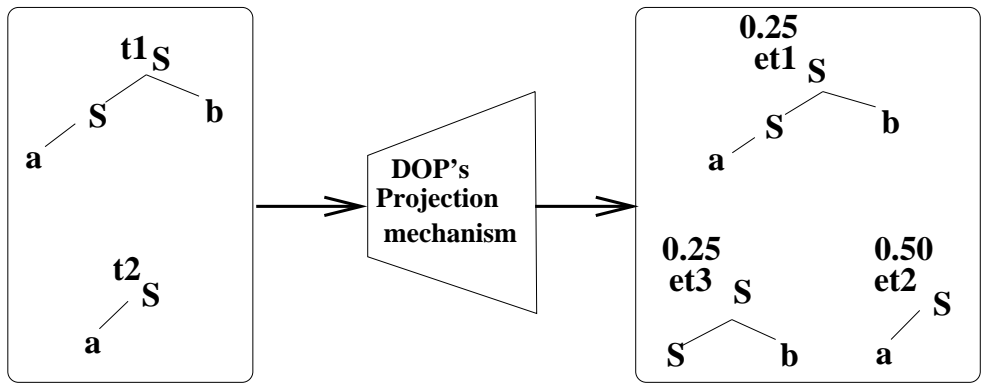}
}
\end{center}
\caption{An example: STSG projection in DOP}
\label{DOPINF}
\end{figure}

\begin{figure}[htb]
\begin{center}
\mbox{
\epsfxsize=100mm
\epsfbox{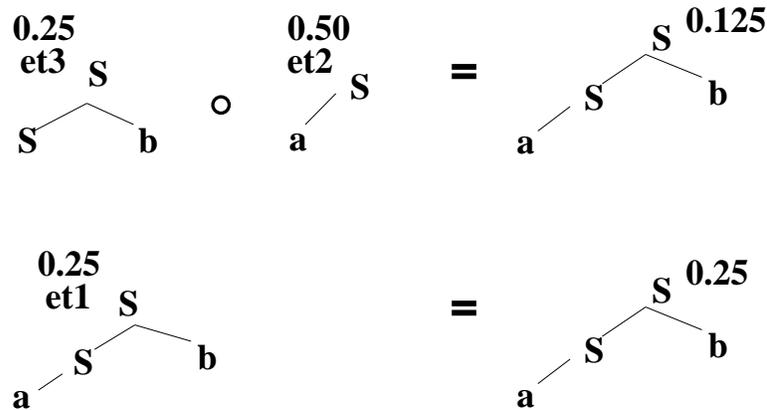}
}
\end{center}
\caption{Two derivations for the same parse}
\label{TwoDOPDers}
\end{figure}

%
\newtheorem{AexCHDinL}{Example}[chapter]
\newcommand{\exampleDinLxA}{
\EExample{AexCHDinL}
{
Figure~\ref{DOPINF} shows the ``classical" example (due to Bod) of the DOP projection 
mechanism on a toy tree-bank of two trees. 
The tree-bank trees, at the left hand side of the figure, have only one single 
non-terminal $S$ and two terminals $a$ and $b$. The resulting set of elementary-trees, 
at the right side of the figure, has three members $t_{1}$, $t_{2}$ and $t_{3}$.
Each of the elementary-trees $et1$ and $et3$ occurs only once in the tree-bank 
trees, while $et2$ occurs twice (once as a tree and once as a result of 
cutting $t1$); the total number of occurrences of elementary-trees with a root
labeled $S$ is 4, leading to the probabilities shown in the figure.
}{DinLA}  }
\newtheorem{BexCHDinL}[AexCHDinL]{Example}
\newcommand{\exampleDinLxB}{
\EExample{BexCHDinL}
{
Figure~\ref{TwoDOPDers} exhibits two derivations of the sentence ``a~b" based on
the DOP model projected in example~\ref{DinLA}. The two derivations
result in the same parse-tree with different probabilities.
}{DinLB}  }
\newtheorem{CexCHDinL}[AexCHDinL]{Example}
\newcommand{\exampleDinLxC}{
\EExample{CexCHDinL}
{In figure~\ref{DOPnotML}, a training tree-bank of two trees is shown. 
 The STSG learned by DOP is shown in figure~\ref{DOPgrammar}. And in figure~\ref{MLgrammar}, 
 another STSG is shown. The latter STSG assigns a greater total 
 probability, to the two trees of the training tree-bank of figure~\ref{DOPnotML}, than the 
 STSG learned by DOP. 
 Notice that both STSGs generate the same string- and tree-languages as the CFG underlying 
 the tree-bank. 
}{DinLC} }

\chapter{Background} 
\label{CHDOPinML}
This chapter provides the background knowledge for the rest of this thesis.
It supports the formal foundations of the other chapters through supplying
the main necessary definitions and notation. It also provides background knowledge
concerning two main pillars on which this thesis rests:
the Data Oriented Parsing (DOP) model and Machine Learning paradigms.
\section{Introduction}
\label{Intro}
%
The study of computational models of learning from past experience, Machine Learning,
is currently a vibrant field of research. It brings together many disciplines from  
many different corners of the scientific world. Its subjects are as diverse as the skills 
in which humans exhibit learning and improvement on the basis of experience, 
e.g. games such as chess, expertise such as medical diagnosis or car-reparation,
vision and linguistic capacities such as speaking, writing and reading. 

In its short history, Machine Learning presented various theoretical accounts, referred~to 
as paradigms, of various ways of learning from past experience, e.g. Inductive Learning, 
Analytical Learning and Memory-Based Learning. These paradigms are considered the abstractions 
of most, if not all, computational models that can be developed for modeling the many 
skills in which learning takes place. They abstract away from all the skill-specifics and 
constitute the subject-matter of theoretical and empirical studies concerning the
capabilities and limitations of the kinds of learning they represent. The results of these 
theoretical studies are, then, immediately applicable to all specific computational models 
that fall under these paradigms. 

In the field of natural language processing, the subject of computational language learning is 
currently gaining serious momentum. Although the current picture of human natural language 
learning is still a patternless collection of scattered ideas, some of these 
ideas, when cast into computational models, can be significant for language 
technology, where the specific application-requirements and the importance of empirical 
results delimit the range of the viable models. 
One such recent idea on human natural language disambiguation is the 
Data Oriented Parsing (DOP) model~\cite{Scha90}.

Data Oriented Parsing is a so called performance model of natural language processing, 
as opposed to so called linguistic theories and models of language competence. Roughly
speaking, the latter theories and models are concerned mainly with questions of 
coverage and adequacy of general representations, e.g. grammars, of ``idealized" human 
language use.
In contrast, performance models are concerned mainly with the question of how to simulate 
non-idealized human language use. One of the most persistent problems in language modeling
that is tackled more explicitly by performance models than competence models is the problem 
of ambiguity. In language interpretation, it is often very hard to distinguish the most 
adequate interpretation of an utterance due to the dependency of that interpretation on 
various extra-linguistic factors such as world-knowledge. Currently, the most popular approach 
to constructing performance models that tackle the ambiguity problem is through enhancing
natural language grammars ({\em not necessarily competence grammars}) probabilistically. 
In the probabilistic corpus-based approach,
a natural language is modeled as a triangle that consists of a set of utterances, a set of 
analyses and a stochastic correspondence between members of these two sets.
In general, this stochastic correspondence is achieved through spanning a probability 
distribution over the set of analyses and another related distribution over the set of 
utterances. The probability distributions are not defined directly on pairs of sentences
and analyses, rather they are defined through assigning probabilities to the production
units (e.g. rules) of a natural language grammar. What grammar to use and how to enhance it 
probabilistically are currently central themes in computational linguistics research.
In any event, when a probabilistic model is prompted to analyze an utterance, 
the model responds by 
emitting the {\em most probable analysis}\/ that corresponds to that utterance according to its 
distribution\footnote{In some cases,
the utterance is either not in the set of utterances of the model or it does not have
any corresponding analysis at all. In these cases the model simply emits the symbol of
failure.}. This most probable analysis is considered the best bet the model can make on what 
the most suitable analysis should be. \\
%

Machine Learning paradigms, Data Oriented Parsing and probabilistic grammars play a central
role in this thesis. This chapter provides the reader with the main part of the necessary 
background knowledge on these subjects.
It is of course impossible to define every basic term and notion
that is borrowed from another field which is encountered during the discussion.
Therefore, the discussion in this chapter, and the rest of the thesis, assumes that the 
reader is familiar with the {\em most basic notions}\/ common in Computer Science 
(e.g.\ in graph theory, formal language theory, automata theory, parsing technology, 
complexity theory), Probability Theory, Information Theory, Machine Learning and 
Linguistics. For textbooks and references on some of these subjects the reader is advised to
consult~\cite{ShannonWeaver,AHO,Papadi,Johnson,PapoulisBook,YoungBloot,Mitchell97}. 
For an excellent introduction to current probabilistic computational linguistics, the 
reader is referred to~\cite{SamuelssonCompendium}. 

This chapter is organized as follows. Section~\ref{SecGrammars} lists some definitions
and notation on grammars common to the subsequent sections and chapters.
Section~\ref{SecDOPOverview} provides an overview
of the DOP framework. Section~\ref{SecElemsML} briefly discusses Machine
Learning and provides short introductions to Bayesian Learning, Explanation-Based Learning
and the notion of Entropy. 
And finally, section~\ref{ThisThesis} states the goals of this thesis, in the light of 
the background knowledge that the preceding sections provide.

\newcommand{\SPRINGHERE}{\linebreak}
\newcommand{\DST}{\displaystyle}
\newcommand{\lmdr}[1]{\stackrel{#1}{\lmd}}
\section{Stochastic grammars}
\label{SecGrammars}
%
In this section, we review briefly some of the formal devices that underly the
probabilistic models that the other chapters assume, namely
Stochastic Finite State Machines (SFSMs), Stochastic Context-Free Grammars (SCFGs)
and Stochastic Tree-Substitution Grammars (STSGs). 
Needless to say, the list of definitions here is not meant to be exhaustive; 
only the main notions and terminology are listed in order to facilitate a more accurate
and concrete discussion. Other basic common notions might be used in the sequel even 
though they do not appear in this list. 
%
\paragraph{Global assumption:}
For convenience, throughout this work we assume that all involved grammars are 
proper and $\epsilon$-free.
\paragraph{String notation:}
  A string (or sequence) of symbols $w_{i},\cdots,w_{j}$, where $i < j$ are natural numbers,
  is denoted in the sequel as $w_{i}^{j}$.
%
\subsection{Stochastic Finite State Machines and word-graphs}
Finite State Machines (FSMs) also called Finite State Automata (FSAs) are formal devices
that generate Regular languages. Other equivalents for FSMs are Regular
Expressions, and Right/Left Linear Context-Free Grammars.

{\it
\begin{description}
\item [Finite State Machine (FSM):]
{An FSM is a quintuple \FSM, where $Q$ is a finite set 
of symbols called the alphabet, $\Sigma$ is a finite set of states,
$S \in \Sigma$ is the start-state, $F \in \Sigma$ is the target or final state\footnote{
It is possible to have FSMs with sets of final states. However, for every FSM with a set
of final states there is an equivalent FSM with a single final state, i.e. both accept
the same language -~both even have, up-to a homomorphism, the same set of derivations. 
},
and $T$ is the finite set of transitions, i.e tuples \mbox{$\LA s1,s2,w\RA$} 
where \mbox{$s1, s2 \in \Sigma$} and \mbox{$w \in Q$}.
}
\item [Stochastic FSM (SFSM):]
{An SFSM is a six-tuple \SFSM~ that extends the FSM 
 \FSM~ with the probability function \mbox{$P:T\lraN (0,1]$}, such that\linebreak
 $\forall l\in\Sigma$: \mbox{$\displaystyle \sum_{r\in\Sigma, w\in Q} P(\LA l,r,w\RA) = 1$}.
}
\item [Word-graph:]
{In the context of speech recognition, the output of a speech recognizer\footnote{
 Often the word-graphs output by a speech-recognizer do not fully abide by the formal
 definition of an SFSM that is given here because, for example, the probabilities on 
 the transitions that emerge from the same state might not sum up to one (due to e.g.
 pre-pruning of the word-graph).
 In this work we abstract away from such small inconveniences and assume that the word-graphs
 output by a speech-recognizer are SFSMs. In the sequel, whenever these differences become 
 important we supply a special treatment of word-graphs output by speech-recognizers.
 } is an SFSM referred to with the more casual term word-graph or word-lattice.
 Therefore, in the sequel, we will use the terms SFSMs and word-graphs as synonyms.
}
\item [Path:]
{In an FSM \FSM, a sequence $\LA s_{0},s_{1},w_{1}\RA, \cdots, \LA s_{n-1},s_{n},w_{n}\RA$ 
 of transitions from $T$ 
 is called a {\em path}. Such a path may also be indicated by means of
 the shorter notation \mbox{$s_{0} s_{1}\cdots s_{n}$ \lmdir $w_{1}^{n}$.}
}
\item [Derivation:]
{A path
 $ \LA s_{0},s_1,w_{1}\RA, \LA s_{1},s_{2},w_{2}\RA, \cdots, \LA s_{n-1},s_{n},w_{n}\RA,$ 
 where \mbox{$s_{0} = S$} and \linebreak
\mbox{$s_{n} = F$}, 
 is called a {\em derivation}\/ of $w_{1}^{n}$. 
}
\item [String accepted by FSM:]
{A string \mbox{$w_{1}^{n} \in Q^{+}$} is\footnote{$Q^{+}$ denotes the union 
                                  of all $Q^{n}$, $n\geq 1$. $Q^{*}$ denotes 
                                  the set $\{\epsilon\}\cup Q^{+}$.}
 said to be {\em accepted}\/ by the FSM\linebreak
 \FSM~\/ iff there is a derivation 
 $\LA S,s_1,w_{1}\RA, \LA s_{1},s_{2},w_{2}\RA, \cdots \LA s_{n-1},F,w_{n}\RA$\linebreak 
 where \mbox{$\forall\/ 1\leq i \leq n-1$}, \mbox{$s_{i} \in \Sigma$} 
 and \mbox{$\forall\/ 1\leq i \leq n$}, \mbox{$w_{i}\in Q$}.
}
\item [Language accepted by an FSM:]
{The language accepted by an FSM is the set of all strings from $Q^{+}$ that the 
 FSM {\em accepts}.
}
\item [Path probability:]
{The probability of the path \mbox{$\LA s_{0},s_1,w_{1}\RA,\cdots, \LA s_{n-1},s_{n},w_{n}\RA$} 
 is defined by \mbox{$\prod_{i=1}^{n} P(\LA s_{i-1},s_{i},w_{i}\RA)$}. 
 This also defines the probability of a derivation since it is a special case
 of a path.
}
\item [Probability of a string:]
{The probability of a string under an SFSM is the sum of the probabilities of all
 its derivations in that SFSM.
}
\item [Language accepted by an SFSM:]
{The language accepted by \SFSM~ is a set of pairs \mbox{$\LA {\tt string}, {\tt probability}\RA$} 
 such that\/ ${\tt string}$ is in the language of \linebreak
 \FSM\/ and ${\tt probability}$ is the probability of ${\tt string}$.
}
\end{description}
}
%
\subsection{Stochastic Context Free Grammars (SCFGs)}
{\it
\begin{description}
\item [CFG:]
  A Context-Free Grammar (CFG) is a quadruple~\CFG, where \VN\/ is the finite set of
  non-terminals, \VT\/ is the finite set of terminals, $S\in$\VN\/ is the start non-terminal
  and \Rules~ is the finite set of production rules (or simply rules), which are pairs\footnote{
  We are not interested in $\epsilon$ production-rules in this work, so we assume that
  $\epsilon$, the empty string, is not on the right hand side of any rule.}
  from \mbox{\VN$\mul V^{+}$}, where the symbol $V$ denotes \mbox{\VN$\cup$\VT}.
  A rule \mbox{$\LA A,\alpha\RA\in \RulesN$} is written \mbox{$\RuleM{A}{\alpha}$},  
  $A$ is called the 
  left hand side (lhs) and $\alpha$ the right hand side (rhs) of the rule.
\item [SCFG:] An SCFG\footnote{Also known as Probabilistic CFG (PCFG).}
       is a quintuple~\SCFG, where \CFG\/ is a CFG and 
      \mbox{$P:\RulesN$\lra$(0,1]$} is a probability function such that
      for all \mbox{$N \in$ \VN:}\linebreak
         \mbox{$\sum_{\alpha: \RuleM{N}{\alpha}\in \RulesN} P(\RuleM{N}{\alpha}) = 1$}.
\item [Notation for SCFGs:] 
         We employ capital letters such as $A, B, C, N, S$ to denote non-terminal symbols 
         and small letters such as $a, b, c, w$ to denote terminal symbols. 
         Greek letters such as $\alpha, \beta, \gamma, \delta$ denote strings of symbols 
         that can be either terminals or non-terminals (i.e. from $V^{+}$).
         Adding numerical subscripts to a symbol results in another symbol of the same type.
\item [Leftmost derivation step:]
       A {\em leftmost\footnote{ In CFGs it does not matter whether one
assumes leftmost, rightmost or any other order of derivation steps when defining a
partial-derivation or derivation.  The choice for leftmost order is convenient for 
some parsing techniques.} 
       derivation step (lmd-step)}\/ of a CFG\linebreak
       \CFG~\/ is a triple\footnote{
         Usually an lmd step is defined as a pair where the rule identity is obscured.
         However, in this work we deal also with Tree-Substitution Grammars (TSGs). For
         TSGs it is necessary to specify the rule identity in derivation steps.
         In order to keep the discussion homogeneous, it is more convenient to make the 
         rule identity explicit also in CFG lmd derivation steps.
       } 
       \mbox{$\LA \delta A\beta, \delta\alpha\beta, \RuleM{A}{\alpha}\RA$}\/ such that
       \mbox{$\RuleM{A}{\alpha}\in \RulesN$,}\/ \SPRINGHERE
       \mbox{$\delta\in$\VT$^{*}$}, \mbox{$\alpha\in V^{+}$} and \mbox{$\beta\in V^{*}$.}\
       The triple\/ \mbox{$\LA\delta A\beta, \delta\alpha\beta, \RuleM{A}{\alpha}\RA$}\/ is 
       denoted by \linebreak
       $\delta A\beta\lmdr{\RuleM{A}{\alpha}} \delta\alpha\beta$. 
%
\item [Partial-derivation:]
      A {\em (leftmost) partial-derivation}\/ is a sequence of zero or more lmd steps
  \mbox{$ \alpha_{0}\lmdr{r_{1}}\alpha_{1}\lmdr{r_{2}}\alpha_{2}\cdots\lmdr{r_{n}}\alpha_{n}$},\/ 
      where $n\geq 0$.
      A shortcut notation that obscures the lmd steps and the rules involved in 
      partial-derivations of CFGs is embodied by the symbols \lmdir/\lmpdir~ that 
      denote respectively 
      {\em zero or more}/{\em one or more}\/ derivation steps.
\item [Derivation:]
      A {\em (leftmost) derivation}\/ of a CFG is a partial-derivation that starts with 
      the start symbol $S$ and terminates with a string consisting of only terminal symbols
      (i.e.  no lmd steps are possible any more).
\item [Subsentential-form:]
      A {\em subsentential-form}\/ is a string of symbols $\alpha\in V^{+}$ achievable
      in a partial-derivation~ $\cdots$\lmdir$\alpha$\lmdir$\cdots$. 
\item [Sentential-form:]
      Every subsentential-form in a derivation is called a {\em sentential-form}.
\item [Partial-parse:]
      A {\em partial-parse}\/ is an abstraction of a partial-derivation obtained by
      obscuring the rule identities of that partial-derivation. The partial-parse is
      said to be {\em generated}\/ by the partial-derivation it is obtained from.
\end{description}
}
      Note that in CFGs it is possible to reconstruct the partial-derivation from 
      the partial-parse. Therefore, the notions of a partial-parse and a partial-derivation 
      are equivalent in CFGs.
{\it
\begin{description}
\item [Parse:]
      A {\em parse}\/ is a partial-parse obtained from a derivation.
\item [Partial-parse tree:]
      A convenient representation of a partial-derivation~/ partial-parse of
      a CFG is achieved by employing the well known representation from graph theory: 
      a tree. 
     Because the tree representation of a partial-parse is 
     a popular one, often a partial-parse is called {\em partial parse-tree}\/
     or shortly {\em partial-tree}.
\end{description}
\paragraph{\bf Additional Terminology:} 
     In a partial-parse tree $t$, the node which no other node points to is 
     called the {\em root} of $t$, or shortly $root(t)$. And the nodes from which no edges 
     emerge are called the {\em leaves}\/ of the partial-parse tree. The last 
     subsentential-form in a partial-derivation (i.e. the ordered sequence of symbols that 
     label the leaves) is called the {\em frontier}\/ of the partial-derivation and of the 
     partial-parse tree that is
     generated by that partial-derivation. A node $A$ that has an edge emerging from it 
     that points to another node $B$ is called the {\em parent}\/ of $B$; 
     $B$ is called a {\em child}\/ of $A$. 
\begin{description}
\item [Parse tree:]
     As a special case of a partial-parse tree, 
     a {\em parse-tree}\/ (shortly parse or tree) is the tree 
     representation of a parse/derivation in CFGs.
\item [Substitution-site:] A {\em substitution-site}\/ is a leaf node of 
       a partial-parse that is labeled by a non-terminal.
\item [Substitution:] In some cases it is convenient to employ a definition
       of the notion of a derivation which involves the term-rewriting operation of 
       substituting partial-parses for substitution-sites of other partial-parses. 
       A leftmost {\em substitution}\/ of\linebreak
       partial-parse $t2$ in another partial-parse $t1$ is defined only when the root
       of $t2$ is labeled with the same non-terminal symbol $N$ as the leftmost
       substitution-site in the frontier of $t1$. When this
       operation is defined, a new partial-parse is obtained, denoted as $t1\circ t2$, by 
       replacing substitution-site $N$ with partial-parse $t2$. 
\item [Sentence and string-language:] The frontier of a derivation/parse/parse-tree in a CFG 
      is called a {\em sentence}\/ of that CFG ({\em generated}\/ by that derivation). 
      The set of all sentences of a CFG is called its {\em string-language}.
\item [Tree-language:] The set of all parse-trees generated by derivations of a CFG is
      called the {\em tree-language}\/ of that CFG.
\item [Probability of a partial-derivation:] The probability of a partial-derivation
      \mbox{$r_{1}, r_{2}\cdots r_{n}$}, 
      \SPRINGHERE 
      $n\geq 1$, of a given SCFG is defined by
      $\prod_{i=1}^{n} P(r_{i})$. 
\item [Probability of a subsentential-form:] 
       The probability of a subsentential-form under a given SCFG 
       is the {\em sum of the probabilities of all partial-derivations for 
       which it is the frontier}.
\item [Probability of a sentence:] As a special case of the preceding definition, 
    the probability of a sentence under a given SCFG is the {\em sum of the probabilities 
    of all derivations that generate it}. 
\end{description}
}
\newcommand{\VNX}{$V_{N}^{'}$\/}
\newcommand{\VTX}{$V_{T}^{'}$\/}
\newcommand{\RulesX}{${\cal R}^{'}$\/}
\newcommand{\CFGX}{(\VNX,~\VTX,~$S^{'}$,~\RulesX)} 
\subsection{Stochastic Tree-Substitution Grammars (STSGs)}
Stochastic Tree-Substitution Grammars (STSGs) may be viewed as generalizations 
of SCFGs where the rules have internal structures, i.e. are partial-trees. 
Therefore, the terminology and the definitions of term-rewriting notions in STSGs 
correspond to a large extent to those in SCFGs.
However, some of the STSGs notions differ radically from those in SCFGs.
To define STSGs and their relevant term-rewriting notions, let be given 
a CFG \mbox{$G =$ \CFGX}:
{\it
\begin{description}
\item [TSG:]
     A TSG based on\/ $G$ is a quadruple \TSG, where 
     \VN$\subseteq$\VNX, \VT$\subseteq$\VTX,\/ $S\in $\VN\/
     and \Corp\/ is a finite set of partial-parse trees of\/ $G$ over the symbols in~\VT$\cup$\VN.
     Each element of \Corp\/ is called an {\em elementary-tree}.
\item [CFG underlying TSG:]
     The CFG \CFG\/ is called {\em the CFG underlying a TSG}\/ \TSG\/ iff the set \Rules\/
     contains all and only those rules involved in the\linebreak
     elementary-trees in \Corp.
\item [Substitution-site:] Recall that a {\em substitution-site}\/ is a leaf node of a 
      partial-parse tree labeled by a non-terminal; this carries over to elementary-trees
      of course.
\item [STSG:]  An STSG is a five tuple \STSG\/ which extends the TSG\linebreak
       \TSG~\/ with a function\/ $PT$;\/ $PT$ assigns to every\/ \mbox{$t\in\CorpM$}\/ 
       a value \linebreak 
       $0<~PT(t)\leq~1$ such that\/ $\forall\/ N\in$\VN:
      \mbox{$\sum_{{t\in{\cal C}},~{root(t) = N}}PT(t) = 1$}.
\item [Leftmost TSG derivation step:]
       Let be given\/ \mbox{$t\in$\Corp}\/ such that its root node is labeled\/ $A$ and its
       frontier is equal to the string\/ $\alpha$. 
       A {\em leftmost TSG derivation step}\/ (or derivation step)
       of a TSG \TSG\/ is a triple\/ $\LA\delta A\beta, \delta\alpha\beta, t\RA$ such that 
       $\delta\in$\VT$^{*}$, \mbox{$\alpha\in V^{+}$} and\/ \mbox{$\beta\in V^{*}$}.
       As before, the triple\/ $\LA\delta A\beta, \delta\alpha\beta, t\RA$ is 
       written as \linebreak
       $\delta A\beta\lmdr{t} \delta\alpha\beta$. 
\item [TSG Partial-derivation:]
      As in CFGs, a {\em (leftmost) TSG partial-derivation}\/ (or simply partial-derivation) 
      is a sequence of zero or more lmd steps
      \[ A\lmdr{t_{1}}\alpha_{1}\lmdr{t_{2}}\alpha_{2}\cdots\lmdr{t_{n}}\alpha_{n} \]
      where\/ $n\geq 0$.
      A short cut notation that obscures the lmd steps and the elementary-trees involved in 
      partial-derivations of TSGs is embodied by the symbols \lmdir/\lmpdir\/ that 
      denote respec.\
      {\em zero or more}/{\em one or more}\/ derivation steps.
      Note that the ordered sequence of elementary-trees involved 
      in a (leftmost) partial-derivation of a TSG uniquely determines
      the partial-derivation. Therefore, a partial-derivation of a TSG will often be 
      represented by the ordered sequence of elementary-trees involved in it.
\item [TSG Derivation:]
      As in CFGs, a {\em (leftmost) TSG derivation}\/ (or shortly derivation) of a TSG is a 
      partial-derivation that starts with 
      the start symbol\/ $S$ and terminates with a string consisting of only terminal symbols,
      i.e. no lmd steps are possible any more.
\item [Unfolded TSG partial-derivation:]
      The {\em unfolded TSG derivation step}\/ which corresponds to
      the TSG derivation step \mbox{$\alpha\lmdr{t}\beta$} is the leftmost partial-derivation 
      which generates\/ $t$ in the CFG underlying the TSG.
      An {\em unfolded TSG partial-derivation}\/ (also unfolded partial-derivation)
      is obtained from a TSG partial-derivation by replacing every TSG derivation step
      by the corresponding unfolded TSG derivation step\footnote{
      Note that an unfolded TSG partial-derivation is a partial-parse of the CFG
      underlying the TSG.
      }.
\item [Unfolded TSG derivation:]
      An {\em unfolded TSG derivation}\/ is the unfolded \linebreak
      partial-derivation of a TSG derivation.
\item [Subsentential-form:]
      A {\em subsentential-form}\/ is a string of symbols\/ $\alpha\in V^{+}$ achievable
      in an {\em unfolded}\/ TSG partial-derivation~\/ $\cdots$\lmdir$\alpha$\lmdir$\cdots$. 
\item [Sentential-form:]
      Every subsentential-form in an {\em unfolded}\/ TSG derivation is called a 
      {\em sentential-form}.
\item [TSG partial-parse:]
      A {\em TSG partial-parse}\/ (also partial-parse) is an abstraction of an unfolded 
      TSG partial-derivation obtained by obscuring the rule identities. The TSG partial-parse is
      said to be {\em generated}\/ by the (unfolded) TSG partial-derivation it is obtained from.
      Crucially, it is {\em not}\/ always possible to reconstruct a TSG partial-derivation 
      from a given TSG partial-parse, since there can be many TSG partial-derivations
      (involving different elementary-trees) that generate that TSG partial-parse.
\item [TSG parse:]
      A {\em TSG parse}\/ (also parse) is a TSG partial-parse obtained from an
      unfolded TSG derivation. 
      Note that there can be more than one TSG derivation that generates the same TSG parse.
\item [TSG partial-parse tree:]
      Just as for CFGs, a convenient representation of a TSG partial-parse 
      is achieved by employing the tree representation from graph theory. 
     As before we will employ the terms TSG partial-parse, 
     {\em TSG partial-parse tree}\/ and {\em TSG partial-tree} as synonyms.
     The terms root and frontier of a TSG partial-parse tree are defined exactly 
      as for CFGs.
\item [Substitution:] As in the case of CFGs, a TSG partial-derivation can be seen also 
      in terms of the operation of substituting elementary-trees in other TSG partial-trees.
      Therefore, a leftmost TSG partial-derivation involving the ordered sequence of
      elementary-trees\/ $t_{1},\cdots,t_{n}$ can be written in terms of substitution as
      \linebreak
      $(\cdots(t_{1}\circ t_{2})\circ\cdots)\circ t_{n}$ or simply  
     \/ $t_{1}\circ t_{2}\circ\cdots\circ t_{n}$. 
\item [TSG parse tree:]
     As a special case of a TSG partial-parse tree, a {\em TSG parse-tree}\/ 
     (shortly parse or tree) is the tree representation of a TSG parse.
\item [Sentence and string-language:]The frontier of a TSG derivation/parse  
      of some TSG is called a {\em sentence}\/ of that TSG (that is said to be
      {\em generated}\/ by that TSG derivation). 
      The set of all sentences of a TSG is called its {\em string-language}.
\item [Tree-language:] The set of all TSG parse-trees generated by the derivations of a TSG is
      called the {\em tree-language}\/ of that TSG.
\item [Probability of a TSG partial-derivation:] The probability of a TSG partial-derivation 
      \mbox{$t_{1}, t_{2}\cdots, t_{n}$},\/ $n\geq 1$, of a given STSG, is defined to be equal
      to\/ $\prod_{i=1}^{n} PT(t_{i})$. 
\item [Probability of a TSG parse:]
      The probability of a TSG parse is equal to the sum of the probabilities of all
      TSG derivations that generate it.
\item [Probability of a sentence:] 
      The probability of a sentence is equal to the sum of the probabilities of all
      TSG derivations that generate it.
\end{description}
}
For formal studies on TSGs and the related formalism Tree-Adjoining Grammars (TAGs),
the reader is referred to TAG literature
e.g.~\cite{Joshi85,JoshiSchabes91,Schabes92,SchabesWaters93}.
And for a comparison between the stochastic weak/strong generative power of STSGs 
and SCFGs, the reader is referred to~\cite{RENSDES}.
\subsection{Ambiguity}
{\it
\begin{description}
\item [Ambiguous grammar:]
  A grammar (e.g. CFG, TSG, SCFG, STSG) is called ambiguous iff there is a sentence in its 
  string-language that has more than one parse.
\item [Inherently ambiguous language:] A language\/ $L$ (i.e. set of strings) is called inherently
      ambiguous with respect to some class of formal grammars iff there exists no unambiguous
      instance grammar in that class that has a string-language which is equal to~$L$.
\end{description}
}
The terminology and definitions given above form the common basis for the subsequent
sections and chapters. Other definitions and terminology on SCFGs and SFSMs will be 
introduced whenever necessary. 

\section{Data Oriented Parsing: Overview}
\label{SecDOPOverview}
%
Informally, the intuitive notion of ambiguity in natural language can be described
as {\em the inability to discriminate between various analyses of an utterance}\/
due to the lack of essential sources of information, e.g. discourse and domain of 
language use.
The ambiguity of a natural language grammar goes beyond the intuitive ambiguity of a natural 
language since the grammar usually assigns extra analyses, not perceived by a human, to some 
utterances of the language. This ``extra" ambiguity, which is due to 
the imperfection of the grammar, is referred to as ``redundancy", 

It is widely recognized that the intuitive ambiguity of natural language can be 
resolved only by access to so called ``extra-linguistic" resources that surpass the power 
of existing formal grammars.
Grammar imperfection (i.e. redundancy), in contrast, is usually considered 
the result of an unfortunate choice of grammar-type or an incompetent grammar 
engineering effort; both ``misfortunes" that lead to redundancy seem to suggest that 
the problem is solvable by smarter grammar writing.
Currently, however, an opposition to this view is developing within the
natural language processing community, which believes it to be a genuine problem
that cannot be eliminated by better and smarter grammar engineering, since it is 
virtually impossible to engineer a non-overgenerating grammar (for a serious 
portion of a language) without introducing undergeneration. Compared to the ``curse" 
of substantial undergeneration, reasonable overgeneration can be considered a ``blessing". 


In any event, the view that it is necessary to involve extra-linguistic
resources for resolving ambiguity prevails in the community.
One available resource, which enables ambiguity resolution, is statistics
over a large representative sample from the language. This is exactly the 
motivation behind statistical enrichments of grammatical descriptions
in their various forms. Data Oriented Parsing (DOP) is one such statistical
enrichment of linguistic descriptions, which poses critical questions on
how to collect and how to employ the statistics obtained from a language 
sample. As we shall see below, DOP introduces its own manner of enriching
linguistic descriptions with statistics. Rather than simply enriching a
predefined competence grammar, DOP adopts a {\em stochastic memory-based}\/
approach to specifying a language and defines an ordering on the 
analyses that are assigned to every sentence in that language.

\subsection{Data Oriented Parsing}
Data Oriented Parsing (DOP), introduced by Scha in~\cite{Scha90},
is a model aimed at performance phenomena of language, in particular 
at the problem of ambiguity in language use.
In~\cite{Scha90}, Scha describes the DOP model as follows (page~14):
\begin{quotation}
The human language-interpretation-process has a strong preference for recognizing
sentences, sentence-parts and patterns that occurred before. More frequently
occurring structures and interpretations are preferred to not or rarely perceived
alternatives. All lexical elements, syntactic structures and ``constructions" that
the language user ever encountered, and their frequency of occurrence, can influence
the processing of new input. Thus, the database necessary for a realistic 
performance model is much larger than the grammars that we are used to. The language
experience of an adult language user consists of a large number of utterances. And
each utterance is composed of a large number of constructions; not only the whole
sentence, and all its constituents, but also all patterns which we can extract from
it by introducing ``free variables" for lexical elements or complex constituents.
\end{quotation}
Scha also instantiates this abstract model with an example employing the substitution
operation on what he calls ``patterns", and suggests to construct this model in 
analogy to existing ``simple" statistical models. 
Bod~\cite{RENS92} is the first to work out a formalization of an instance of Scha's 
detailed description in a computational model. 
Therefore, the DOP model is strongly associated with Bod's formalization to
the extent that the latter has become a synonym for Scha's DOP model. 
In this thesis, merely for convenience, every reference to the DOP model 
is a reference to Bod's formalization, unless stated otherwise.

\subsection{Tree-banks}
As the description of DOP suggests, it is necessary to have a tree-bank
simulating ``the language experience of an adult language user". 
Clearly, it is impractical to wait for the construction of tree-banks of general 
language use that represent the experience of an adult language user.
Therefore, it seems most expedient to employ a tree-bank limited to a specific 
domain of language use.

Preceding work involving tree-banks does not define the notion of a tree-bank.
Although it is almost always clear what the notion ``tree-bank" involves, it seems 
appropriate to define this notion explicitly here. The definition here is general 
in the sense that it can be instantiated to any kind of a linguistic theory which 
employs a formal grammar.
\DEFINE{Tree-bank:}
{A tree-bank annotated under some formal grammar\/ $G$\/ is a pair \mbox{$\LA G,A\RA$}
 where\/ $A$\/ is called the {\em analyses-sample}. 
We will refer to the formal grammar\/ $G$\/ in the tree-bank pair \mbox{$\LA G,A\RA$} with the more 
common term\footnote{
In the practice of annotating tree-banks, the term annotation scheme often refers to
a set of linguistic guidelines rather than a strict formal grammar. This set of guidelines
leaves much freedom to the annotator to fill in the gaps and even improve the annotation scheme
itself during the annotation process. In the terminology we are using here, the term annotation
scheme may refer to the sum of the set of {\em formal}\/ linguistic guidelines and 
the formalized part of the linguistic knowledge of the annotator that is involved 
in the annotation process. 
} ``annotation scheme".
An analyses-sample is a {\em sample}\/ of {\em correct}\/ analyses (assigned by an oracle). 
Each analysis is associated with a sentence. The analysis of a sentence must be
a member of the set of analyses which the formal grammar\/ $G$\/ can assign to the sentence.
}
In conformity with its use in the community, in the sequel the term {\em tree-bank}\/ 
may also be used to refer in particular to the analyses-sample of the tree-bank at hand.
\subsection{Bod's instantiation}
Given a tree-bank annotated under a CFG (i.e. every analysis in the analyses-sample
is a parse-tree of that CFG) Bod's DOP model obtains a probabilistic grammar from 
this tree-bank by extracting so called subtrees from 
the tree-bank trees by cutting them in all possible ways, as described by Scha, and 
assigning relative frequencies to them in a way similar to other existing stochastic 
models, e.g. SCFGs~\cite{Jelinek,ViterbiCFG}. 
The following constitutes a restatement of Bod's instantiation, with some additions
in order to suit the discussion in this thesis.
\DEFINE{Subtree:}{A subtree\/ $pt$\/ of some tree\/ $t$\/ is a connected subgraph of\/ $t$\/
that fulfills
           1)~every node\/ $N$\/ in\/ $pt$\/ corresponds to a node in\/ $t$, denoted\/
           $C_{t}(N)$, labeled with the same symbol, 
           2)~for every non-leaf node\/ $N$\/ in\/ $pt$, the ordered sequence of symbols that 
           label its children is identical to the ordered sequence of symbols that label the 
           children of\/ $C_{t}(N)$.
} 
\DEFINE{Subtree of tree-bank:}{
 Each subtree of a tree in the given tree-bank is also called a subtree 
 of that tree-bank or simply subtree.
}
\DEFINE{CFG underlying tree-bank:}{
The CFG \CFG~\/ is called the CFG underlying the tree-bank\/ ${\TB}$\/ iff\/
\VN\/ and \VT\/ are respectively the sets of non-terminals and terminals that label the
nodes of the trees in\/ ${\TB}$, $S$\/ is the non-terminal that labels the roots of the trees
in\/ ${\TB}$, and \Rules\/ contains all the rules that participate in the (derivations of the)
trees in the tree-bank.
}
Note that the CFG underlying the tree-bank generates a language that is partial to 
that of the annotation scheme CFG, since its productions are only those that are 
used for annotating the trees of the tree-bank.
The annotation scheme CFG might have other productions
that have not been used for annotating the tree-bank at hand; this can be due
to the fact that the tree-bank represents only a small portion of a limited 
domain of language use that the annotation CFG represents.

A subtree of a tree-bank is thus simply a partial-parse, which is related to 
(i.e. is part of) one or more trees of that tree-bank.
The notion of a subtree of a tree-bank formalizes Scha's notion of a ``construction" 
or a ``pattern". Often in the sequel, we loosely use the term {\em subtree}\/ and the 
term {\em partial-tree}\/ as synonyms; this does not entail any confusion due to the
clear contexts where these terms are used and the minor difference between the notions 
which they denote.

Let be given a tree-bank \mbox{${\TB} = \LA G^{'}, A\RA$} and 
let\/ $G=$\CFG\/ be the CFG underlying it:
\DEFINE{UTSG of a tree-bank:}
{The TSG \TSG, where \Corp\/ is equal to the set 
 of all subtrees of the trees of\/ ${\TB}$, is called
 the {\em Union}\/ TSG (UTSG) of\/ ${\TB}$.
}
\DEFINE{DOP STSG:}{
The DOP model projects from\/ ${\TB}$\/ an STSG~\STSG\
such that the TSG ~\TSG\/ is the UTSG of\/ ${\TB}$\/ and
\[\forall\/ t\in\CorpM: PT(t) = \frac{freq(t)}{\sum_{x\in\CorpM: root(x)=root(t)}{freq(x)}}\]
where\/ $freq(x)$ denotes the frequency (occurrence-count) of\/ $x$\/ in the 
analyses-sample\/ $A$. Note that in this definition the probabilities of the elementary-trees 
that have roots labeled with the same non-terminal sum-up to one; they are also proportional
to their occurrence-count in the analyses-sample.
}
%
\exampleDinLxA
%
Clearly, the notions of partial-derivation, derivation, partial-parse, parse and their
probabilities are thereby defined under a DOP STSG as they are defined under STSGs in general.
\DEFINE{A parse generated for a sentence:}{A parse is said to be {\em generated by an 
STSG for a given sentence}\/ if the frontier of that parse is equal to the given sentence.
}
Therefore, an alternative definition of the probability of a sentence, which makes
the issue of ambiguity more obvious, is: 
\DEFINE{Sentence probability:}{The probability of a given sentence is the sum 
of the probabilities of all parses that the STSG can generate for it.
}
A few observations concerning DOP are important to mention here. In contrast to SCFGs, in STSGs
there can be various partial-derivations that generate the same partial-parse.
This ``redundancy" is an essential component of the DOP model~\cite{Scha90}; each
derivation represents an informal process of combining some patterns that originate 
from different sentence-analyses that have been encountered in the past (and have
been stored in memory). The probability of a derivation is in essence a kind of weight
based on the frequencies of patterns obtained from all sentences encountered in the past.
And the probability of a parse reflects the weighted sum of the weights of all such processes 
(representing derivations) that result in that parse.

\exampleDinLxB
%
As in all stochastic grammars, in DOP the goal of parsing and disambiguation is to 
select a ``distinguished entity", e.g. a prefered parse, by optimizing some probabilistic 
function that is stated in terms of the probability which DOP assigns to that entity. 
For example, in parsing and disambiguating an input sentence,
the goal of DOP can be to compute the Most Probable Parse (MPP) that the STSG 
can generate for it. Other tasks concern computing 
the Most Probable Derivation (MPD) and the probability of an input sentence.\\ 

In his description, Scha suggests to employ a ``matching process", for parsing
and probabilistic disambiguation, which is biased towards
{\em larger and more frequently occurring subtrees}; he suggests to achieve this 
effect as follows (\cite{Scha90}, page~15): 
``{\em Statistically speaking, this effect
can be achieved in an elegant implicit manner by searching in the corpus
``at random" for matching constructions}". The matching process, which Scha
describes, betrays his wish to employ a randomized stochastic memory-based process for
parsing. An instantiation of this idea is the algorithm of 
Monte~Carlo Parsing~\cite{RENSMonteCarlo} for computing the 
MPP and the MPD of an input sentence; these Monte~Carlo algorithms are 
{\em non-deterministic exponential-time}~\cite{GoodmanDES}. 
In principle, these algorithms can achieve good approximations of the MPP and the MPD.
However, in practice, they turn out to be extremely time-consuming to the degree that 
they can be considered of theoretical interest only. Substantially more efficient algorithms 
are presented in chapter~\ref{CHOptAlg4DOP} of this thesis. And a proof that, among others, the 
problem of computing the MPP is NP-Complete is provided in chapter~\ref{CHComplexity}.
\subsection{The DOP framework}
\label{DOPframework}
As mentioned earlier, Bod's instantiation is only one of many possible instantiations 
of Scha's description of the DOP model. A DOP framework, which restates Scha's description,
is provided in Bod's dissertation~\cite{RENSDES}. An improved version of this framework is 
provided in~\cite{BodScha96}. The latter version states, in most general terms,
that a DOP model is obtained by indicating the following components:
\begin{enumerate}
\item a definition of a formal representation for utterance-analyses 
      (e.g. parses of a CFG), 
\item a definition of the fragments of the utterance-analyses that 
      may be used as units in constructing an analysis of a new utterance (e.g. subtrees
      of parses),
\item a definition of the operations that may be used in combining fragments (e.g. substitution
      of partial-trees in other partial-trees),
\item a definition of the way in which the probability of an analysis of a new utterance
      is computed on basis of the occurrence-frequencies of the fragments in the corpus.
\end{enumerate}
The DOP framework does not commit itself to any specific linguistic theory or formal language, 
nor to any grammatical framework; it can be instantiated under any choice of a formal 
representation and under any choice of a linguistic theory. 
Since Scha's introduction of the DOP model~\cite{Scha90}, there have been various 
other instantiations of it, some involving CFG-based annotations with different constraints on 
the probabilistic grammar learned from the 
tree-bank~\cite{SekineGrishman95,MyRANLP95,Charniak96,Goodman96,MyRANLPBook,Tugwell95,GoodmanDES},
some involving a semantic extension of 
PSGs~\cite{vdBergEtAl,BonnemaScr,BonnemaEtAl96,BonnemaEtAl97},
and others committed to the Lexical Functional Grammar (LFG) linguistic 
theory~\cite{BodKaplanSchaI,BodKaplan98}.


\section{Elements of Machine Learning}
\label{SecElemsML}
This section describes, in short, both Bayesian learning
and Explanation-Based Learning (EBL) (also called analytical learning).  
In order to introduce the necessary terminology, this section starts with a
brief informal description of the general setting for machine learning (for a good 
introduction on the subject, we refer the reader to~\cite{Mitchell97}). 
Subsequently, it describes, in most general terms, the paradigm of Bayesian learning 
and discusses the Information Theoretic measure of {\em entropy}. Finally, it discusses 
EBL by contrasting it to the better known paradigm of inductive learning. 
\subsection{Learning}
The notion of a {\em learning algorithm}\/ is defined, informally, 
in~\cite{Mitchell97} as an algorithm which improves the 
performance of a system, according to some measure, {\em through experience}. 
%
\paragraph{Concept learning:}
In general, learning involves acquiring {\em general concepts}\/ from specific
training examples that represent experience. Humans continuously
learn concepts such as ``bird" from examples seen in the past.
A {\em concept}\/ can be represented as a function~$f$ from some 
domain $\Sigma$ to some range $C$. For example, the concept ``bird" is a function 
from any object (in some predefined set of objects) to a Boolean value stating TRUE
if the object is indeed a bird and FALSE otherwise.
A linguistic example: the concept
``a word sequence accepted by some formal grammar of the English language" is a function 
from sequences of English words to a Boolean value. Similarly, there can be 
concepts such as ``propositional phrase (PP)" or ``verb phrase (VP)" etc.$\dots$
%
\paragraph{Instances and classes:}
For a concept \mbox{$f:\Sigma\rightarrow C$}, each member of $\Sigma$ is called 
an {\em instance}\/ and each member of $C$ is called a {\em class} (or a classification).
Usually, an instance is represented as a tuple of attribute-value pairs (or feature-value
pairs); the tuple of attributes is called the {\em instance-scheme}.
Note that the choice of an instance-scheme to represent some concept immidiately delimits 
the domain of that concept.
For example, the concept ``rainy-day" tells whether a certain day is
rainy or not.  Instances (i.e. days) of the concept ``rainy-day", 
can be represented as quadruples of values for the four attributes in the
instance-scheme \mbox{$\LA$month, temperature, cloudiness, humidity$\RA$}. 
The concept ``rainy-day" is a function that maps such instances, 
each representing a day, into one of the two classes of a day:
either TRUE (i.e.\ rainy) or FALSE (i.e.\ not-rainy).
\paragraph{Hypotheses-space:}
A {\em hypotheses-space}\/ is a pair of sets \mbox{$\LA\Sigma, C\RA$}, 
where $\Sigma$ is the set of instances and $C$ is the set of classes. 
For a learning algorithm, a hypotheses-space defines both the domain and the range of all 
functions ``eligible" for that learning algorithm (i.e. all functions 
from~$\Sigma$ into~$C$). Each function in the hypotheses-space is called 
a {\em hypothesis}.
\paragraph{Training instances:} 
The {\em training instances}\/ or {\em training examples}\/ for some concept 
constitute a finite multi-set of instances; this multi-set represents the 
experience of the system with identifying instances of that concept. 
The training instances serve as examples of the space of instances that 
the system may expect in the future. Training instances can be either (a priori) 
classified or not classified depending on the type of learning, respectively 
{\em supervised}\/ or {\em unsupervised}\/ learning. In this work we
only deal with supervised learning (from {\em positive}\/ examples); therefore, the 
training instances are pairs \mbox{$\LA i,c\RA \in \Sigma\mul C$}. 
\paragraph{Learning as search:}
In general, it is convenient to view learning as {\em search}\/ in a space of
functions delimited by the hypotheses-space.
The goal of learning is to {\em hypothesize}\/ or {\em estimate}\/ some unknown 
concept, called the {\em target-concept}.
For hypothesizing on the target-concept, the learner conducts a search through the 
hypotheses-space. The search is directed in part by {\em training instances}\/ 
(or training examples) that are provided to the learner, and in part by some 
{\em inductive bias}\/ (prior knowledge) that might be used for limiting the domain of 
the  hypotheses-space (e.g. only linear functions) or for directing the search algorithm. 
An estimate of the target-concept is found during the search as a hypothesis
that satisfies both the training-instances and the inductive bias.
In most learning paradigms, the inductive bias is essential for successful generalization 
over the training instances; without inductive bias there is no way to learn a hypothesis 
that is able to classify {\em unseen}\/ instances in a rational manner. 
\subsection{Inductive learning}
\label{SecIndLearning}
Many well known Machine Learning algorithms belong to the
paradigm of {\em inductive learning}\/;  Decision-Tree learning, 
Neural Network learning and Bayesian learning are the most 
prominent examples of inductive learning. The aim of inductive learning
is to generalize a set of previously classified instances 
(i.e. training instances) of a certain concept into a hypothesis 
on that concept. In search terminology, inductive learning seeks 
a hypothesis that fits the training instances and generalizes them 
according to some prior knowledge of the domain (inductive bias).
In most cases, inductive learning is employed to estimate a
target concept that is not fully defined beforehand. More accurately,
often, the only known description of the target concept 
is a partial description, given extensionally as the set of training 
instances together with some intuitive inductive bias (which is {\em believed}\/ to
lead to the target-concept). Thus, inductive learning relies only on 
the inductive bias (i.e. prior knowledge) for generalizing over the training 
material; this makes the inductive bias a crucial factor in inductive learning.
inductive learning can be expressed within various frameworks
that aim at providing a way of combining and expressing the inductive bias 
and the training data. One such framework, which is of importance here, is 
Bayesian Learning. 
\subsubsection{Bayesian Learning}
Bayesian learning assumes a probabilistic approach to inductive learning.
In Bayesian learning, the learner searches for the {\em most probable hypothesis 
given the data}. In other words, the search, in the hypotheses space, is for the 
hypothesis $h$ which maximizes the term $P(h | D)$, where $D$ denotes
the training data and $P(h | D)$ denotes the conditional probability of $h$ given
$D$. Since it is hard to compute the probability of the hypothesis given the data 
directly, it is more attractive to use Bayes rule for optimization. 
This results in the standard equation of Bayesian Learning\footnote{
The notation $argmax_{x} f(x)$ or \mbox{$arg_{x} max f(x)$} denotes 
``that $x$ for which $f(x)$ is maximal". Similarly there is 
$argmin_{x} f(x)$ which denotes ``that $x$ for which $f(x)$ is
minimal".
}:
\[
  argmax_{h} P(h | D) = argmax_{h} \frac{P(D | h) P(h)}{P(D)}
\]
$P(h | D)$, called posterior probability,
is the probability of a hypothesis given the data. 
$P(D | h)$, called likelihood, is the probability of the 
data given the hypothesis. And $P(h)$, called the {\em prior}\/ (of $h$),
and $P(D)$ are the prior probabilities of respectively the hypothesis and 
the data. Note that Bayesian learning sees the probability $P(h|D)$ as
a combination of how much a hypothesis is consistent with the data 
and how probable it is prior to seeing the data.
A common simplification of the above optimization term is achieved 
by removing the constant $P(D)$, resulting in 
\mbox{$ argmax_{h} P(h | D) = argmax_{h} P(D | h) P(h) $}.

Note that $P(h)$ represents our prior knowledge of the hypotheses
space, i.e. this is the distribution over the hypotheses-space which
is assumed to be known beforehand, i.e. independently of the data. 
In fact $P(h)$ is exactly the place which Bayesian learning reserves 
for the inductive bias.
In practice, often the prior is implicitly embedded in the learning algorithm by 
assuming a certain search order or some preference beyond likelihood
e.g. smallest maximum likelihood hypothesis. In other cases
$P(h)$ is assumed to have no preference, i.e. a uniform distribution
over the hypotheses, leading to the Maximum Likelihood (ML) formula: 
\mbox{$ argmax_{h} P(h | D) \approx argmax_{h} P(D | h)$}. 

\subsubsection{Minimum description length}
Bayesian learning has its own interpretation of the Minimum Description Length (MDL) 
principle~\cite{MDLPrinciple}. 
In MDL, the goal of learning is to find the hypothesis $h_{MDL}$
that minimizes the sum of its length together with the length of the data given that hypothesis;
this is equal to saying that the sum of the length of the ``irregularities" in the data
(i.e.\ data not covered by $h_{MDL}$) together with the length of $h_{MDL}$ is the 
minimum over the hypotheses-space.
Crucial in MDL, however, is that the length of an entity must be obtained using an optimal 
coding scheme. Many practical MDL algorithms measure the length of entity $X$ using 
Shannon's~\cite{ShannonWeaver} optimal code length (i.e. optimal expected message length), 
which is equal to $-log_{2} P(X)$ (see next the definition of entropy). 
This interpretation of the MDL principle can be stated by the following equation
%
\[  h_{MDL} = argmin_{h} -log_{2}(P(h)) - log_{2}(P(D | h)) = 
              argmin_{h} -log_{2}(P(h) P(D | h)) \] 

It is easy to see that this is equivalent to: \mbox{ $h_{MDL} = argmax_{h} P(h) P(D | h)$}.
Thus, the MDL principle is, in fact, interpretable as learning the hypothesis 
which maximizes the posterior probability of the hypothesis given the training 
data \mbox{$P(h | D)$}.

For an excellent introduction to Bayesian learning we recommend
the chapter in \cite{Mitchell97}. For examples of applications
of Bayesian learning and Maximum Likelihood learning we refer the
reader to~\cite{SamuelssonCompendium}.
\subsubsection{Entropy}
Having mentioned Shannon's optimal expected message length, it is time
to explain the term entropy as well, since we will need it in the sequel.
Suppose we have a source $X$ (or random variable), which is capable of 
emitting any of the words in the alphabet $X_{1}\ldots X_{n}$ under 
a probability distribution $P$. If we are to encode the next word emitted by $X$, 
using binary code, what is the expected number of bits which we should 
reserve for this code~? 

If $P$ is uniformly distributed then we have to reserve exactly 
$log_{2}(n)$ bits, or equally $-log_{2}(\frac{1}{n})$, for encoding the next 
word. A uniform distribution expresses our worst case scenario concerning 
predicting the next word which $X$ will emit. And the fact that we need
as many as $-log_{2}(\frac{1}{n})$ bits expresses that.

However, $P$ is not necessarily uniformly distributed. Thus, $P$ might be
``easier than uniform", i.e. it might be easier to predict the next word of $X$. 
In that case, we should be able to reserve less bits for the next word. 
Now let $P_{i}$, for all $i$, denote $P(X_{i})$. 
According to a proof by~\cite{ShannonWeaver},
if we reserve exactly \mbox{$-log_{2}(P_{i})$} bits for $X_{i}$, then this will
constitute a generalization of the uniform case. 
This is called the optimal code length~\cite{ShannonWeaver}.

The {\em expectation value} (or mean), denoted ${\tt E}[]$, of 
the optimal code length for the next word, which $X$ will emit, 
is given by the equation
\[ {\tt E}[-log(P)] = - \sum_{i}P_{i} log(P_{i}). \]
Exactly this value is called the entropy of source (or random variable) $X$
and is usually denoted by $H(X)$.
Entropy can be seen as a measure of how hard it is to predict the next word which
source $X$ will emit. A high entropy value simply implies very hard predictions,
i.e. many possible words or a spread out distribution with little preference.
Entropy can also be seen as a measure of the uncertainty we have by not
knowing the next word $X$ will emit. Alternatively, it can be seen as the measure
of the information on $X$ which we gain by knowing the next word it will emit.

%
\subsection{Explanation-Based Learning}
\label{SecEBL}
In contrast to other learning paradigms, in Explanation-Based Learning (EBL), 
the learner has access to a so called background-theory\footnote{
A considerable portion of the Machine Learning literature employs the term
``domain-theory" rather than the term ``background-theory". In the context of
this thesis, we prefer the latter due to the different interpretation of the
term ``domain" in computational linguistics. A better term might be ``task-theory"
but this is also confusing due to the various possible interpretations 
of the term ``task". 
}, in addition to the set of training instances. 
The background-theory consists of knowledge (e.g. rules, facts, assertions)
about the target concept (and possibly other concepts); in many cases
the background-theory is in fact an oracle that is able to explain ``how
it arrives at the classification of a given instance".

During the learning (or training) phase, the background-theory provides EBL with 
a class for each training instance {\em as well as an explanation of how 
it arrived at that class}. An explanation is (often) a proof tree or a 
derivation-sequence; it shows explicitly all derivation-steps, involving
assertions and rules from the background-theory, that lead to  
the class of the instance at hand. 
For example, a CFG can be seen as a set of assertions, asserting that sequences on
the right hand side of a rule are of the type on the left hand side of that rule.
A derivation of some sequence of words in some CFG, explains how that CFG proves that 
this sequence of words is a sentence (i.e. class non-terminal $S$). A set of training 
instances is, for example, a tree-bank annotated under that CFG.

The goal of EBL is to learn a hypothesis which, on the one hand, generalizes over the 
training instances, and on the other hand has {\em partial coverage}\/ compared to the 
background theory. Therefore, the target concept which EBL aims at, is always in the hypotheses
space of the learner, since the background theory delimits that space. The main 
learning problem for EBL is that this space contains many hypotheses that generalize over 
the training instances.
Therefore, EBL searches for a target concept that also satisfies some requirements that 
make it a more efficient (or more {\em operational}) general representation of the training 
instances than the background theory.

\begin{figure}[htb]
\begin{center}
\mbox{
\epsfxsize=130mm
\epsfbox{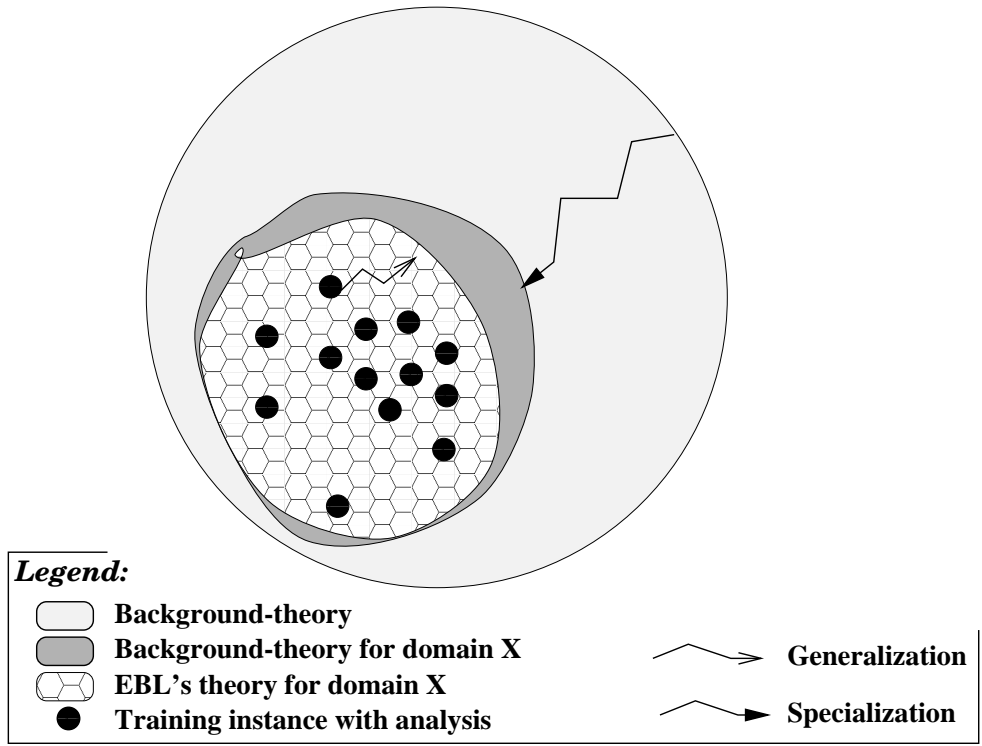}
}
\end{center}
\caption{Two views of EBL: generalization and specialization}
\label{FigEBLGoal}
\end{figure}

Thus, in contrast to inductive learning, the goal of EBL is to generalize 
the set of examples in the light of the background-theory, rather than to find
a generalization from ``scratch" in the light of some inductive-bias.
Since the background-theory at hand is assumed to be a good theory of the concept,  
there is no risk of assuming inaccurate inductive bias (as is the case in inductive 
learning). In fact, the contrast between the two paradigms becomes clearer
if one looks at the result of learning from the point of view of the background-theory;
from this point of view, the result of EBL is a {\em specialization}\/ of the given 
background-theory to the given set of training examples, usually representing some
specific domain of application of that background-theory. The goal of EBL with this
specialization is to deal more {\em efficiently}\/ with future examples, that are 
similar in their features and distribution to the given set of training instances of 
the target concept. 
Thus, the background-theory serves as part of EBL's prior knowledge of the target 
concept and the goal is to infer a hypothesis on the target concept that 
provides a more specialized description of the training instances
than the background-theory does. Figure~\ref{FigEBLGoal} depicts the two points
of view on EBL's kind of learning i.e. 1)~as generalization in the light of a 
background-theory, and 2)~as specialization of a background-theory to 
a given domain\footnote{EBL in fact 
assumes that there is a theory for the given domain, which is more specialized than
and consistent with the background-theory. One could refer to this theory, a theory
specialized for the domain at hand, with the term ``domain-theory". EBL's goal in
learning is then to learn this ``domain-theory" given a finite set of examples from the
domain, that are analyzed by the background-theory.
To avoid confusion around terminology, we refrain from using the term domain-theory here.
}.

A most vivid summary of the difference between inductive learning and EBL
is given in~\cite{Mitchell97} page~334:{\sl ``Purely inductive learning 
methods formulate general hypotheses by finding empirical regularities
over the training examples. Purely analytical methods use prior knowledge
to derive general hypotheses deductively"}.

For specializing the background-theory, EBL resorts to analyzing
the given explanations of the training instances. This analysis
involves the extraction of the set of features (from the instance-representation
which is a tuple of features) that is essential and sufficient for classifying each instance.
For example, if the instance of a day contains features that do not
influence the concept Rainy-day (e.g. year), then EBL's conclusion
is to ``exclude" (i.e. mask) these features for this target concept. 
Besides this specialization step, EBL generalizes over the
values that the features may take by computing for each explanation
(proof tree) the weakest conditions under which the proof still holds.
In other words, EBL transforms an explanation of many derivation-steps
into a complex rule (also called macro-rule) which has the form: 
{\sl generalized-instance implies target-concept}.
The {\sl generalized-instance}\/ is the most general form of
the instance, according to the background-theory, for which the 
explanation (proof) still holds.

Since EBL does not aim at learning a hypothesis in the absence
of one, but rather to improve the performance of an existing background-theory,
one often tends to underestimate EBL's contribution. But consider for 
example the game of chess, which has a perfectly defined background-theory which 
consists of all rules of the game. It is clear that good chess players
do not try all possibilities in their heads before they make a move.
Instead, they develop some expertise through experience under the
full awareness of the rules of the game. These good players
seem to develop a more specialized and operational ``theory"  of 
chess that enables them to arrive faster at better results. EBL seems 
to provide part of the computational explanation of phenomena such as
human learning from explanations for specialization and improving efficiency.
Here it is worthwhile emphasizing that the term ``efficiency" does not necessarily
mean speed-up. Efficiency can mean an improvement in any aspect of a
system's performance such as time, space or overgeneration. 
There are no limitations in EBL that prevent employing it for improving
any desired performance measure.

Finally, the EBL paradigm makes two major assumptions~\cite{Mitchell97}: 
1)~the background-theory is correct (i.e. ``sound") and complete (i.e. has 
full coverage) with respect to the instance space and the target-concept at hand,
2)~The training material does not contain any errors or inconsistencies.
In cases where the background-theory is not perfect (i.e.
incorrect and incomplete), Mitchell~\cite{Mitchell97} suggests to combine EBL 
with inductive learning. 
%
\subsubsection{EBL: specification}
The specification of  the general scheme for an EBL algorithm consists of four 
preconditions and one postcondition. The preconditions are defined as follows:
\begin{description}
\item [Background theory:]
                        A description language for the task at hand together with
                        rules, assertions and facts about the task. 
                        For instance, in natural language parsing, 
                        the description languages used in the background-theories are 
                        usually grammars e.g. CFGs. Facts about the task are provided
                        by a human annotator who is responsible for providing the 
                        right non-ambiguous descriptions. 
\item [Training examples:]
			A history of explicit explanations that are
                        given by the background-theory to example instances of the
                        concept being learned.
                        In parsing, for example, this is a tree-bank annotated under a CFG,
                        i.e. a derivation or a parse-tree constitutes an explanation for
                        why a sequence of words is a sentence in the language of the
                        (background theory) CFG.
\item [Target concept:] 
                        A formal description, {\em in terms of the alphabet of the 
                        background-theory},
                        of the domain and the range of the function that is to-be-learned
                        (this also defines the instance-scheme). 
                        For instance, when given a tree-bank annotated under a CFG 
                        \mbox{$G$ = \CFG},
                        a target concept can be the notion of ``constituency"; this concept
                        is defined by $V^{+}$ as domain and by \VN$\cup\{NOT\}$ as range,
                        where NOT denotes sequences that are not constituents.
                        As an example, the instance-scheme could be in the
                        form: XP($x_{1}\cdots x_{m}$), $m\geq~1$, expressing a
                        CFG rule $\RuleM{XP}{x_{1}\cdots x_{m}}$ 
                        (not necessarily a member in $\RulesN$).
\item [Operationality criterion:]
                       A requirement on the form of the target concept (i.e. its domain 
                       and range), in order to render the learned hypothesis ``operational"
                       with respect to some measure.
                       In parsing, this can be a formal requirement on any CFG rule that
                       can be learned. For example, one could set an operationality criterion 
                       that limits the length of the right hand side of CFG rules, in order
                       to limit the size of the CFG that is learned.
                       Or, one could demand that the CFG rule be Right-Linear if one
                       intends to parse using an FSM.
\end{description}
The postcondition is:
\begin{description}
\item [Postcondition:] To {\sl find a generalization of the instances of 
the target concept given in the training-examples, that satisfies 
the operationality criterion and the background theory. 
In our example on parsing, the generalization consists of CFG rules (or subtrees of the 
tree-bank trees, if one maintains also an internal structure for these rules) that satisfy 
the operationality criteria.}
\end{description}

The operationality criteria represent in fact extra bias in order to
render the learned knowledge, i.e. the target concept, {\em operational}. 
These are requirements on the form of the target-concept, which may originate
in part from the background-theory and in part from knowledge of the
``machinery", which will be employed for exploiting the learned target-concept.
For example, if one is planning to run the learned knowledge on a Finite-State
Machine (FSM) then it is worth considering an operationality criterion stating that
the target-concept must be a regular expression.
\subsubsection{EBL: short literature overview}
Historically speaking, EBL~\cite{DeJong81,DeJong+Mooney86,MitchellEtAl86} 
is the name of a unifying framework for various methods that learn 
from explanations of examples of a certain concept using a background-theory. 
In most existing literature, the main goal of EBL is faster 
recognition of concepts than the background-theory;
EBL learns ``shortcuts" in computation (called macro-operators or ``chunks"), 
or directives for changing the thread of computation. EBL stores the learned 
macro-operators in the form of partial-explanations to previously seen 
input instances, in order to apply them in the future to ``similar" 
input instances (in EBL, also ``similarity" is assumed to be provided by 
the background-theory). However, as mentioned above, Speedup Learning is in fact 
only one area where EBL can be applied. 

Past experience in Machine Learning cast doubt on the feasibility of improving 
performance by using EBL~\cite{Minton}. In his paper, Minton explains that EBL 
does not guarantee better performance, since the cost of applying the learned 
knowledge might outweigh the gain since EBL has no mechanism 
for measuring the utility of the learned knowledge. For this reason, Minton discusses 
a formula for computing the utility of knowledge during learning.
Generally speaking, this formula is not part of EBL; 
it is an extension to the EBL scheme by e.g. statistical inference over large sets of 
training examples.

For an overview of the literature on EBL the reader is referred
to chapter~4 of \cite{ShavlikD} and chapter~11 of \cite{Mitchell97}.
For a formal framework and discussion of Speedup Learning the reader is
referred to~\cite{TadNat96}. For a study of the relation between
EBL and partial-evaluation the reader is referred to \cite{Harmelen+Bundy}.
%

\section{Thesis goals and overview}
\label{ThisThesis}
The focus of this thesis is on the ambiguity problem in natural languages and 
their grammars. The primary goal is to present efficient solutions for 
the problem of ambiguity resolution in natural language parsing. 
The starting point for this study is that the solutions must be algorithmic and general,
do not necessitate human intervention except for supplying the input, 
and are amenable to empirical assessment. The main vehicles that carry the present 
solutions are Machine Learning and Probability Theory.

As in various preceding work, the task of natural language parsing is conceptually divided 
in two modules:
1)~the {\em parser}\/ which is the module that generates the linguistic analyses that
are associated with the input (i.e. the parse-space), and 2)~the {\em disambiguator}\/ which is 
the module that selects one preferred analysis from among the many that the parser generates.
The parser is based on a grammar, e.g. a CFG or a TSG (not necessarily a competence grammar). 
The disambiguator is based on a probabilistic enrichment of the relations (e.g. 
subsentential-forms, partial-derivations, partial-trees) that can be expressed by the grammar 
of the parser. We refer to this probabilistic enrichment with the term {\em probabilistic
model}. Despite of this conceptual devision of labour between both modules,
it is easy to see that the parser also involves ``disambiguation" because the grammar that 
underlies it already delimits the parse-space of the input even before probabilistic
disambiguation starts. Therefore, the grammar itself constitutes another location where 
ambiguity resolution can take place.

For acquiring (or training) both the grammar and the probabilistic model, this work adopts 
the Corpus-Based view that it is most expedient to employ an unambiguously and correctly 
annotated (i.e.\ by a human) sample of real-life data as training material, i.e. a tree-bank.
This offers the possibility that the resulting parser and disambiguator will answer to similar 
distributions as the sample that was used for training, which is considered representative of 
some domain of language use.\\

This thesis concentrates on the inefficiency aspect of existing performance models 
of ambiguity resolution.
The thesis defends the hypothesis that the task of ambiguity resolution in parsing limited 
domains of language use can be implemented more efficiently and more accurately 
by integrating two {\em complementary and interdependent}\/ disambiguation methods: 
\begin{enumerate}
\item Through the automatic learning of a less ambiguous grammar from a tree-bank representing
      a specific domain. The main goal here is to cash in on the measurable biases in that domain
      (which are properties of human language processing in limited domains).
      We refer to this as {\em off-line partial-disambiguation} or {\em grammar specialization
      through ambiguity reduction}.
\item Through utilizing a probabilistic model that imposes a complete ordering
      on the space of analyses and allows the selection of a most probable analysis.
      This can be called {\em on-line full-disambiguation}.
      The DOP model is the most suitable candidate for this task since it generalizes over
      most existing probabilistic models of disambiguation.
\end{enumerate}
These two manners of disambiguation are {\em complementary}: on-line disambiguation is applied
only where it is impossible to disambiguate off-line without causing undergeneration.
And they are {\em interdependent}\/ since a less ambiguous grammar, acquired off-line, can serve
as the linguistic annotation scheme for the tree-bank from which the probabilistic model 
is obtained.

Although on-line full disambiguation seems sufficient for ambiguity resolution, 
there are reasons to believe that off-line partial disambiguation by grammar
specialization is an essential element of disambiguation:
\begin{itemize}
\item None of the existing probabilistic models has the property that 
      {\em frequently occurring and more grammatical input is processed faster}.
      This is the main property accounted for by specialization by ambiguity reduction, 
      i.e. off-line disambiguation as implemented in this thesis.
\item A grammar that extremely overgenerates on some domain forms a bottleneck for
      any probabilistic model which is obtained from a tree-bank representing that domain and
      annotated under that grammar. Such an overgenerating grammar imposes an extreme strain on 
      the probabilistic model: applying probabilistic computations that are based on a 
     large table of probabilities to a large parse-space implies a very inefficient 
     parser. This situation is most evident when parsing word-graphs output by 
     a speech-recognizer using a DOP STSG as probabilistic model. 
\item When a performance model such as DOP employs a tree-bank that has an extremely 
      overgenerating underlying grammar, it acquires a probabilistic grammar 
      that has a large number of statistical parameters. 
      It is common wisdom in probabilistic modeling that the larger the number of parameters, the
      more data are necessary for estimating them from relative frequencies. 
      Therefore, large probabilistic grammars are more prone to sparse-data effects than 
      smaller ones. Extreme overgeneration is a source of data-sparseness in
      probabilistic models.
\end{itemize}
Specialization by ambiguity-reduction provides solutions to these problems and combines
very well with probabilistic models such as DOP.

The order of the subsequent chapters reflects the course of events that
led to the shift in my subject of interest from parsing-algorithms to the more general 
subject of learning how to parse efficiently. After proving that some problems of 
probabilistic disambiguation are intractable (chapter~\ref{CHComplexity}) I arrived at 
the conclusion that it is necessary to develop non-traditional methods for improving 
the efficiency of parsing (chapter~\ref{CHARS}) and to combine them together with 
optimized (relatively) traditional methods (chapter~\ref{CHOptAlg4DOP}).
Next I elaborate on the contents of these chapters.
\paragraph{Chapters~\ref{CHComplexity} and~\ref{CHOptAlg4DOP}:}
The choice for DOP for the task of on-line disambiguation implies 
a time- and space-bottleneck, because it employs huge probabilistic STSGs and
involves complex computations.
Chapters~\ref{CHComplexity} and~\ref{CHOptAlg4DOP} study the computational aspects 
of disambiguation under the DOP model.
Chapter~\ref{CHOptAlg4DOP} presents efficient parsing and disambiguation algorithms 
that provide useful solutions to some of the problems of disambiguation under DOP.
The chapter presents optimized deterministic polynomial-time algorithms for computing the
MPD (and the total probability) of a tree, a sentence, 
an FSM (word-graph without probabilities) and an SFSM
(a word-graph as output by speech-recognizers). Moreover, it also suggests
effective heuristics for the application of DOP, and exhibits extensive experiments 
with the DOP model on various domains and for various applications.
These algorithms and heuristics have been fully implemented in the Data 
Oriented Parsing and DISambiguation\footnote{
DOPDIS currently serves as the kernel of the Speech-Understanding Environment of the
Probabilistic Natural Language Processing~\cite{DOPNWORep} in the OVIS system,
developed in the Priority Programme Language and Speech Technology of the 
Netherlands Organization for Scientific Research (NWO).
} environment (DOPDIS)~\cite{MyRANLP95}.
%

In contrast to chapter~\ref{CHOptAlg4DOP}, chapter~\ref{CHComplexity} brings the
(often negatively interpreted) news that some disambiguation problems under DOP 
are (currently and most probably in the future) not solvable in 
deterministic polynomial-time.
It supplies a study of the computational complexity of  (on-line)
probabilistic disambiguation, under SCFGs and STSGs. 
It provides proofs that the following problems are NP-Complete: the problem 
of computing the MPP of a sentence or a word-graph under STSGs, the problem of computing 
the Most Probable Sentence (MPS) of a word-graph under STSGs and SCFGs. These proofs imply 
that it is highly unlikely that efficient deterministic polynomial-time solutions 
can be found. This redirects the research on finding solutions to these problems in
non-conventional ways, e.g. off-line disambiguation as presented in chapter~\ref{CHARS}.
%
\paragraph{Chapter~\ref{CHARS}:}
For off-line disambiguation, chapter~\ref{CHARS} presents a brand new view on the 
subject of grammar specialization, embodied by the Ambiguity Reduction Specialization (ARS) 
framework. Rather than learn a grammar that has shorter derivations, as preceding
work on grammar specialization expresses its goals, the ARS framework suggests to
learn a less ambiguous grammar without jeopardizing coverage. 
Chapter~\ref{CHARS} presents the ARS 
framework and derives from it two learning algorithms for grammar specialization. 
Grammar specialization under ARS improves the efficiency of parsing and disambiguation, 
especially on more frequent and longer input. 

An interesting aspect of ARS is that it relates the following issues on parsing 
and disambiguation to each other: efficiency, partial-parsing, partial-disambiguation, 
and grammar and DOP specialization to specific domains. Grammar specialization 
in ARS is equal to reducing ambiguity only where possible (i.e. partial-disambiguation). 
The specialized-grammar in ARS is always partial to the original grammar 
(partial-parsing). Projecting a DOP STSG from the specialized tree-bank results 
in a specialized DOP model. Chapter~\ref{CHARS} elaborates each of these issues. 

A novel property of ARS is that it learns partial-parsers
that can be combined with a full parser in a complementary manner; the full parser 
is engaged in the parsing processes only where and when necessary.
The partial-parser can be implemented in various ways among which the most natural is
a Cascade of Finite State Transducers. Chapter~\ref{CHARS} discusses these issues of
partial-parsing and full-parsing in detail.
\paragraph{Chapter~\ref{CHARSImpExp}:}
The ARS learning and parsing algorithms are implemented in the {\sl Domain 
Ambiguity Reduction Specialization}\/ (DOARS) system. Chapter~\ref{CHARSImpExp}
discusses implementation issues of DOARS and exhibits an empirical study of the
ARS algorithms on various domains. DOARS is evaluated on its own and in combination
with DOPDIS on sentence and word-graph disambiguation.

\paragraph{Chapter~\ref{CHDiscussion}:}
Chapter~\ref{CHDiscussion} summarizes and discusses the general conclusions of this thesis.\\

Each of the subsequent chapters relies to some extent on the definitions and 
background information discussed in the present chapter. Nevertheless, to avoid a
large background chapter, each chapter contains its own necessary definitions and
background information. Therefore, reading the subsequent chapters can take place 
in any desired order.


%

\chapter{Complexity of Probabilistic Disambiguation}
\label{CHComplexity}
%
Probabilistic disambiguation, represented by the DOP model, constitutes one 
of the two methods of disambiguation that this thesis combines. 
A major question in applying probabilistic models is {\em whether it is possible}\/ 
and {\em how}\/ to transform optimization problems of probabilistic disambiguation 
to {\em efficient}\/ algorithms~?\/
The departure point in answering this question lies in classifying these 
problems according to the time-complexities of their solutions. A problem that
does not have solutions of some desirable time-complexity, constitutes a
source of inconvenience and demands special treatment. 

The present chapter provides a proof that some of the common problems of probabilistic 
disambiguation, under DOP and similar models, belong to a class of problems for which we
do not know whether we can devise {\em deterministic polynomial-time algorithms}. 
In fact, there is substantial evidence that the problems that belong to this class, the
{\em NP-complete class}, do not have such algorithms. For NP-complete problems, the only known 
deterministic algorithms have exponential-time complexity. This is, to say the least, 
inconvenient since exponential-time algorithms imply a serious limitation on the kinds 
and sizes of applications for which probabilistic models can be applied. \\

For obvious reasons, the problems considered in this chapter are stated in a form that 
generalizes over the case of the DOP model. All these problems involve some form
of probabilistic disambiguation under SCFG-based models~\cite{Jelinek,Black,Charniak96} 
or STSG-based models~\cite{SekineGrishman95,RENSDES}. Moreover, the results of the present 
proofs apply also to TAG-based models~\cite{Schabes92,SchabesWaters93,Resnik} and the 
proofs themselves have been adapted to prove that other similar problems of disambiguation 
are also NP-complete~\cite{GoodmanDES}. The applications that face these hard disambiguation
problems range from applications that involve parsing and interpretation of text to 
applications that involve speech-understanding and information retrieval.
\section{Motivation}
As mentioned earlier, an important facility of probabilistic disambiguation is that 
it enables the selection of a single analysis of the input. Parsing and disambiguation of some
input, e.g. under the DOP model, can take place by maximizing the 
probability of some entity (in short ``maximization-entity").
Chapter~\ref{CHOptAlg4DOP} exhibits efficient deterministic algorithms for 
some problems of disambiguation under DOP. Among these problems, there are
the problems of computing the Most Probable Derivation (MPD) and computing the 
probability of an input tree/sentence/word-graph.
These algorithms serve as useful disambiguation tools for
various tasks and in various applications. 
However, there are many other applications for which the DOP model prescribes to compute 
other entities. For example, under current DOP, 
the disambiguation of sentences can (theoretically speaking) better take place by computing 
the Most Probable Parse (MPP). 
Another example is syntactic disambiguation in speech-understanding applications where 
the Most Probable Sentence (MPS) of a word-graph, produced by the speech-recognizer, 
is the desired entity. 

In the present chapter we consider the problems of computing the MPP or the MPS
for input sentences or word-graphs under STSGs in general. We provide proofs that these 
problems are NP-complete. This means that, {\em as far as we know}, there are no deterministic 
polynomial-time algorithms for computing these entities in an exact manner.
For this reason, NP-complete problems are treated in practice as {\em intractable}\/ problems.
Surprisingly, the ``intractability" of the problem of computing the MPS of a word-graph 
carries over to models that are based on SCFGs. 

This work is not only driven by mathematical interest but also by the
desire to develop efficient algorithms for these problems. As mentioned above, such 
algorithms can be useful for various applications that demand disambiguation facilities,
e.g. speech-recognition and information retrieval. The proofs in this
chapter serve us in various ways. They save us from investing time in searching
for deterministic polynomial-time algorithms that do not exist.
They provide an explanation for the source of complexity, an insight that can
be useful in developing models that avoid the same intractability problems. 
They place our specific problems in a general class of problems, the NP-complete class, 
which has the property that if one of its members ever becomes solvable, due to unforseen
research developments, in deterministic polynomial-time, all of its members will be 
solvable in deterministic polynomial-time and the algorithms will be directly available.
And they form a license to redirect the research for solutions towards 
non-standard methods, discussed in the conclusions of this chapter.
%

The structure of this chapter is as follows. Section~\ref{CHComplBackG} provides a short overview
of the theory of intractability, its notions, and its terminology. 
Section~\ref{SecProblems} states
more formally the problems that this chapter deals with. Section~\ref{NPCProofs} provides the
proofs for each of the problems. Finally, section~\ref{CHCompConcs} discusses the conclusions
of this study and the questions encountered that are still open.

\section{Tractability and NP-completeness}
\label{CHComplBackG}
This section provides a short overview of the notions of tractability
and NP-completeness. The present discussion is not aimed at providing the reader
with a complete account of the theory of NP-completeness. Rather, the aim is to provide
the basic terminology and the references to the relevant literature. Readers
that are interested in the details and formalizations of tractability and NP-completeness
are referred to any of the existing text books on this subject 
e.g.~\cite{Johnson,Hopcroft,Papadi,DavisWeyuker,Barton}. 

The present discussion is formal only where that is necessary. It employs the terminology which
is common in text books on the subject e.g.~\cite{Johnson}. For a formalization of 
this terminology and a discussion of the limitations of  this formalization, the reader is 
referred to chapters~1~and~2 of~\cite{Johnson}. 

\paragraph{Decision problems:}
The point of focus of this section is the notion of a {\em tractable decision problem}.
Informally speaking, a problem is a pair: a generic instance, stating the formal devices 
and components involved in the problem, and a question asked in terms of the generic instance.
A {\em decision problem}\/ is a problem where the
question can have only one of two possible answers: {\em Yes}\/ or {\em No}. For example,
the well known 3SAT (3-satisfiability) problem\footnote{
The 3SAT problem is a restriction of the more general satisfiability problem SAT
which is the first problem proven to be NP-complete (known as Cook's theorem).
} is stated as follows:
\begin{description}
\item{INSTANCE:}
   A Boolean formula
   in 3-conjunctive normal form (3CNF) over the variables $u_{1},\ldots,u_{n}$. 
   We denote this {\em generic instance}\/ of 3SAT with the name INS. Moreover,
   we denote the formula of INS by the generic form: 
  \[(d_{11}\vee d_{12}\vee d_{13}) \wedge 
    (d_{21}\vee d_{22}\vee d_{23}) \wedge\cdots\wedge
    (d_{m1}\vee d_{m2}\vee d_{m3}) 
  \] 
  where $m \geq 1$ and $d_{ij}$ is a literal\footnote{
  A literal is  a Boolean variable (e.g.~$u_{k}$), or the negation of
  a Boolean variable (e.g.~$\overline{u_{k}}$).
  } over \{$u_{1},\ldots,u_{n}$\}, for all $1\leq~i~\leq~m$ and all $1\leq~j~\leq~3$. 
  This formula will also be denoted $C_{1}\wedge C_{2}\wedge\cdots\wedge C_{m}$, 
  where $C_{i}$ represents $(d_{i1}\vee~d_{i2}\vee~d_{i3})$, for all
  $1\leq~i~\leq~m$. 
\item{QUESTION:}
   Is the formula in INS {\em satisfiable}~? i.e.~is there an assignment
   of values {\bf true} or {\bf false} to the Boolean 
   variables $u_{1},\ldots,u_{n}$ such that the given formula is true~? 
\end{description}
%
%

Decision problems are particularly convenient for complexity studies
mainly because of the natural correspondence between them and the formal 
object called ``language" (usually languages and questions in terms of set-membership 
are the formal forms of decision problems).
The size or length of an instance of a decision problem is the main variable in 
any measure of the time-complexity of the algorithmic solutions to the 
problem (in case these exist). Roughly speaking, this length is measured with
respect to some {\em reasonable encoding}\/ (deterministic polynomial-time computable)
from each instance of the decision problem to a string in the corresponding language.
In order not to complicate the discussion more than necessary,
we follow common practice, as explained in~\cite{Johnson}, and assume measures of length
that are more ``natural" to the decision problem at hand (knowing that it is at most a
polynomial-time cost to transform an instance to a string in the language corresponding
to the decision problem). For 3SAT, for example, the length of an instance 
is~\mbox{$3m+(m-1)+2m$}, i.e. the number of symbols in the formula is linear in the number 
of conjuncts~$m$.

\paragraph{Tractable problems and class P:}
While Computability Theory deals with the question whether a given problem has
an algorithmic solution or not, the theory of NP-completeness deals, roughly speaking,
with the question whether the problem has a general solution that is 
computationaly attainable in practice.
%
In other words:
\begin{verse}
Is there a (deterministic) algorithmic solution, which computes the answer
to every instance of the problem and for every input to that instance, of 
length $n \geq 1$, in a number of computation steps that is proportional to 
a ``cheap" function in $n$~?
\end{verse}
The problem with defining the term ``cheap" lies in finding a borderline
between those functions that can be considered expensive  and those that can be
considered cheap. A first borderline that has been drawn by a widely accepted
thesis (Cook-Karp) is between {\em polynomials}\/ and {\em exponentials}. 
Problems for which there is a deterministic polynomial-time solution are called
{\em tractable}. Other problems for which there are only deterministic 
exponential-time solutions are called intractable.

The main motivation behind the Cook-Karp descrimination between 
polynomials and exponentials is the difference in {\em rate of growth}\/ between 
these two families. Generally speaking, exponentials tend to grow much faster than polynomials.
Strong support for the Cook-Karp thesis came from practice: 
most practically feasible (``naturally occuring") problems (in computer science 
and natural language processing) have deterministic polynomial-time solutions where 
the polynomial is  of low degree\footnote{In a polynomial $n^{j}$,  $j$ is known as 
the degree.}. Only very few problems with exponential-time solutions 
are feasible in practice; usually these problems have a small expected input length.
Moreover, the overwhelming majority of problems that are not solvable in deterministic 
polynomial-time turn out to be not feasible in practice.
For further discussions on the stability of this thesis (also in the face of massively 
parallel computers) the reader is reffered to the text books listed above, 
especially~\cite{Johnson,Barton}.
%
\paragraph{Class P and class NP-hard:}
As stated above, according to the Cook-Karp thesis, a decision problem (i.e. every one 
of its instances) that 
has a deterministic polynomial-time solution in the length of its input is considered 
{\em tractable}. All other decision problems are considered {\em intractable}.
The tractable decision problems, i.e. those that have a  polynomial-time deterministic
algorithmic solution (a so called Deterministic Turing Machine (DTM)),
are referred to with the term {\em class P} problems. All other problems, that are 
intractable, are referred to with the term NP-hard problems (see below for the reason
for this terminology).

\paragraph{Class NP:}
Interestingly, there exist problems that are solvable in polynomial-time {\em provided
that the algorithmic solution has access to an oracle, which is able to guess the right 
computation-steps without extra cost}\/ (a so called Non-deterministic Turing Machine (NDTM)). 
Note that every problem in class~P is found among these so called Non-deterministic 
Polynomial-time solvable problems, also called class~NP problems (i.e. \mbox{P $\subseteq$ NP}). 
The question is, of course,  are there more problems in class~NP than in class~P~?
This is where we enter the gray zone in the theory of NP-completeness: nobody 
yet knows the answer to this question. Most computer scientists suspect, however, 
that \mbox{P $\neq$ NP}.

\paragraph{NP-complete problems:}
Strong evidence to the hypothesis \mbox{P $\neq$ NP} is embodied by the discovery of
many practical and theoretical problems that are known to be in class NP but 
{\em for which nobody yet knows}\/ how to devise deterministic polynomial-time solutions;
these problems are in class~NP but are not known to be in class~P.
In fact, if \mbox{P$\neq $NP} is true then these problems are indeed NP-hard (and also
in NP). This set of problems is called the class of NP-complete problems. 
Thus, NP-complete problems are those that lie in
the difference between class P and class NP, if indeed \mbox{P$\neq$NP} is true.

Now, the term {\em NP-hard}\/ can be explained as denoting those problems that are 
{\em at least as hard as any problem that is in NP}\/ (the formal definition of this
relies on the notion of deterministic polynomial-time reducibility, which we discuss 
below). And the NP-complete problems are the {\em hardest}\/ among all problems 
that are in~NP.

\paragraph{Proving NP-completeness:}
To prove that a given problem $L$ is NP-hard, it is sufficient to show that
another problem that is known to be NP-complete is {\em polynomial-time reducible}\/ to $L$.
This is done by providing a polynomial-time reduction (i.e. function)  
from the NP-complete problem to problem~$L$. Such a reduction shows how every instance
of the NP-complete problem can be transformed into an instance of problem~$L$. Naturally,
the reduction must be {\em answer-preserving}, i.e. for every instance of the 
NP-complete problem and for every possible input, the instance answers {\em Yes}\/ to that 
input iff the $L$-instance (resulting from the reduction) also does answer {\em Yes}\/ to the
transformed-form of that input. Note that the {\em reducibility relation between problems}\/ 
is a transitive relation.

Thus, once we lay our hands on one NP-complete problem, we can prove other problems
to be NP-hard. Fortunately, a problem that has been proven to be NP-complete is the 
3SAT problem stated above. Therefore 3SAT can serve us in proving other problems to
be NP-hard. In addition, proving that a problem is NP-hard and that it is also in NP,
proves it to be NP-complete.

\paragraph{The NP-complete ``club":}
Note that all NP-complete problem are polynomial-time reducible to each other. 
This makes the NP-complete problems an interesting class of problems: 
either all of them can be solved in deterministic polynomial-time or none will ever be.
Discovering one NP-complete problem that has a deterministic polynomial-time solution
also implies that \mbox{P = NP}.

Currently there are very many problems that are known to be NP-complete, but none has 
been solvable in deterministic polynomial-time yet. The efforts put into the study of
these problems in order to solve them in deterministic polynomial-time have been immense
but without success. This is the main evidence strengthening the hypothesis that 
\mbox{P$\neq$NP} and that NP-complete problems are also intractable.
In the current situation, where we don't know how to (and whether we can) solve NP-complete 
problems in deterministic polynomial-time, we are left with a fact: all current
solutions to these problems (in general) are not feasible, i.e. these problems
are ``practically intractable". 
This might change in the future, but such change seems highly unlikely to happen soon.
In any event, the main motivation behind proving that a new problem is NP-complete lies 
in saving the time spent on searching for a deterministic polynomial-time solution that
most likely does not exist.
Of course, proving a problem to be NP-complete does not imply that the problem should 
be put on the shelf.  Rather, it really provides a motivation to redirect the effort 
towards other kinds of feasible approximations to that problem. 
\paragraph{Optimization problems and NP-completeness:}
In this work the focus is on optimization problems rather than decision problems.
In general, it is possible to derive from every optimization problem a decision problem 
that is (at most) as hard as the optimization problem~\cite{Johnson}. 
Therefore, it is possible to prove NP-completeness of optimization problems through 
proving the NP-completeness of the decision problems derived from them.
To derive a suitable decision problem from a given maximization/minimization problem, 
the QUESTION part of the maximization/minimization problem is stated differently: 
{\em rather than asking for the maximum/minimum, the question asks whether there is 
a value that is larger/smaller than some lower/upper bound that is supplied as an 
additional parameter to the problem}.
For example, the problem of computing the maximum probability of a sentence
of the intersection between an SCFG and an FSM, is transformed to the problem of deciding
whether the intersection between an SCFG and an FSM contains a sentence of probability
that is at least equal to~$p$, where \mbox{$0 < p \leq 1$} is an extra input 
to the decision problem that serves as a ``threshold".

\newcommand{\PROBLEM}[2]{\begin{enumerate} \item [] \hspace*{-2em}{\bf #1} #2 \end{enumerate}}
\newcommand{\MPP}{{\bf MPP}}
\newcommand{\MPPWG}{{\bf MPPWG}}
\newcommand{\MPS}{{\bf MPS}}
\newcommand{\MPSCFG}{{\bf MPS-SCFG}}

\section{Problems in probabilistic disambiguation}
\label{SecProblems}
To start, the generic devices that are involved in the problems, which
this chapter deals with, are SCFGs, STSGs and word-graphs (SFSMs and FSMs).
%
For word-graphs (SFSMs and FSMs), it is more convenient to consider a special case, 
which we refer to with the term ``sequential word-graphs", defined as follows:
\DEFINE{Sequential word-graph (SWG):}
{A sequential word-graph over the alphabet $Q$ is\linebreak
 \mbox{$Q_{1}\mul\cdots\mul Q_{m}$}, where \mbox{$Q_{i}\subseteq Q$}, for all $1\leq~i\leq~m$. 
 We denote a sequential word-graph with $Q^{m}$ if $Q_{i}~=~Q$, for all $1\leq~i\leq~m$.
}
Note that this defines a sequential word-graph in terms of cartesian products of
sets of the labels on the transitions (every $Q_{i}$ is such a set); 
transforming this notation into an FSM is rather easy to do ($Q$ is the set of transitions
where the set of states is simply $\{0,\cdots,m+1\}$ and all transitions are between states 
$i$ and $i+1$, state~$0$ is the start state and state $m+1$ is the final state).
In the sequel, we refer to an SWG with the more general term {\em word-graph}. This need
not entail any confusion, this chapter refers to SWGs only, and any statement in the proofs
that applies to SWGs automatically applies to word-graphs in general.

\subsection{The optimization problems}
Now it is time to state the optimization problems that this study concerns.
Subsequently, these problems are transformed into suitable decision problems
that will be proven to be NP-complete in section~\ref{NPCProofs}. 

The first problem which we discuss concerns computing the Most Probable Parse of
an input sentence under an STSG. This problem was put forward in~\cite{RENSMonteCarlo} 
and was discussed later 
on in~\cite{MyNEMLAP94,RENS95,MyNEMLAPBook}. These earlier efforts tried to supply algorithmic
solutions to the problem: none of the solutions turned out to be deterministic polynomial-time.
Then came the proof of NP-completeness~\cite{MyCOLING96}, which forms the basis for the proof 
given in this chapter. The problem \MPP\ is stated as follows:
\PROBLEM{Problem \MPP:}
{\begin{description}
  \item [INSTANCE:]
         An STSG $G$ and a sentence $w_{0}^{n}$.
  \item [QUESTION:]
         What is the MPP of sentence $w_{0}^{n}$ under STSG $G$~? 
 \end{description}
}
The role of the MPP in the world of applications is rather clear:
in order to derive the semantics of an input sentence, most Natural Language Processing 
systems (and the linguistic theories they are based on) assume that one needs a syntactic
representation of that sentence. Under some linguistic theories, the syntactic 
representation is a parse-tree. DOP and many other probabilistic models select
the MPP as the parse that most probably reflects the right syntactic structure of 
the input sentence. Applications in which problem \MPP\ is encountered include 
many systems for parsing and interpretation of text.

A related problem is the problem of computing the MPP for an input word-graph, rather
than a sentence, under an~STSG:
\PROBLEM{Problem \MPPWG:}
{\begin{description}
  \item [INSTANCE:]
         An STSG $G$ and a sequential word-graph $SWG$.
  \item [QUESTION:]
         What is the MPP of $SWG$ under\footnote{
         A {\em parse generated for a word-graph by some grammar}\/ is a parse generated by the
         grammar for a sentence that is accepted by the word-graph. We also say that 
         a given sentence is {\em in a word-graph}\/ iff it is a member in the language 
         of that word-graph (i.e. accepted by the word-graph)
         (see chapter~\ref{CHDOPinML}).} 
         STSG $G$~? 
 \end{description}
}
Applications of this problem are similar to the applications of problem \MPP.
In these applications, the input sentence is not a~priori known and the parser must
select the most probable of the set of parses of all sentences which the word-graph
accepts. By selecting the MPP, the parser selects a sentence of the word-graph also.
Typical applications lie in Speech Understanding, morphological analysis,
but also in parsing sequences of words {\em after PoS-tagging by a tagger that provides 
at least one (rather than exactly one) PoS-tag per word} (packed in a word-graph).

The third problem is the following:
\PROBLEM{Problem \MPS:}
{\begin{description}
  \item [INSTANCE:]
         An STSG $G$ and a sequential word-graph $SWG$.
  \item [QUESTION:]
         What is the Most Probable Sentence (MPS) in $SWG$ under STSG $G$~? 
 \end{description}
}
This problem has applications that are similar to problem~\MPPWG. In
Speech Understanding it is often argued that the language model should
select the MPS rather than the MPP of the input word-graph. 
Selecting the MPS, however, does not entail the selection of a syntactic
structure for the sentence, a necessity for interpretation of the spoken
utterance.

And finally:
\PROBLEM{Problem \MPSCFG:}
{\begin{description}
  \item [INSTANCE:]
         An {\em SCFG}\/ $G$ and a sequential word-graph $SWG$.
  \item [QUESTION:]
         What is the MPS in $SWG$ under SCFG $G$~? 
 \end{description}
}
This problem is equal to a special case of problem \MPS:
an SCFG is equal to an STSG in which the maximum depth of subtrees 
is limited to~1 (see section~\ref{SecHeuristics}).
Applications that face this problem are similar to those that face problem \MPS\
described above.
\subsection{The corresponding decision problems}
The decision problems that correspond to problems \MPP, \MPPWG, \MPS\ and \MPSCFG\
are given the same names:
\PROBLEM{Decision problem \MPP:}
{\begin{description}
  \item [INSTANCE:]
         An STSG $G$, a sentence $w_{0}^{n}$ and a real number \mbox{$0 < p\leq 1$}.
  \item [QUESTION:]
         Does STSG~$G$ generate for sentence $w_{0}^{n}$ a parse for which it assigns
         a probability greater or equal to $p$~? 
 \end{description}
}
\PROBLEM{Decision problem \MPPWG:}
{\begin{description}
  \item [INSTANCE:]
         An STSG $G$, a sequential word-graph $SWG$ and a real number \mbox{$0 < p\leq 1$}.
  \item [QUESTION:]
         Does STSG~$G$ generate for $SWG$ a parse for which it assigns
         a probability greater or equal to $p$~? 
 \end{description}
}
\PROBLEM{Decision problem \MPS:}
{\begin{description}
  \item [INSTANCE:]
         An STSG $G$, a sequential word-graph $SWG$ and a real number \mbox{$0 < p\leq 1$}.
  \item [QUESTION:]
         Does $SWG$ accept a sentence for which STSG $G$ assigns 
         a probability greater or equal to $p$~? 
 \end{description}
}
\PROBLEM{Decision problem \MPSCFG:}
{\begin{description}
  \item [INSTANCE:]
         An {\em SCFG}\/ $G$, a sequential word-graph $SWG$ and a real number 
         \mbox{$0 < p\leq 1$}.
  \item [QUESTION:]
         Does $SWG$ accept a sentence for which SCFG $G$ assigns a
         probability greater or equal to $p$~? 
 \end{description}
}
In the sequel we refer to the real value $p$ in these decision problems with the
terms ``threshold" and ``lower bound". 
%

\newcommand{\SNUM}{(\frac{1}{2n+1})}
\newcommand{\Half}{(\frac{1}{2})}
\newcommand{\Third}{(\frac{1}{3})}
\newcommand{\ovu}[0]{\overline{u}}
\newcommand{\True}{{\bf T} }
\newcommand{\False}{{\bf F} }
\newcommand{\X}{\sum_{i=1}^{n}\Half^{n_{i}}}
\newcommand{\PZERO}{[1-2\theta\X]}
\begin{figure}[bth]
\epsfxsize=15.1cm
\center{
\epsfbox{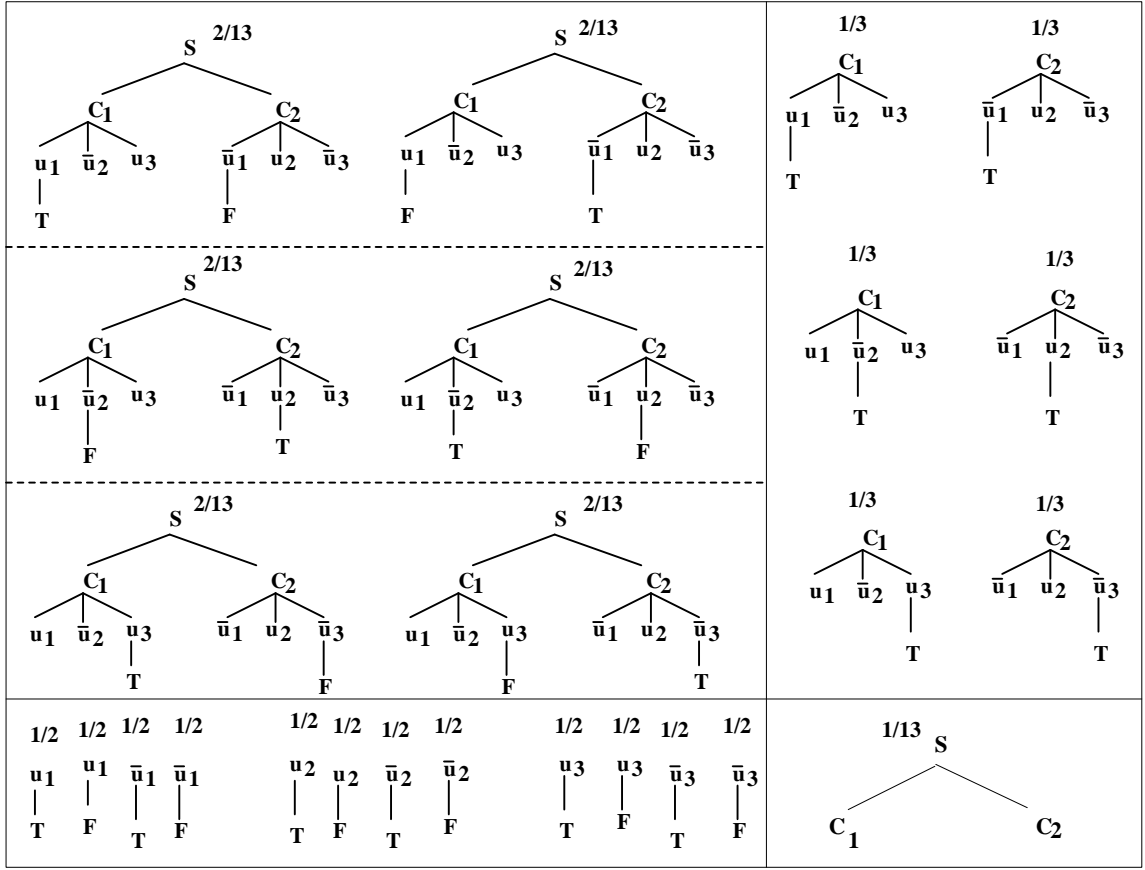}
}
\caption{The elementary-trees for the example 3SAT instance}
\label{ExampleRed}
\end{figure}
\section{NP-completeness proofs}
\label{NPCProofs}
As section~\ref{CHComplBackG} explains, 
for proving the NP-completeness of some problem it is necessary to prove 
that the problem is NP-hard and is a member of class~NP. For proving NP-hardness,
it is necessary to exhibit a suitable reduction from every instance of 3SAT to every
instance of the problem at hand. This is what we do next for each of the decision problems
listed in the preceding section. To this end, we restate the generic instance INS of 3SAT 
here:
\begin{description}
\item{INSTANCE:}
   A Boolean formula in 3-conjunctive normal form (3CNF) over the 
   variables $u_{1},\ldots,u_{n}$:
  \[(d_{11}\vee d_{12}\vee d_{13}) \wedge 
    (d_{21}\vee d_{22}\vee d_{23}) \wedge\cdots\wedge
    (d_{m1}\vee d_{m2}\vee d_{m3}) 
  \] 
  where $m \geq 1$ and $d_{ij}$ is a literal
  over \{$u_{1},\ldots,u_{n}$\}, for all $1\leq~i~\leq~m$ and all $1\leq~j~\leq~3$. 
  This formula is also denoted $C_{1}\wedge C_{2}\wedge\cdots\wedge C_{m}$, 
  where $C_{i}$ represents $(d_{i1}\vee~d_{i2}\vee~d_{i3})$, for all
  $1\leq~i~\leq~m$. 
\item{QUESTION:}
   Is the formula in INS {\em satisfiable}~? \\
\end{description}
%
%
\newtheorem{CHComplexA}{Proposition}[chapter]
\EExample{CHComplexA}
{Decision problems \MPP, \MPPWG,  \MPS\ and \MPSCFG\ are NP-complete.
}{CHCompPropA}
\subsection{A guide to the reductions}
The reductions in the next section are structured as follows. 
The first reduction is conducted {\em from 3SAT to \MPPWG}\/ and 
{\em from 3SAT to \MPS}\/ simultaneously, i.e. the same reduction serves proving
both problems to be NP-hard. Then the reductions {\em from 3SAT to \MPP}\/
and {\em from 3SAT to \MPSCFG}\/ are obtained from the preceding reduction 
by some minor changes. 
\subsection{3SAT to \MPPWG\ and \MPS\ simultaneously}
\label{Complex}
In the following, a reduction is devised which proves
that both \MPPWG\ and \MPS\ are NP-hard. For convenience, the discussion 
concentrates on problem \MPPWG, but also explains why the same reduction 
is suitable also for \mbox{\MPS.}

The reduction from the 3SAT instance INS to an \MPPWG\ instance
must construct an STSG, a word-graph and a threshold value
in deterministic polynomial-time.
Moreover, the answers to the \MPPWG\ instance must correspond exactly to the
answers to INS.
The presentation of the reduction shall be accompanied
by an example of the following 3SAT instance~\cite{Barton}:
\[ (u_{1}\vee \overline{u}_{2}\vee u_{3}) 
\wedge (\overline{u}_{1}\vee u_{2}\vee \overline{u}_{3}), \]
where $u_{1}$, $u_{2}$ and $u_{3}$ are Boolean variables.

Note that a 3SAT instance is satisfiable iff at least one of the
literals in each conjunct is assigned the value {\em true}. Implicit in
this, but crucial, the different occurrences of the literals of the
same variable must be assigned values {\em consistently}. 
These two observations constitute the basis of the reduction. The reduction 
must capture these two ``satisfiability-requirements" of INS in the problem-instances
that it constructs. For example, for MPPWG, we will construct an STSG and a WG.
The WG will be $WG~=~\{\True,~\False\}^{{\bf 3m}}$, where $3m$ is the number of
literals in the formula of INS. The STSG will be constructed such that it has 
two kinds of derivations for every path in WG that constitutes a solution for INS
(if there is such a solution): one kind of derivations takes care of the consistent
assignment of truth values, and the other takes care of assigning the value {\em true}\/
for exactly one literal in every conjunct. The derivations will have such probabilities 
that will enable us to know whether a path in WG is a solution for INS by inspecting 
the probability of that path; the probability of a path in WG will tell us whether 
the STSG derives that path by {\em enough derivations of each kind}\/ in order for that 
path to be a solution for INS.
\subsubsection{The reduction:}The reduction constructs an STSG and a word-graph. 
The STSG has start-symbol labeled $S$, two terminals represented
by \True and \False, non-terminals which include (beside $S$) all $C_{k}$,
for \mbox{$1\leq k \leq m$}, and both literals of each Boolean variable of the
formula of INS. The set of elementary-trees and probability function
and the word-graph are constructed as follows:
\begin{enumerate}
\item The reduction constructs for each Boolean variable $u_{i}$, $1\leq~i~\leq~n$,
  two elementary-trees that correspond to assigning the values {\em true}\/ and {\em false}\/ 
  to $u_{i}$ {\em consistently}\/ through the whole formula. 
 Each of these elementary-trees has root
 $S$, with children $C_{k}$, $1\leq~k~\leq~m$, in the same order as
 they appear in the formula of INS; subsequently the children of 
 $C_{k}$ are the non-terminals that correspond to its three 
 disjuncts $d_{k1}$,~ $d_{k2}$ and $d_{k3}$. And finally, the assignment
 of {\em true}\/ ({\em false}) to $u_{i}$ is modeled by creating a child terminal
 \True (resp. \False) to each non-terminal $u_{i}$ and \False (resp. \True)
 to each $\overline{u}_{i}$.
 The two elementary-trees for $u_{1}$, of our example, 
 are shown in the top left corner of figure~\ref{ExampleRed}.
%
\item 
   The reduction constructs three elementary-trees for each conjunct $C_{k}$.
   The three elementary-trees for conjunct $C_{k}$ have the same internal
   structure: root $C_{k}$, with three children that correspond to
   the disjuncts $d_{k1}$,~ $d_{k2}$ and $d_{k3}$. 
   In each of these elementary-trees exactly one of the disjuncts
   has as a child the terminal \True; in each of them this is a different 
   one.
   Each of these elementary-trees corresponds to the conjunct where 
   one of the three possible literals is assigned the value \True.
   For the elementary-trees of our example see the top right corner 
   of figure~\ref{ExampleRed}. 
%
\item The reduction constructs for each of the literals of each 
      variable $u_{i}$ two elementary-trees 
    where the literal is assigned in one case \True and in the other \False.
   Figure~\ref{ExampleRed} shows these elementary-trees for variable $u_{1}$ 
   in the bottom left corner.
\item The reduction constructs one elementary-tree that has root
     $S$ with children $C_{k}$, $1\leq~k~\leq~m$, in the same order as
     these appear in the formula of INS (see the bottom right corner 
     of figure~\ref{ExampleRed}).
\item The reduction assigns probabilities to the elementary-trees that were constructed
      by the preceding steps.
      The probabilities of the elementary-trees that have the same root
      non-terminal must sum up to~$1$. 
      The probability of an elementary-tree with root $S$ 
      that was constructed in step~1 of this reduction is
      a value $p_{i}$, $1\leq~i\leq~n$, 
      where $u_{i}$ is the only variable of which
      the literals in the elementary-tree at hand are lexicalized
      (i.e. have terminal children).
      Let $n_{i}$ denote the number of occurrences of both literals 
      of variable $u_{i}$ 
      in the formula of INS. Then $p_{i}$~=~$\theta~\Half^{n_{i}}$, for some
      real $\theta$ that has to fulfill some conditions which will 
      be derived next.
      The probability of the tree rooted with $S$ and constructed at
      step~4 of this reduction must then be
      $p_{0}~=~[1~-~2\sum_{i=1}^{n}p_{i}]$. 
      The probability of the elementary-trees of root $C_{k}$ 
      (step 2) is $\Third$, and
      of root $u_{i}$ or $\overline{u}_{i}$ (step 3) is $\Half$. 
      For our example some suitable probabilities are shown 
      in figure~\ref{ExampleRed}. 
\end{enumerate}
      Let $Q$ denote a threshold probability that shall be derived below.
      The \MPPWG\ (\MPS) instance produced by this reduction is:
       \begin{description}
         \item [INSTANCE:] The STSG produced by the above reduction (probabilities are
              derived below), the word-graph $WG~=~\{\True,~\False\}^{{\bf 3m}}$,
              and a probability value~$Q$ (also derived below).
         \item [QUESTION:] 
         Does this STSG generate for the word-graph \mbox{$WG = \{\True,~\False\}^{{\bf 3m}}$}
         a parse (resp. a sentence) for which it assigns a probability greater than
         or equal to $Q$~? 
       \end{description}
%
\paragraph{Deriving the probabilities and the threshold:}The parses generated 
by the constructed STSG 
differ only in the sentences on their frontiers.
Therefore, if a sentence is generated by this STSG then {\em it has exactly 
one parse}. This justifies the choice to reduce 3SAT to \MPPWG\ and \MPS\ simultaneously.

By inspecting the STSG resulting from the reduction,
one can recognize two types of derivations in this STSG:
\begin{enumerate}
\item The first type corresponds to substituting for
   a substitution-site (i.e literal) of any of the $2n$ elementary-trees constructed in 
   step 1 of the reduction.
   This type of derivation corresponds to assigning values 
   to all literals of some variable $u_{i}$ in {\em a consistent manner}.
   For all \mbox{$1\leq i\leq n$} the probability of a derivation of this
   type is
 \[ p_{i}\Half^{3m-n_{i}}~=~\theta\Half^{3m}\]
\item The second type of derivation corresponds to 
  substituting the elementary-trees that $C_{k}$ as root in 
    \Rule{$S$}{$C_{1}\ldots~C_{m}$}, and subsequently substituting in the 
  substitution-sites that correspond to literals.
  This type of derivation corresponds to assigning to at least one literal 
  in each conjunct the value {\em true}.
  The probability of any such derivation is
  \[p_{0}\Half^{2m}\Third^{m}~=~\PZERO\Half^{2m}\Third^{m}\]
\end{enumerate}

Now we derive both the threshold $Q$ and the parameter $\theta$.
Any parse (or sentence) that fulfills both the ``consistency of 
assignment" requirements and the requirement that each conjunct has
at least one literal with child \True, must be generated by $n$ derivations 
of the first type and at least one derivation of the second type. 
Note that a parse can never be generated by more than $n$ derivations 
of the first type. Thus the threshold $Q$ must be set at:
\[ Q = n\theta\Half^{3m}~+~\PZERO\Half^{2m}\Third^{m} \]
However, $\theta$ must fulfill some requirements for our reduction to 
be acceptable:
\begin{enumerate}
\item For all i: \mbox{$0 < p_{i} < 1$}. This means that
      for $1\leq~i\leq~n$: $0~<~\theta\Half^{n_{i}}~<~1$,
      and \mbox{$0 < p_{0} < 1$}. However, the last
      requirement on $p_{0}$ implies that 
      \[ 0~<~2\theta\sum_{i=1}^{n}\Half^{n_{i}}~<~1,\] 
      which is a stronger
      requirement than the other $n$ requirements.
      This requirement can also be stated as
      follows:   \mbox{$0 < \theta < \frac{1}{2\X}$}.
\item Since we want to be able to know whether a parse is generated by
      a second type derivation only by looking at the probability of the
      parse, the probability of a second type derivation must be 
      distinguishable from first type derivations.
      Moreover, if a parse is generated by more than one derivation of the
      second type, we do not want the sum
      of the probabilities of these derivations to be
      mistaken for one (or more) first type derivation(s).
      For any parse, there are at most $3^{m}$ second type derivations
      (e.g. the sentence \True$\dots$\True). Therefore we require that:

 \[3^{m}\PZERO\Half^{2m}\Third^{m}~<~\theta\Half^{3m}\] 
 
   Which is equal to demanding that:  $\theta~>~\frac{1}{2\X~+~\Half^{m}}$.
\item For the resulting STSG to be a probabilistic model,
      the ``probabilities" of parses and sentences must be  
      in the interval~$(0,1]$. 
      This is taken care of by demanding that the sum of the
      probabilities of elementary-trees that have the same root
      non-terminal is 1, and by the definition of the 
      derivation's probability, the parse's probability, and the sentence's
      probability.
\end{enumerate}
\paragraph{Existence of $\theta$:}
There exists a $\theta$ that fulfills all these requirements 
because the lower bound
 $\frac{1}{2\X~+~\Half^{m}}$\/  is always larger 
than zero and is {\em strictly smaller}\/ than the 
upper bound~ $\frac{1}{2\X}$. 
\paragraph{Polynomiality of the reduction:}
This reduction is deterministic polynomial-time in $m$ and $n$ 
(note that \mbox{$n\leq 3m$} always). It constructs not more than 
$2n+1+3m+4n$ elementary-trees, each consisting of at most $7m+1$ nodes.
And the computation of the probabilities and the threshold 
is also conducted in deterministic polynomial-time.
%
%
\paragraph{The reduction preserves answers:}The proof that this reduction preserves
answers concerns the two possible answers Yes and No:
\begin{enumerate}
\item [Yes:] If INS's answer is Yes then there is an assignment 
to the variables that is consistent and where each conjunct has 
at least one literal assigned {\em true}.
Any possible assignment is represented by one sentence in $WG$.
A sentence which corresponds to a ``successful" assignment must 
be generated by $n$ derivations of the first type and at least 
one derivation of the second type; this is because the sentence
$w_{1}^{3m}$ fulfills $n$ consistency requirements 
(one per Boolean variable) and has at least 
one \True as $w_{3k+1}$, $w_{3k+2}$ or $w_{3k+3}$, for all $0\leq~k~<~m$.
Both this sentence and its corresponding parse have 
probability $\geq~Q$. Thus \MPPWG\ and \MPS\ also answer Yes.
\item [No:]
If INS's answer is No, then all possible assignments are 
either not consistent or result in at least one conjunct with 
three false disjuncts, or both.
The sentences (parses) that correspond to non-consistent 
assignments each have a probability that cannot result in a 
Yes answer. This is the case because such sentences have fewer than $n$
derivations of the first type, and the derivations of the second
type can never compensate for that (the requirements on $\theta$
take care of this).
For the sentences (parses) that correspond to consistent assignments, 
there is at least some $0\leq~k~<m$ such that $w_{3k+1}$, $w_{3k+2}$
and $w_{3k+3}$ are all \False.
These sentences do not have second type derivations.
Thus, there is no sentence (parse) that has a probability that can result 
in a Yes answer; the answer of \MPPWG\ and \MPS\ is NO~~~$\Box$
%
\end{enumerate}
To summarize, we have deterministic polynomial-time reductions, that preserve
answers, from 3SAT to \MPPWG\ and from 3SAT to \MPS. We conclude that 
\MPPWG\ and \MPS\ are both NP-hard problems. Now we prove NP-completeness
in order to show that these problems are {\em deterministic polynomial-time solvable 
iff \mbox{P = NP}}.

\subsubsection{NP-completeness of \MPS\ and \MPPWG}
Now we show that \MPPWG\ and \MPS\ are in NP. 
A problem is in NP if it is decidable by a non-deterministic 
Turing Machine in polynomial-time. In general, however,
it is possible to be less formal than this. It is sufficient to exhibit a 
suitable non-deterministic algorithm, which does the following: it proposes 
some entity as a solution (e.g. a parse for \MPPWG\ and a sentence for \MPS), and
then computes an answer (Yes/NO) to the question of the decision problem, on
this entity, in deterministic polynomial-time (cf.~\cite{Johnson,Barton}).
The non-deterministic part lies in guessing or proposing an entity as a solution.

For \MPPWG, a possible algorithm proposes a parse, from the set of all parses
which the STSG $G$ assigns to $WG$, and computes its probability in deterministic
polynomial-time, using the algorithms of chapter~\ref{CHOptAlg4DOP}, and verifies
whether this probability is larger or equal to the threshold~$p$.
Similarly, an algorithm for \MPS\ proposes a sentence, from those accepted by the
word-graph, and computes its probability in polynomial-time, using the algorithms
in chapter~\ref{CHOptAlg4DOP}, and then verifies whether this probability is larger 
or equal to the threshold~$p$. In total, both algorithms are non-deterministic
polynomial-time, a thing that proves that \MPPWG\ and \MPS\ are both in class~NP.\\
%

In summary, now we proved that decision problems \MPPWG\ and \MPS\ are both NP-hard 
and in~NP, therefore both are NP-complete problems. Therefore, the corresponding 
optimization problems \MPPWG\ and \MPS\ are NP-hard.
\subsection{NP-completeness of \MPP}
The NP-completeness of \MPP\ can be easily deduced from the proof in section~\ref{Complex}. 
The proof of NP-hardness of \MPP\ is based on a reduction from 3SAT to \MPP\ 
obtained from the preceding reduction by minor changes. The main idea now is to construct
a sentence and a threshold, and to adapt the STSG in such a way, that the sentence has 
many parses, each corresponding to some possible assignment of truth values to the 
literals of the 3SAT instance. As in the preceding reduction, the STSG will have at most
two kinds of derivations and suitable probabilities; again the probabilities of 
the two kinds of derivations enable inspecting whether a parse is generated by
enough derivations that it corresponds to an assignment that satisfies the 3SAT instance, i.e.
a consistent assignment that assigns the value {\em true}\/ to at least one literal in every
conjunct. 

The preceding reduction is adapted as follows. In the reduction, the terminals of 
the constructed STSG are now fresh new symbols $v_{ij}$, $1\leq~i~\leq~m$ and $1\leq~j~\leq~3$, 
instead of \True and \False; the symbols \True and \False become non-terminals in the 
STSG of the present reduction. Then, some of the elementary-trees and some of the 
probabilities, constructed by the preceding reduction, are adapted as 
follows (any entity not mentioned below remains the same):
\begin{enumerate}
 \item In each elementary-tree, constructed in step~1 or step~2 of the preceding
       reduction (with root node labeled either $S$ or $C_{k}$), the leaf node labeled 
       \True/\False, which is the child of the $j$th child (a literal) of $C_{k}$,
       now has a child labeled $v_{kj}$.
 \item Instead of each of the elementary-trees with a root labeled with a literal
       (i.e. $u_{k}$ or $\overline{u}_{k}$), constructed in step~3 of the preceding reduction,
       there are now $3m$ elementary-trees, each corresponding to adding a terminal-child 
       $v_{ij}$, $1\leq~i~\leq~m$ and $1\leq~j~\leq~3$, under the 
       (previously leaf) node labeled \True/\False. 
 \item The probability of an elementary-tree rooted by a literal (adapted in the
       preceding step) is now $\frac{1}{6m}$.
 \item The probabilities of elementary-trees rooted with $C_{k}$ do not change.
 \item The probabilities of the elementary-trees that has a root labeled $S$ are adapted
       from the previous reduction by substituting for every $\Half$ the value $\frac{1}{6m}$.  
 \item The threshold $Q$ and the requirements on $\theta$ are also updated accordingly,
       and then derived as done in the preceding reduction.
 \item The sentence, which the reduction constructs, is $v_{11}\ldots~v_{m3}$. 
 \item And the decision problem's question is {\em whether there is a parse generated by 
       the newly constructed STSG for sentence $v_{11}\ldots~v_{m3}$, that has probability 
       larger than or equal to $Q$}~?
\end{enumerate}
The proofs that this reduction is polynomial-time and answer-preserving are very similar to 
that in section~\ref{Complex}. It is easy also to prove that \MPP\ is in class~NP 
(very much similarly to \MPPWG). Therefore, the decision problem \MPP\ is NP-complete.
\subsection{NP-completeness of \MPSCFG}
The decision problem \MPS\ is NP-complete also under SCFG, i.e. \MPSCFG\ is NP-complete.
The proof is easily deducible from the proof concerning \MPS\ for STSGs. 
The reduction is a simple adaptation of the reduction for \MPS.
Every elementary-tree of the \MPS\ reduction is now simplified by ``masking" its
internal structure, thereby obtaining simple CFG productions.
Crucially, each elementary-tree results in one unique CFG production.
The probabilities are kept the same and also the threshold. The word-graph is also 
the same word-graph as in the reduction of \MPS. The present decision-problem's question
is:
 {\em does the thus created SCFG generate a sentence with probability
      $\geq~Q$, accepted by the word-graph $WG~=~\{\True,~\False\}^{{\bf 3m}}$}.

Note that for each derivation, which is possible in the STSG of problem \MPS\
there is one corresponding unique derivation in the SCFG of problem \MPSCFG. 
Moreover, there are no extra derivations. Of course,
each derivation in the SCFG of problem \MPSCFG\ generates a different parse. 
But that does not affect the probability of a sentence at all: it remains
the sum of the probabilities of all derivations that generate it in the SCFG.
The rest of the proof follows directly from section~\ref{Complex}.
Therefore, computing the \MPS\ of a word-graph for SCFGs is also NP-complete. 
%

\section{Conclusions and open questions}
\label{CHCompConcs}
We conclude that optimization problems \MPP, \MPPWG\ and \MPS\
are NP-complete. This implies that computing the maximization entities
involved in these problems (respectively MPP of a sentence, MPP of a word-graph and
MPS of a word-graph) under STSG- and STAG-based models is (as far as we know) not
possible in deterministic polynomial-time; examples of such models 
are respectively~\cite{SekineGrishman95,RENSDES}\ and~\cite{Schabes92,SchabesWaters93,Resnik}. 
In addition, we conclude that problem \MPSCFG\ is also NP-complete; note that
\MPSCFG\ concerns SCFGs, i.e. the ``shallowest" of all STSGs.
This is, to say the least, a serious inconvenience for applying probabilistic
models to Speech-Understanding and other similar applications. 
In particular, the models that
compute probabilities in the same manner as DOP does, clearly suffer from this
inconvenience, e.g. SCFG-based models~\cite{Jelinek,Black,Charniak96},
STSG-based models~\cite{SekineGrishman95,RENSDES}, and TAG-based
models~\cite{Schabes92,SchabesWaters93,Resnik}.

The proof in this chapter helps to understand why these problems are so hard to 
solve efficiently.
The fact that computing the MPS of a word-graph under SCFGs is also NP-complete implies 
that the complexity of these problems is not due to the kind of grammar underlying the
models. Rather, the main source of NP-completeness is the following common structure 
of these problems:
\begin{verse}
 Each of these problems searches for an entity that maximizes the {\em sum}\/ of 
 the probabilities of processes that are defined in terms of that entity.  
\end{verse}
For example, in problem \MPSCFG, one searches for the sentence, which maximizes the 
{\em sum}\/ of the probabilities of {\em the derivations that generate that sentence}; 
to achieve this, it is necessary to maintain for every sentence, of the (potentially) 
exponentially many sentences accepted by the word-graph, its own space of derivations 
and probability. In contrast, this is not the case, for example, in the problem of computing 
the MPD under STSGs for a sentence or even a word-graph, or in the problem of computing the 
MPP under SCFGs for a sentence or a word-graph. In the latter problems there is one
unique derivation for each entity that the problems seeks to find.

The proof in this paper is not merely a theoretical issue. It is true that an 
exponential algorithm can be a useful solution in practice, especially 
if the exponential formula is not much worse than a low degree polynomial for 
realistic sentence lengths. However, for parse-space generation, for example,
under natural language grammars (e.g. CFGs), the common exponentials are much 
worse than any low degree polynomial; in~\cite{MartinChurchPatil}, the authors
conduct a study on two corpora and conclude that the number of parses for a sentence,
under a realistic grammar, can be described by the Catalan series and in some cases by
the Fibonnachi series. The authors conclude:
\begin{quotation}
$\cdots$, many of these series grow quickly; it may be impossible to enumerate these
numbers beyond the first few values.~$\cdots$
\end{quotation}

A further complication in the case of DOP STSGs is the huge size of the grammar.
For example, the exponential $|G|~e^{n}$ and the polynomial $|G|~n^{3}$ are comparable for 
$n\leq~7$ but already at $n~=~12$ the polynomial is some 94 times faster.
If the grammar size is small and the actual comparison is between execution-time of
respectively 0.1 seconds and 0.001 seconds for actual sentence length, then
polynomiality might be of no practical importance.
But when the grammar size is large and the comparison is between 
60 seconds and 5640 seconds for a sentence of length~12 then things become different. 
For larger grammars and for longer sentences the difference can acquire ``the width of
a crater". While the polynomial gives hope to be a practical solution in the near future,
the exponential does not really warrant hopeful expectations since future applications 
will become larger and more complex. \\

Of course, by proving these problems to be NP-complete we did not solve them yet.
The proof, however, implies that for efficient solutions to these problems it is 
necessary to resort to non-standard  and non-conventional methods, as:
\begin{itemize}
 \item to model observable efficiency properties of the human linguistic system.
       This can be done by employing learning methods prior to probabilistic parsing in 
       order to reduce the {\em actual complexity} encountered in practice in a way that 
       enables {\em sufficiently efficient computation most of the time}, e.g. Ambiguity 
       Reduction Specialization as presented in chapter~\ref{CHARS},
 \item to approximate the DOP model by allowing more suitable assumptions, and by inferring
       such STSGs in which the MPP can be approximated by entities that have deterministic 
       polynomial-time solutions, e.g.~MPD, 
 \item to approximate the search space delimited by an instance of any of these problems
       through ``smart" sampling to improve on the brute-force Monte-Carlo 
       algorithm~\cite{RENSMonteCarlo,RENSDES}.
\end{itemize}  
These solutions, especially the first, might offer these problems a way out of the trap of 
intractability. However, 
it is also necessary to incorporate other crucial disambiguation sources (based on e.g.
semantics, discourse-information and other world-knowledge) that are missing in the 
existing performance models.

\chapter{Specialization by\ Ambiguity~Reduction}
\label{CHARS}
\begin{figure}[hbt]
\epsfxsize=12cm
\center{
\epsfbox{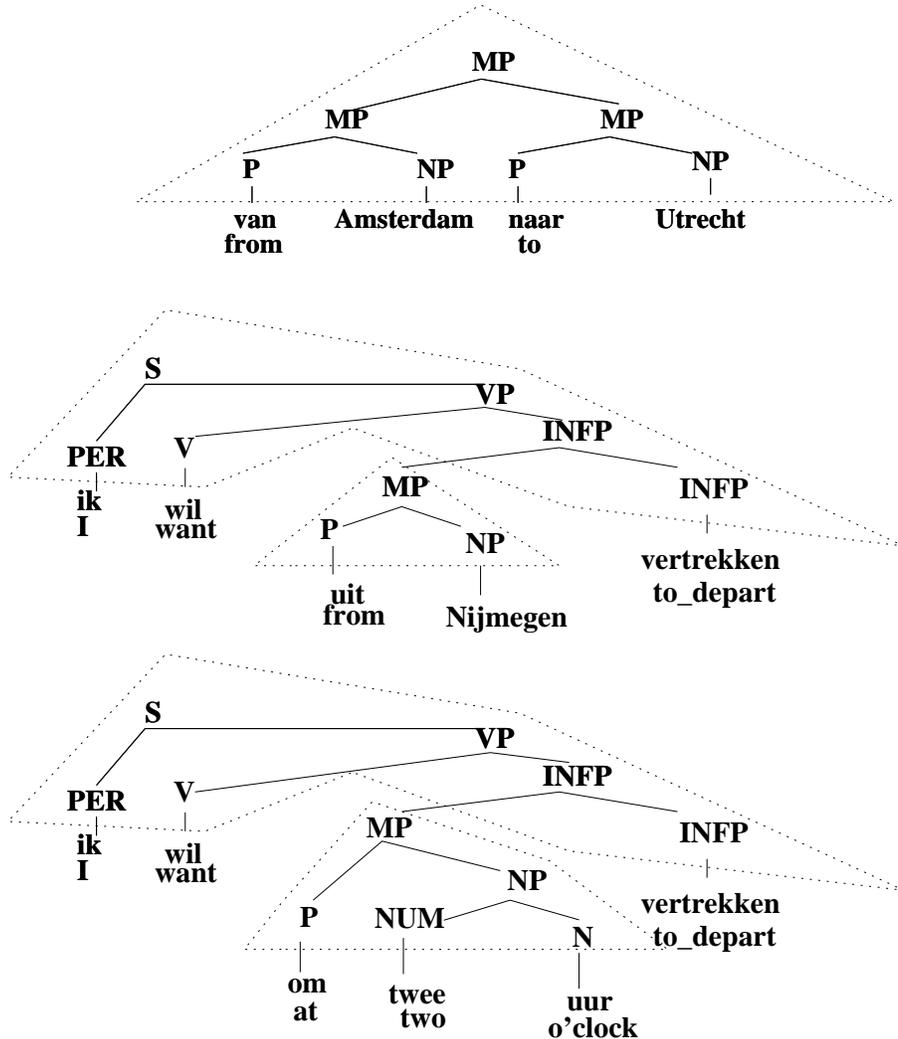}
}
\caption{A toy example tree-bank}
\label{CHGrSpuone}
\end{figure}

\begin{figure}[hbt]
\epsfxsize=12cm
\center{
{\epsfbox{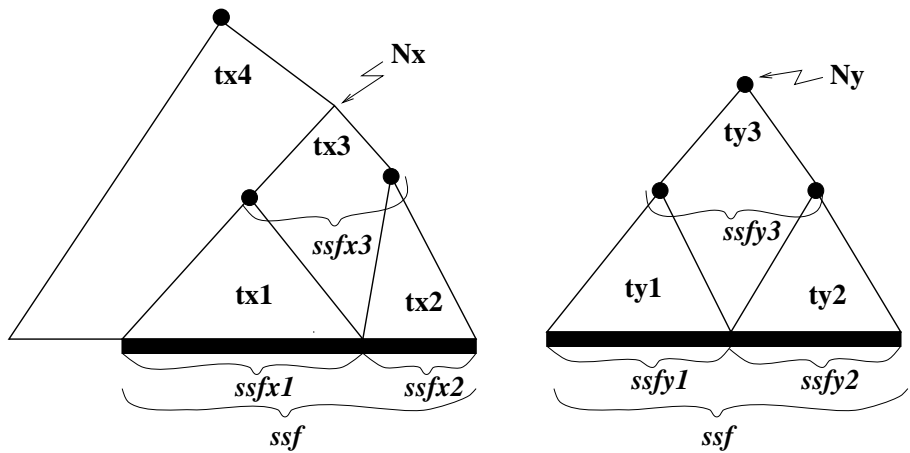}}
}
\caption{A compositional SSF}
\label{CHGrSputhree}
\end{figure}

%
\newtheorem{exampleA}{Example}[chapter]
\newcommand{\exampleI}{
\Example{exampleA}
{In figure~\ref{CHGrSpuone}, three example trees are shown. 
 Examples of SSFs are the sequences \mbox{($P$~$NP$~$P$~$NP$)}, 
 $(from~NP~to~NP)$ and \mbox{($P$~$NP$)}, taken from the top-most tree in the figure. 
 The first of these two SSFs is also a sentential-form, since it is 
 derived from the start non-terminal of the grammar. 
 Some example SSFs in the middle tree are 
 ($P~NP$), ($V~P~NP~INFP$), ($P~NP~INFP$), ($PER~V~MP~INFP$) and ($PER~V~P~NP~INFP$).
 The last two are also sentential-forms. The sequence $NP~INFP$, for example,
 is not an SSF.   In this toy tree-bank, each SSF has only one subtree
 associated with it. The ambiguity sets of these SSFs are, therefore, singletons.

 An example ambiguity set which contains two partial-trees is seen in the classical
 linguistic example: {\tt I saw the man with the telescope}. This sequence of words is 
 an SSF with which we can associate two structures:
%

\begin{tabular}{l}
\\
{\tt S(~I VP(~VP(saw NP(the man)) PP(with the telescope)~)~)} \\
{\tt S(~I VP(~saw NP(NP(the man)~~PP(with the telescope)))~)}\\
\\
\end{tabular}

\noindent
that correspond to two different ways of attachment for 
{\tt (with the telescope)}, i.e. verb phrase vs.\/ noun phrase attachment.
}}
\newtheorem{exampleB}[exampleA]{Example}
\newcommand{\exampleII}{ 
\Example{exampleB}
{We assume only in this example that the SSFs that our algorithm is allowed to learn 
 do not contain terminals (this assumption is not inherent to the algorithm).
 Consider again the trees in figure~\ref{CHGrSpuone}. These trees are cut-up into
 the partial-trees, delimited with dotted-lines, by the present SC EBL learning 
 algorithm. 
 In the top-most tree, there is only one SSF that reduces the whole tree, 
 namely \mbox{($P$~$NP$~$P$~$NP$)}, which corresponds to ``from~Amsterdam~to~Utrecht". 
 Therefore, this tree was entirely reduced after one iteration of the algorithm.
 In the middle tree, there are two SSFs, ($P~NP$) and ($PER~V~MP~INFP$). 
 The first of these two was learned in the first iteration, resulting in reducing its
 associated subtree. And the second was learned in the second iteration. A similar
 situation holds for the tree at the bottom of the figure.
} }
\newtheorem{exampleC}[exampleA]{Example}
\newcommand{\exampleIII}{ 
\Example{exampleC}
{Figure~\ref{CHGrSputhree} shows two trees from a tree-bank. The ``bullets" $\bullet$
indicate cut nodes marked by the specialization algorithm. In the right tree, 
the SSF $ssf$ is a combination of three SSFs ($ssfy1$, $ssfy2$ and $ssfy3$);
the associated subtree in this case is\footnote{As usual $\circ$ denotes 
left most substitution.} \mbox{$ty3\circ ty1\circ ty2$}.
In contrast, in the left tree $ssf$ was not learned at all. 
Furthermore, $ssfx3$, the SSF which enables recognizing $ssf$ in a 
compositional way (using $ssfx1$ and $ssfx2$), has not been learned either. 
The subtree associated with $ssf$ in the lhs tree is \mbox{$tx3\circ tx1\circ tx2$}. 
The ambiguity set of $ssf$ misses the latter subtree. Therefore, during parsing
a sentence similar to that of the lhs tree, the specialized grammar might be unable 
to construct the lhs tree. In many cases the specialized grammar might contain many other 
partial-trees so that it might be able construct other wrong trees for that sentence.
To avoid this situation, the partial-tree under node $Nx$ should be in the specialized
grammar. This is achieved by the algorithm for completing the ambiguity sets, which
is applied in this case to $ssf$; since node $Nx$ is unmarked but node $Ny$ is marked,
then the partial-tree \mbox{$tx3\circ tx1\circ tx2$}, with $Nx$ as root,
is added to the specialized grammar.
} }

This chapter addresses the task of {\em specializing}\/ general purpose grammars, 
called Broad-Coverage Grammars (BCGs), and DOP STSGs to limited domains.
It presents a new framework for grammar specialization, called the Ambiguity Reduction 
Specialization (ARS) framework, and two different algorithms that instantiate it.


The present chapter is organized as follows. 
Section~\ref{SecMotivation} discusses the motivation behind this work.
Section~\ref{SecEBLOthers} provides an analysis of other contemporary work on BCG 
specialization and motivates the need for a new approach. 
Section~\ref{SecARSFramework} presents the ARS framework and sketches,
in most general terms, its application to BCG-specialization, DOP-specialization and 
parsing. Sections~\ref{SCEBL} and~\ref{SecSpecAlgs} instantiate the ARS framework into
two different ARS algorithms and discuss the related subjects of specializing
DOP and related parsing algorithms in the light of these ARS algorithms.
Finally, section~\ref{CHARSConcs} summarizes the results of this chapter 
and lists the unanswered questions of this study.
%
\section{Introduction}
\label{SecMotivation}
The development of linguistic grammars is a major strand of 
research in linguistics. Although most linguistic research concerns 
itself with specific isolated competence phenomena of language, 
still the developed grammars make explicit the major variables and 
factors that play a role in understanding 
language utterances. In some cases, these grammars are adapted and combined 
together into a so called Broad-Coverage Grammar (BCG). Prominent examples of 
such BCGs are the XTAG grammar~\cite{XTAG}, the CLE grammar~\cite{Alshawi92}
and the CG~system~\cite{ENGCG}. 

At the other end of the spectrum of linguistic research, one finds efforts
to exploit existing knowledge of linguistic grammars in order
to annotate large corpora. Often, these efforts involve much creativity in 
filling the gaps in linguistic grammars. These efforts result in
real-life linguistic grammars, so called ``annotation schemes".
In most cases these annotation schemes are broad-coverage, as their BCG
counterparts, in the sense that they do not encapsulate domain specifics. 

In computational and empirical linguistics, the principle subject of study 
is the ambiguity problem. In this line of research, the exploitation of BCGs and 
broad-coverage annotation schemes (in the sequel loosely BCGs) 
for constructing tree-banks enables the extraction of performance models of language that 
attach probabilities to grammatical relations (whether elementary relations such as CFG 
rules or complex ones such as partial-trees). These probabilities enable the resolution
of ambiguities by selecting a most probable analysis according to the probabilistic model.\\

The feasibility and usefulness of BCGs is of major interest to linguists as well 
as language-industry. The task of {\em specializing}\/ these BCGs to specific domains 
is the next step in the enterprise of exploiting linguistic knowledge 
for tasks that involve language. Usually, a BCG recognizes sentences or 
generates analyses that are not plausible in a given domain. The specialization 
of a BCG to a given domain amounts to reducing its redundancy (i.e. overgeneration)
with respect to that domain; this often involves restricting its descriptive power 
to fit only utterances that are from that domain. The gain from specialization of BCGs
is in improving time and space consumption, and in the case of probabilistic models,
which are based on these linguistic grammars, in minimizing the effects of data-sparseness.

Research on automatic grammar specialization has been initiated by Rayner~\cite{Rayner88} 
who incorporates manually extracted domain specifics into an EBL 
(Explanation-Based Learning) algorithm. Other 
attempts at automatic BCG~specialization using EBL followed Rayner's
with success~\cite{Samuelsson94,RaynerCarter96,Srinivas97,Neumann94}.
These works, without exception, concentrated on the speed-up of the classical form of
parsing, i.e. parse-space generation\footnote{The assignment of a set of parses (the 
parse-space) to an input utterance using a grammar.}.
 But current linguistic parsing involves more 
than mere parse-space generation. It also involves probabilistic disambiguation using
models that rely on extensive tables of probabilities of linguistic relations.
This is exactly the case for the DOP model, where probabilistic disambiguation 
is by far the main source for time and space consumption. 


This chapter presents a new framework for the automatic specialization of 
linguistic grammars. This framework, called the Ambiguity Reduction Specialization (ARS)
framework, construes grammar specialization as learning the smallest least ambiguous grammar 
which assigns to every constituent which it recognizes a parse-space which contains
all its ``correct structures"; roughly speaking, a structure is correct for some 
constituent in a given domain if it is partial to a correct structure of some sentence 
from that domain.
The latter property of the ARS framework enables a novel way of integrating the 
specialized grammar and the BCG.

The idea behind grammar specialization through ambiguity reduction 
is to exploit the statistics of a given tree-bank in order to cash in on the 
specifics of the domain which it represents. The resulting specialized grammar 
should be able to quickly span a {\em smaller (but sufficient) parse-space}\/ than 
the original grammar.
Since typical grammars span the parse-space of a sentence with little cost 
(time and space), relative to the cost of probabilistic 
disambiguation\footnote{In DOP models this is often some 1\% of the total 
cost of parsing and disambiguation -~see chapter~\ref{CHOptAlg4DOP}.},
the cost of applying the expensive probabilistic disambiguation module is 
smaller when using the specialized grammar. The net effect of specialization on the total 
parsing process can be large: both time and space costs are reduced. 

As mentioned above, in many performance models of language, 
e.g. \cite{RENSDES,Charniak96,SekineGrishman95}, the relationship between 
the two modules, the grammar and the disambiguator, is strong: the grammar
provides the basis for the stochastic relations present in the disambiguator.
By specializing a BCG through ambiguity reduction, we obtain a new, less ambiguous, 
grammatical description, which, in turn, may serve as the basis for new, smaller,
probabilistic models; these models are obtained by reannotating the tree-bank
using the specialized grammar. Therefore, successful specialization should result in a new 
space of probabilistic relations and new probability distributions, which are 
good approximations of the originals. The specialized probabilistic models should be 
smaller yet (practically) as powerful as the original.

Theoretically speaking, a very attractive property of ARS specialization, which is missing 
in Bod's DOP model~\cite{RENSDES}, is that more frequent input in a given domain 
is represented in the specialized DOP model as un-ambiguously as possible in that
domain. For the DOP model, specialization implies that the speed-up on more frequent 
input is larger than on rare input. This brings in a property described by 
Scha~\cite{Scha90,Scha92} 
(page~16) as follows: 
\begin{quotation}
{\sl
We expect that, in the present processing model, the most plausible sentences
can be analyzed with little effort, and that the analysis of rare and less
grammatical sentences takes significantly more processing time.
}
\end{quotation}
Note that grammaticality in the DOP model is strongly associated with frequency 
in the training tree-bank, and that the only interpretation of the term 
``plausible sentence" in DOP is through probability. We will refer to this 
desirable property with the name {\em the Frequency-Complexity Correlation 
Property (FCCP)}.\\

In essence, grammar specialization techniques share a common goal with
techniques for dynamic pruning of the parse 
space~\cite{RaynerCarter96,GoodmanDES}. 
However, while pruning is a useful technique, it is by no means a substitute for
good specialization methods. The problem with pruning is the main concept of
pruning itself: to prune some of the partial-analyses, it is necessary to generate them
in the first place. Apart from the fact that generating analyses and then pruning them 
is time-consuming (the more analyses to prune the larger the time-cost), pruning does not 
provide a solution for the problem of grammar redundancy.
Nevertheless, in practice pruning techniques can complement off-line specialization 
methods as~\cite{RaynerCarter96} show. We maintain this view and also argue that the 
theoretical study of how to specialize a theory to specific domains is in itself 
very interesting, let alone the fact that it is rewarding on the practical side.\\
%

Throughout this chapter, we employ the Machine Learning terminology defined in 
section~\ref{SecElemsML}. Since the EBL paradigm is a central player in this
chapter, we briefly restate it here: {\em to construct an EBL algorithm it is necessary 
to have a background-theory, a set of training-examples, a definition of the 
target-concept and a definition of the operationality criterion. The product of an 
EBL algorithm is a function that generalizes the instances of the target-concept 
that are found in the training-examples, and that satisfies the operationality 
criterion and the background-theory}. 
\section{Analysis of other work}
\label{SecEBLOthers}
As mentioned earlier, grammar specialization using EBL was introduced to 
natural language parsing by Rayner~\cite{Rayner88}.
In Rayner's work and joint work together with Samuelsson,
e.g.~\cite{RaynerSamuelsson90,SamuelssonRayner91},
and other work on combining grammar specialization with pruning 
techniques~\cite{RaynerCarter96}, the operationality criterion 
for an EBL algorithm is specified manually based on intuitions and knowledge 
of the domain at hand. The first attempt to automatically compute the operationality 
criterion in grammar specialization is due to Samuelsson~\cite{SamuelssonThesis,Samuelsson94}, 
who uses the measure of entropy for this purpose, thereby extending EBL with inductive learning 
by collecting statistics over large sets of explanations. 
A more conventional application of EBL is found in~\cite{SrinivasJoshi95}, 
where Lexicalized Tree-Adjoining Grammar (LTAG)~\cite{Joshi85} is used as
the background-theory. And an effort that involves different methods of 
generalization is due to Neumann~\cite{Neumann94} who combines manually specified 
operationality criteria 
on syntax (in line with~\cite{SamuelssonRayner91}) with generalizations implied by
Head-driven Phrase-Structure Grammar (HPSG) as the background-theory (in line 
with~\cite{SrinivasJoshi95}).

In this section, we review the different efforts on BCG specialization and
provide a short analysis of their goals, features, capabilities and shortcomings. 
The discussion starts with a short overview of each method, listing the EBL elements
that constitute it, followed by a joint analysis of these methods. 

%
%
\subsection{CLE-EBL: Rayner and Samuelsson}
The CLE-EBL~\cite{Rayner88,RaynerSamuelsson90,SamuelssonRayner91,RaynerCarter96}
referres to various grammar specialization schemes that share a common basis.
We try here to describe the common parts of these schemes.

In CLE-EBL, the goal of specialization is to trade-off coverage for speed in
parsing. To achieve this, the training tree-bank trees are ``cut"
into partial-trees and then employed for parsing new input.   
In this, CLE-EBL assumes that {\em in some contexts, some non-terminals, 
i.e. constituent-types, are much less generating than other types, and hence 
can be considered internal to the latter}. If a constituent-type generates extremely 
little compared to others, its impact on coverage should be very small.

The domain of application for CLE-EBL is the ATIS domain~\cite{HemphilEtAl90}. 
The training tree-bank for EBL consists of a manually corrected output\footnote{
In early work the CLE-EBL did not employ corrected training tree-banks, probably due
to their absence at those times. It resorted to more complex ways for directing
the EBL generalizer in obtaining a suitable training parse-tree. See~\cite{SamuelssonRayner91}.
} of the SRI Core Language Engine (CLE)~\cite{Alshawi92}. 
In many cases, the annotations of the training tree-bank are based on unification grammars 
that employ feature-structures.  
The target-concept is a constituent-category represented by various non-terminals of the grammar
(including a special category for lexicon entries).  The operationality criteria are manually 
specified. The CLE-EBL learning scheme is based on the general 
Prolog generalizer specified in e.g.~\cite{Harmelen+Bundy}. However, CLE-EBL extends 
this scheme in two ways that later became common use in most work on grammar specialization:
\begin{enumerate}
\item There can be multiple levels of target-concepts, i.e. the target-concepts form a
      hierarchy. This means that the operationality criteria refer to labeled-nodes at 
      various levels in the parse trees. This is contrast to earlier work on EBL that 
      employed operationality criteria that refer to target-concepts at the same level.
\item The learnt rules are indexed in special ``data-bases" that enable fast recognition and
      retrieval of the learnt rules.
\end{enumerate}
The operationality criteria in the CLE-EBL are schemes for identifying nodes 
in a training tree that should constitute the lhs and the rhs symbols of a learnt macro-rule.
Each of the operationality criteria consists of three parts: a grammar category that
serves a target-concept and as the lhs of the learnt rule, other grammar categories 
that serve as the rhs of the learnt rule, and conditions on when to learn such a
rule. For example, one operationality criterion that was used in this work marks 
as operational all NPs, PPs and lexical categories that are in a partial-tree that 
has a root labeled~S when~S is specified as the target-concept. This criterion results
in learning rules that have~S at their lhs and sequences consisting of symbols that are
either NPs, PPs or lexical categories at their rhs.
The various CLE-EBL operationality criteria are specified in a hierarchy of 
target-concepts such that the categories that are operational under some target-concept 
are specified lower in the hierarchy than that target-concept.

The literature on CLE-EBL contains various extensions to the general approach briefly
described here. One such extension for instance allows discriminating between recursive 
and non-recursive categories (e.g. NPs) in defining operationality criteria.  
For example, non-recursive NPs, PPs and lexical categories are operational when the 
target-concept is a recursive NP. 

It is worth noting that currently the CLE-EBL is the most extensively tested and 
widely applied grammar specialization method. Its success has inspired 
the other grammar specialization methods that followed later including the 
present work.
\subsection{Samuelsson's entropy thresholds}
Samuelsson~\cite{Samuelsson94,SamuelssonThesis} is the first to explore a 
fully data-driven function for inducing the operationality criteria. His method employs 
exactly the same setting as in the CLE-EBL: the ATIS is the domain of application 
and the CLE is the BCG. The target-concept is also the concept of constituency.
However, the main assumption here is more refined than that of the CLE-EBL: 
{\em there are constituent-types for which a rule application\footnote{A single 
derivation step starting from that constituent type.} is, in some contexts 
(e.g. preceding partial-derivation), quite easy to predict. 
These ``easy-to-predict" constituent types (in the corresponding contexts) 
are considered internal to the other types}.

Samuelsson's scheme can be summarized as follows:
\begin{enumerate}
\item Compile the training trees into a decision-tree called AND-OR tree,
    which represents them compactly. To do so, first represent every training tree 
    by an ``explicit tree" that shows at every node not only the lhs of the rule
    that is applied but the rule itself. Then store these explicit trees in the
    AND-OR tree one at a time. 

    The AND-OR tree has a special root node. An OR-node corresponds to some grammar 
    rules that have the same lhs symbol. And an AND node corresponds to the rhs 
    \mbox{$X_{1}\ldots X_{n}$} of a grammar rule \mbox{$A\rightarrow X_{1}\ldots X_{n}$}.
    We describe the construction of the AND-OR tree recursively.
    Let $AOT$ denote the current AND-OR tree and let be given an explicit-tree 
    \mbox{$rule\rightarrow sub_{1}\cdots sub_{n}$}, where $rule$ is a grammar rule and
    each $sub_{i}$ is a subtree of the explicit tree\footnote{We assume the tree-bank trees
    have the same root label $S$.}. To store this explicit tree in $AOT$, we first deal 
    with $rule$: if there is an arc emerging from the root of $AOT$ and is labeled $rule$ 
    then we follow it to arrive at an AND-node that is labeled with the rhs of $rule$;
    otherwise, we add a new arc labeled $rule$ that leads to an AND-node labeled by the rhs
    of $rule$. In any event, for the $ith$ symbol $XP_{i}$ in the rhs of $rule$
    an arc emerges (either already existing or newly added) from this AND-node 
    and leads to an OR-node $COR_{i}$ labeled by $XP_{i}$; 
    the arc is labeled by the number $i$ of that symbol in the rhs of $rule$.
    Subsequently we deal with each of $sub_{i}$ recusively from the OR-node $COR_{i}$. 
    The recursion terminates if $sub_{i}$ is a lexical rule \mbox{$A\rightarrow word$} in 
    which case a special rule identifier $lex$ is used and is added under the OR-node.
\item Define at every OR-node in the AND-OR tree a probability function as follows.
      At an OR-node $N$ for symbol $X$, the probability function assigns 
      to every arc emerging from that OR-node and that is labeled by a 
      rule $X\rightarrow\gamma$, a probability value conditioned on the OR-node $N$. 
      This probability is computed as the ratio between the frequency of 
      $X\rightarrow\gamma$ and the frequency of the nodes labeled~$X$ that 
      correspond to OR-node $N$ in the training tree-bank explicit trees. 
\item Compute the complexity (as measured by the entropy) of the choice 
      at each OR-nodes. 
      The hardest points of choice (largest complexity) in the tree are marked 
      as cut points.
      In practice, a threshold on the 
      entropy is set according to the desired coverage such that all nodes that 
      exceed that threshold are considered operational i.e. are cut nodes. 
\item Then cut up the tree-bank trees by matching each of them against the
      AND-OR tree, thereby resulting in the specialized grammar.
\end{enumerate}
In~\cite{Samuelsson94,SamuelssonThesis} this scheme\footnote{
Other probability distributions and definitions of how to calculate the entropy of
an OR-node are defined and tested in~\cite{Samuelsson94}.} 
is extended by an iterative mechanism: after each iteration, the set of OR-nodes 
is partitioned into equivalence classes each corresponding to the non-terminal symbol 
of the OR-node and some local context in the AND-OR tree. The entropies of the OR-nodes are
recomputed: the entropy of an OR-node is now the sum of the entropies of all 
OR-nodes that are together with it in the same equivalence class. And then the
procedure is repeated. The iteration stops when the change in 
the set of cut-nodes is not significant any more (according
to a predefined measure) (for detail, see the discussion on finding the cutnodes 
in~\cite{Samuelsson94}).
This iterative training procedure is not guaranteed to stop, i.e. it does not
always arrive at a preferred set of cut nodes~\cite{Samuelsson94}. 
In~\cite{Samuelsson94}, this entropy scheme 
is further augmented with manually specified limitations on the training algorithm, 
in order to achieve better results.

It is worth noting that Samuelsson's EBL is, strictly speaking, not pure EBL; 
it involves inductive learning by collecting statistics over large sets of explanations. 
As we shall see in the next section, this extension is inevitable if one is to employ
data-driven learning in the absence of background-theories that supply the 
operationality criteria and the generalization power. But more importantly, it is 
inevitable because it addresses {\em statistical properties of samples}: properties
that are not addressed by linguistic theories. 
\subsection{LTAG-EBL: Srinivas and Joshi}
The application of EBL in the context of the LTAG theory is probably the
most according-to-recipe EBL algorithm. It relies totally on the strong  background-theory, 
which it assumes, i.e. LTAG theory; the generalizations it achieves and 
also the indexing scheme are directly taken from the LTAG representation. 
The target-concept here is the notion of a sentence rather than the more general concept
of a constituent-type. The explanations are provided by LTAG (manually selected)
as LTAG-derivations to sentences in the training tree-bank.

LTAG incorporates natural language syntax into the lexicon. Each combination of a word 
and a specific syntactic environment it might appear in is represented by a structure,
called elementary-tree, which makes explicit its necessary and sufficient arguments. 
To account for long-distance behaviour, LTAG ``factors" out recursion, i.e. adjuncts 
and modifiers, from the representation of other kinds of words. It then allows 
recursive trees, representing these modifiers and adjuncts, to ``adjoin" in prespecified 
points in the elementary trees. The LTAG theory is implemented in a system which allows
also for morphological representations, rather than only phrase-structure representations.
For more on LTAG and its properties see~\cite{Srinivas97}.

LTAG-EBL exploits these properties of LTAG as follows. Firstly, for every tree in
the training-set, its LTAG-derivation is generalized by  un-instantiating the 
morphological descriptions, i.e. features, as well as the specific words that
it incorporates. Secondly, every generalized derivation is stored indexed by 
the sequence of PoSTags which forms its frontier. Then each of these sequences of PoSTags 
is generalized by representing the recursion, due to modifiers and adjuncts, 
into a regular expression, i.e. an FSM. With every FSM there is 
a set of associated generalized parse-trees, i.e. it is a Finite State
Transducer (FST).  For parsing a new input sentence, firstly its PoSTag sequence is 
obtained from a part-of-speech tagger, then it is processed by the FSTs to
obtain generalized parses.  These generalized parses are then instantiated by the features of 
the current sentence to result in full-parses. 
\subsection{HPSG-EBL: Neumann}
The background-theory in~\cite{Neumann94} is HPSG and the domain is
Appointment-Scheduling. The operationality criteria
are of two types and are both manually specified: 
1)~syntactic criteria in the spirit of~\cite{SamuelssonRayner91}, and
2)~feature-structure uninstantiation criteria in the spirit of~\cite{SrinivasJoshi95}.

In~\cite{Neumann94}, the tree-bank is assumed to contain HPSG-trees: each tree is 
a combination of a syntactic CFG-backbone-tree (shortly bare-tree) and a feature structure 
(with the correspondences between the nodes in the bare-tree and the features
in the feature structure). The learning process is not complicated and can be
summerized (with some minor simplifications) as follows:
\begin{enumerate}
\item The CFG-backbone-trees (shortly bare-trees) of the training HPSG-trees 
      are cut into partial-trees at some manually predefined syntactic categories,
\item The feature structures associated with every tree are then ``taken apart"
      such that with every bare partial-tree (resulting from the preceding step)
      the right feature structure remains associated,  
\item The features in the fearture structures of a partial-tree 
      that correspond to substitution sites at the frontier of the bare partial-tree
      are {\em uninstantiated}\/ by introducing suitable variables instead of their
      values.
\item The resulting partial-trees are stored in a discrimination-tree indexed by 
      the feature structures that correspond to the terminals on their frontiers.
      Note that partial-trees that have no terminals on their frontiers are not 
      indexed and are not used during the phase of applying the learned knowledge 
      to new input.
\end{enumerate}
During the application phase, the words of the input sentence receive the corresponding
feature structure entries in the lexicon (after morphological analysis) and the resulting 
sequence is used as the index of the input sentence. For every part of the sentence the
discrimination-tree is tarversed to retrieve the associated partial-trees. Subsequently,
the feature structures of the retrieved partial-trees are instantiated by (and unified with) 
the values found in the lexicon-entries of the words. The retrieved and instantiated 
partial-trees are stored in a Earley-type chart. However, rather than employing the 
Earley-parser, Neumann employs a {\em deterministic}\/ Earely-type parser that prefers 
larger retrieved partial-trees to smaller ones; the algorithm neither backtracks nor 
computes the whole parse-space of the input sentence. He also describes a way to combine 
the specialized parser obtained by EBL with the original HPSG parser such that the 
retrieved partial-trees are completed by the HPSG-parser. However, this manner of combining 
the two parsers is based on a best-first search heuristic, 
i.e.\ it is not based on a quality of the specialization algorithm that enables complementary 
roles for the two parsers.

\subsection{Analysis}
From the overview given above we see that there are currently three types
of EBL from tree-banks: 1)~manually specified generalization rules 
based on knowledge of the specific domain, 2)~manually specified generalization
rules that are theory specific, and 3)~automatically inferred generalization rules 
using statistics. Strictly speaking, the third type is in fact a combination of EBL 
together with inductive learning. 

The following points are common to the work described above:
\begin{itemize}
\item Either the result of parsing is still ambiguous or additional intuitive
      heuristics (e.g. prefer largest partial-trees) are employed to
      disambiguate it.
      Thus, all these works conceptually divide the parsing system into a parser 
      which generates the possible tree-space for the input, and a disambiguator 
      which selects the preferred tree (in these works the disambiguator is 
      embodied by the heuristics).
\item Speed-up of parse-space generation is the goal of learning. 
\item None of these methods takes the cost of full 
      probabilistic disambiguation into account {\em during learning}.
\item None of these methods
    is able to integrate the specialized grammar, a partial-parser,
      with the original BCG parser in a manner in which the failure of the partial-parser 
      does not always imply full recomputation of the parse-space for the input sentence.
      If the partial-parser fails to parse some input, the BCG parser always has to 
      do the whole job from scratch, accumulating the processing times of the two parsers.
\end{itemize}
In addition to these common features, each of the four efforts
has its own specific strengths and weaknesses. 
The specific strengths and weaknesses of LTAG-EBL are:
\begin{description}
\item [Strengths:]
          Relies on a strong linguistic theory that offers elegant generalization 
          capabilities.
          Features a simple and fast learning algorithm.
\item [Weaknesses:]
       It is specific to the LTAG theory and the XTAG system. And 
       currently it is limited to learning on the sentential level only; therefore,
       the coverage of the resulting specialized grammars is usually too limited.
\end{description}
For CLE-EBL the situation is the following:
\begin{description}
\item [Strengths:]
     It features a fast learning algorithm. Relies on a strong linguistic theory.
     The learning algorithms and operationality criteria are well-tested
          and are (claimed to be) general.
\item [Weaknesses:]
      It is based on manual specification which depends on the intuitions of linguists.
     It does not provide a {\em direct}\footnote{
      A mechanism is indirect in the sense that it is based on the paradigm of
      ``generate and test", i.e. {\em learn it, test it, accept it or else discard 
      and loop again}. In an indirect mechanism, to discover whether a newly introduced 
      operationality criterion conserves the tree-language coverage, it is necessary to test 
      its effect on data after learning and decide only after learning whether to employ it
      or to repeat the learning again. A direct mechanism is built into the training algorithm
      and provides a guarantee that the results of learning will be satisfying (under
      suitable conditions on the kind of training tree-bank).}
      mechanism to control {\em tree-language coverage},
      where ``tree-language coverage" is, roughly speaking, a measure of coverage 
      of a parser related to well known {\em recall} measure; 
      this measure indicates the expected percentage of sentences for
      which the parser is able to assign a parse-space that contains the {\em correct}\/
      analysis (rather than any analysis). 
      We define the term ``tree-language coverage" more precisely in the next
      section but for now it is sufficient to say that the CLE-EBL does not provide 
      a mechanism that guarantees (to any desired extent) covering the {\em correct}\/
      partial-trees that a constituent might be associated with, in the domain.
      Note that if a category is not operational in some contexts, other 
      categories, that depend on it, may not be able to produce some of the parses 
      that are necessary for the coverage of some sentence. As a consequence, this
      sentence would get a {\em non-empty}\/ parse-space that does not contain the right 
      parse. 
\end{description}
Neumann's HPSG-EBL has the following strengths and weaknesses:
\begin{description}
\item [Strengths:]
          Relies on a strong linguistic theory HPSG.
          Features a simple and fast learning algorithm.
          Combines various generalization capabilities inspired by the other three approaches.
\item [Weaknesses:]
  It is based on manual specification which depends on good knowledge of the domain
      and on the intuitions of linguists.
     It does not offer a direct mechanism for controlling the
      tree-language coverage of the learnt partial-parser.
\end{description}
As for the novel entropy-thresholds EBL~\cite{Samuelsson94}:
\begin{description}
\item [Strengths:]
         Generally applicable due to automatic estimation of the operationality 
         criteria by means of information theoretic measures that rely on statistics
         over large bodies of training-explanations.
    Allows control of the desired balance between coverage and efficiency.
\item [Weaknesses:]
      It does not offer a direct mechanism for controlling the tree-language coverage 
       of the learnt partial-parser.
\end{description}
To summarize, none of the above mentioned learning methods takes into account
the cost of full disambiguation during learning. The speed~up which these methods have
addressed, is in fact partial (or complementary in the best case) to the speed~up we aim at,
namely the speed-up of the sum of parsing and disambiguation. Moreover, most of the 
above mentioned methods do not have a (direct) mechanism to control 
the tree-language coverage on real-life data, and this should be a major concern for 
grammar specialization algorithms.

In the next section we present a new framework for grammar-specialization,
which specifically deals with these two shortcomings.
In this framework, the goal of learning is a small less ambiguous grammar,
which provides a special mechanism for controlling the tree-language coverage. 
The new framework benefits from some of the insights of preceding work. 
In particular it shares with Samuelsson's method two major insights:
1)~in order to capture efficiency properties of human processing in limited domains
it is necessary to combine EBL with statistical knowledge,
and 2)~specialization is most fruitful when it reduces the ambiguity of the grammar.
However, the present framework differs from Samuelsson's framework in major issues,
especially the way it views specialization. In Samuelsson's framework, 
the task of specialization is stated as a problem of off-line filtering/pruning 
by thresholds; in contrast, the present framework states specialization 
as a constrained-optimization problem. 
%
%

\newcommand{\DOM}{{\cal D}}
\section{Ambiguity Reduction Specialization}
\label{SecARSFramework}
This section presents a theoretical framework for specializing BCGs by reducing their 
ambiguity, called the Ambiguity Reduction Specialization (ARS) framework. 
The ARS framework states the necessary and sufficient requirements for successful
BCG specialization algorithms. Actual specialization algorithms that operationalize 
the ARS framework are presented in sections~\ref{SCEBL}
and~\ref{SecSpecAlgs}.

In addition to presenting the ARS framework, this section also discusses how to exploit 
the specialization result under the ARS framework on two fronts: parsing 
(i.e. parse-space generation) and probabilistic disambiguation under DOP.
%
%
\subsection{The ARS framework}
\label{ARSframe}
As mentioned in chapter~\ref{CHDOPinML}, learning can be seen as
search in a space of hypotheses, using inductive bias (training-data
and prior knowledge) as a guide in the search. Let us consider the elements
of learning as search that underly the ARS framework.

The departure point of the ARS framework is that an ARS learning algorithm must 
have access to {\em training material}.
As training material we assume a manually annotated and corrected tree-bank, 
which represents a specific domain\footnote{Note that we assume that the tree-bank 
contains only the correct structures.}.
We also assume that the BCG is represented by the grammar underlying 
the tree-bank, thereby limiting our knowledge of the background-theory only to what 
the tree-bank contains.

Two notions are central to the ARS framework: {\em tree-language coverage}\/ and
{\em ambiguity}. We will now define the notion of {\em tree-language coverage}\/
with respect to a domain of language-use. This definition can be operationalized
only by approximation: tree-language coverage is measured on samples (e.g. tree-banks)
representing the domain of language-use. Theoretically speaking, in the limit when
the tree-bank is a good sample of the domain, the approximation approaches this
theoretical measure of tree-language coverage\footnote{Note that the term 
``tree-language coverage" is completely different from the more common term ``coverage".
While the latter usually implies that a sentence be assigned {\em any}\/
structure, the former implies that it is assigned a set containing all {\em correct}\/
structures for that sentence.}.
\DEFINE{Domain:}{A domain of language use $\DOM$ is a bag (or multi-set) of 
                pairs $\tuple{s,t}$, where $t$ is a parse-tree assigned to sentence~$s$
                by some annotation scheme (formal grammar)~$G$.
}
\DEFINE{Correct structure:}{
A partial-tree $pt$ is {\em correct}\/ for a string of symbols $sq$ under domain $\DOM$ 
iff $sq$ is identical to $pt$'s yield and 
\mbox{$\exists \tuple{s,t}\in \DOM$} such that $pt$ is a subtree of $t$.
}
%
\DEFINE{Tree-language coverage on a constituent:}{
The tree-language coverage of a grammar $G$ on a constituent $C$ with respect to some
domain $D$, is measured as the ratio between the correct structures which 
$G$ assigns to $C$ and the total number of correct structures associated 
with $C$ in domain $D$. 
}
\DEFINE{Grammar's tree-language coverage:}{
The Tree-Language Coverage (TLC) of a grammar $G$ with respect to a given domain $D$ is 
the expectation value of the tree-language coverages of $G$ on each of the constituents 
which it recognizes in $D$. The expectation value is computed on a tree-bank representing
(i.e. a sample from) domain~$D$.
}
The tree-language coverage of a grammar is a measure that is strongly related to another
measures used in the community: recall. The tree-language coverage of a grammar can be seen
as ``subtree-recall" on constituents. 

In the ARS framework, the {\em concept-definition}\/ is identical to the {\em grammar type}\/ 
of the annotation scheme underlying the training tree-bank. The {\em target-concept}\/ 
is called a {\em specialized grammar}; this is a grammar that is partial\footnote{
A grammar $G1$ is called partial with respect to another grammar $G2$ if the string-language 
(tree-language) of $G1$ is partial to the string-language (tree-language) of $G2$.
} with respect to the grammar underlying the training tree-bank.

\subsubsection{Requirements and biases}
The specialized grammar is identified by an ARS learning algorithm as the
grammar that satisfies the following requirements:
\begin{description}
\item [1. Tree-language coverage:]
     The specialized grammar must {\em provide a satisfying tree-language coverage 
     for each constituent it recognizes}. In other words, the learning
     algorithm must provide a mechanism for favoring grammars with a satisfying 
     tree-language coverage.
\item [2. Ambiguity reduction:]
     The specialized grammar must be as unambiguous as possible, taking the other 
     constraints into consideration. 
\item [3. Compactness:]
      The specialized grammar must be as small as possible, taking the other constraints 
      into consideration. 
\item [4. Recognition power:] The specialized grammar must recognize as many of the
       sentences in the domain as possible, taking the other constraints into consideration.
\end{description}
Besides these requirements, the algorithm may embody some further biases:
\begin{enumerate}
\item Any desired application and/or domain dependent biases.
\item A general objective and task-independent learning-bias.
\end{enumerate}
The first requirement is, on the one hand, the most straightforward,
and, on the other, the most tricky. The problem lies in the definition of 
tree-language coverage of a constituent. As we saw in the analysis of preceding work on
BCG-specialization, often there is no guarantee that a grammar encapsulates
(with high probability) all correct structures that might be associated
with a constituent in the given domain. This requirement states exactly 
that the specialized grammar should provide sufficient tree-language coverage i.e. 
if the specialized grammar recognizes a constituent,
it also provides (with high probability) all its correct structures. 
Note that if a grammar has a satisfying tree-language coverage, by the virtue of
the compositionality property of grammars, it is able to guarantee, to some 
extent, satisfying tree-language coverage on full sentences\footnote{Of course,
there are the famous troublesome idiomatic structures, idiomatic with respect 
to the grammar at hand, that do not abide by the compositionality assumption.
The discussion in the next sections suggests a solution for this problem.
 }. 

The second requirement states that the specialized grammar be minimally 
redundant, within the borders delimited by the other constraints.
As we shall see later on, it is possible to employ various frameworks to
define a measure of redundancy, e.g. entropy, probability. The source for 
ambiguity reduction is domain-specific bias exploited by the learning algorithm. 
Crucial in this respect:
\DEFINE{Measuring ambiguity:}{
     We measure ambiguity on a given training tree-bank which constitutes a 
     sample of a certain domain.
     Theoretically, in the limit, when the tree-bank is infinitely large, 
     the ambiguity of the grammar on that tree-bank is a realistic 
     estimation of the ambiguity of the grammar on the given domain.
}

The third requirement excludes the situation where the  
the specialized grammar is so large that the cost of parse-space generation 
cancels the gain from ambiguity reduction. The predicate ``smallest" grammar 
refers to some measure of the size of a grammar, which should be specified by 
the learning algorithm. We exemplify this in the next sections.

The fourth requirement (besides the third requiremnt) guarantees that ARS learning 
generalizes over the training tree-bank. Without this requirement, grammar specialization 
might result in learning the whole tree-bank as a specialized grammar, a clear case of 
overfitting. 

Since grammar specialization often has an application-dependent character,
the ARS framework states that beyond these three requirements, the designer
may add own biases based on prior knowledge of the domain and the application.
For example, one such bias could be a requirement on the form of the grammar 
rules, limiting the grammars to those that can be represented by some kind
of machines e.g. FSMs, LR-parsers.

Note that the four requirements stated above do not make the ARS-framework a 
complete prescription for grammar-specialization. They only state the safety 
requirements for successful specialization. For implementing the ARS-framework 
into learning algorithms, it is necessary to specify a learning paradigm, e.g.
Bayesian Learning or Minimum-Description Length (MDL), which provides the 
{\em objective}\/ principles of learning, independent of the task of
grammar-specialization. In the ARS-framework, this is incorporated in the task-independent 
learning-bias which enables generalization over the training-data by selecting a 
preferred paradigm. The preferred paradigm enables combining the other biases into 
full-fledged learning algorithms. For example, if we choose Bayesian Learning, 
the likelihood of the data as well as the prior probability provide places for 
expressing the other biases that are specified by the ARS requirements. We exemplify 
this in the next section, where we present various ARS specialization algorithms. \\
 
\subsubsection{Hypotheses-space}
Having stated the training material, the target-concept and the inductive bias, which 
guide learning, the next step is to identify explicitly the
search space (or hypothesis space) for an ARS algorithm: a space of grammars. 
In the general case, there are no limitations on the choice of this space. 
However, since the goal of ARS,
in this thesis, is the specialization of tree-bank annotations\footnote{Recall that
in ARS we assume that the relevant features of a background-theory are explicitly represented 
in the tree-bank, i.e. the tree-bank annotation is our background-theory.},
by Explanation-Based Learning, the space of grammars is determined by the expressive 
power of the BCG underlying the tree-bank. Currently, tree-bank annotations are limited 
to Phrase Structure Grammars (PSGs), i.e. have a CFG backbone. 
Therefore, the grammar space of the ARS framework is 
currently (a subset of) the {\em Tree-Substitution Grammar (TSG) space}\/ of the tree-bank;
a TSG is in the TSG-space of a tree-bank iff each of its elementary-trees 
is a subtree of trees in that tree-bank. This constraint on the 
learning space is another major reason, besides using a tree-bank, to qualify this 
learning framework as involving EBL. In contrast, we could allow some inductive 
generalization over the number of modifications in a constituent, analogous to 
LTAG-EBL~\cite{SrinivasJoshi95}, extending the learning space beyond the TSG-space,
and thus beyond the capabilities of pure~EBL.
\subsection{Parsing under ARS}
\label{ParARS}
Before we specify and implement the ARS framework into actual 
algorithms, let us first sketch how we shall use the properties of the specialized 
partial-grammar resulting from learning. Two aspects of this grammar provide a 
unique possibility for a novel method of parsing, which integrates the specialized 
grammar and the BCG in a natural way. Firstly, as the first requirement states,
if this grammar spans a non-empty space for a domain's constituent, this space is 
(with high probability) complete in the sense that it contains all correct
partial-trees that can be associated with the constituent in this domain.
And secondly, the parse-space of a constituent, according to this
grammar, is smaller than or equal to the space spanned by the BCG for the 
same constituent.

The integration of the specialized grammar and the original BCG
results in a continuous two phase parser. The specialized grammar is allowed
to parse all constituents of the input string resulting in a parse-space for every
such constituent, including the input string as a whole. As the first property
guarantees, a non-empty parse-space must be complete in the sense described above. 
Thus, for constituents that receive a non-empty space, the BCG-parser need not do anything. 
For the other possible constituents we can employ the BCG-parser to span the 
parse-space. This results in a complementary role for the BCG-parser with regard to 
the role of the specialized parser (as partial-parser): 
the BCG-parser recognizes constituents with empty space (according to the specialized grammar)
and integrates the results of the specialized parser together with its own results into 
a parse-space for the whole input string. 
And by the second property of the specialized grammar, the integrated
parser overgenerates as little as possible\footnote{The integrated parser achieves its
worst-case overgeneration when the grammar underlying the training tree-bank is ideally
tuned for the domain. In that case, the integrated parser overgenerates as the grammar
underlying the tree-bank.}.
We elaborate more on this integrated parsing algorithm in the sequel.

\subsection{Specializing DOP with ARS}
Beyond BCG specialization, ARS provides a way to specialize probabilistic models 
such as DOP. To achieve this, one should consider the partial-trees of the specialized 
grammar as atomic units. This can be achieved by identifying these partial-trees
in the trees of the tree-bank and marking them as atomic units; then DOP is allowed
to cut the tree-bank in any way which does not violate the atomicity of these
partial-trees. The resulting DOP models should be smaller than the original one and 
have less wide coverage.

However, due to the possibility of complementing the specialized grammar with the 
BCG grammar, it is possible to complement the specialized DOP model with the
original DOP model. This is achieved by parsing the input with the integrated parser.
In case the whole parse-space is determined solely by the specialized grammar
then it is sufficient to employ the specialized DOP model for 
disambiguation\footnote{
Note that the parse-space which the specialized DOP model can evaluate probabilistically
is exactly equivalent to the parse-space spanned by the specialized grammar. This is simply
because each elementary-tree of a specialized DOP STSG is either an elementary-tree of
the specialized grammar or is constructed of a combination of elementary-trees of
the specialized grammar.}.
In all other cases, where the BCG grammar complements the space, one employs the
original DOP model for disambiguation. Note that even in this last case the cost of
disambiguation is smaller than when using only the DOP model. This is simply because
the parse-space is expected to be smaller due to employing the specialized parser.

An interesting question in specializing DOP models concerns the faithfulness of the
probabilistic distributions of the specialized models. The answer to this question,
in the ARS framework, is a combination of two major properties of specialized grammars
and specialized DOP models:
\begin{itemize}
\item The probabilistic subtrees of specialized DOP models are combinations of\linebreak 
      specialized-grammar rules (i.e. partial-trees). 
\item Each recognizable constituent has a complete parse-space in the sense discussed
      in the first requirements of ARS.
\end{itemize}
Due to the first property, a specialized DOP model should be able to evaluate 
probabilistically all structures in the parse-space of any constituent, which the 
specialized parser is able to recognize. And due to the second property,
it is clear that ARS discards only incorrect structures of constituents.  
Since the inference of probabilities of specialized DOP models is exactly the 
same as that for the original DOP models, the ``specialized" distributions 
should be satisfying approximations to the original ones.

To see this, we note that ARS specialization of DOP models does not introduce 
a new rule for redistributing the probabilities of a discarded incorrect
structure; therefore, the probability of a discarded structure is redistributed 
among the remaining structures, that fall into the same distribution, proportionally to the
original probabilities. 
Moreover, the impact of this redistribution on the remaining distributions is limited 
due to a property of DOP: incorrect structures receive much lower probability 
relative to correct ones. 

\subsection{Summary}
The ARS framework for BCG specialization: 
\begin{itemize}
\item provides a means for controlling the tree-language coverage,
\item learns a grammar as small and as unambiguous as possible, 
\item enables combining the specialized grammar together with the BCG into a novel
      parsing algorithm, which provides high coverage,
\item and specializes DOP and similar stochastic models. 
\end{itemize}
In the rest of this chapter we present various computational learning and parsing algorithms
that implement the ARS-framework and operationalize the ideas presented above.
Chapter~\ref{CHARSImpExp} exhibits experimental results to support this theoretical discussion 
of grammar specialization under the ARS.

\newcommand{\MEASGM}{\MEASM_{g}}
\newcommand{\MEASG}{\MEASN$_{g}$}
%
\section{An instance ARS algorithmic scheme}
\label{SCEBL}
The various ARS algorithms presented in the next sections are all based on
inductive extensions to EBL. They all share a common learning strategy, a 
common choice of target-concept, training-data and background-theory. 
The only aspect where these algorithms differ 
is the inductive learning technique, which defines the measure of ambiguity 
and the operationality criteria. This section presents the 
algorithmic scheme that encompasses exactly the common aspects of these algorithms,
where the measures of ambiguity and size are left unspecified. The scheme is derived from the
ARS framework through the adoption of various assumptions and approximations, which
our presentation here aims at exposing. In addition to this scheme,
this section presents a novel parsing algorithm and a novel method for specializing DOP,
thereby instantiating our preceding theoretical discussion on parsing and DOP-specialization 
under the ARS framework.
\subsection{A sequential covering EBL scheme}
A precondition to the present scheme is access to an adequate tree-bank 
representative of a given domain, both in the statistical and in the
linguistic sense. As explained earlier, in this it already assumes access 
to a background-theory represented by the annotation scheme of the tree-bank at
hand.

The target concept of this scheme is a formalization of the concept of {\em constituency}
where the notion of a Sub-Sentential Form (SSF) (defined in chapter~\ref{CHDOPinML})
plays a major role:
\DEFINE{SSF:}
 {Any ordered sequence of terminal and non-terminal symbols, which is the frontier
  of a subtree of a tree in the tree-bank is called an SSF of that tree-bank. 
  The set of all SSFs of a tree-bank $TB$ is denoted $SSF_{TB}$.
 }

The term SSF is derived from the better known term ``sentential-form": any ordered sequence 
of symbols, which results from a finite set of derivation steps starting from the
start symbol of a grammar (not necessarily leading to a sequence of terminals). 
Thus, a sentential-form is a sequence which forms the frontier of some partial-tree 
with as its root the start symbol of the grammar.

\DEFINE{Subtree associated with an SSF:}
{If an SSF $ssf$ constitutes the frontier of a subtree $pt$ of a tree in the tree-bank,  
 $pt$ is called a {\em subtree (or partial-tree) associated with $ssf$}. 
}
Note that there can be many subtrees associated with the same SSF. 
Therefore the following definition.

\DEFINE{Ambiguity set of an SSF:}
{The ambiguity set of $ssf$ over a tree-bank is the set of all subtrees associated with
 $ssf$ in that tree-bank. The ambiguity set of $ssf$ over tree-bank $TB$ is denoted
 as $[[ssf]]_{TB}$. When the tree-bank can be unambiguously determined from the context
 we may omit the subscript $TB$. The set of the ambiguity-sets of all SSFs
 in $SSF_{TB}$ is denoted $AS_{TB}$.
}

\DEFINE{Target concept:}
{The target concept of the present learning algorithmic scheme is a
 function \FUN{${\cal T}$}{$D$}{$R$}, where the domain $D$ is a subset of the set 
 $SSF_{TB}$ and the range $R$ is a subset of the set $AS_{TB}$ that contains the
 ambiguity-sets of all SSFs in $D$. The function ${\cal T}$ assigns to every
 SSF $ssf\in D$ its ambiguity-set $[[ssf]]\in R$.
}
\paragraph{Generalization of the target-concept:}
This definition of the target concept does not take into account that under some 
linguistic theories different SSFs are considered variations of the same entity;
for example in LTAG, two SSFs that are equivalent up to a different number of the
same kind of adjunctions are variations of the same entity. Under such linguistic theories,
the training tree-bank's set of SSFs is partitioned into {\em equivalence classes}\/
according to some notion of linguistic equivalence between SSFs. For these linguistic
theories, the definition of the target concept is a straightforward generalization 
of the definition stated above to equivalence classes of SSFs.
For clarity of the presentation here we will restrict the discussion to a target concept 
which is based on individual SSFs. We will return back to this point during the 
discussion of the implementation detail.
%

%
\exampleI              

Having defined the target-concept, we now turn to deriving the learning algorithm.
For this we need the following definitions:
%
\DEFINE{Constituency Probability:}
{The Constituency-Probability of a sequence of symbols $sq$, denoted $CP_{TB}(sq)$, 
 over a given tree-bank $TB$, is the probability of $sq$ being a subsentential-form 
 in the tree-bank; it is equal to the ratio between the number of times $sq$ is a 
 subsentential-form (denoted $Freq_{C,TB}(sq)$) to the total number of times it 
 appears in the tree-bank (denoted $Freq_{TB}(sq)$). When the tree-bank is known 
 we may omit the subscript $TB$ from the notation: 
 \mbox{$CP(sq) \defEqual \frac{Freq_{C}(sq)}{Freq(sq)}$}.
 The value of $CP(sq)$ is undefined if \mbox{$Freq(sq) = 0$} in the training tree-bank.
}

The constituency probability of a sequence of symbols $X$ is the probability that
$X$ is an SSF. Beside this constituency probability we also define:

\DEFINE{Ambiguity-Set Distribution (ASD):}
{This is a discrete probability function $ASD_{ssf}$ over the ambiguity-set of an SSF $ssf$.
 Each structure $t$ in this ambiguity-set has probability $ASD_{ssf}(t)$ equal to
 the ratio of the frequency of the structure $t$ in the tree-bank (denoted $Freq(t)$)
 to the total frequency of $ssf$ as an SSF in the tree-bank,
 i.e.  \mbox{$ASD_{ssf}(t) \defEqual \frac{Freq(t)}{Freq_{C}(ssf)}$}.
 The value of $ASD_{ssf}(t)$ is undefined if \mbox{$Freq_{C}(ssf) = 0$}.
}
Note that the ASD distribution assumes that a structure associated with $s$, 
{\em given that $s$ is an SSF}, is a random variable. Let  $struct(ssf)$
denote this random variable. The ASD of $ssf$ describes the various probabilities 
\mbox{$ASD_{ssf}(struct(ssf) = t)$}
for every $t$ in its ambiguity set. Moreover, the value \mbox{$CP($ssf$) \mul ASD_{ssf}(t)$}, 
for any $t$ in the ambiguity set of $ssf$, denotes the total probability of assigning 
$t$ as a structure to $ssf$ prior to knowing whether it is an SSF or not; it combines the
two decisions 1)~whether $ssf$ is an SSF and 2)~what structure to assign to $ssf$ 
given that it is an SSF.

In the light of the definitions above, it is worth pausing shortly at this point and 
considering the ARS framework's requirement concerning tree-language coverage (first bias-rule).
It turns out that this requirement can be implemented as follows, under 
the important assumption that we have access to an infinitely large 
tree-bank (i.e. in the limit):
\begin{verse}
 If the specialized grammar can recognize a certain SSF in the domain, 
 it must be able to generate its ambiguity set.  
\end{verse}
Since the limit case is only a theoretical situation, we have to use a
good approximation, i.e. a sufficiently large tree-bank. If the tree-bank is not
sufficiently large, we should be able to decide whether a given SSF has 
a ``sufficiently complete" ambiguity set in that tree-bank or not. We will delay 
the discussion of this problem  until section~\ref{SecSufComp}
and assume, for the time being, 
that we are able to decide which SSFs in the tree-bank have a sufficiently 
complete ambiguity set. Given this and assuming an EBL-based learning algorithm, the 
first requirements specification of the present algorithmic scheme can be stated 
as follows:
\begin{verse}
To cut the tree-bank trees in such a way that the obtained set of partial-trees
is the union of only sufficiently complete ambiguity sets of SSFs. Moreover,
this set of partial-trees is both the smallest and least ambiguous among 
those that fulfill the condition.
\end{verse}
Needless to say, the currently vague requirement ``smallest and least ambiguous" still 
needs to be specified and quantified. As mentioned above, the present 
algorithmic scheme encapsulates only the commonalities of the various ARS algorithms,
which we present.  Since these algorithms differ exactly in the measures
of ambiguity and grammar size (and the ways of weighing them against each other),
the specification of this requirement is left to the discussions in the next 
sections.

\subsubsection{The search strategy: sequential-covering}
An exhaustive search for the way of cutting the tree-bank trees which fulfills the 
requirements stated above, is virtually impossible due to time and space limitations.
Therefore, we need to limit the search in order to find good approximations. Our choice 
here is for the so called Sequential Covering search strategy~\cite{Mitchell97}. 
 Sequential covering is an iterative strategy which can be summarized
as follows: {\em learn some rules, remove the data they cover and iterate
this process}. Hence the name Sequential Covering (SC) scheme.
Due to its stepwise reductive nature, the SC strategy of learning reduces the space 
of hypotheses (i.e. grammars in our case) drastically after each iteration,
thereby facilitating faster learning from larger bodies of data.
Needless to say, the SC strategy is a greedy approach that might be suboptimal in comparison
to exhaustive search.

The SC strategy is incorporated in the present algorithmic scheme in such a way
that at each iteration the tree-bank trees are reduced in a {\em bottom-up}\/
fashion only. Of course, there are many other reduction strategies which might
be as good. But our choice here for a bottom-up fashion has to do with 
the way we wish to integrate the resulting specialized grammar and the initial Broad-Coverage
Grammar. This will become clearer when we discuss the parsing algorithm in 
section~\ref{ParAlgSch}.
In essence, the present algorithmic scheme learns in iterations until
the tree-bank is empty. At each iteration, the following actions are taken:
\begin{enumerate}
\item A set of SSFs, each with a sufficiently complete ambiguity set, is learned.
      The SSFs which are considered at the current iteration are only those that
      are on the frontiers of the tree-bank partial-trees (i.e. we proceed in a bottom-up fashion).
      Moreover, the union set of the ambiguity sets of the learned SSFs must be the 
      smallest and least ambiguous possible.
\item All instances of the partial-trees in the ambiguity-sets of learned SSFs
      are removed from the current tree-bank partial-trees, resulting in the 
      tree-bank of the next iteration.
      The removal of the instances of the partial-trees takes place only bottom-up,
      i.e. the iterations ``nibble" on the tree-bank trees from their 
      lower parts upwards.
\end{enumerate}
%

\exampleII

\subsubsection{The operationality criterion}
At this point we make another convenience assumption: each SSF competes only with 
a limited number of other SSFs on a place in the learned grammar. This limits the number 
of combinations of SSFs, since if an SSF is determined to be more suitable than 
any of its competitors, it will be part of the learned grammar. 
A suitable definition of a competitor seems the following:
\DEFINE{Competitors of an SSF:}
{In tree-bank $TB$, $ssf_{c}$ is called a competitor of $ssf$ 
 if and only if there is a tree $t$ in $TB$ such that 
 there are $st_{c}\in [[ssf_{c}]]$ and $st\in [[ssf]]$ which are subtrees
 of $t$ and $st_{c}$ is a subtree of $st$.
 The {\em set of competitors}\/ of SSF over a tree-bank contains all competitors 
 of SSF in that tree-bank. 
}
Note that the relation ``competitor" is asymmetric.
The motivation behind this definition is that if we choose to limit the free 
competition between SSFs on a place in the learned grammar, we should take care that this
competition does not harm the coverage of the learned grammar too much.
This definition can be used by the learning algorithm to guarantee that an SSF is
learned only if it ``beats" all SSFs in the tree-bank that are either subsequences 
of it or of which it is a subsequence. And this means that either an SSF is 
represented as a whole (either on its own or as a subsequence of another SSF), 
or subsequences of it are represented.  

Since the algorithm learns from a given tree in the tree-bank, in a bottom-up 
fashion, it considers the SSFs on the frontier of that tree and the competitors
of these SSFs in the whole tree-bank. For every such SSF, we have to decide, on the basis
of global information from the whole tree-bank, on whether to learn it or not.
An SSF is learned from a given tree,  if and only if it has a sufficiently
complete ambiguity set and is the ``best choice" among all its competitors. 
The predicate ``best choice" must be defined and quantified to facilitate the
choice of the ``smallest and least ambiguous" grammar possible. This amounts to 
localizing these measures to individual SSFs rather than applying it to a whole grammar.
Instead of choosing the smallest and least ambiguous grammar, we now choose the grammar
which is the union of the smallest and least ambiguous ambiguity sets of SSFs, where 
the comparison is only between an SSF and its competitors, as defined above. \\
\paragraph{Summary:} To summarize, the present algorithmic scheme assumes:
\begin{itemize}
\item a macro-rules EBL-algorithm extended with inductive learning,
      i.e. the goal is to cut the tree-bank trees into a set of macro-rules that
      forms the specialized-grammar,
\item an (iterative) sequential covering strategy of learning,
      i.e. the specialized-grammar is the union of a finite sequence of
      macro-rule sets, each in its turn being the union of ambiguity sets of the
      SSFs learned at a certain iteration. Let $SG$ denote the specialized grammar,
      $i$ a counter of iterations and $j$ a counter of SSFs learned at the same iteration,
      then: 
     \[ SG = \bigcup_{i}\bigcup_{j}~[[ssf_{i,j}]],      \]
      In other words, $SG$ is the union of ambiguity-sets of SSFs that are learned
      from the tree-bank\footnote{In fact, $SG$ can be partitioned into (mutually-exclusive)
      equivalence classes (the ambiguity sets), each associated with a different SSF.},
\item the space of SSFs in the tree-bank is divided into (not necessarily
      mutually exclusive) sets of competitors,
\item each set of competitors contributes its best choices of SSFs
      to the learned grammar; the best choices are determined on each tree
      in the tree-bank individually, using measures of size and ambiguity of SSFs,
      which will be defined in the sequel. 
\end{itemize}
We will refer to the grammar resulting from this learning process, 
i.e. the specialized-grammar, also as the partial-grammar; this grammar
is partial with respect to the original BCG. And we will refer to the
learned SSFs, i.e. the SSFs that underly the specialized-grammar,
with the term {\em specialized-grammar SSFs}.

\newcommand{\SSF}{{\cal SSF}}
\begin{figure}[tbh]
\hrule
{\sl
\begin{description}
\item [0.] $i := 0$;
\item [\bREPEAT]
\item [~~1.] Compute ${\SSF}_{i}$,
\item [~~2.] $\forall ~ssf\in {\SSF}_{i}$ compute\\
          ~ ~~~the frequencies that are necessary for \MEAS($ssf$), and\\
          ~ ~~~the $Competitors_{i}(ssf)$,
\item [~~3.] $\forall ~ssf\in {\SSF}_{i}$: $Viable(ssf):= true$ iff\\
          ~  ~~~$ssf$ has a sufficiently complete ambiguity-set in ${\TB}_{i}$, and\\
          ~  ~~~$\forall ssf2\in Competitors_{i}(ssf)$: \MEAS($ssf$) $\geq$ \MEAS($ssf2$),
\item [~~4.]
\begin{tabbing}
\shftl
$\forall$ $t\in {\TB}_{i}$, \\
\shftl
$\forall$ node address $N$ in $t$:  \\
\shftl
~~N \= is {\it marked as cut node}\/ {\bf iff} \\
    \> $Viable($\Front$(N))$ is true  \AND\\
    \> $\forall$ $Nx\neq N$ in $t$: 
       ~$Viable($\Front(Nx)$)$~\lra \Front(N)$\notin$ Competitors(\Front(Nx))
\end{tabbing}
\item [{\bf ~5.}] $i := i+1;$
\item [{\bf ~6.}] 
      ${\TB}_{i}$ := (${\TB}_{i-1}$ after reducing all partial-trees under marked nodes); 
\item [\bUNTIL] ((${\TB}_{i} == \emptyset$)  \OR~ (${\TB}_{i} == {\TB}_{i-1}$));
\end{description}
}
\hrule
\caption{An implementation of the present algorithmic scheme}
\label{AlgScheme}
\end{figure}

%
\subsubsection{Operationalizing the algorithmic scheme}
Figure~\ref{AlgScheme} contains a functional specification of the present algorithmic
scheme. The specification in figure~\ref{AlgScheme} assumes the following notation and
definitions:
\begin{itemize}
\item 
$N$ denotes a unique address for each node of a tree $t$ in the tree-bank.
\item 
${\TB}_{i}$ denotes the tree-bank obtained after $i$ iterations; ${\TB}_{0}$ thus
denotes the initial tree-bank.
\item 
\Front($N$) denotes the sequence of leaf nodes dominated by $N$.
\item 
$Competitors_{i}(ssf)$ denotes the set of all competitors of $ssf$ in
         tree-bank ${\TB}_{i}$.
\item 
${\SSF}_{i}$ denotes the set of all SSFs in tree-bank ${\TB}_{i}$.
\item 
 \MEAS() denotes the combined measure of ambiguity and grammar size: 
\begin{verse}
\mbox{\MEAS: SSFs \lra \REALS} assigns a larger (real) value to SSFs that 
result in less ambiguous and smaller grammars. 
\end{verse}
\end{itemize}
In section~\ref{SecSpecAlgs}, function \MEAS() will be defined in various ways,
according to different choices of the inductive learning paradigms and measures 
of ambiguity and grammar size.

In the specification, the goal is to mark the {\em cut nodes}\/ in the 
tree-bank trees; the {\em cut nodes}\/ in the tree-bank trees denote the borders of the
partial-trees associated with the learned SSFs. By cutting\footnote{
Cutting a tree at the marked nodes is simple: duplicate every cut node in the
tree by making it a pair of nodes (labeled exactly as the original node) such 
that one of the two nodes is connected to the parent node and the other to the 
children nodes of the original node. The duplicate nodes are not connected and
thus the resulting graph is not connected. The subgraphs are subtrees of the
original tree.
} the trees of the tree-bank at the cut nodes we obtain a set of subtrees of 
the tree-bank trees. This set is the union of the Ambiguity-Sets of the learned SSFs. 

Cut node marking implements the present algorithmic scheme in one of the many
possible (and equivalent) ways.
A node with address $N$ is marked in tree $t$ of the tree-bank iff 1)~its 
frontier SSF (\Front$(N)$) has a larger \MEAS() value than all its competitors in 
the tree-bank and 2)~for every other node $Nx$ in $t$, if the frontier of $Nx$ in $t$
has a larger \MEAS() value than all its competitors \Front(N) is not one of
them. This is equal to traversing the tree $t$ by a depth-first traversal from the root 
downwards and stopping the in-depth traversal at those nodes that have a $Viable$ 
frontier SSF.
%
\subsubsection{Sufficient completeness}
\label{SecSufComp}
The question whether an SSF has a sufficiently complete ambiguity-set
can be stated as follows. Given an SSF $ssf$ of a training tree-bank $TB$,
{\em what is the probability that the ambiguity-set of
$ssf$ over $TB$ is not complete~?},
or equivalently: {\em what is the probability that a new sentence of the domain
(i.e. not in $TB$) has a parse $t$ in which $ssf$ is the frontier of a 
subtree $st$ of $t$}\/
{\em such that $st$ is not in the ambiguity-set of $ssf$ over $TB$~?}.
It is not hard to see that the latter question can be reduced to the older well-known
research question: 
{\em does the CFG underlying the training tree-bank generate for a sufficient portion of 
the domain sentences the right parse-trees~?}. 

In general, all methods that employ tree-banks for learning assume that the training 
tree-bank is sufficient in the sense that it is a {\em sample}\/ of the domain it 
represents, i.e. the distributions in the tree-bank are good approximations. In many
cases this is an incorrect assumption, and probabilistic methods try to compensate for this
by using smoothing or reestimations methods that improve the distributions 
obtained from the tree-banks e.g. Good-Turing~\cite{Good}, Back-Off~\cite{Katz} and Successive 
Abstraction~\cite{SamuelssonSA}.

The question that we deal with can be solved by the same methods that are used
to reestimate the probabilities of subtrees of the tree-bank trees e.g.~\cite{Good,RENSDES}. 
These methods ``reserve" some probability-mass for subtrees that did not occur in the 
tree-bank and adjust the probabilities of the subtrees that did occur in the tree-bank to
allow for this. Suppose now that this method is applied to a DOP model obtained from the 
training tree-bank. Now we have the original DOP model, denoted $O$ and the 
reestimated DOP model, denoted $E$. 
In any DOP model, the probability of the ambiguity-set $[[ssf]]$ is the sum of the 
probabilities of all partial-derivations that generate any of the subtrees in~$[[ssf]]$. 
Denote the probability of the ambiguity-set of $ssf$ with respect to model 
\mbox{$X\in \{O, E\}$} with $P_{X}([[ssf]])$. 
Then ${P_{O}([[ssf]])-P_{E}([[ssf]])}$ is an estimate of 
the probability that a subtree is missing from the ambiguity-set of~$ssf$.

Clearly, this procedure can be very expensive if it is applied to every SSF in the 
training tree-bank.
The number of SSFs in a tree-bank is as large as the number of DOP subtrees; for the
tree-banks we are dealing with, the space which is necessary for computing these subtrees exceeds
a few Giga-bytes. And the number of SSFs runs in the millions. This costs a huge amount of
time to compute if it is possible to do so at all. Therefore, in the sequel I will assume that
the training tree-banks are sufficiently large such that all SSFs have sufficiently complete
ambiguity-sets. However, the issue of how to estimate the sufficient 
completeness of SSFs in a practical way remains an open question for future research.
%
\subsection{Completing composed ambiguity sets}
\label{SecQofCDisc}
Let us now consider the important question whether the above instantiation of 
the ARS framework fulfills the ARS requirement concerning a satisfying tree-language coverage.
This requirement states that
{\em if the specialized grammar is able to recognize a certain 
constituent, it is able to generate its ambiguity set}. 
To consider this requirement in the light of the present learning algorithm, 
we may distinguish between two cases in applying the learned grammar to recognizing
constituents:
\begin{description}
  \item [i.] a constituent is recognized in (among others) a derivation that consists of
        a single derivation-step.
  \item [ii.] a constituent is recognized only by derivations that are compositions of at
         least two derivation-steps (denoted composed derivations). 
\end{description}
Let us now consider whether the requirement concerning the tree-language coverage is
fulfilled in both case:
\begin{enumerate}
 \item For the first case, since the ambiguity sets, that constitute the specialized-grammar,
       are all assumed sufficiently complete by the learning algorithm, the requirement 
       is immediately fulfilled.
 \item Assume we are given a constituent $cssf$ of the second type. 
       We note that $cssf$ is recognized by a composition of at least two SSFs from the 
       specialized-grammar. The composition of these SSFs is governed by the composition 
       of subtrees from their ambiguity sets; composition here is limited to the substitution 
       of one tree in another. 
       The result of composing subtrees from the ambiguity sets of SSFs from the
       specialized-grammar is a composed partial-tree, which has $cssf$ as its frontier. The set 
       of composed partial-trees that can be constructed in this way is called
       the {\em composed ambiguity set}\footnote{Under the given specialized 
       grammar and in the domain.}\/ of $cssf$. 

       For this type of constituents it is not immediately clear that the requirement of
       a satisfying tree-language coverage is fulfilled. The issue here is on the one hand,
       the ability of a grammar to represent language in a compositional manner, and on 
       the other hand, the compositional nature of language use in the given domain.
       As usual in language modeling, it is reasonable to assume that this subrequirement is
       fulfilled to a large extent but not quite the whole way. This is because a problem 
       might arise with the so called idiomatic and semi-idiomatic constructions, which are not 
       compositional. In general, these constructions are much less frequent than the 
       compositional constructions of language, but they are still not negligible. Thus, 
       to some small extent, the above algorithmic scheme, as is, fails to fulfill the 
       requirement.
\end{enumerate}
To tackle the problem caused by constituents of the second type, we complement 
our learning algorithm with another simple algorithm, which aims at {\em completing the 
composed ambiguity sets}\/ by a second round of learning from the tree-bank. 
%
%
%
\subsubsection{An algorithm for completing ambiguity sets}
\label{CompoAmbS}
The only source of information we can employ for completing the composed
ambiguity sets using an automatic algorithm, is the training tree-bank and the
markings of the cut nodes provided by the specialization algorithm. The goal here is 
to detect structures which are missing from ambiguity sets of composed
specialized-grammar SSFs.

In essence, the algorithm for completing the ambiguity sets consists of the
following steps:
\begin{enumerate}
 \item It simply scans the tree-bank output by the specialization  
       algorithm (cut nodes are marked). It collects all possible 
       SSFs on the {\em frontiers}\/ of trees in the tree-bank, and maintains for every 
       SSF a set of pairs.
       Each pair in the set for $ssf$ points to a node in a tree in the tree-bank,
       which is the root of a subtree that has $ssf$ at its frontier. 
       A pair \mbox{$\tuple{Marked, \mu}$} consists of the unique address of the node, $\mu$, 
       and a truth-value $Marked$, stating whether the node is marked or not.
 \item For every SSF on the frontier of a tree in the domain:
  \begin{enumerate}
   \item The SSF is marked {\em incomplete}\/ iff its set of pairs (i.e. nodes)
       contains unmarked as well as marked nodes. Marked nodes indicate that
       the SSF can be constructed from one or more of the SSFs, which were 
       learned by the specialization algorithm. And unmarked nodes, in contrast,
       indicate subtrees which should be associated with that SSF but which
       have not been learned by the specialization algorithm.
   \item The subtrees associated with an incomplete SSF which have unmarked root 
        nodes are extracted and kept aside. 
  \end{enumerate}
 \item The specialized grammar now consists of the ambiguity 
       sets learned by the specialization algorithm complemented by all extracted 
       subtrees learned in the present algorithm.
\end{enumerate}

\exampleIII
Notice that the algorithm considers only SSFs that are on the frontiers of 
trees (i.e. those that are constituents) in the tree-bank. This is because
the aim here is to learn idiomatic structures that were missed by the original 
learning algorithm. And intuitively, idiomatic constructions can be expected to involve
the lowest levels of the trees most (if not all) of the time.
\subsection{A novel parsing algorithm}
\label{ParAlgSch}
As described in section~\ref{ParARS}, the parsing algorithm under the 
ARS scheme integrates the specialized grammar and the BCG in
a two phase parser. Below we instantiate this parsing algorithm
for the SC~EBL-scheme described above. But let us first consider the
types of specialized grammars which result from the present learning algorithm.
\subsubsection{The specialized grammar}
There are three ways to view the specialized grammar: as a Tree-Substitution
Grammar (TSG) or as a Cascade of Finite State Transducers (CFSTs):
\begin{description}
\item [TSG:]
Consider the ambiguity-sets of the SSFs that are learned from a given
tree-bank, with our scheme. The union of these sets is a CFG with rules
that are partial-trees, i.e. a TSG. The start-symbol, 
the set of non-terminal symbols and the set of terminal symbols of this 
TSG are exactly those of the BCG underlying the tree-bank.
%
\item [CFSTs:]
The TSG implementation masks the iterative process in which the
SSFs were learned and results in a (possibly) recursive grammar. An
alternative is to keep the sets of SSFs, and their ambiguity sets, 
for each iteration {\em apart}. For each iteration, the set of SSFs and their 
ambiguity sets learned at this iteration are considered as a 
set of Finite State Transducers (FSTs); each SSF is a regular expression
(i.e. FSM) which, upon recognition, emits its ambiguity set. 
When parsing a given input, the set of FSTs of iteration $i$ is 
applied only to the output of iteration $i-1$, for all \mbox{$i\geq 1$}. 
The set of FSTs learned at iteration~$1$ is applied to the given input 
sentence or word-graph. 
\item [Union-FST:] An FST can be obtained by taking the union of the FSTs that 
      are constructed in the CFST implementation. This Union-FST can be applied
      in a feed-back construction to an input sentence. The feed-back iterations
      stop when the last output is equivalent to the last input.
\end{description}
There are two major differences between the TSG and the CFST: 
\begin{enumerate}
\item Due to the finite number of iterations, the CFSTs implementation 
      does not allow unlimited recursion. 
\item Due to the requirement that the set of FSTs of each iteration be
      applied only when its turn comes, the CFST implementation imposes 
      a further constraint on the substitution of partial-trees. In a TSG,
      it is sufficient that the root of one partial-tree $t$ be labeled with the
      same non-terminal symbol as a substitution-site in another partial-tree
      $tx$, in order for the substitution of $t$ in $tx$ to take place. 
      In the CFST, $t$ and $tx$ must fulfill an extra requirement: $t$ and $tx$
      must be in the ambiguity sets learned, respectively, at iteration $i$ 
      and $i+j$, where \mbox{$j\geq 1$}.
\end{enumerate}
These differences imply that both the language and the tree-language of the
CFSTs implementation maybe proper subsets of those of the TSG implementation.
In fact, both the string-language and tree-language of the CFST are finite 
(since the number of iterations and the ambiguity sets are also finite).
Nevertheless, because of its close fit with the process of learning, the CFSTs 
implementation might provide a better tree-language coverage than the TSG. 
Similar differences exist between the Union-FST implementation and the TSG and CFST
implementation.

Because of the substantial additional effort that might be involved
in implementing the CFSTs implementation and the Union-FST, we pursue only 
the TSG implementation in this thesis. 
\subsubsection{Integrating the two parsers}
\label{SecCombPars}
As mentioned before, the present parsing algorithm integrates the TSG~specialized parser
(denoted by the term partial-parser) with the original BCG parser in
a pipeline construction; the input sentence is fed to the partial-parser 
and the result is then fed to a {\em constrained}\/ BCG-based parser.
For implementing the partial-parser we employ the CKY parsing algorithm~\cite{CKY}
extended and optimized for TSGs as described in detail in chapter~\ref{CHOptAlg4DOP} 
of this thesis~\footnote{For accuracy we may note that
the algorithm as described in chapter~\ref{CHOptAlg4DOP} assumes STSGs;
it is trivial that it can be used for TSGs also.}.  
And for combining it with the BCG-parser we employ a novel 
{\em constrained}\/ version of the CKY algorithm described next.

\begin{figure}[htb]
\epsfxsize=13.4cm
\center{
{\epsfbox{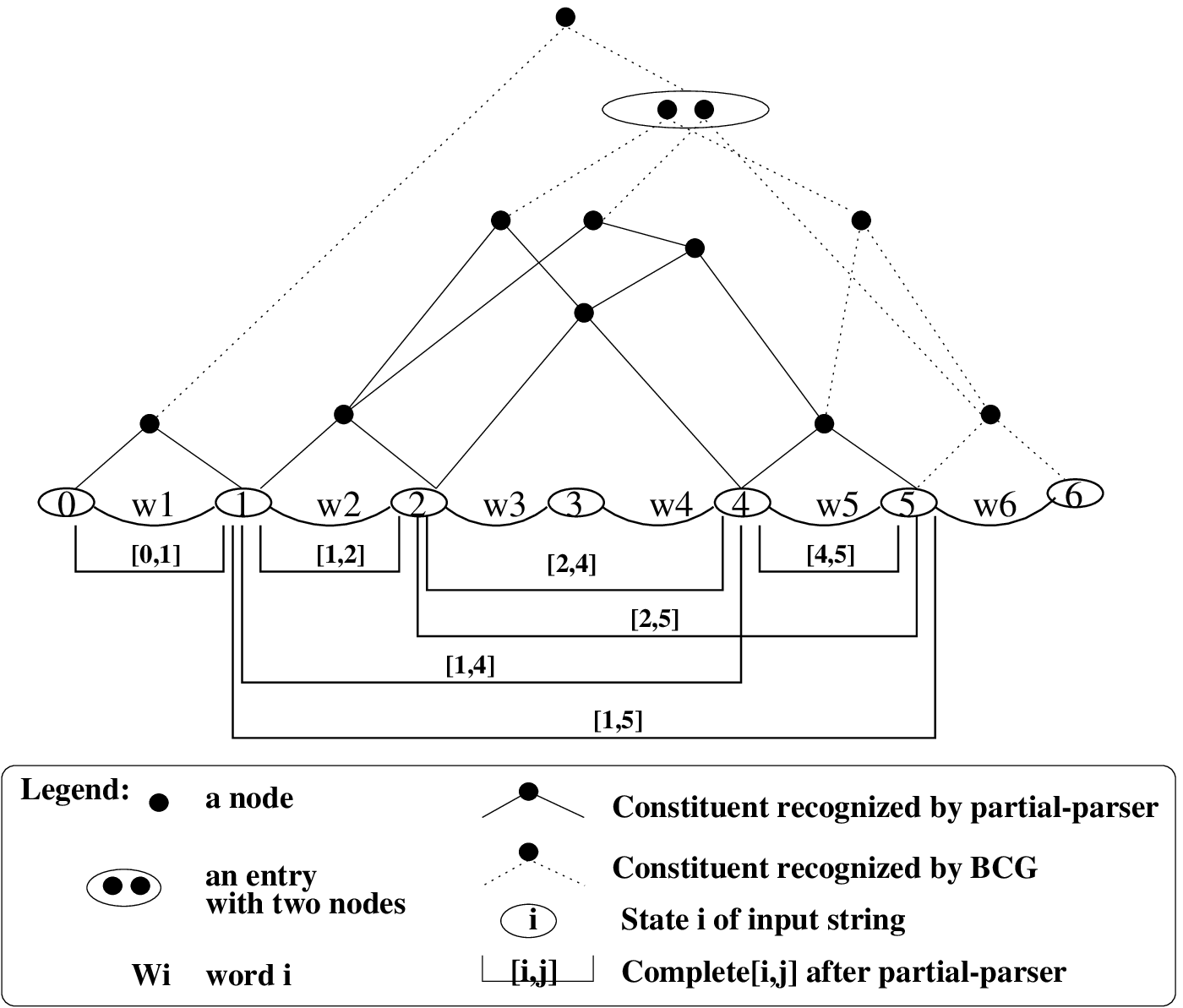}}
}
\caption{A simplistic sketch of combined parsing}
\label{CHGrSputwo}
\end{figure}

In the CKY algorithm\footnote{The reader interested in the details of the CKY algorithm
is advised to read section~\ref{CKYViterbi}. In particular, figure~\ref{FigCKYAlg} 
provides a specification of the CKY algorithm for CFGs.}, 
an input sentence \mbox{$w_{1}\cdots w_{n}$} of 
length $n$ is considered as a sequence of $n+1$ different 
states\footnote{I use the same terminology as that of word-graphs in order to keep
the discussion as general as possible.}; 
a state before the first word, a state after the last word and a state 
between every two consecutive words (see figure~\ref{CHGrSputwo}).
These states are numbered \mbox{$0\cdots n$}. Now, for every combination 
of two states in this sequence \mbox{$i < j$}, a parse-table (or so called 
well-formed substrings table or chart) contains an entry \mbox{$[i, j]$}; 
entry \entry{i}{j} holds all nodes, representing roots of 
structures spanned between states $i$ and $j$, i.e. for the word sequence 
\mbox{$w_{i+1}\cdots w_{j}$}. A node labeled with the start symbol of the grammar
in entry \entry{0}{n} implies a parse of the whole input sentence.

\begin{figure}[htb]
\hrule
{\sl
\begin{tabbing}
\comment{
$root_{TSG}(r)$ denotes the predicate: $r$ is a root of~~~~~~~~~~}\\
\comment{
a partial-tree of the specialized grammar (a TSG).~~~~}\\
\comment{
CKY\_Parser$_{G}$(sentence, i, j) denotes the CKY parser,~}\\
\comment{
instantiated for grammar $G$ (either a BCG CFG or a }\\
\comment{
specialized TSG), applied to entry \entry{i}{j} of input sentence $sentence$.}\\
\end{tabbing}
\begin{tabbing}
{\bf 1.}
$\forall$ \= $1\leq k \leq n$ ~\AND~ $\forall$ $0 \leq i\leq (n-k)$\\
          \> CKY\_Parser$_{TSG}$($w_{1}\cdots w_{n}$, i, i+k);\\
{\bf 2.}
$\forall$ \= $1\leq k \leq n$ ~\AND~ $\forall$ $0 \leq i\leq (n-k)$\\
          \> \bIf ($\ITEMN{r}{\alpha}{}\in$\entry{i}{i+k} ~\AND~ $root_{TSG}(r) )$ \\
          \> \bThen $Complete(i,i+k) = true$; \\ 
{\bf 3.} 
$\forall$ \= $1\leq k \leq n$ ~\AND~ $\forall$ $0 \leq i\leq (n-k)$\\
          \> \bIf (Complete(i,i+k) == false) \\
          \> \bThen CKY\_Parser$_{BCG}$($w_{1}\cdots w_{n}$, i, i+k); \\
\end{tabbing}
\hrule }
\caption{Integrated parsing algorithm}
\label{CHGrSpuComPar}
\end{figure}

The integration of the partial-parser with the BCG-parser employs a single CKY 
parse-table. The integrated parser uses this parse-table as follows (figure~\ref{CHGrSpuComPar}
provides a specification):
\begin{enumerate}
\item The partial-parser is employed first in order to recognize as much as it 
     can from the input. The structures built by the partial-parser, are placed 
     in the table. These structures are {\em combinations of partial-trees of the TSG},
    i.e. combinations of partial-trees from ambiguity-sets of the SSFs which were 
    acquired during the learning phase.
   \item Every entry $[i,~j]$, which contains a node corresponding to the root of 
      a partial-tree of the partial-parser's TSG, is marked as ``complete"; indicated 
      by the proposition $Complete(i,~j)$ in figure~\ref{CHGrSpuComPar}. Note that this is
      justified by the {\em tree-language coverage}\/ property of the partial-parser.
   \item 
      All and only those entries that are {\em not}\/ marked {\em complete}\/ are 
      reparsed by the BCG-based CKY parser. This involves building structures for 
      these entries using the CKY algorithm, i.e. by\/: 
    \begin{enumerate}
     \item [i.] exploiting the structures in entries marked as complete,
     \item [ii.] building new structures from scratch when there are no such structures, 
     \item [iii.] and combining all these structures together as the CKY algorithm dictates. 
    \end{enumerate}
\end{enumerate}
Figure~\ref{CHGrSputwo} depicts this process using a hypothetical example.
The figure shows only parses of the whole string but other partial-parses 
might be in the chart as well. 
\subsubsection{Complexity of the integrated parser:}
The time complexity of this parsing algorithm is still equal to that of an ordinary
CKY for TSGs, i.e. \mbox{$O(A n^{3})$}, where $A$ denotes the total number of nodes in the
elementary-trees of the TSG and $n$ denotes the length of the input sentence.
The space complexity of the algorithm is also equal to that of the CKY for TSGs,
i.e. for recognition this is \mbox{$O(A n^{2})$} and for parse-forest generation 
this is \mbox{$O(A^{2} n^{2})$}.

The main gain from combining the partial-parser together with the BCG-based parser is in
the fact that on the one hand the partial-parser is able to provide a satisfying 
tree-language coverage of the sentence portions that it is able to recognize, and on the other hand, 
it spans a smaller parse-space for these portions. The integrated parser's output is 
therefore less ambiguous 
than the BCG-based parser (in the limit) without loss of tree-language coverage.
%
%
\subsection{Specializing DOP}
\label{SecSpecDOP}
As mentioned in the preceding chapter, we assume that a parser consists of two modules,
a parse-space generator (the parser) and a probabilistic parse-space evaluator (the
disambiguator). Chapter~\ref{CHOptAlg4DOP} describes a two phase DOP STSG parsing and 
disambiguation algorithm based on this assumption:
the parsing phase employs the CFG underlying the STSG to span the parse-space of the input,
and the disambiguation phase applies the DOP STSG's probability computations on this phase.

\subsubsection{Integrating the partial-parser and the DOP STSG}
The integrated parser described in the preceding section allows integrating the 
specialized grammar and the CFG underlying the DOP STSG during the parsing phase.
In this construction, henceforth referred to as the {\em partial-parser+DOP STSG}\/
(abbreviated shortly by ParDOP), the role of the partial-parser is simply to limit the 
parse-space prior to the disambiguation phase. The result is that the parse-space 
generation produces smaller parse-spaces; the disambiguation phase, i.e. the DOP STSG, 
does not change.
\subsubsection{Acquiring Specialized DOP STSGs (model SDOP)}
Apart from the specialized grammar, the learning algorithm results in marking cut nodes
in the tree-bank trees. These cut nodes are used for acquiring the 
{\em Specialized DOP (SDOP) STSG}\/ from the tree-bank. The idea here is that the specialized
grammar's partial-trees are atomic units, i.e.  rules, and thus their internal nodes should
not be used for acquiring the SDOP STSG. Given a tree-bank in which the specialization algorithm
marked the cut nodes, acquiring the SDOP STSG is done as follows: 
in the tree-bank trees, only nodes that are marked as cut nodes qualify for the extraction 
of subtrees (which become the SDOP STSG's elementary-trees). In other
words:
\begin{verse}
a subtree~$st$ is extracted from a tree-bank tree~$t$ iff 1)~its root node corresponds
to a marked node of $t$, and 2)~its leaf nodes correspond either to marked nodes or 
to terminal nodes of~$t$.
\end{verse}
The Specialized DOP (SDOP) STSG is simply the STSG that has a set of elementary-trees that 
contains all subtrees extracted only from nodes marked as cut nodes in the tree-bank. Note that
every elementary-tree of the SDOP STSG is a combination of partial-trees of 
the specialized grammar\footnote{
Of course not all combinations of the partial-trees of the specialized grammar are in the
SDOP STSG. Only those combinations that actually occur in the tree-bank are
there.
}. Specialized DOP STSGs are much smaller but (at least) as accurate as the STSG's
obtained according to the original DOP model~\cite{RENSDES}. 
\subsubsection{Integrating the SDOP with Par+DOP (model ISDOP)}
\begin{figure}[bht]
\epsfxsize=14.3cm
\center{
\epsfbox{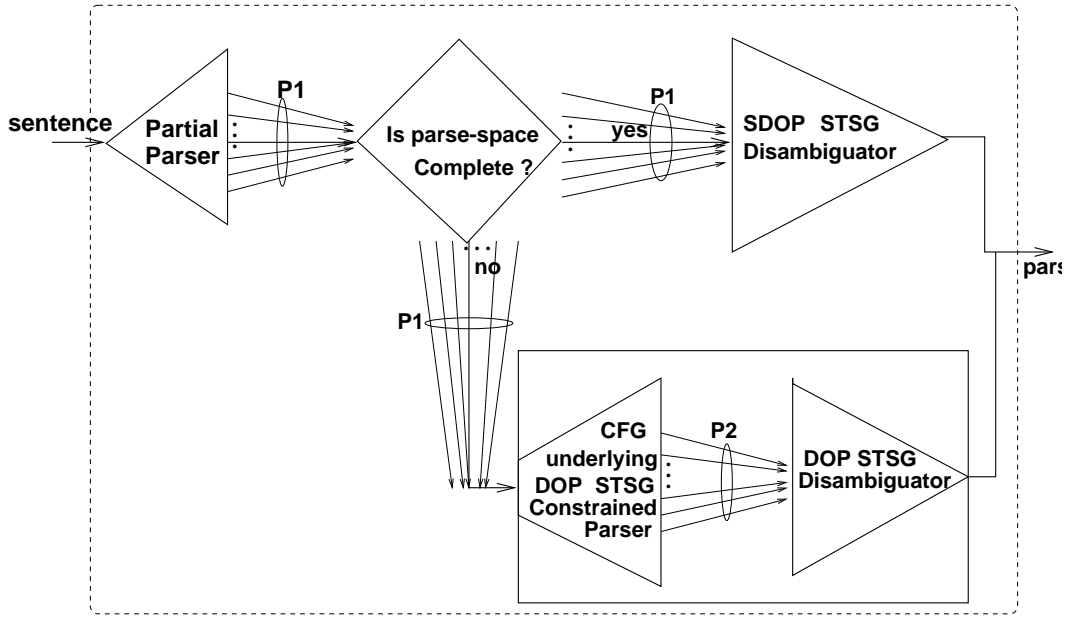}
}
\caption{Integrating SDOP with DOP on basis of the partial-parser}
\label{FigCombDOP}
\end{figure}

In the same manner as the specialized grammar is integrated with the BCG, the 
Par+DOP model is integrated together with the SDOP in a complementary manner;
the partial-parser (based on the specialized grammar) forms the basis for the
integration. The integration has the following modules:
\begin{description}
\item [Par+DOP:] the partial-parser (specialized grammar) is integrated with the
      CFG underlying the (original) DOP STSG for the parsing phase; the disambiguation
      phase simply applies the DOP STSG to the parse-space resulting from the parsing
      phase.
\item [SDOP:] the Specialized DOP STSG acquired as described above.
\end{description}
The integration of these modules, called Integrated Specialized DOP (ISDOP), 
operates as follows:
\begin{enumerate}
\item Parse the input sentence with the partial-parser. This results in a CKY table
      containing a parse-space.
\item Mark entries in the table with $Complete()$ as explained in section~\ref{SecCombPars}.
\item If $Complete(0,n)$ is true then apply the SDOP STSG for disambiguation of
      the parse-space (see chapter~\ref{CHOptAlg4DOP} for the details of this).
      Else apply the CFG underlying the (original) DOP STSG for completing the parse-space
      as described in section~\ref{SecCombPars}, and then apply the DOP STSG for disambiguation
      of the resulting parse-space.
\end{enumerate}
Figure~\ref{FigCombDOP} depicts the ISDOP construction, where the parse-space generated
by the partial-parser is denoted {\bf P1}, and the parse-space {\bf P1} complemented by the 
CFG underlying the DOP STSG is denoted {\bf P2}. The following properties
of this integration make it attractive:
\begin{itemize}
\item The integration does not result in wasted parsing time since the partial-parser's
      work is not lost in any case: in case the parse-space contains parses of the whole
      input sentence, the parse-space is used as is, and in the other case it is complemented 
      by the CFG underlying the DOP STSG.
\item The DOP STSG is applied {\em only when it is sure that the input sentence is not in the
      language of the SDOP}. This implies that the large DOP models (in general)
      are applied to input sentences that deviate in unexpected ways from the training
      tree-bank sentences.
\item The tradition in acquiring DOP STSG by limiting the maximum depth\footnote{
     Section~\ref{SecHeuristics} describes this and other heuristics in detail.
     The depth of a partial-tree is the length of the longest path from the root
     to a leaf node of the partial-tree. The length of a path is equal to one less
     than the number of nodes on that path.
     } of subtrees is exploited in acquiring SDOPs to improve the accuracy of the
    acquired SDOP STSG. In acquiring SDOP STSGs, only marked nodes are counted
    in calculating the depth of subtrees. This results in a smaller number of subtrees
    that are much deeper than DOP STSG's subtrees. This captures many probabilistic
    relations that cannot be captured by limited depth DOP STSGs.
\end{itemize}

%
\section{Measures of ambiguity and size}
\label{SecSpecAlgs}
In this section we discuss a two different measures of the ambiguity and the size of an SSF,  
that instantiate the function \MEAS of the ARS specialization algorithm of section~\ref{SCEBL}.
The first instantiation is the based on Information Theoretic measures; it is theoretically
the better instantiation. And the second one is based on straightforward heuristic measures 
of length and ambiguity that are less expensive to compute; this is the practically more
attractive instantiation.
\subsection{Entropy minimization algorithm}
\label{SecEntMin}
In this section we derive an algorithm which mirrors the
assumptions made by the specialization algorithm presented in section~\ref{SCEBL},
and employs {\em entropy}\/ as a measure for ambiguity. This derivation introduces new
assumptions necessary for arriving at a computationally attractive formula.

The goal of the specialization algorithm is to learn a specialized grammar,
which is on the one hand satisfactorily small, and on the other hand satisfactorily
less ambiguous than the BCG. This can be implemented in a constrained optimization 
algorithm, which tries to minimize ambiguity while satisfying the requirements on size.
To express this optimization algorithm, we observe that the ambiguity of a sentence with
respect to a given grammar can be seen as uncertainty about what structure should the
grammar assign to that sentence. The measure of entropy, introduced to grammar specialization 
by Samuelsson~\cite{Samuelsson94}, is strongly associated with the concept of uncertainty. 
Therefore, it is a candidate for measuring the ambiguity of a sentence with respect to
a given grammar. However, for computing the entropy of a sentence with respect to a given
grammar, we need first to extend that grammar with probabilities that are computed from 
relative frequencies collected from a tree-bank. 

Assume for the time being that we know how to assign suitable probabilities
to grammars in order to measure ambiguity by entropy; we will come back to this issue a bit later.
Let $Cor$ denote the tree-bank sentences 
\mbox{$S_{1}\cdots S_{N}$}, and let $TB$ denote the tree-bank trees \mbox{$T_{1}\cdots T_{N}$}.
The constrained optimization problem can be stated as follows\footnote{Note the close 
resemblance of this optimization problem and the Bayesian interpretation of the 
Minimum-Description Length (MDL) Principle.
See section~\ref{SecDiscussion} on this issue further.}:
\begin{eqnarray}
\left\{
\begin{tabular}{rcl}
$SG$  & = & $argmin_{G} H(Cor | G)$ \\
$G$   & : & $ L(G) \leq \chi$  
\end{tabular}
\right.
& &
\label{FLH}
\end{eqnarray}
where $H(Cor | G)$ expresses the entropy of the sentences of the
tree-bank given that grammar, $L(G)$ expresses the size of grammar $G$,
and $\chi$ is an upper-bound on the sizes of the grammars considered in
the optimization. This optimization algorithm expresses the
wish to find the least ambiguous grammar of which the size is less 
than~$\chi$. 
%
%

Optimization algorithm~\ref{FLH} concerns measures of
ambiguity and size of grammars, while the function \MEAS is based on measures of
SSFs. To see how this algorithm can be fitted to SSFs, we concentrate our derivation 
first only on \mbox{$H(Cor | G)$} and come back later to $L(G)$ and 
how to determine~$\chi$.
%
\paragraph{Derivation for $H(Cor | G)$:}Next we will derive a sequential-covering
 algorithm that approximates algorithm~\ref{FLH}. To start, we note the assumption 
 of independence between the sentences of the tree-bank:
\[ H(Cor | G) = \sum_{i=1}^{N} H(S_{i} | G). \]
The next assumption concerns restricting the search space to the
TSG-space of the tree-bank. Since the CFG underlying the tree-bank
is in that space and forms a good starting point, we may as well express this
in our algorithm as the wish to improve on it, i.e. search for a
less ambiguous grammar. However, we need the stochastic version of
that CFG which enables measuring the ambiguity of an SSF. 
This is the version which assigns a probability to every rule 
{\em conditioned on the right-hand side of that rule}\/ (rather than the left 
hand side as in SCFGs, since we intend to measure the ambiguity of SSFs,
which are constructed from the right-hand sides of rules).
This stochastic CFG is denoted CFGS in order to stress the conditioning of its
probabilities on the right hand sides of rules.
Thus, the entropy optimization formula in algorithm~\ref{FLH} is approximated as:
\begin{equation}
 argmin_{STSG} \sum_{i=1}^{N} (H(S_{i} | STSG) - H(S_{i} | CFGS)) 
\label{InvCFGP}
\end{equation}
\newcommand{\STSGU}{STSG^{\cup}}
Note that the term $H(S_{i} | CFGS)$ is a constant that does not affect the
optimization. However, in the next derivation of a greedy approximation of this optimization
algorithm, the term $H(S_{i} | CFGS)$ is going to become essential for expressing a 
more attractive approximation. We state it here already only for convenience.

Let $x$ denote the number of iterations of the Sequential Covering EBL
specialization algorithm. Our next assumption is that each STSG 
in the search space is the union of ambiguity-sets of SSFs. For simplicity 
we also assume that the algorithm learns only a single SSF at each iteration.
This assumption about the learned STSG  $STSG$ can be expressed by
\mbox{$STSG = \bigcup_{1\leq j\leq x} [[ssf_{j}]]$} and denoted by\footnote{
This defines only the set of elementary-trees of the STSG.
Its start-symbol, set of non-terminals and terminals are those of the
CFG underlying the tree-bank.
} $\STSGU$.

Assume that $TB_{0}$ and $Cor_{0}$ denote respectively the original tree-bank 
and the corresponding corpus of sentential-forms.
After iteration $j$, for all \mbox{$1\leq j\leq x$}, 
the algorithm learns $ssf_{j}$ and its ambiguity set $[[ssf_{j}]]$, and reduces 
(bottom-up) the current tree-bank partial-trees $TB_{j-1}$ and sentential-forms 
$Cor_{j-1}$ to result in respectively $TB_{j}$ and $Cor_{j}$. 
Let \mbox{$Cor_{j-1} - Cor_{j}$} denote the difference between the sentential-forms 
before iteration $j$ and those in the situation after it, i.e. the part reduced by 
learning $ssf_{j}$.
Analogous to this, let \mbox{$TB_{j-1} - TB_{j}$} denote the difference between 
the partial-trees before iteration $j$ and those in the situation after it, i.e. the part 
reduced by learning $[[ssf_{j}]]$.

Assume independence between $ssf_{j}$ and the other sentential-forms in
$Cor_{j}$, and between $[[ssf_{j}]]$ and the other partial-trees in $TB_{j}$,
for all $j$. Due to this assumption we can write, for all $G$:
\[\forall j: H(Cor_{j-1} | G) = H(Cor_{j} | G) + H(Cor_{j-1} - Cor_{j} | G) \]
Taking all iterations into consideration and knowing that $TB_{x}$ and $Cor_{x}$ 
are empty, we can translate the sequential covering nature of our algorithm into
entropy notation:
\[H(Cor_{0} | G) = \sum_{j=1}^{x}~ H(Cor_{j-1} - Cor_{j} | G) \]
By writing \mbox{$Cor_{j-1} - Cor_{j}$} as \mbox{${\Delta Cor}_{j}$},
and assuming independence between $ssf_{i}$ and $ssf_{j}$ and between $[[ssf_{i}]]$
and $[[ssf_{j}]]$, for all \mbox{$0\leq i < j \leq x$}, we can write for the
STSG $SG$, which our algorithm learns:
\[ H(Cor_{0} | SG) = \sum_{j=1}^{x}~ H({\Delta Cor}_{j}~|~[ssf_{j}]) \]
where $[ssf_{j}]$ denotes an STSG rather than a mere set of partial-trees\footnote{ 
The STSG $[ssf_{j}]$ has:
\begin{itemize}
\item as start symbol a newly introduced non-terminal $SSS$, 
\item as set of elementary-trees it assumes the union of
     \begin{itemize} 
       \item the set $R([[ssf_{j}]])$, where $R$ is
         a unique renaming of each symbol in a partial-tree in $[[ssf_{j}]]$, 
         except for the frontier symbols, 
       \item a new set of rules, each of the form \mbox{$SSS$\lra $N$}, where
         $N$ is the (renamed) root of some partial-tree in $R([[ssf_{j}]])$,
     \end{itemize}
\item as set of non-terminals, $SSS$ together with all symbols of the 
         partial-trees in $R([[ssf_{j}]])$ except for those in the 
         sequence $ssf_{j}$ (i.e. the common frontier of the partial-trees),
\item and as set of terminals all symbols in $ssf_{j}$.
\end{itemize}
}.
By incorporating the latter facts into optimization algorithm~\ref{InvCFGP}, we
obtain:
\begin{equation}
SG = argmin_{\STSGU} \sum_{j=1}^{x}~ H({\Delta Cor}_{j}~|~[ssf_{j}]) 
                                     - H({\Delta Cor}_{j}~|~CFGS)
\label{OptBySTSGU}
\end{equation}
To simplify even further, we incorporate a greedy search strategy, i.e. minimize
during each iteration rather than on the total sum of all iterations.
%
\begin{equation}
SG = \bigcup_{j=1}^{x} argmin_{[ssf_{j}]}~ 
        H({\Delta Cor}_{j}~|~[ssf_{j}]) - H({\Delta Cor}_{j}~|~CFGS)
\label{OptBySSF}
\end{equation}
Note that here the term $H({\Delta Cor}_{j}~|~CFGS)$ is not a constant but
an important factor in this greedy optimization. Without it, the greedy optimization
would result in extreme overfitting: by learning at each iteration a minimum entropy
SSF, entropy equal to zero, the algorithm learns a tree of the tree-bank.
And since this is not what we want from this greedy strategy, a good way to prevent
this is to improve on the CFGS underlying the tree-bank as expressed by 
formula~\ref{OptBySSF}.

To summarize, the derivation arrives at optimization algorithm~\ref{OptBySSF},
which represents our specialization algorithm. Now we still need to explain 
how to compute the term $H(Cor_{j-1} - Cor_{j} | XXX)$, for 
\mbox{$XXX\in \{CFGS, [ssf_{j}]\}$}. And that is exactly what we do next.\\

Recall that $Cor_{j-1} - Cor_{j}$ denotes the part of $Cor_{j-1}$ which
was reduced by learning $ssf_{j}$, i.e. by reducing all occurrences of the
partial-trees in $[[ssf_{j}]]$). If we assume independence between all
these occurrences then:
\[ H(Cor_{j-1} - Cor_{j} | XXX) = Freq_{c}(ssf_{j}) \mul H(ssf_{j} | XXX) \]
Let $ssf$ be the string $w_{1}\cdots w_{n}$ and let $H_{XXX}()$ be
the conditional entropy $H(~|XXX)$. Now consider the general case of an 
$\STSGU$~$XXX$ and denote with $T^{N}_{ssf}$ a partial-tree $T$ with $ssf$
as frontier and $N$ as root label, where $T$ is derivable within $XXX$. 
Then we can write:
\begin{equation}
 H_{XXX}(ssf) = \sum_{T} H_{XXX}(T^{N}_{ssf}) 
\label{EntrOfssf}
\end{equation}
\newcommand{\elemtree}{N\stackrel{\tau}{\longrightarrow}~N_{k}N_{l}}
\newcommand{\elemtreeT}{N\stackrel{\tau}{\longrightarrow}~w_{t}}
\newcommand{\elT}[3]{T_{w_{#2}\cdots w_{#3}}^{#1}}
\newcommand{\subt}[3]{#1(w_{#2}\cdots w_{#3})}
This is to say that the entropy of the SSF according to grammar $XXX$ is the 
sum of the entropies of the trees assigned to it by $XXX$.

For grammar $[ssf_{j}]$, the computation of this entropy is direct:
\begin{equation}
 H_{[ssf_{j}]}(ssf_{j}) = \sum_{t\in [[ssf_{j}]]} H_{[ssf_{j}]}(t) 
\label{EntrOfssfSSF}
\end{equation}
where the definition of entropy is the common definition:
\[H_{XXX}(Y) = - P_{XXX}(Y) \mul \log{P_{XXX}(Y)} \]
and where we employ the definition\footnote{Note that this is equal to 
$ASD_{ssf_{j}}(t)\mul CP(ssf_{j})$.}:
\begin{equation}
 P_{[ssf_{j}]}(t) = \frac{Freq(t)}{Freq(ssf_{j})} 
\label{ProbSSF}
\end{equation}

For other grammars than $[ssf_{j}]$, in particular $CFGS$,
we develop a different computation.
It can be assumed without loss of generality that we are dealing only with 
Chomsky Normal Form (CNF) STSGs. We can write the term in
equation~\ref{EntrOfssf} in a recursive manner employing the 
elementary-trees (i.e. rules) of the STSG $XXX$. 
In CNF there are too cases:
\begin{description}
\item [Binary:] elementary-trees with two non-terminals on the
         rhs.  Denote such an elementary-tree with $\elemtree$, 
         where $\tau$ is a unique label for each elementary-tree of $XXX$ 
         and $N$, $N_{k}$ and $N_{l}$ are non-terminals of $XXX$. Then:

  \[ H_{XXX}(ssf) = \sum_{j,k,l,\tau} H_{XXX}(~N(~\subt{N_{k}}{1}{j} ~\subt{N_{l}}{j+1}{n})~) \]
  or in other words:
  \begin{eqnarray*}
   H_{XXX}(ssf) & =  & \sum_{j,k,l,\tau} H_{XXX}(\subt{N_{k}}{1}{j}) +\\
               &    & H_{XXX}(\subt{N_{l}}{j+1}{n}~|~\subt{N_{k}}{1}{j}) +\\
               &    & H_{XXX}(\elemtree~|~\subt{N_{l}}{j+1}{n}~\subt{N_{k}}{1}{j}) 
  \end{eqnarray*}
   As usual in these cases, we assume independence between the two spaces of partial-trees
   under $N_{k}$ and $N_{l}$, and limited dependence of $\elemtree$ on them
   two, i.e. only on their roots. This way we arrive at the point of recursion
   in our computation:
  \begin{eqnarray}
  H_{XXX}(ssf) & \approx & \sum_{j,k,l,\tau} H_{XXX}(\elemtree | N_{k}N_{l}) +\nonumber \\
               &         & ~~~~~~~ H_{XXX}(\subt{N_{k}}{1}{j}) + \nonumber\\
               &         & ~~~~~~~ H_{XXX}(\subt{N_{l}}{j+1}{n})~ 
  \label{BinaryEqn}
  \end{eqnarray}
\item [Terminal:] elementary-trees with a single terminal symbol
      on their rhs:
   \[ H_{XXX}(\elemtreeT) = H_{XXX}(\elemtreeT | w_{t}) + H_{XXX}(w_{t}) \]
    And since $w_{t}$ is given, its entropy is zero and we arrive at:
   \begin{eqnarray}
    H_{XXX}(\elemtreeT) & = & H_{XXX}(\elemtreeT | w_{t}) 
   \label{TermEqn}
   \end{eqnarray}
\end{description}
The above recursive computation of $H_{XXX}(ssf)$,
expressed in equations~\ref{BinaryEqn} and~\ref{TermEqn},
takes place on the parse-space of $ssf$ spanned by $XXX$.  

The probabilities involved in computing these entropies
are defined as follows:
\begin{eqnarray}
 P_{XXX}(\elemtree | N_{k} N_{l}) & = & \frac{Freq(\elemtree)}{Freq(N_{k} N_{l})} \nonumber\\
 P_{XXX}(\elemtreeT | w_{t}) & = & \frac{Freq(\elemtreeT)}{Freq(w_{t})} 
\label{ProbAmbEqn}
\end{eqnarray}
where $Freq$ denotes the frequency in the given tree-bank.\\

Now that we know how to compute the entropy of an SSF and we are aware of the
assumptions made by our specialization algorithm, we need to deal with the
question of how to measure size of grammars. This is our next derivation.

\paragraph{Computing $L(G)$:}
A suitable measure of the size of a grammar might be the optimal expected
description length according to Shannon (see chapter~\ref{CHDOPinML}).
This measure is independent of any coding-scheme and provides an independent
estimate of the size of the grammar.
 
The grammars $G$ dealt with in our algorithm are all STSGs of the training tree-bank.
Moreover, in the sequential covering scheme, the algorithm learns 
ambiguity sets of SSFs rather than single trees, thereby constraining
the space to the STSGs denoted~$\STSGU$. 

\DEFINE{Grammar size:}
{The size of a grammar $\STSGU$ is the sum of the lengths of 
 each of the ambiguity-sets that constitute it.
}
%
%
Let $L([[ssf_{j}]])$ denote the length of the ambiguity set of 
$ssf_{j}$, the SSF learned at iteration $j$ of the algorithm.
%
  %
Then:~
%
$L(\STSGU) = \sum_{j=1}^{x} L( [[ssf_{j}]] )$,\/ 
where $\STSGU = \bigcup_{j=1}^{x} [[ssf_{j}]]$. 
  %
  %
The quantity $L([[ssf_{j}]])$ is defined as the sum of the lengths of
all partial-trees in $[[ssf_{j}]]$, i.e.\/
$ L( [[ssf_{j}]] ) = \sum_{t\in [[ssf_{j}]]} l(t) $,\/
where $l(t) = -\log P(t)$ and $P(t)$ is the probability of $t$ among all
possible partial-trees of the tree-bank, or alternatively:
\begin{eqnarray}
 L([[ssf_{j}]]) & = & - log P([[ssf_{j}]]) \nonumber \\
 P([[ssf_{j}]]) & = & \frac{Freq_{C}(ssf_{j})}{\sum_{ssf\in Corp} Freq_{C}(ssf)} 
\label{SizeProbEqn}
\end{eqnarray}
Note here that we employ Shannon's optimal code length for defining $L$. 
   %
%
\paragraph{Combining ambiguity and size:}The combination of the results of the
above derivations leads to the constrained optimization: 
\begin{eqnarray}
\left\{
\begin{tabular}{l}
$SG$  =  $\bigcup_{j=1}^{x} argmax_{[ssf_{j}]}~ 
        H({\Delta Cor}_{j}~|~CFGS) - H({\Delta Cor}_{j}~|~[ssf_{j}])~~~~~~$ \\
 \\
$L(SG)$ $\leq$ $\chi $ 
\end{tabular}
\right.
& &
\label{SizeAndAmbA}
\end{eqnarray}
where the length \mbox{$L(SG) = \sum_{j=1}^{x} - \log{P([[ssf_{j}]])} $}, the entropy $H$
is defined in equations~\ref{TermEqn}, ~\ref{BinaryEqn} and~\ref{EntrOfssfSSF}
using the probabilities defined in equations~\ref{SizeProbEqn}, \ref{ProbAmbEqn},
and~\ref{ProbSSF}, and $\chi$ is an a priori set upper-bound on the size of the learned grammar.

The upper-bound $\chi$ is still on the size of the whole grammar. Since our
algorithm learns in a sequential covering iterative manner, it would be more
convenient to express the size-constraint locally on the size of  
$ssf_{j}$ rather than on the whole grammar. Therefore, we assume a sequence
$\chi_{j}$ of upper-bounds on the sizes of SSFs, each for an iteration 
of the algorithm:
\begin{eqnarray}
\left\{
\begin{tabular}{l}
$SG$  =  $\bigcup_{j=1}^{x} argmax_{[ssf_{j}]}~ 
        H({\Delta Cor}_{j}~|~CFGS) - H({\Delta Cor}_{j}~|~[ssf_{j}])~~~~~~ $\\
 \\
~~~$\forall j$  :  $- \log{P([[ssf_{j}]])} \leq \chi_{j}$ \\
\end{tabular}
\right.
& &
\label{EqnSizeAmb}
\end{eqnarray}

The question now is how to determine $\chi_{j}$ in a sensible manner.
One way to do so is to make an informed estimate of the average size
of $[[ssf_{j}]]$ by inspecting the training tree-bank.
This involves determining two values: the number \mbox{$| [[ssf_{j}]] |$} of partial-trees 
in $[[ssf_{j}]]$ and their average probability $AP_{j}$ in the training tree-bank.
For example, we could say that, on average, the partial-trees should be expected 
to be as probable as \mbox{$AP_{j} = 1\%$} and that \mbox{$| [[ssf_{j}]] | = 2$},
thereby setting \mbox{$\chi_{j} = - 2\mul \log{0.01}$}. 

\subsubsection{Entropy-Minimization and MDL}
\label{SecDiscussion}
In this subsection we clarify the relationship between the
Entropy-Minimization algorithm and the Minimum Description Length (MDL) principle,
and highlight some of the approximations that it embodies. This is done in the following
list of issues.
\begin{itemize}
\item Consider again algorithm~\ref{FLH} and its approximation algorithm~\ref{EqnSizeAmb}. 
      Both algorithms can be expressed slightly
      differently in a less constrained form as follows:
\begin{eqnarray}
\begin{tabular}{rcl}
$SG$  & = & $argmin_{G} H(Cor | G) + ( L(G) - \chi )$  
\end{tabular}
& &
\label{FLHMDL}
\end{eqnarray}
\begin{eqnarray}
\left\{
\begin{tabular}{rl}
$SG$~= $\bigcup_{j=1}^{x} argmin_{[ssf_{j}]}~$ & $[~H({\Delta Cor}_{j}~|~[ssf_{j}])~-~ $\\
        &~~$H({\Delta Cor}_{j}~|~CFGS)$~] ~{\bf +}~ \\
        &~~$[~-\log{P([[ssf_{j}]])} - \chi_{j}~]$
\end{tabular}
\right.
& &
\label{MDLLike}
\end{eqnarray}
This shows much similarity to the Bayesian interpretation of
the Minimum Description Length (MDL)~\cite{MDLPrinciple} principle. By removing
$\chi$ and $\chi_{j}$ from the sum, we obtain an algorithm which tries to minimize
the sum of the size and the ambiguity measures. 
This is a very interesting and theoretically attractive algorithm which has one 
single disadvantage: it has to scan the whole space of grammars rather than a 
constrained space. 
\item The derivation of the Entropy-Minimization algorithm can be expressed in the terminology
      of the Bayesian Learning paradigm, i.e. with the Bayes formula as a starting
      point. The Bayesian derivation is parallel to the present derivation and does not
      add any news. Therefore, we do not work it out here. 
\item
An interesting aspect of optimization algorithm~\ref{EqnSizeAmb} is that it
enhances the search by improving on the $CFGS$ underlying the tree-bank. Consider
the optimization algorithm which searches the whole space:
\begin{eqnarray}
\left\{
\begin{tabular}{rcl}
$SG$ & = & $\bigcup_{j=1}^{x} argmin_{[ssf_{j}]}~ H({\Delta Cor}_{j}~|~[ssf_{j}])$ \\
& & \\
$\forall j$ & : & $- \log{P([[ssf_{j}]])} \leq \chi_{j}$ \\
\end{tabular}
\right.
& &
\label{NoCFGP}
\end{eqnarray}
It might seem that algorithm~\ref{EqnSizeAmb} is in fact 
algorithm~\ref{NoCFGP} but involving an extra
``constant" $H({\Delta Cor}_{j}~|~CFGS)$. However, this is not the case since 
\mbox{${\Delta Cor}_{j}$} depends on $ssf_{j}$. This implies a major difference
between these, essentially similar, algorithms: the first profits from a 
better search start point, which enables it to arrive faster at a better
result (it is better equipped to avoid some of the local minima, which 
algorithm~\ref{NoCFGP} might fall into). 
\end{itemize}

%
\subsection{Reduction Factor algorithm}
Among the measures of the ambiguity of an SSF $ssf$ one clearly can 
identify the Constituency Probability (i.e. $CP(ssf)$) measure;
it expresses the probability that $ssf$ is a constituent.
The earliest and simplest algorithm within the ARS framework,
presented in~\cite{MyECMLWkSh97,MyACLWkSh97}, relies on this simple 
measure of ambiguity. The function \MEAS is defined in this
algorithm as follows:
\begin{tabbing}
~~~~~~~\= \bIf~ ($CP(ssf) \leq \delta$)~ \bThen~ \MEAS$(ssf) \defEqual 0$;\\
       \> \bElse~ \MEAS$(ssf) = GRF(ssf)$;
\end{tabbing}
where $GRF$ is called the Global Reduction Factor, a measure of
size of the grammar, which will be defined below, and $\delta$ is
a probability threshold chosen prior to training\footnote{
In some of the implementations, the threshold $\delta$ is allowed 
to vary during iterations of the specialization
algorithm; starting from the value $1.0$, $\delta$ will be reduced 
by a fixed amount whenever the algorithm does not learn new SSFs
any more, until the value of $\delta$ reaches a lower bound 
$\theta$ set beforehand. 
}. This definition
simply says that we prefer SSFs which are SSFs
in at least \mbox{$\delta\mul 100$\%} of their occurrences in the training 
tree-bank. This
way, during parsing an input, if the specialized grammar recognizes an SSF,
 the chance that it is not an SSF is less than~$1-\delta$.

The measure $GRF(ssf)$ expresses the amount by which an $ssf$ is able 
to reduce the tree-bank if it is learned by the specialization algorithm.
We employ this measure to express our preference for reducing the tree-bank
trees the fastest way, thereby expressing a wish for a smaller grammar.
When a subtree associated with an SSF in a tree in the tree-bank is
reduced to its root, we obtain a partial-tree which has a frontier
shorter than the original's tree frontier by the length of the SSF minus
one. This is exactly the Reduction Factor ($RF$) of the SSF:
\DEFINE{Reduction Factor:}{ $RF(ssf) \defEqual |ssf| - 1$ }
Since an SSF is reduced simultaneously in all places where it 
appears in the whole tree-bank, in fact it reduces the tree-bank 
by the amount equal to its Global Reduction Factor:
\DEFINE{Global RF:}{
$GRF(ssf) \defEqual RF(ssf) \mul Freq_{C}(ssf)$
}
A reminder: $Freq_{C}(ssf)$ expresses the frequency of the sequence
$ssf$ as an SSF in the tree-bank, i.e. the sum of the frequencies of
all subtrees associated with $ssf$.

\paragraph{Discussion:}
In words, this algorithm learns the smallest
grammar of which the SSFs are not more ambiguous than a predefined measure.
The definition of \MEAS does not take into account the Ambiguity-Set Distribution
of an SSF (defined in section~\ref{SCEBL}).
This means that the measure of ambiguity in \MEAS is not optimal.
Moreover, weighing the two measures of ambiguity and size of an
SSF in \MEAS is not possible. Nevertheless, as the experiments in the next section
exhibit, it is an effective, simple and conceptually clear algorithm.
%
\subsubsection{A back-off approximation of GRF} 
The function \MEAS$()$ is context-free in the sense that its value does not depend
on the context of its parameter; for any sequence of symbols $S$: \mbox{\MEAS$(S) > 0$} iff 
\mbox{$CP(S) > \delta$}.
      Often however, due to data-sparseness,  the context-free requirement that 
      \mbox{$CP(S) > \delta$} is too rigid. Therefore, the implementation of the 
      GRF-based learning algorithm employs a back-off technique on local context in 
      the computation of the value of \MEAS$(S)$. Here the local context is limited to 
      two grammar symbols to the left and two to the right of the sequence~$S$. 
   
      Let the operator ``$\cdot$" denote the infix concatenation operator on sequences of symbols.
      To redefine \MEAS, we introduce the following definitions:
      \begin{itemize}
      \item Let $C_{TB}(S)$ denote the set of all pairs (contexts) $\tuple{LC,RC}$, where $LC$
            and $RC$ are each a sequence of two grammar symbols, with which sequence $S$ is 
            encountered in tree-bank~$TB$. 
      \item Let $GC^{i}(S)$ denote the set of pairs $\tuple{GLC,GRC}$, where $0\leq i\leq 4$ 
            and \linebreak 
            $\tuple{GLC,GRC}$ is obtained from a context $\tuple{LC,RC}\in C_{TB}(S)$ by 
            replacing exactly $i$ symbols in \mbox{$LC\cdot RC$} by the wild-card symbol $*$.
            Note that $GC^{4}(S)$ is the singleton set $\{\tuple{**,**}\}$ and that $GC^{0}(S)$ is
            in fact $C_{TB}(S)$. 

            If \mbox{$G = \tuple{GLC,GRC}$} is obtained from \mbox{$I=\tuple{LC,RC}$} by replacing 
            symbols with wild-cards, $G$ is called {\em a generalization}\/ of $I$ and 
            $I$ is called {\em an instance}\/ of $G$. Two contexts are called {\em unrelated}\/ 
            iff they are neither generalizations nor instances of one another.
            A context $C1$ is called {\em more general}\/ than context $C2$ iff $C1$ 
            contains more wild-cards than $C2$.
      \item Let $Viable(LC,S,RC)$  denote the proposition that is true iff the frequency 
            of $\tuple{LC\cdot S\cdot RC}$ exceeds the threshold $\Phi$
            and \mbox{$CP(LC,S,RC) > \delta$}, where $CP(LC,S,RC)$ is the Constituency 
            Probability of the sequence of symbols \mbox{$LC\cdot S\cdot RC$}. 
      \item $MGUVC(S)$ is the set of Most General Unrelated Viable Contexts 
            of a sequence of symbols $S$, defined by\/
            $\tuple{LC,RC}\in MGUVC(S)~~ {\bf iff}$ 
            \begin{enumerate} 
              \item there is some $0\leq i\leq 4$ such that $\tuple{LC,RC}\in GC^{i}(S)$, 
              \item $Viable(LC,S,RC)$ is true, 
              \item \mbox{$\forall \tuple{LC1,RC1}\in MGUVC(S)$}: if $LC\neq LC1$ or $RC\neq RC1$
                    then $\tuple{LC,RC}$ and $\tuple{LC1,RC1}$ are unrelated, and 
              \item for every~$i$ and every $\tuple{LC1,RC1}\in GC^{i}(S)$, if $Viable(LC1,S,RC1)$ 
                    is true then $\tuple{LC1,RC1}$ is an instance of or identical to 
                 $\tuple{LC,RC}$.
            \end{enumerate}
      \item Let
            \MEASG$(LC,S,RC)$ denote the following generalization of the original \MEAS$(S)$,
            where $\tuple{LC,RC}\in GC^{i}(S)$ for some \mbox{$0\leq i\leq 4$}:
\begin{tabbing}
~~~~~~~\= \bIf~ ($Viable(LC,S,RC)$ == false)~ \bThen~ \MEASG$(LC,S,RC) = 0$;\\
       \> \bElse~ \MEASG$(LC,S,RC) = GRF(LC\cdot S\cdot RC)$;
\end{tabbing}
         Note that if $\tuple{LC,RC}\in GC^{i}(S)$ for $i > 0$, then the frequencies of 
         \mbox{$\tuple{LC\cdot S\cdot RC}$} are the sums of the corresponding frequencies
         of the instances of $\tuple{LC,RC}$ in the set $GC^{0}(S)$.
      \end{itemize}
\noindent Now we redefine the function \MEAS by:~~\/ 
        $ \MEASM(S) = \sum_{\tuple{LC,RC}\in MGUVC(S)} \MEASGM(LC,S,RC)$. 
      This redefinition sums over the GRF values of all unrelated most general contexts
      of the sequence~$S$. Note that in the case \mbox{$MGUVC(S) = \{\tuple{**,**}\}$} the value
      of this context-sensitive function $\MEASM(S)$ is equal to that of the 
      original $\MEAS(S)$. And in case $\tuple{**,**}$ is not $Viable$ then the new definition of
      \MEAS(S) backs-off to the set $MGUVC(S)$ of unrelated most general contexts 
      of $S$ that are viable.

\section{Summary and open questions}
\label{CHARSConcs}
In this chapter I presented a new framework for specializing broad-coverage
grammars (BCG) and DOP-probabilistic-grammars to specific domains represented by
tree-banks. The framework,
Ambiguity Reduction Specialization (ARS), specifies the requirements that
specialization algorithms must satisfy in order to guarantee any degree of
success in learning specialized grammars. Of these requirements two are
central: 1)~a specialized grammar must be able to span all structures expected
to be associated with any constituent it is able to recognize, and 2)~the
specialized grammar must be less ambiguous than the BCG it specializes and
must have a size which does not cancel the gain from its ambiguity reduction.
I also provided an analysis of preceding work on specialization and concluded that 
these efforts are less suitable for specializing probabilistic performance models.
I also presented two new learning algorithms based on the ARS framework, namely the 
Global-Reduction Factor (GRF) algorithm and Entropy Minimization algorithm. 
In addition, I discussed how to integrate the learned specialized grammar and the 
original BCG into a novel parsing algorithm. 

In this chapter I also encountered problems and questions for which no solutions
are provided in this thesis. These problems might constitute the subject of future work
on the present framework:
\begin{itemize}
\item An interesting aspect of the present learning and parsing algorithms is
      that a domain-language is partially modelled by a finite set of SSFs and their
      ambiguity-sets. It is evident that in practice the present learning algorithms
      face a problem: due to the huge number of different lexical entries (i.e. words)
       it is hard to imagine them as part of SSFs. Thus, generally speaking SSFs will 
      not contain words of the language. Moreover, a complication that all learning
      methods face is that no tree-bank whatsoever contains all the lexical entries that 
      are probable in the domain. Thus, the use of a lexicon in some way or another is inevitable.
      Here we envision that the presence of a lexicon provides information that might
      enable local-disambiguation of the ambiguity sets of SSFs. Methods for learning
      this kind of local-disambiguation in the presence of a lexicon provide, not
      only theoretically but also in practice, a way to {\em lexicalize}\/ the ARS framework. 
      A related issue is the issue of learning from richer descriptions that contain e.g.
\item In this work we assume that the tree-bank contains analyses of sentences. These
      analyses can be syntactic but also more elaborate descriptions involving
      e.g. semantic formulae or feature-structures. For feature structures in particular,
      the ambiguity-sets of the learned grammar would consist of partial-analyses that 
      also contain partially instantiated feature-structures learnt from the tree-bank. 
      The issue of how to uninstantiate feature-structures during learning has been 
      exemplified by earlier work on grammar specialization e.g.~\cite{Neumann94,Srinivas97}.
\item As discussed in section~\ref{ParAlgSch}, the specialized grammar can be 
      implemented as a Cascade of Finite State Transducers (CFSTs). There have been
      many earlier attempts at parsing using 
      CFSTs~\cite{EjerhedChurch83,Ejerhed88,Koskenniemi90,Abney91,Hindle94}, 
      all specifying
      the transducers manually. The CFSTs implementation of a specialized grammar 
      represents a finite language. This language can be extended by 
      generalizing over the SSFs of the CFST with a Kleene-star, as in the
      work of~\cite{SrinivasJoshi95}. This generalizes over the BCG. 
      Open questions in this regard  are:
       \begin{itemize}
         \item What is the expected coverage and accuracy of a CFSTs implementation
               compared to the current TSG implementation~?
         \item How does post-learning inductive generalization over the learned SSFs improve
               coverage and accuracy~?
               How does pre-learning inductive generalization of the tree-bank annotations
               (i.e. the BCG) enable acquiring generalized SSFs automatically,
               and how does this affect coverage and accuracy compared to an
               TSG implementation~?
       \end{itemize}
%
%
\item Given an SSF $ssf$ of a tree-bank $TB$ representing some domain $X$,
       what is the probability that a new sentence of domain $X$
       (i.e. not in $TB$) is assigned a parse $t$ in which $ssf$ is the frontier of a 
       subtree $st$ of $t$ such that $st$ is not in the ambiguity-set 
       of $ssf$ over $TB$~?.
\end{itemize}


\chapter{Efficient algorithms for DOP}
\label{CHOptAlg4DOP}
%
Since its birth in~\cite{Scha90}, Data Oriented Parsing was considered by many as an unattainable
goal; interesting but only as a theoretical account of probabilistic disambiguation.
For many it was unthinkable that parsing natural language could take place on basis
of such huge probabilistic grammars. Bod was the first to run experiments~\cite{RENS1,RENSDES}
on the small Penn-Tree-bank ATIS (total of 750 trees). Bod's experiments raised conflicting
feelings: the reported accuracy results were ``good enough" to warrant the 
effort of looking into the model, but the fact that parsing one sentence took on 
average 3.5~hours seemed to support the argument that DOP is not feasible. 
In the absence of more efficient alternatives, Bod conducted these experiments using a 
Monte-Carlo algorithm~\cite{RENSMonteCarlo}, an approximation based on iterative random sampling, 
on Sun~Sparc~2 machines\footnote{
On the machines that are currently available (e.g. SGI~Indigo2) the average 3.5~hours 
per sentence would be reduced to about 20~minutes per sentence.}.
In retrospect, Bod can be credited for one thing: ``daring to look the beast in the eyes
with so much patience". 

Nevertheless, daring as they were, Bod's limited and unstable experiments 
exhibit only the theoretical possibility of the Data Oriented Parsing approach.
They demonstrate neither its feasibility nor its usefulness for real-life applications. 
In the world of applications, the natural habitat for performance models such as DOP,
there are various other requirements besides correct modeling. 
Efficiency is one major requirement: given 
the current limitations on space and time, what is the scope of applicability 
(e.g. language-domains, input-size, kind of application) of a given method~?

For DOP to be viable in the applications-world, it is necessary to have the 
algorithmic means that facilitate more efficient disambiguation. These algorithmic means
can be expected to go through various stages of development before they can achieve their
full potential. In this process of development, both the algorithms and the model are expected
to evolve: while the algorithms become more efficient, often they also reshape their models 
through acquiring new insights into these models. 

This chapter thus considers the question of how to disambiguate under DOP in 
an efficient manner. It provides efficient algorithmic solutions to some problems 
of disambiguation under DOP; typical to all these solutions is that they are, unlike
Monte-Carlo parsing, {\em deterministic polynomial-time and space}.
The present algorithms are ``pure" solutions, in the sense 
that they do not incorporate any assumptions or approximations that do not originate from
the DOP model. To improve the behavior of these algorithms in practice, this chapter
also presents approximations and heuristics that proved very effective and useful in
empirical studies of the DOP model. Some of these heuristics and approximations moderately
reshape the DOP model.
%
%
%
\section{Motivation}
An important facility of probabilistic disambiguation is that it enables the selection of 
a single analysis of the input. Parsing and disambiguation of an
input under the DOP model can take place by means of maximization of the probability of some 
entity (in short ``maximization-entity"), e.g. for input sentences DOP could be made to 
select the Most Probable Parse (MPP), Most Probable Derivation (MPD) etc.. 
A central question in research on DOP has become the question: what entity should one
maximize~?
\subsection{What should we maximize~?}
In his presentation of the DOP model, Bod~\cite{RENSDES} singles out the MPP as the 
best choice for disambiguation under the DOP model when parsing input sentences. 
According to Bod, the MPP is, at least theoretically, superior to the MPD. This statement
is supported by some empirical results on the small Penn-TreeBank ATIS domain
(750 trees in total)~\cite{RENSDES,BodScha96}. 

The choice of the MPP, rather than the MPD, as maximization-entity is unique among 
existing probabilistic parsing models. Apparently this is due to the major role that SCFGs always 
played in these models: in SCFGs, there is no such difference between the MPD and the MPP
(since every derivation generates exactly one unique parse-tree). Theoretically speaking,
the MPP is superior to the MPD in the sense that it reduces the probability of commiting
errors. However, as we prove in chapter~\ref{CHComplexity}, the problem of computing the
MPP under STSGs is an NP-Complete problem. And, as we argue below, the existing 
non-deterministic algorithms (particularly Monte-Carlo parsing~\cite{RENSDES})
do not constitute practical solutions to this problem.
The crucial question that rises here is: is it always necessary to maximize the
theoretically most accurate entity~? In other words, are there situations when parsing
sentences where maximizing another entity than the MPP, e.g. the MPD, is sufficient~?

To answer this question we note first that in any situation the desired maximization 
entity should commit as little errors as possible. Generally speaking, error-rates are measured 
with respect to {\em subsequent interpretation of an analysis}; eventually, the interpretation 
module is the one that determines how fine-grained an analysis should be. And in its turn,
the amount of information contained in an analysis determines the kind of measure of error-rate
(e.g. exact-match, labeled bracketing precision).
In its turn, the measure of error determines the maximization entity that might be sufficient 
for the situation at hand. Clearly, in parsing sentences, whatever interpretation module 
we employ after parsing, the MPP remains the one that minimizes the chance of errors.
But there might be another entity that achieves as good (or comparable) results when the 
measure of error is less stringent than exact-tree match.

As a concrete example, consider the task of (non-labeled) bracketing
sentences. In such a task, one aims at minimizing the errors in assigning a bracketing 
to the input sentence. To this end, it is sufficient to maximize an entity that 
minimizes the chance of committing a wrong bracket. Although the MPP achieves this
goal perfectly, a less complex maximization entity achieves it equally well. 
This argument is the main motivation behind this and other work on developing efficient
algorithms for DOP~\cite{Goodman96,GoodmanDES}. 
\subsection{Accurate + efficient $\approx$ viable}
Bod's Monte-Carlo algorithm computes only {\em approximations}\/ of the 
Most Probable Derivation (MPD) and the Most Probable Parse (MPP) of an input sentence.
For achieving good approximations, it is necessary to apply the sampling procedure iteratively;
a quality of Monte-Carlo is that the result approaches the MPP/MPD as the number of iterations 
becomes larger. Bod claims that Monte-Carlo approximation is  {\em non-deterministic
polynomial-time}~\cite{RENSDES}; the polynomial is cubic in sentence length and square 
in error-rate. However, Goodman~\cite{GoodmanDES} argues\footnote{
Although the argument in~\cite{GoodmanDES} does not constitute a full proof, it is sound and
convincing. Goodman also constructed an efficient implementation of the Monte~Carlo algorithm.
For the small ATIS tree-bank~\cite{HemphilEtAl90} with 770~trees, Goodman  shows that 
this implementation runs (although exponential-time) as fast as his polynomial-time DOP 
parser (see below). Although this is a remarkable achievement, it is clear that this does
not necessarily scale up to larger tree-banks and longer sentences.
} that the error-rate in the time-complexity of the Monte-Carlo algorithm
is actually not a constant but rather a function of sentence length; 
this implies that Monte-Carlo parsing has in fact exponential time-complexity. 
Chapter~\ref{CHComplexity} supplies a proof that
the problem of computing the MPP for an input sentence under DOP is indeed NP-Complete,
making the Monte-Carlo algorithm one of the few ``brute-force" possibilities available. 

The fact that the Monte-Carlo algorithm obtains good approximations only when the number
of iterations is relatively large, makes it more of a prototyping-tool rather than 
a practical alternative. Future developments in multi-processor 
parallel computing and high-speed computing might change this but {\em only to a limited 
extent}. Given the fact that there will always be larger applications and larger domains,
it will always remain necessary to develop alternative efficient algorithms that constitute 
good solutions. As far as we can see, the presence of constraints on space and time is an
invariant in the evolution of computer programs. 
Thus, if we wish to apply DOP within the given time and space limitations
rather than wait for a miracle to happen, we are obliged to develop efficient algorithms
for {\em as accurate as possible}\/ disambiguation under DOP. Of course, we are also obliged 
to remain aware of the ultimate accuracy of the original DOP model and strive for improving
the efficiency of the algorithmic means for achieving it.
\subsection{Efficient algorithms}
The goal of the present chapter is to present efficient and accurate algorithms for
disambiguation under large STSGs. We present: 
\begin{itemize}
 \item algorithms for computing the MPD under STSGs of an input 1)~parse-tree, 2)~sentence, 
       3)~FSM and 4)~SFSM (i.e. word-graph - see chapter~\ref{CHDOPinML}).
 \item algorithms for computing the probability under STSGs of an input 1)~parse-tree,
       2)~sentence, 3)~FSM and 4)~SFSM.
\end{itemize}
These algorithms are based on the same general algorithmic scheme.
Common to the present algorithms and other algorithms developed earlier within the
Tree-Adjoining\linebreak
Grammar (TAG) framework (see section~\ref{SecOthersAlgs})
is that they are all based on extensions of parsing and recognition algorithms that 
originally stem from the CFG tradition. 
The present algorithms differ from the algorithms developed in the TAG framework
in that they are two-phase (rather than single phase) algorithms. They are most suitable
for large DOP STSGs: their superiority is enhanced as the ratio between the size of 
the DOP STSG and the size of the CFG underlying it becomes larger, a typical
situation in natural language DOP STSGs.

In short, the present algorithms have two phases. The first phase spans a good 
{\em approximation}\/ of the parse-space of the input; this approximation is spanned 
by parsing under a simplified CFG that is an approximation of the CFG underlying the 
DOP STSG. The second phase computes the maximization entity (and the probabilities) 
on the parse-space that was spanned in the first phase.
Crucially, the parse-space spanned in the first phase constitutes a substantial limitation of 
the space that the DOP STSG would have to explore when parsing from scratch. Moreover,
in a typical DOP STSG, the CFG employed in the first phase is very small relative to the
large DOP STSG. Therefore, the costs of the first phase in space and time are negligible
relative to the second phase, and the second phase is computed in linear time-complexity
in grammar-size with as little ``failing" derivations as possible\footnote{
A common problem in parsing algorithms are actions taken by the algorithm that 
eventually do not lead to parses of the whole input but only to parses of part of the input.
These are often seen as failing derivations that cost time. Serious effort has been put into
reducing the failing derivations in CFG parsing as the next section explains.
}. The sum of the time/space costs of the two phases constitutes a substantial improvement
on the alternative one-phase algorithms. \\

The present algorithms compute the exact values rather than approximations as Monte-Carlo does.
Moreover, except for the algorithm for computing the MPD for an input sentence, none of these 
algorithms have been developed before within the Monte-Carlo framework.
In fact, for some of the above listed entities, it is highly questionable whether the 
Monte-Carlo algorithm can be adjusted to provide good approximations in reasonable 
computation-times (e.g. computing the probability of a parse-tree, a sentence 
or a word-graph).

An essential quality of the present algorithms, is that they have time-complexity 
linear in STSG size and cubic in input length. Crucially, linearity in STSG 
size is achieved without sacrificing memory-use, a typical situation in all other alternative 
algorithms that could be used for parsing DOP, 
e.g.~\cite{CKY,Earley,SchabesJoshi88,SchabesWaters93}. In section~\ref{CKYViterbi} I explain
why in the case of DOP STSGs the Cocke-Kasami-Younger (CKY) algorithm~\cite{CKY,AHO} does not 
succeed in providing
a good compromise between time and space needs for achieving time-complexity 
linear in STSG size. The same argument carries over to the other 
alternative algorithms that are listed above.\\ 

This chapter is structured as follows. Section~\ref{SecOthersAlgs} lists references to
and discusses related work on parsing and disambiguation under tree-grammars.
Section~\ref{CKYViterbi} mainly provides the background
to the rest of this chapter: it discusses CKY parsing under CFGs and the Viterbi optimization 
for computing the MPD/MPP of input sentences under SCFGs. It also argues that
the CKY algorithm and the Viterbi for CFGs are not suitable for parsing the huge DOP STSGs.
Section~\ref{SecOptAlg} presents deterministic polynomial-time algorithms and 
optimizations thereof.
Section~\ref{SecHeuristics} presents useful heuristics for improving the time- and
space-consumption of DOP models.
And finally section~\ref{SecConcs} summarizes the conclusions of this chapter.
%

\section{Overview of related work}
\label{SecOthersAlgs}
Historically speaking, parsing Tree-Grammars has been developed mainly within the 
TAG framework. Most work has concentrated on the recognition problem under TAGs 
e.g.~\cite{VijayJoshi85}, \cite{VijayThesis}, \cite{Vijay}. To a much smaller extent, 
some work concentrated on probabilistic disambiguation and probabilistic training 
of some restricted versions of STAGs, e.g.~\cite{SchabesWaters93}. 

The algorithm presented in~\cite{SchabesWaters93} can be used to compute the MPD
of an input sentence under STSGs. Similarly, existing SCFG algorithms based
on CFG parsing algorithms~\cite{ViterbiCFG,Jelinek} can be extended to compute the
MPD for an input sentence under STSGs, as the next section explains. However,
as I argue in that section, none of these algorithms can 
achieve time-complexity linear in STSG size without resulting in huge memory-use.
Moreover, since none of these algorithms is really tailored for the huge DOP STSGs, 
none of them suitably exploits the properties that are typical to these STSGs.\\ 

%
Recently, Goodman~\cite{Goodman96,GoodmanDES} presented a new DOP projection mechanism:
rather than projecting a DOP STSG from the tree-bank, Goodman projects from the tree-bank an
equivalent SCFG. The projected SCFG is equivalent to the DOP STSG of that 
tree-bank in the sense that both generate the same parse-space\footnote{
Actually, up to node renaming, there is a homomorphism between the trees of the two grammars.}
and the same parse-probabilities. However, the derivation-spaces of the STSG and the SCFG 
differ essentially. The SCFG implementation of DOP (or DOP SCFG) does not have the notion 
of a subtree and thus prohibits the computation of such entities that depend on subtrees 
e.g. the MPD (of the STSG). Moreover, as we proved in chapter~\ref{CHComplexity}, computing
the MPP of a sentence as DOP prescribed is NP-hard under STSGs; switching to Goodman's 
SCFGs, of course, does not change this.
However, the SCFG implementation enables fast parsing and computation of other maximization 
entities of an input sentence under the SCFG implementation of DOP.

Besides the SCFG projection mechanism, \cite{Goodman96,GoodmanDES} presents efficient algorithms 
for computing new maximization entities under the DOP SCFG and studies the plausibility of Bod's
empirical results. The main maximization entity in that work concerns the so-called General 
Labeled Recall Parse (GLRP); the GLRP is the 
parse-tree which maximizes the {\em expectation value of the Labeled Recall rate}\footnote{
A labeled constituent is a  triple $\tuple{i,j,A}$, where $w_{i}^{j}$ are the words covered by
the constituent and $A$ is the label of that constituent.
Let be given a parse-tree $T$ selected by a parser for an input sentence and also the correct 
parse-tree $CT$ (i.e. the ultimate goal of the parser) for that sentence.
The Labeled Recall rate is the ratio between the number of correct labeled constituents 
in $T$ and the total number of nodes in $CT$. Roughly speaking, Goodman's algorithm computes
for every labeled constituent in the chart of the input sentence a ``weight";
the weight of a given labeled constituent is the ratio between {\em the product of its 
Inside and Outside probabilities}\/ and the total probability of the input sentence. 
Subsequently, Goodman's algorithm computes the GLRP as that candidate parse (of the input 
sentence) which has the maximum sum of ``weights" of all its labeled constituents.
}. Goodman shows that on the small ATIS, the GLRP achieves exactly the same accuracy as the 
MPP of the DOP STSG (which is equal to the MPP of the DOP SCFG).

A nice property of Goodman's SCFG instantiation of DOP is that it employs the whole
training tree-bank {\em as is}\/ for parsing; this is equal (in the sense mentioned above)
to the DOP STSG as defined in chapter~\ref{CHDOPinML}. This nice quality is also the bottleneck 
of Goodman's implementation\footnote{
Goodman's SCFG has approximately $8\mul N\mul T$ rules, where $T$ is the number of trees
in the tree-bank and $N$ is the average number of words per sentence. For a relatively
small tree-bank of 10000 trees, on average 6 words per sentence, the SCFG has~480000 rules
(about half a million). 
}. Goodman does not offer a way to limit the grammar's subtrees;
this is an essential property for obtaining DOP models under larger tree-banks and 
for avoiding sparse-data effects.  An attempt\footnote{
In personal communication, Goodman has shown that there are such possibilities but that
the number of rules of the SCFG remains very large. Limiting the substitution-sites 
is in fact only a theoretical possibility rather than a practical one.
} at limiting the depth and the number of substitution-sites 
(see section~\ref{SecHeuristics}) of subtrees in Goodman's SCFG, 
immediately results in large SCFGs (hundreds of thousands of different rules). 
As section~\ref{CKYViterbi} explains, such large SCFGs currently form a serious problem 
for parsers.

Although Goodman's algorithms provide efficient solutions to some problems of
disambiguation under DOP, Goodman's solutions are only suitable for DOP models that 
are represented as SCFGs. In this chapter, I provide efficient solutions to other 
disambiguation problems under STSGs in general, and in particular under DOP STSGs.
%
%

\section{Background: CKY and Viterbi for SCFGs}
\label{CKYViterbi}
In this section I discuss the classical CKY
algorithm for (deterministic) polynomial-time parsing of CFGs.
I argue that direct use of this algorithm for parsing DOP STSGs 
is impractical because it does not successfully compromise the two 
conflicting requirements of extensive memory-needs and high computation costs.
The direct application of the CKY algorithm for DOP is therefore
only a theoretical possibility.
\subsection{CFG parsing with CKY}
For the definition and notation of CFGs the reader is 
referred to chapter~\ref{CHDOPinML}. Some extra definitions and assumptions are necessary
here:
\begin{description}
\item [CNF:] 
  To simplify the presentation of CKY we assume CFGs in Chomsky Normal Form (CNF);
  a CFG is said to be in CNF iff each of its rules
  is in one of two possible forms: the binary form $\RuleM{A}{B C}$ or 
  the terminal form $\RuleM{A}{a}$.
\item [Sizes:] $|X|$ denotes the cardinality of a given set $X$.
       $|\alpha|$\ denotes the number of symbols in string $\alpha$.
       And $|G|$ denotes the size of a CFG $G$=\CFG, not necessarily in CNF,
       where by definition:
        $|G|$=\mbox{$\sum_{\RuleM{A}{\alpha}\in \RulesN}  |\alpha| + 1$}.
\item [Finitely ambiguous:]
   Throughout this work we assume that the CFGs we deal with are {\em finitely ambiguous}\/
   i.e. the CFG derives any finite string it can recognize only in a finite
   number of different left-most derivations\footnote{This is equivalent to saying that the
   CFG does not allow cycles, e.g. $A$\lmpdir$A$.}. 
\item [Items:] In parsing theory, an item is a CFG rule with a dot inserted 
            somewhere in its right-hand side,
             e.g. \ITEM{A}{\alpha}{\beta} where $\RuleM{A}{\alpha\beta}\in$ \Rules. 
            Items \ITEM{A}{\alpha}{} are called final. 
            Roughly speaking, in general, the dot in the item
            is a marker, which identifies the stage of a derivation beginning from the rule
            underlying the item:
            $\alpha$ is the part already recognized and $\beta$ is the part that still 
            has to be recognized. In the following characterization of the CKY algorithm
            we show exactly what the dot marks.
\end{description}
The CKY algorithm \cite{CKY} is one of the earliest and simplest algorithms
for polynomial-time recognition of arbitrary Context-Free Languages (CFLs). CKY
employs a data-structure, called well-formed substrings table, with entries $[i,j]$ for
all $0\leq i < j\leq n$, where $n$ is the number of terminals in the input
sentence. An entry is a store of items that are introduced during the parsing process.
Let \mbox{$w_{1}\cdots w_{n}$} denote the input sentence, and let $w_{0}=\epsilon$.
The invariant of CKY is to maintain the following condition for every 
entry $[i,j]$ in the table at all times:
\begin{verse}
  \ITEM{A}{\alpha}{\beta}$\in [i,j]$ iff 
  \mbox{$\RuleM{A}{\alpha\beta}$ \lmdir $w_{i}^{j}\beta$}\/ holds in the CFG at hand.
\end{verse}
This condition implicitly explains the notation of an item; CKY recognizes the 
input sentence iff there is an item \ITEM{S}{\alpha}{} in entry $[0,n]$. 

A specification\footnote{
This specification is rather inefficient. It does not optimize the order 
of steps of the CKY parser. 
} of the CKY algorithm is given in figure \ref{FigCKYAlg}.
For more information on the CKY and on how to implement it efficiently, the reader is 
referred to \cite{AHO,GrahamEtAl80}.
\begin{figure}
\hrule
\begin{tabbing}
{\bf Init:} \=                                           \\
            \>$\forall$~ \= $0 < i \leq n$  	 \\
            \>           \> $\forall$ \= ~$\RuleM{A}{w_{i}} \in$ \Rules      \\
            \>           \>           \> ~~~add \ITEM{A}{w_{i}}{} to $[i-1,i]$ \\
            \>           \> $\forall$ \= ~$\ITEMN{B}{\beta}{} \in [i-1, i]$  \AND~ $\forall~ \RuleM{A}{B C} \in$ \Rules \\
            \>          \>            \> ~~~add \ITEM{A}{B}{C} to $[i-1, i]$  \\
{\bf Deduce:}\=                  					\\
             \>$\forall~ 0 \leq i < n$       			\\
             \>$\forall$~\= $i < j \leq n$			\\
             \>          \> ~~~$\forall$\= ~$i < k < j$  \AND~  $\forall$~ \ITEM{A}{B}{C}$\in [i, i+k]$ 		\\
             \>          \>             \> ~~~~~\bIf~ $\exists$ \ITEM{C}{\alpha}{} $\in [i+k, j]$\\
             \>          \>             \> ~~~~~\bThen~ add \ITEM{A}{BC} to $[i, j]$ \\
             \>          \> ~~~$\forall$ \= ~$\ITEMN{B}{\beta}{} \in [i, j]$  \AND~ $\forall~ \RuleM{A}{B C} \in$ \Rules \\
             \>          \>              \> ~~~~~add \ITEM{A}{B}{C} to $[i, j]$  \\
\end{tabbing}
\hrule
\caption{CKY algorithm for CFGs in CNF}
\label{FigCKYAlg}
\end{figure}
As figure~\ref{FigCKYAlg} shows, the CKY {\em recognition}\/ algorithm has time-complexity 
proportional to $|G|\mul n^{3}$. However, to achieve time-complexity linear in grammar 
size $|G|$ it is necessary to be able to decide on the membership condition 
               ``whether \mbox{$\exists$ \ITEM{C}{\alpha}{}$ \in [i+k, j]$}"
in $O(1)$. This is easily achieveable using very efficient implementations of entries
as arrays. Note that such an array, however, has to be of size $|V_{N}|$\ in order to
allow this. Good hashing functions approximate this behaviour using smaller arrays or 
lists.
\begin{figure}
\begin{center}
\mbox{
\epsfxsize=138mm
\epsfbox{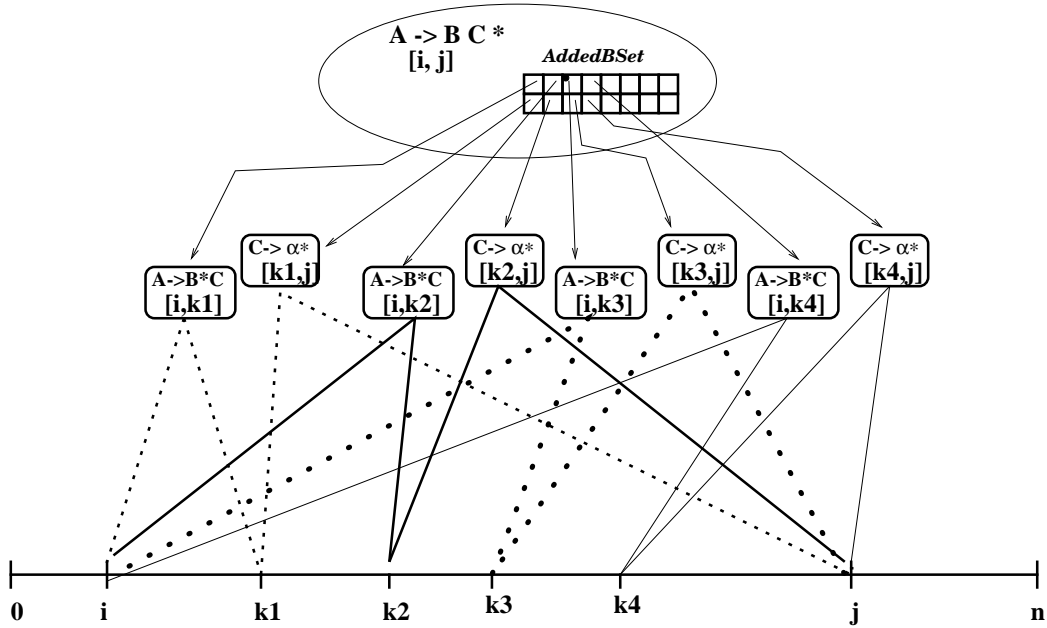}
}
\end{center}
\caption{Illustration of a possible implementation of a parse-forest} 
\label{FigAddedBySet}
\end{figure}

The recognition algorithm is in itself not very useful since it does not offer an 
efficient way for retrieving parse-trees for the input sentence. For this reason, 
the algorithm is usually extended at the {\bf Deduce} phase as follows: when adding item 
$I$ to entry \entry{i}{j}, a set of ``pointers", referred to as
$AddedBySet_{I, i, j}$, is attached to item $I$ (see illustration in figure~\ref{FigAddedBySet}). 
Each pointer in $AddedBySet_{I, i, j}$
points to one or more items, in this or other entries, that resulted in adding $I$
to \entry{i}{j} during the {\bf Deduce} phase of the algorithm. 
Note that a final binary item \ITEM{A}{BC}{}
is added to \entry{i}{j} by pairs of items \ITEM{A}{B}{C} $\in$ \entry{i}{i+k} and 
\ITEM{C}{\gamma}{} $\in$ \entry{i+k}{j}; therefore, $AddedBySets$ of final binary items 
are always pairs of pointers. For simplicity we refer to 
$AddedBySet_{I, i,j }$ as the $AddedBy$ set of $I$. The CKY table together with the $AddedBy$
sets is called a parse-forest, a compact and nice representation of the parse-space (set of
all parse-trees) of the input sentence. A depth-first traversal of the pointers, 
starting at any item \mbox{\ITEM{S}{\alpha}{}$ \in [0, n]$}, leads to retrieving parses
for the input sentence.

A complication, which occurs in implementing a CKY that constructs a parse-forest
in time linear in~$|G|$ is exactly in maintaining the $AddedBy$ sets of binary items. For the item
\ITEM{A}{BC}{}$ \in [i, j]$, there are usually many items \ITEM{C}{\beta}{}$ \in [k, j]$,
for each $i < k < j$ and any $\beta$, that resulted in adding it to the table. Moreover,
usually there are many items \ITEM{C}{\beta}{} in every entry \entry{k}{j} 
that result in adding \ITEM{A}{BC}{} into \entry{i}{j}.
For achieving time-complexity linear in grammar size it is necessary
to maintain for every non-terminal $C$ and for every \entry{k}{j}, \mbox{$i < k < j$},
a set of all items \ITEM{C}{\beta}{}, for any $\beta$ value, in a special set, which
we denote with $Final(C, k, j)$.
Again, this can be achieved only if we implement every entry $[k, j]$ as an array of 
length $|V_{N}|$. At array-entry $C$, of table entry $[i, j]$, we gather 
together all items \ITEM{C}{\beta}{}. Even good hashing-functions can lead to 
sacrificing linearity in grammar-size just because a ``bucket" (in the hash-table) 
containing \ITEM{C}{\beta}{} might
also contain other items \ITEM{D}{\gamma}{}, making the construction of the $AddedBy$ sets,
during {\bf Deduce}, go deep into these buckets in order to filter them. This can lead, 
in the worst case, to time-complexity \mbox{$|G|^{2} \mul n^{3}$}.

The space-complexity of the CKY parser for CFGs is \mbox{$O(|G|^{2} \mul n^{3})$}
(this includes the $O(|G|\mul n^{2})$ recognition and for every item in the recognition
 chart there can be at most $O(|G|\mul n)$ pairs of items in its $AddedBy$ list).
Usually, very efficient storage methods such as bit-vectors can be used
to implement table entries and items, in order to minimize the actual 
memory-consumption for large $|G|$ values. However, these methods have a 
hidden overhead (retrieval time-costs) that might defeat their utility in many 
cases.

For general CFGs, the CKY algorithm has some  known disadvantages:
\begin{itemize}
 \item For many grammars, the transformation of the CFG into CNF 
       results in expanding the grammar dramatically. In some cases,
       the size of the CNF grammar is square the size of the
       original grammar $|G|$~\cite{GrahamEtAl80}. This sets the algorithm
       back at time-complexity proportional to $|G|^{2}$. 
       \cite{GrahamEtAl80} offers solutions for this but in practice, as I argue below, 
       these solutions cannot avoid this problem for very large CFGs.
 \item The characterization of the CKY algorithm given above means that
       an item \ITEM{A}{\alpha}{} is added to \entry{i}{j} regardless of
       whether it participates in a derivation of the string 
       \mbox{$w_{k}^{i} w_{i}^{j} w_{j}^{l}$}, for any \mbox{$0 \leq k < i < j < l \leq n$}.
       Consequently, many of the items added to the table are ``useless" since they do
       not contribute to derivations of the whole input sentence.
       The run-time and memory costs, which these ``useless" items
       imply, can be reduced to a certain degree with simple optimization methods
       discussed also in~\cite{GrahamEtAl80}.
\end{itemize}
Next I argue that, in practice, for real-life DOP STSGs, the above two 
disadvantages imply either very slow speeds or very huge memory costs or both. 
\subsection{Computing MPP/MPD for SCFGs}
An algorithm for the computation of the Most Probable Parse (MPP) (and at the same time 
the Most Probable Derivation - MPD) based
on CKY for SCFGs is presented in~\cite{ViterbiCFG,Jelinek}. The computation
does not alter the time or space complexity of the CKY. It is based on
the Viterbi~\cite{ViterbiAlg} observation that two partial-derivations, 
starting from the same root non-terminal $N$ and ending in the same 
portion $w_{i}^{j}$ of the input sentence,
can be extended to derivations of the whole sentence exactly in the same ways. 
Therefore, if the one partial-derivation has a lower probability than the other,
it can be discarded. 

Be given an SCFG~\SCFG~ and an input sentence $w_{1}^{n}$.
Let \mbox{$MaxP(item, i, j)$} denote the probability of the MPP (MPD) starting
from an item \mbox{$item \in$ \entry{i}{j}}. Figure~\ref{FigMPDCFGs} shows the 
specification of an extension of the CKY algorithm for computing $MaxP$ for
every item in the CKY table. The operator $\sumMPD$ computes the maximum of a set
of reals. The probability of the MPP of the input sentence $w_{1}^{n}$ is then
found as \linebreak 
\mbox{$\sumMPD \SETM{MaxP(\ITEMN{S}{\alpha}{}, 0, n)}{\ITEMN{S}{\alpha}{} \in [0, n]}$}.
\begin{figure}
\hrule
\begin{tabbing}
$MaxP(\ITEMN{A}{a}{}, i-1, i) = ~P(\RuleM{A}{a})$ \\
\\
$MaxP(\ITEMN{A}{B}{C},$$i, j) =$
  $\sumMPD_{\beta}~ \SETM{MaxP(\ITEMN{B}{\beta}{}, i, j)}{\ITEMN{B}{\beta}{} \in [i, j]}$ \\
\\
$MaxP(\ITEMN{A}{B C}{~},$
\= $i, j) = P(\RuleM{A}{B C}) \mul$\\
\> $\sumMPD_{i < k < j}$ {\bf \{} \= $MaxP(\ITEMN{A}{B}{C}, i, k) \mul$ \\
\>                               \> $\sumMPD_{\gamma}$ \{$MaxP(\ITEMN{C}{\gamma}{}, k, j) ~|~\ITEMN{C}{\gamma}{} \in [k, j]$\} {\bf \}} \\
\end{tabbing}
\hrule
\caption{The MPD algorithm for CFGs in CNF}
\label{FigMPDCFGs}
\end{figure}

Note that in SCFGs a derivation and the parse it generates have exactly the same
probability simply because every parse can be generated by one single derivation.
The algorithm described above can be used for computing the probability of 
the input sentence by replacing every $\sumMPD$ operator with the $\sum$ operator. 
For computing the MPP (MPD) of an input sentence, the algorithm of figure~\ref{FigMPDCFGs}
should be extended with a storage that keeps track of all entities that result in
the maximum value in every $\sumMPD$ term (i.e. replace every $\sumMPD$ by \mbox{$arg max$}). 
\subsection{Direct application to DOP STSGs}
\label{SecLargePCFG}
An STSG \STSG\ can be seen as a SCFG, where each elementary-tree
$t$ is considered a production rule ``\Rule{root(t)}{frontier(t)}".
A problem can arise if two internally different elementary-trees result
in the same CFG production rule. To avoid this every elementary-tree
receives a unique address and the address is attached to the rule.
So if the number of elementary-trees in the STSG at hand is denoted by 
\TSGsize, the size of the rule set of the corresponding CFG is also \TSGsize.
Also let $|A|$ denote the number of internal nodes of all elementary-trees
of the STSG. In practice, for real-life DOP STSGs, the number of elementary-trees 
runs in the hundreds of thousands (these figures are obtained on 
limited domain tree-banks such as the OVIS~\cite{DOPNWORep} and 
the ATIS~\cite{HemphilEtAl90}). For future tree-banks and applications,
these numbers are expected to grow. In any case, let us consider what
happens when applying the CKY directly for such huge CFGs that are
obtained from DOP STSGs:
\begin{itemize}
 \item The CFG is not in CNF. As explained earlier,
       a transformation to CNF expands the number of rules and non-terminals
       by the size of that CFG. The size of the CFG, 
       defined above, is the sum of the lengths of its productions. For this
       CFG this is equal to \mbox{\TSGsize $\mul \mu(|rule|)$}, where $\mu(|rule|)$
       is the average length of a rule of the CFG. For linguistic annotations,
       \mbox{\TSGsize $\mul \mu(|rule|)$}
       is of the same order of magnitude as the total number of nodes
       in the DOP STSG's elementary-trees, denoted by $|A|$. 
       In the CKY, to achieve time complexity proportional to linear in $|A|$, 
       each entry of the algorithm must be represented as an array of length $|A|$,
       as the number of non-terminals of the CNF CFG is also in the order of magnitude 
       of $|A|$. In the current limited domain applications the number $|A|$ runs
       also in the hundreds of thousands. Clearly, maintaining such huge entries
       implies such memory-needs that are beyond those that are currently available,
       especially if the application involves large tables for long input sentences or 
       word-graphs (e.g. with up to 70~states).
       Therefore, linearity of the CKY in $|A|$, the STSG size,
       has to be sacrificed and we fall back to $|A|^{2}$ worst case complexity.
       A second issue in CNF transformation is that it may even result in squaring
       the size of the STSG~\cite{GrahamEtAl80}, a thing that also sets us back at the
       $O(|A|^{2} \mul n^{3})$ time-complexity.
 \item The second problem is the huge number of failing derivations in such
       a large grammar. As explained in the preceding section, this implies extra 
       work that leads to useless nodes, useless partial-derivations and useless
       partial-parses.
       Of course this is inevitable in CFG parsing but it is important to minimize its
       effects. Methods that use lookahead variables and rely on knowledge of the 
       grammar can reduce this a bit. But for such a large STSG, this remains a serious
       burdon that begs for an alternative solution.
\end{itemize}
This argues against direct application of the CKY parser for real-life
DOP STSGs. In section~\ref{SecOptAlg} I introduce an algorithm that exploits the
CKY algorithm but avoids these problems. The idea is to first parse with a
very small grammar which approximates the STSGs parse-space. And subsequently to
apply the large STSG on this parse-space in order to conclude the parsing
process with the parse-forest of the STSG.
%
\section{An optimized algorithm for DOP}
\label{SecOptAlg}
This section presents an alternative to the direct application of the CKY\linebreak
algorithm  \cite{CKY} for parsing DOP STSGs. Among the advantages of this alternative
is that this algorithm achieves an almost linear time-complexity 
in STSG-size with substantially less memory costs than the CKY algorithm.

The structure of this section is as follows. Subsection~\ref{SubSecTwoPhase} provides an 
introduction and some necessary definitions. Subsection~\ref{SubSecCNFapprox} describes a simple
approximation of the CNF transformation. Subsection~\ref{SubSecDerRecognition} discusses
the algorithm for recognizing STSG derivations of an input sentence, which constitutes the 
basis for the present optimization algorithm.
Subsection~\ref{SubSecCompMPD} presents the algorithm for 
computing the MPD of an input sentence and explains how with minimal changes it can be adapted 
for computing sentence probability and the MPD of an input parse-tree.
Subsection~\ref{SubSecOptimiz} presents the optimization
that leads to linear time-complexity in grammar-size. And finally, 
subsection~\ref{SecOptAlg4WGs} presents an extension of this optimized algorithm for
disambiguation of word-graphs.
\subsection{A two-phase parser}
\label{SubSecTwoPhase}
The algorithm is based on the idea that it is possible to span a good
approximation of the parse-forest of an STSG using a relatively small grammar. 
The exact computation of the STSG's parse-forest takes place on this approximated
parse-forest rather than from scratch. This results in reducing the number of useless 
derivations and the storage costs substantially. Subsequently, a simple optimization,
based on a nice property of sets of paths of STSGs and CFGs, brings the time complexity 
of the two-phase algorithm back to linear in STSG size, without extra 
storage costs.

The small grammar of the first phase is an {\em approximation}\/ of the CNF of the
CFG underlying the given STSG (shortly {\em original CFG}). 
Rather than transforming the elementary-trees
of the STSG into CNF in the usual manner, the transformation is simplified
such that the tree- and string sets of the CNF~CFG, underlying
the CNF~STSG resulting from the transformation,
are supersets of those of the {\em original~CFG}.
This simplified transformation has an advantage over the regular CNF transformation:
the size of the resulting CNF CFG is guaranteed to be of the same order of magnitude 
as that of the original CFG\footnote{Recall that the regular CNF transformation might
expand the number of non-terminals and rules drastically. We discuss this in detail below.}.
Crucially, transforming the STSG elementary-trees to CNF, in the simplified way, 
does not change its (weak and strong) generative power. A suitable correspondence between 
the CNF~CFG rules and the CNF~STSG elementary-trees' nodes allows applying the second 
phase of the parsing algorithm in an efficient manner.

Below we need the following definitions:
\begin{description}
\item [String language:] The {\em string-language}\/ of a CFG, SCFG, TSG or STSG $G$,
      denoted ${\cal L}(G)$, is the set of all strings of terminals that it can generate. 
\item [Tree language:] The {\em tree-language}\/ of a CFG, SCFG, TSG or STSG $G$,
      denoted ${\cal T}(G)$, is the set of all parse-trees that it can generate for all
      strings in its string-language.
\end{description}
Moreover, the following property of STSGs plays a central role in our discussion:
\begin{verse}
     {\sl The string/tree language of an STSG is always a subset of the string/tree 
     language of the CFG underlying~it.}
\end{verse}
Furthermore, for the rest of this section we assume that we are given a finitely 
ambiguous and $\epsilon$-free STSG $G_{stsg}$, and that the CFG underlying 
$G_{stsg}$, denoted $G_{cfg}$, is also finitely ambiguous and $\epsilon$-free.
%
%
\newcommand{\stsg}{$G_{stsg}$}
\newcommand{\cfg}{$G_{cfg}$}
\subsection{CNF approximation}
\label{SubSecCNFapprox}
The CNF transformation discussed here is a simple approximation of the
well-known CNF transformation of CFGs. Although often the ``traditional" CNF
transformation can also be applied to the CFG underlying the STSG without 
introducing too many rules, I prefer here the following approximation.

The transformation is applied to the elementary-trees of the STSG \stsg. 
Two observations underly this Approximated CNF (ACNF):
\begin{itemize}
 \item Every elementary-tree of the STSG is an atomic unit. 
       Every parser must treat the elementary trees as such, in one way or another, 
       in order to compute the parse-forest of the STSG.
 \item Based on the first observation, it does not matter in what way we transform the 
       internal structure of an elementary-tree into CNF, provided that we take care that 
       its atomicity is preserved. In the present parsing algorithm, the second phase takes 
       care of the atomicity of elementary-trees.
\end{itemize}
\begin{figure}
\begin{center}
\mbox{
\epsfxsize=125mm
\epsfbox{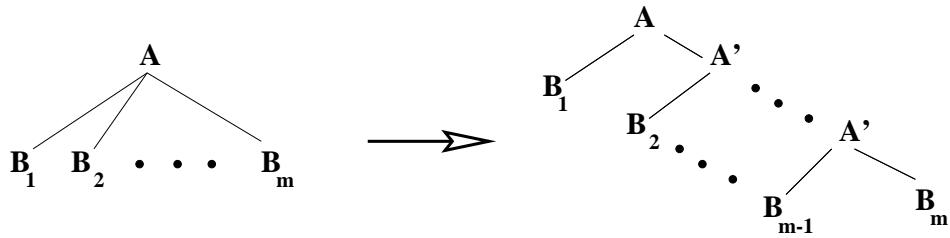}
}
\end{center}
\caption{An approximation of the CNF transformation} 
\label{CNFTrans}
\end{figure}
Given these two observations, the ACNF transformation starts by assigning to every 
elementary-tree of the STSG a unique address from a domain~$\Pi$.
During the ACNF transformation, every elementary-tree retains its address.
The ACNF transformation has the following steps:
\begin{itemize}
  \item Transform the elementary-trees of \stsg\/ to elementary-trees containing no
        unary productions. 
  \item Transform the elementary-trees, resulting from the preceding transformation,
        to a form where every node, 
        in every elementary-tree, has either children that are all non-terminals 
        or has one single terminal child. 
  \item Transform every resulting elementary-tree to CNF by traversing it
        from its root to its leaves in a breadth-first traversal. Under every
        node in the elementary-tree, encountered during the traversal, there is
        now a production rule; if the production rule is not in CNF then it
        must be a rule $\RuleM{A}{B_{1}\cdots B_{m}}$, where $m > 2$ and all 
        $B_{i}$ is a non-terminal. Transform this rule into a right linear (or left
        linear) structure by introducing exactly $m-2$ nodes labelled 
        with the \underline{same fresh symbol $A^{'}$}. Figure~\ref{CNFTrans} depicts
        this step.
\end{itemize}
The tree- and string-sets of the resulting CNF~CFG are supersets of 
the respective sets of the CNF~STSG.
In fact, the ACNF transformation results in a CNF~CFG that behaves in two different ways: 
internal to elementary-trees as a right-linear grammar (a regular grammar) 
and external to elementary-trees as a CFG.

It is important here to restate two simple facts, mentioned above, and sketch their proof 
informally:
\newtheorem{CHOptAlgA}{Proposition}[chapter]
\EExample{CHOptAlgA}
{The ACNF transformation results in a CNF~STSG with the same 
string language as the original STSG \stsg. Moreover, up to a simple 
reverse transformation, the tree language of the CNF~STSG is equal to that
of \stsg.}{CHOptPropA}
\DEFINE{Proof~\ref{CHOptPropA}:}
{The transformation alters only the {\em internal
structure}\/ of elementary-trees, i.e. it does not alter the root node or the leaf nodes
and their labels. Exactly the same derivations, substition sequences of elementary-trees
together with their unique addresses, are still possible in the CNF~STSG as the original STSG.
Thus the string language of the CNF~STSG is the same as that of the original STSG.
A simple reverse-transformation gives back the same tree language also. Text books
on parsing exhibit for the first two transformations 
reverse transformations, e.g.~\cite{AHO}. 
For the third step, a reverse transformation simply 
removes every node labeled with $A^{'}$, where $A$ is a non-terminal of the STSG
and uses the unique addresses in order to retrieve the original structures~~$\Box$
}
\newtheorem{CHOptAlgB}[CHOptAlgA]{Proposition}
\EExample{CHOptAlgB}
{The string/tree language of the CFG underlying the STSG \stsg, i.e. \cfg,
 is a subset of (respectively) the string/tree language of the CNF~CFG 
 (underlying the CNF~STSG resulting from the approximate CNF transformation).
}{CHOptPropB}
\DEFINE{Proof~\ref{CHOptPropB}:}
{It can be eaily observed that every derivation of \cfg~ is still possible in 
 the ACNF~CFG resulting from the transformation. However, the ACNF~CFG might have extra
 non-terminals introduced during the transformation. This introduces new derivations
 for (possibly) other strings and other trees~~$\Box$
}
As mentioned earlier, the string/tree language of an STSG is always a subset of 
(respectively) the string/tree language of the CFG underlying it. Together with the above two
facts this means that the tree/string language of the original STSG is respectively
a subset of the tree/string language of the CNF~CFG. 

The ACNF transformation introduces new rules. The number of the newly introduced
rules is at most equal to the sum of the lengths of the right~hand sides of the rules of
the CFG underlying the STSG, i.e. $O(|G_{cfg}|)$. The number
of non-terminals has grown but it is at most double that of the CFG \cfg.
In this way, we have a small CNF grammar of the same order of magnitude as
the original CFG \cfg. For linguistic CFGs, the right hand sides of rules are 
much shorter than the lengths of frontiers of elementary-trees of DOP STSGs.
And besides, some of the newly introduced rules have more chance to be identical than
in transforming the STSG into CNF, implying even a smaller grammar.
Most importantly, the ACNF transformation, even in the worst-case, does
not result in squaring the size of the CFG.  All in all, this approximation of 
the CNF results in a small grammar that can be parsed very fast using the CKY. 

The transformation results also in a CNF~STSG that has new nodes in its elementary-trees.
As explained above, the number of the new nodes is in the worst case equal to the length
of the frontiers of the elementary-trees, i.e. $O(|A|)$, where $|A|$ is the total
number of nodes in the elementary-trees of the STSG. This doubles the number of nodes
in the worst-case, but the CNF approximation introduces only $O(|G_{cfg}|)$ new
non-terminals (as opposed to $O(|A|)$ in the usual CNF transformation, as explained earlier). 
%
\subsection{STSG-derivations recognition}
\label{SubSecDerRecognition}
Be given an STSG \stsg\/ $=$ \STSG~ in CNF (the transformation to CNF is as
described above). The input sentence is denoted $w_{0}^{n}$, where 
\mbox{$w_{0} = \epsilon$}. The present algorithm has the structure:
\begin{description}
 \item [Phase 1.] Apply the CKY algorithm using the (CNF) CFG underlying
                  \stsg, denoted \cfg, resulting in a parse-forest of the input sentence.
                  As mentioned above, this parse-forest is a superset of the parse-forest
                  which \stsg\/ spans for the same sentence.
 \item [Phase 2.] Apply an algorithm, described below, for computing the parse-forest 
                  of \stsg\/ {\em from the approximate parse-forest}\/ of the previous phase,
                  and compute the MPD on this parse-forest.
\end{description}
Often, I will refer to the first phase with the general term ``parsing phase" 
and to the second phase with the term ``disambiguation-phase". Both terms do not 
reflect the exact task of each phase. They only reflect the main task of each
phase: (approximating) parse-forest generation in the first, and  
computation of the MPD, i.e. the selection of one preferred tree from the exact
parse-forest, in the second.

Below I discuss the second phase, which fulfils three requirements: preserve 
atomicity of \stsg's elementary-trees, preserve the uniqueness of every 
elementary-tree, and recognize exactly \stsg's parse-forest and derivations for
the input sentence from the CFG parse-forest. But first I discuss the issue of how to 
recognize a derivation of \stsg\ when given a parse-forest of \cfg.
The issue of recognizing an STSG derivation is central in the MPD-computation 
algorithm. But first the following definitions.
\DEFINE{Node address:}
{This is a unique address from some domain $\Pi$ assigned to a non-leaf node in an 
 elementary-tree of \stsg. 
}
Every non-leaf node in every elementary-tree of \stsg is assigned such a unique node-address 
from $\Pi$. It is useful if the addresses of the nodes of an elementary-tree enable fast 
checking of the parent-child (and the number of the child if counting from left to right)
and sisterhood relations between nodes in an elementary-tree. 
\DEFINE{Decorated tree:}
{A tree in which every non-leaf node is {\em decorated}\/ with exactly one address, from the
 STSG elementary-trees' node addresses, is called a decorated tree. The collection of
 node addresses in a decorated tree is called the {\em decoration} of that tree.
}
\DEFINE{Derivation-tree:}{Every derivation of the STSG can be characterized by a
          unique decoration of the tree it generates; in this decoration every node 
          in the tree is decorated with the unique address of the corresponding node
          of an elementary-tree. We refer to this decorated tree with the term 
          {\em derivation-tree}. 
}
\begin{figure}[tbh]
\hrule
\center{
\begin{tabular}{c}
\epsfxsize=90mm
\epsfbox{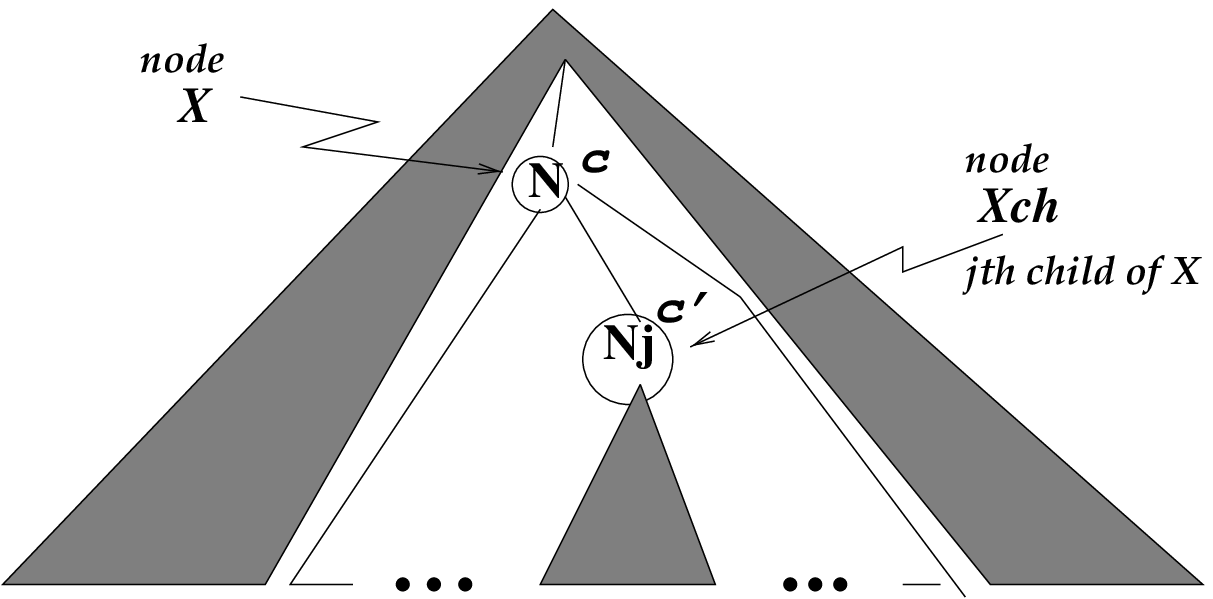} \\
\\
\\
\epsfxsize=130mm
\epsfbox{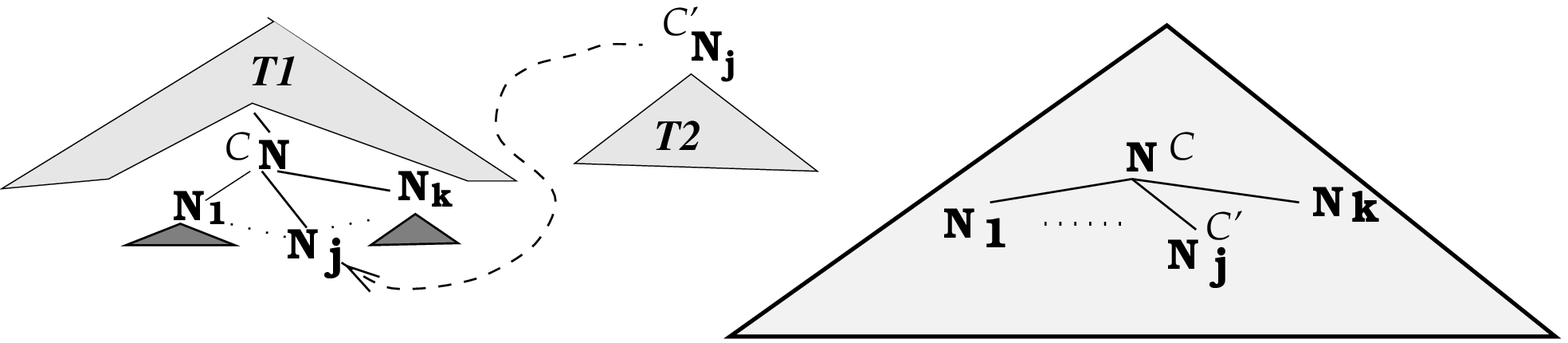} 
\end{tabular}
}
\hrule
\caption{(Top) decorated tree, (bottom) the viability property}
\label{FigViab}
\end{figure}
Note that not every decorated tree is a derivation-tree since some decorations
do not correspond to STSG derivations. We observe that a simple algorithm can be 
used for recognizing whether a decorated tree is a derivation-tree of \stsg\/ or not. 
To this end, we need to define the {\em viability property}, a main element of this
algorithm and the algorithm for computing the MPD.
\DEFINE{Viability property:}
{Let $X$ be a node in a decorated-tree, labeled $N$ and decorated with address $c$, 
 and let $Xch$ be its $j$th child-node, labeled $N_{j}$ and decorated 
 with address $c^{'}$, $j \in \{1,2\}$. We say that the viability property holds for 
 nodes $X$ and $Xch$ if one of the following two holds (see figure~\ref{FigViab}):
 \begin{description}
  \item [Parenthood:] There is some elementary-tree, in \stsg, for which 
       $c$ and $c^{'}$ correspond to two of its non-leaf nodes that are related (respectively)
       as parent and its $j$th child. Moreover the node corresponding to $c$/$c^{'}$ 
       in that elementary-tree is labeled respectively with $N$/$N_{j}$.
  \item [Substitution:] 
        There is some elementary-tree, in \stsg, for which the address $c$ corresponds 
        to one of its nodes that is labeled $N$ and that has a leaf child-node labeled 
        with a non-terminal $N_{j}$, i.e. a substitution-site.
        And there is some elementary-tree, in \stsg, for which the
        address $c^{'}$ corresponds to its root node that is labeled $N_{j}$. 
 \end{description}
}
The algorithm for recognizing whether a decorated tree is a derivation-tree of \stsg\/ 
is shown is figure~\ref{AlgRecDerT}; we will refer to it with the name 
algorithm~\ref{AlgRecDerT}. 
\begin{figure}[h]
\hrule
\vspace*{.3cm}
\begin{description}
  \item [Root:] Check whether the root of the tree is labeled $S$ and is decorated with
    an address which corresponds to the root of an elementary-tree.
    If not then goto {\bf Fail} below.
  \item [Iterate:] For every node in the decorated tree, that is not a leaf node, do:\\
          {\bf 1)}~let that node be the current node, \\
          {\bf 2)}~for the current node $X$, labeled $N$ and decorated 
         with address $c$, and its $j$th child-node $Xch$, labeled $N_{j}$ and decorated 
         with address $c^{'}$, $j \in \{1,2\}$, check that the viability property 
         holds for $X$ and $Xch$, otherwise goto {\bf Fail}.
  \item [Accept:] return accept and exit.
  \item [Fail:] return reject and exit.
\end{description}
\hrule
\caption{Algorithm~\ref{AlgRecDerT}: Recognizing a derivation-tree}
\label{AlgRecDerT}
\end{figure}
The following propositions express the role which the viability property plays 
in recognizing derivation-trees.
\newtheorem{CHOptAlgBB}[CHOptAlgA]{Proposition}
\EExample{CHOptAlgBB}
{If a decorated tree contains a node that does not fulfill the viability property,
 the decorated tree is not a derivation-tree of the STSG at hand.
}{CHOptPropBB}
\newtheorem{CHOptAlgBA}[CHOptAlgA]{Proposition}
\EExample{CHOptAlgBA}
{Algorithm~\ref{AlgRecDerT} recognizes a decorated tree 
 iff it is a derivation-tree of the STSG at hand. 
}{CHOptPropBA}
The proofs of these two propositions are simple to construct 
and will be skipped here. 

As a result to the propositions above, if given a parse-forest, spanned by 
the STSG \stsg\/ for some input sentence, it is possible to determine the set 
of all derivation-trees of the sentence by applying the above algorithm to the 
parse-forest. However, applying this algorithm
blindly would result in exponential time-complexity. It is necessary to apply it
in an efficient manner. Moreover, instead of applying the recognition of the
derivation-trees of a sentence to \stsg's parse-forest, we may apply it to the
CFG's parse-forest; this determines the derivation-trees of the sentence according
to the STSG \stsg\/ and, as a side effect, this delimits \stsg's parse-forest for
the sentence, since trees without \stsg's derivation-trees are not trees of \stsg. 

To obtain a (deterministic) polynomial-time algorithm for the recognition of the set
of derivation-trees of an input sentence, we employ a new compact structure, in analogy
to a parse-forest, called a {\em derivation-forest}\/ of the sentence. To this end, define:
\DEFINE{Rule occurrence:}
{Let rule \mbox{$R \in$ \Rules} be found under a non-leaf node with address $c$ of some
 elementary-tree. The address $c$ is called an occurrence of $R$ in that elementary-tree.
 An occurrence of an item \ITEM{A}{\alpha}{\beta} is an occurrence of the rule
 $\RuleM{A}{\alpha\beta}$.
}
\DEFINE{Occurrence set:}{
 The {\em occurrence set}, denoted $\An(R)$, of rule \mbox{$R \in$ \Rules}
 is the set of all occurrences of $R$ in all elementary-trees of \stsg.
}
\DEFINE{Item's occurrences set:}
{Let \ITEM{A}{\alpha}{\beta} $\in$ \entry{i}{j}. The {\em occurrences set}\/
of this item is defined as 
 $\An(\ITEMN{A}{\alpha}{\beta},~i,~j)$ = $\An(\RuleM{A}{\alpha\beta})$.
}
Given the CFG parse-forest and the CKY table of the input sentence, built by the CKY algorithm,
a decoration of the items in the table enables recognizing the derivation-forest
of that sentence. We maintain with every item its occurrence set.
The derivation-forest is constructed on top of the CFG parse-forest, also in a 
bottom-up fashion by applying an algorithm for recognizing derivation-trees to the 
occurrence sets of the items that are connected together in the parse-forest 
by the $AddedBy$ relation described in the previous section. 
Similar to our presentation of the Viterbi CKY, we do not show exactly how 
to construct the derivation-forest, instead we show how to compute the MPD of the input
sentence. The construction of the derivation-forest extends the MPD algorithm described
below only by associating sets of pointers between occurrences of items. Recall that
if an item $B$ results in adding another item $A$ to entry \entry{i}{j}, during {\bf Deduce}
of CKY, then  $B \in AddedBySet_{A, i, j}$, where this expresses a pointer from $A$ to
$B$. Similarly, a set is associated with every occurrence, $c$, of $A$, denoted 
$ViabilitySet_{c,A}$. \linebreak
If $B\in AddedBySet_{A, i, j}$ and $b$ is an occurrence of $B$,
then $b \in ViabilitySet_{c, A}$ iff a predicate $Viable?(c, b, j)$, 
defined below, is true; except for final binary items, $j$ is always equal to 1.
For final binary items, the value $j \in \{1, 2\}$ allows testing the predicate 
$Viable?(c, b, j)$ for the left (``first child")  or right (``second child") 
items in the pairs of items in $AddedBy$ of the final binary item.
The predicate $Viable?$ tests the viability property defined earlier for 
the algorithm~\ref{AlgRecDerT} for recognizing derivation-trees. 

\subsection{Computing the MPD}
\label{SubSecCompMPD}
For computing the MPD of an input sentence under \stsg, it is useful to precompile
the \stsg\/ into a form that makes the viability property explicit. 
To this end, infer the following predicates and functions from \mbox{\stsg $=$ \STSG}:
\begin{enumerate} 
\item {\bf $Parent?(c,~c^{'},~j)$} denotes the proposition 
      ``{\em $c$ and $c^{'}$ 
      are the addresses of a parent and its $j$-th child in an elementary-tree in \Corp}".
\item {\bf $Root?(c)$} denotes the proposition 
      ``{\em $c$ is the address of the root node of an elementary-tree in \Corp}".
\item {\bf $TreeOf(c)$} denotes the function that returns $t \in $\Corp
       such that $c$ is the address of some node in $t$. 
\item {\bf $SubSite?(c,j)$} denotes the proposition ``{\em child nr. $j$ 
       of the node with address $c$, in some elementary-tree,
        is a substitution-site}\/".
\end{enumerate} 
\newcommand{\BULLET}{\hspace*{-.1cm}$\bullet$~}
\newcommand{\hspx}{\hspace*{-.6em}}
Then infer the eight-tuple \RBSTSG, from \stsg, where:\\
\begin{tabular}{rcl}
\\
\multicolumn{3}{l}{
\BULLET \CFG~ is the CFG underlying~\STSG,}     \\
\\
\BULLET
$\forall$ $R \in \RulesN$: ~${\cal A}(R)$ & =  &
$\{c~| c$ is the address of an occurrence of $R \},$
\\ 
\\
\multicolumn{3}{l}{
\BULLET
$Viable?(c,~c_{j},~j)$  ~=~ 
          $Parent?(c, c_{j}, j)$ ~\OR~ $(~ SubSite?(c, j)$ ~\AND~ $Root?(c_{j})~)$,
} \\
\\
{\bf \A} & = & \SET{\A$(R)$}{$R \in$ \Rules},
\\ 
\\
\multicolumn{3}{l}{
\BULLET
$P$ : $\Pi$ \lra $(\Pi\multi\{1, 2\})$ \lra $[0..1]$,
}\\
\multicolumn{3}{l}{
For $c, c^{'}\in\Pi$ and $j\in\{1, 2\}$: 
}\\
\multicolumn{3}{l}{
\hspace*{-.4cm}
\begin{tabular}{lll}
 $P(c)(c^{'},j)$ & = & $\left\{
        \begin{tabular}{ll}
       \hspx $PT( TreeOf(c) )$ & [$j=2$ \AND~ $Viable?(c, c^{'}, j)$ \AND~ $Root?(c)$] \\
       \hspx {\bf 1} & [$Viable?(c, c^{'}, j)$ \AND~ ($j=1$ \OR~ {\bf not} $Root?(c)$)] \\
       \hspx {\bf 0} & [\bOtherwise],\\
        \end{tabular}
        \right.$
\end{tabular}
}\\
\\
\multicolumn{3}{l}{
\BULLET
$PF$ : $\Pi$ \lra $[0..1]$,}\\
\multicolumn{3}{l}{for $c \in \Pi$:  }\\
\multicolumn{3}{l}{
\begin{tabular}{lll}
 $PF(c)$ & = & $\left\{
        \begin{tabular}{ll}
          $PT( TreeOf(t) )$ & [$Root?(c)$] \\
          {\bf 1} & [~{\bf not} $Root?(c)$] \\
        \end{tabular}
        \right.$
\end{tabular}
}\\
\\
\end{tabular}

The predicate $Viable?()$ formalizes the viability property.
For a decorated tree, in which $c$ decorates a node $X$, 
the term $P(c)$ denotes the probability of $c$~as a function of 
the address $c^{'}$ of the $j$th child of $X$ (counting children is always from left
to right).This definition enables writing the algorithm as a recursive function.
$P$ expresses the probabilities in terms of the ``probabilities" of the rules 
of the CFG underlying \stsg; the probability of a rule of the CFG underlying 
\stsg\/ is a function of its particular occurrence, i.e. address, in the 
elementary-trees set of \stsg. 

\begin{figure}[thb]
\hrule
\[
\begin{array}{rl}
&  \\
P_{mpd}(w_{0}^{n}) \hspf  \hspf = & 
\sumMPD_{\tiny\hspf\TABS
{\ITEM{S}{\alpha}{}$\in \entryN{0}{n}$} 
{$c\in \An(\ITEMN{S}{\alpha}{},~0,~n)$}
{and $Root?(c)$}{}}
Pp(c,~\ITEMN{S}{\alpha}{},~0,~n)\\
\mbox{ }  & \\
Pp(c,item,i,j) \hspf =\hspf  &  \hspf \bCase item~\bOf \\
\mbox{ }  & \\
\hspfth
\ITEMN{A}{w_{i+1}}{} \hspft \hspfth :\hspfth & \hspfth PF(c),~~~$where$ ~$j = i+1$,  \\
\\
\hspfth
\ITEMN{A}{B}{C} \hspft \hspfth :\hspfth &  \hspfth
  \overbrace{
       \STACK{\sumMPD}{\alpha}
       \STACK{\sumMPD}{c1\in \An(\ITEMN{B}{\alpha}{},i,j)}
        P(c)(c1, 1)\mul
        Pp(c1, \ITEMN{B}{\alpha}{}, i, j)}, \\
\\
\hspfth
\ITEMNM{A}{BC}{} \hspft \hspfth :\hspfth& \hspfth
\sumMPD_{i < k < j}~Pp(c, \ITEMN{A}{B}{C}, i, k)~\mul \\
 & 
        \overbrace{
       \sumMPD_{\alpha}
            \sumMPD_{c2\in \An(\ITEMN{C}{\alpha}{}, k, j)}
P(c)(c2, 2)\mul Pp(c2, \ITEMN{C}{\alpha}{}, k, j)}
\end{array}
\]
\caption{Algorithm~\ref{AlgPS} for computing the MPD of a sentence $w_{0}^{n}$}
\label{AlgPS}
\vspace{1em}
\hrule
\end{figure}

Let $\sumMPD_{Pred(x)}A(x)$ denote the maximum of the set \mbox{\{$A(x)$~/~ $Pred(x)$\}}.
The algorithm for computing the MPD of a sentence $w_{1}^{n}$ 
is shown in figure~\ref{AlgPS}, also referred to as algorithm~\ref{AlgPS}. 
It computes the MPD of the input sentence 
on the parse-forest, which was constructed in the first phase by the CKY of the CFG \cfg.
The function $P_{mpd}(w_{0}^{n})$ computes the 
probability of the MPD of the sentence $w_{0}^{n}$, where \mbox{$w_{0} = \epsilon$}. 
The function $Pp:~\Pi\multi~ITEMS\multi~[0,~n]\multi~[0,~n]\rightarrow~[0,~1]$ 
computes recursively the probability of the most probable among the 
``partial-derivations"\footnote{
Below we define formally the probability of a partial derivation-tree, a more
suitable term, in this context, than partial-derivation.
},
that start with address $c$ and generate a partial-tree for $w_{i}^{j}$. 

Note that for SCFGs, i.e. when the elementary-trees of \stsg have no internal nodes, 
the algorithm in figure~\ref{AlgPS} computes the MPD exactly as 
algorithm~\ref{FigMPDCFGs}. This becomes more obvious if one notices that
the probability function $P(c)$ always computes to $PT(t)$ or $1$, due to the fact that
the elementary-trees have no internal nodes at all. For the general case of 
STSGs, a proposition and its proof, following the next definitions, take 
care of the correctness of the present algorithm. 
\DEFINE{Partial derivation-tree:}
{A {\em partial derivation-tree}\/ of an STSG is a partial-tree,
 resulting from a partial-derivation of the CFG underlying that STSG, in which
 each node is decorated with a node-address such that {\em the viability property
 still holds}\/ for each of the nodes.
}
A partial derivation-tree consists of, on the one hand, a sequence 
of elementary-tree substitutions, and on the other hand, other ``pieces" of elementary-trees.
A piece is a decorated subtree, of an elementary-tree, that either contains the root or 
some leaf-nodes of that elementary-tree but not both. The elementary-trees and the pieces
in the partial derivation-tree are combined with substitution under the STSG's 
viability property, i.e. substitution is allowed only of a node that 
has the address of a root of an elementary-tree on a node that originally was a
substitution-site in an elementary-tree. 

We will say that an elementary-tree {\em fully}\/ participates in a partial derivation-tree
$DT$ iff the addresses of all its nodes decorate nodes in $DT$. And we will say that an 
elementary-tree {\em partially}\/ participates in a partial derivation-tree $DT$ iff the
addresses of some of its nodes do not decorate a node in $DT$, but the addresses of
the other nodes do decorate nodes in $DT$.
\DEFINE{Sub-derivation:}
{For every partial derivation-tree there is exactly one corresponding sequence of 
 substitutions, of both the elementary-trees that fully and those that partially participate 
 in that partial derivation-tree;
 this sequence is denoted with the term the {\em sub-derivation} corresponding to the
 partial derivation-tree. 
}
In particular, a derivation of the STSG is a sub-derivation 1)~that starts with an 
elementary-tree with a root node labeled $S$, 2)~that has leaf nodes labeled with
terminals and 3)~in which every elementary-tree fully participates.
\DEFINE{Probability of a sub-derivation:}
{The probability $Pp(DT)$ of a sub-derivation $DT$ is equal to the 
 product of the probabilities of all elementary-trees that {\em fully}\/ 
 participate in it. 
 The probability of a partial derivation-tree is equal to the probability of the 
 corresponding sub-derivation.
}
Thus, a partial derivation-tree corresponds to some ``acceptable" 
decoration of a partial-tree's nodes. The viability property determines the ``acceptable"
decorations that corresponds to actual sub-derivations of the given STSG.
The algorithm in figure~\ref{AlgPS}, in short algorithm~\ref{AlgPS},
exploits the viability property by assigning zero probability to non-acceptable decorations 
and the right probability to the acceptable decorations. 

%
%
\newtheorem{CHOptAlgC}[CHOptAlgA]{Proposition}
\EExample{CHOptAlgC}
{$\forall~ item \in$ \entry{i}{j}, $0 \leq i < j \leq n$, and
\mbox{$\forall ~c \in \An(item, i, j)$}, $Pp$~fulfills one of two:
   \begin{itemize}
    \item \mbox{$Pp(c, item, i, j)$,} is equal to the maximum probability of 
     all possible partial derivation trees $dt$ of $w_{i}^{j}$, 
     that fulfill
     \begin{itemize}
      \item if $item = $\ITEM{A}{\alpha}{} then
         $dt$ has a root node labeled $A$, decorated with $c$, 
         and has children labels that (when concatenated from left to right) 
         form a string equivalent to $\alpha$, 
      \item if $item = $\ITEM{A}{B}{C} then
         $dt$ has a root node labeled $B$ and decorated with $c^{'}$ such that
         $Viable?(c, c^{'}, 1)$ is true.
     \end{itemize}
    \item There are no such partial derivation-trees as in the preceding case
          and \linebreak
          \mbox{$Pp(c, item, i, j)$} is equal to zero. 
    \end{itemize}
}{CHOptPropC}
\DEFINE{Proof~\ref{CHOptPropC}:}
{The proof is by induction on $x$, $1 \leq x = (j-i) < n$. It depends partially on the CKY 
  property that $\ITEMN{A}{\alpha}{\beta} \in [i,j]$ {\bf iff}
  \mbox{$\RuleM{A}{\alpha\beta}$ \lmdir $w_{i}^{j}\beta$}.
\begin{description}
 \item [${\bf x = 1:}$] Then $\alpha = w_{i+1}$. 
   By the CKY property, in this case there is at most one possible partial derivation-tree, 
   i.e. $\RuleM{A}{w_{i+1}} \in \RulesN$ with $A$ decorated with $c$. 
   Therefore, $Pp(c, \ITEMN{A}{w_{i+1}}{}, i, i+1) = PF(c)$. The proposition holds since
   there is only one partial-derivation tree.
 \item [${\bf x = m:}$] Assume the proposition holds for all $1 \leq x \leq m$, $m < n$.
 \item [${\bf x = m+1:}$] Let \ITEM{A}{\alpha \beta}{} $\in$ \entry{i}{j} and 
                  $c \in \An(\RuleM{A}{\alpha\beta})$.
                  By the CKY property there is a partial CFG-derivation (or many such derivations)
                  \mbox{$\RuleM{A}{\alpha\beta}$ \lmdir $w_{i}^{j}\beta$}.
                  Now there are two cases:
     \begin{description}
      \item [\underline{$\alpha = B C, \beta = \epsilon$:}]
                     Again by the same CKY property, for every partial CFG-derivation 
                    \mbox{$\RuleM{A}{B C}$ \lmdir $w_{i}^{j}$}, there 
                    is $i < k < j$ such that there are two 
                    partial CFG-derivations: 
                    \mbox{$\RuleM{A}{B C}$ \lmdir $w_{i}^{k} C$} and
                    \mbox{$C$ \lmdir $w_{k}^{j}$}.
                    The proof now is for every  $i < k < j$ for which this is true.
                    For the current value of $k$, let us consider the probabilities of
                    the possible decorations, of the partial CFG-derivation of $w_{i}^{j}$
                    at hand, that have $c$ decorating $A$ in \ITEM{A}{B C}{}.
                    As explained above, with every such decoration there are (at least) 
                    two decorated partial CFG-derivations for $w_{i}^{k}$ and $w_{k}^{j}$.
                    By the CKY property and the inductive assumption three things hold for these
                    pairs of decorations:
                    \begin{enumerate}
                      \item [a.] 
                       for all $\gamma$, for all $cb \in \An(\RuleM{B}{\gamma})$,
                       \mbox{$Pp(cb, \ITEMN{B}{\gamma}{}, i, k)$}
                       is equal either to zero or to the maximum probability of all partial
                       derivation-trees of $w_{i}^{k}$ starting at 
                       $\RuleM{B}{\gamma}$ with $B$ decorated with address $cb$. 
                      \item [b.] 
                       \mbox{$Pp(c, \ITEMN{A}{B}{C}, i, k)$}
                       is equal either  to zero or to the maximum probability of any partial
                       derivation-tree of $w_{i}^{k}$
                       starting at $\RuleM{B}{\gamma}$ with $B$ decorated with address $cb$ and
                       $Viable?(c, cb, 1)$ is true. 
                      \item [c.] 
                      for all $\delta$, for all $cc \in \An(\RuleM{C}{\delta})$,
                      \mbox{$Pp(cc, \ITEMN{C}{\delta}{}, k, j)$}
                      is equal either to zero or to the maximum probability of any partial
                      derivation of $w_{k}^{j}$
                      starting at $\RuleM{C}{\delta}$ with $C$ decorated with address $cc$. 
                    \end{enumerate}
                     If \mbox{$Pp(c, \ITEMN{A}{B}{C}, i, k)$} is zero for the current value
                     of $k$, then \linebreak
                     $Pp(c, \ITEMN{A}{B C}{}, i, j)$ is also zero. By the
                     inductive assumption we conclude that there is no partial derivation-tree
                     starting at \ITEM{A}{B C}{} and where $A$ is decorated with $c$, for the
                     current value of $k$. If it is not zero, we face two cases
                     for all $\delta$ and for all $cc \in \An(\RuleM{C}{\delta})$:
                    \begin{enumerate}
                     \item $Viable?(c, cc, 2)$ is false: then $P(c)(cc, 2) = 0$
                     \item $Viable?(c, cc, 2)$ is true: 
                            $P(c)(cc, 2)$ is either $PT( TreeOf(c) )$ or $1$, depending on
                             whether $Root?(c)$ is true or not.
                    \end{enumerate}
                     The last case in the switch of algorithm~\ref{AlgPS} 
                     (case \ITEM{A}{BC}{}) takes the maximum of all
                     $P(c)(cc,2) \mul Pp(cc, \ITEMN{C}{\delta}{}, k, j)$,
                     for all $\delta$ and for all $cc \in \An(\RuleM{C}{\delta})$.
                     By the inductive assumption and the viability property and the definition
                     of partial derivation-trees, this is equal  either to zero or to
                     the maximum probability of all partial derivation-trees starting at
                     \ITEM{A}{B C}{} with address $c$, for this value of $k$.
                     Again by the inductive assumption, multiplying this with 
                     $Pp(c, \ITEMN{A}{B}{C}{}, i, k)$ and maximizing this for all values of
                     $k$ proves the proposition for this case.
      \item [\underline{$\alpha = B, \beta = C$:}]
        The proof for this case follows a similar line of argumentation 
        as the preceding case.~~~~$\Box$
      \end{description}
 \end{description}
}
\newtheorem{CHOptAlgD}[CHOptAlgA]{Proposition}
\EExample{CHOptAlgD}
{Algorithm~\ref{AlgPS} computes the MPD of $w_{1}^{n}$ under \stsg.
}{CHOptPropD}
\DEFINE{Proof~\ref{CHOptPropD}:}
{Easy to derive from algorithm~\ref{AlgPS}, proposition~\ref{CHOptPropC} and
 the unique correspondence between derivation-trees and STSG derivations.~~~$\Box$
}
%
\paragraph{Complexity:}
The time complexity of the second phase of the algorithm is \mbox{$O(|\An|^{2} \mul n^{3})$}.
And the time complexity of the first phase is $O(|\RulesN| \mul n^{3})$. 
For DOP STSGs $|\RulesN|$, the size of the CFG underlying the STSG, is usually 
in the order of $1\%$ of $|\An|$. Therefore in practice the algorithm given above
behaves as if having time-complexity \mbox{$O(|\An|^{2} \mul n^{3})$}.
However, a simple optimization, discussed below, reduces the time-complexity of
the second phase to \mbox{$O(|\An| \mul n^{3})$}.
Moreover, the algorithm computes the second phase on the result of the first phase, i.e.
a parse-forest that already pruned many of the failing STSG derivations. This 
saves space and time that are expected to be much larger than the small overhead of 
having a first and a second phase. The net effect is faster computation that 
uses smaller space. 
%

\paragraph{Derived algorithms:}
Algorithm~\ref{AlgPS} can be easily modified to result in other useful algorithms.
To compute the probability of the input sentence, as the sum of the probabilities of 
all its derivations, exchange $\sumMPD$ with the operator $\sum$ everywhere in the
specification of the algorithm. For computing the MPD and the probability of a given 
parse-tree, the first phase of the algorithm is exchanged for a small algorithm that
transforms the parse-tree into a CKY table with items in the entries; the second phase
is applied as is either with $\sumMPD$ or with $\sum$, depending on the algorithm wanted.

\paragraph{Implementation issues:}
It is important to note that, for clarity of presentation, the specification given in 
figure~\ref{AlgPS} is not as efficient as the algorithm can be implemented.
For implementing the second phase of the algorithm, one precompiles the STSG, as explained
above, in order to make the viability property tables explicit.
The second phase is applied on the parse-forest in a bottom-up fashion
(rather than top-down recursion as the specification does)
making use of the $AddedBy$ sets of the items in the parse-forest
of the first-phase. Moreover, failing sub-derivations, which have probability zero, 
are discarded (rather than dragged together with the others all the way).
%
\subsection{Optimization: approaching linearity in STSG size}
\label{SubSecOptimiz}
%
Here we discuss an important optimization of algorithm~\ref{AlgPS}, based on
the following property of STSGs:
\DEFINE{Paths:}{ A sequence of nodes, i.e. pairs of labels and addresses, starting at the root 
          node of a derivation-tree and terminating at a leaf node (labeled with a terminal
          symbol) is called a {\em path}\/ of the derivation-tree.
}
\DEFINE{Path set:}{ The {\em path set}\/ of a derivation-tree is the set of all paths
           in that derivation-tree. The path set of a given {\em tree}\/ is the union of the
           path sets of all derivations that generate the tree under the STSG at hand. 
           Similarly, the path set of a {\em sentence}\/ is the union of all path sets of the  
           derivations that generate it under the STSG at hand. The path set of the
           STSG is the union of the path sets of the sentences in its string language.
}
\begin{description}
  \item [{\bf Relevant property:}] The path set of an input sentence under a TSG is a
          regular language. This property is known to be true for CFGs~\cite{Thatcher71}
          and can be easily proved for TSGs.
\end{description}
This property implies that recognition of the path set, and also the derivations, of
an input sentence under an STSG should be possible in time complexity linear in the
number of possible node-labels and their addresses, i.e. $|\An|$. 
A simple observation, related to the viability property,
makes this even clearer: in an elementary-tree, a node labeled $A$ with address $c$
can be only in one of the following two configurations:
\begin{itemize}
 \item $SubSite?(c, j)$ is true, i.e.\ the $j$th child of $c$ is a substitution-site in
       the elementary-tree (i.e.\ a frontier non-terminal). In this configuration, 
       \mbox{$Viable?(c, cj, j)$} is true only for all $cj$ such that $Root?(cj)$.
 \item $Parent?(c, cj, j)$ is true for some $cj$, i.e. $c$ and $cj$ are both non-leaf nodes
          in the elementary-tree and are respectively the addresses of a parent and its 
          $j$th child in that elementary-tree. Only one single $cj$ fulfills this (due
          to the unique addresses of nodes in elementary-trees).
\end{itemize}
Let $itemP$ denote any item to the left of the semicolon in algorithm~\ref{AlgPS}.
And let $itemCh$ denote any item found in the overbraced term also in 
algorithm~\ref{AlgPS}. The algorithm checks the viability of every
address $c \in \An(itemP)$ with address $cj \in \An(itemCh)$. The above observation says
that this is unnecessary since there are only two complementary configurations in which
$c$ and $cj$ fulfill the viability property. In fact it says even that $c \in \An(itemP)$ 
can be in one of two configurations, and $cj \in \An(itemCh)$ can also be in one of two 
{\em corresponding}\/ configurations.  This motivates partitioning $\An(itemP)$
and $\An(itemCh)$ into:
 \begin{eqnarray*}
 HasSubSite(itemP, j) & = & \{c \in \An(itemP)~|~SubSite?(c, j) ~is~true \},\\
 HasChild(itemP,j) & = &  \An(itemP)~-~HasSubSite(itemP, j), \\
 & & \\
 RootsOf(itemCh) & = & \{cj \in \An(itemCh)~|~Root?(cj)  ~is~true \}, \\
 InternOf(itemCh) & = & \An(itemCh)~-~RootsOf(itemCh) \\
 \end{eqnarray*}
Recall that the viability property states that:
\begin{verse}
$Viable?(c, cj, j)$ is true for $c \in \An(itemP)$ and $cj \in \An(itemCh)$ iff
either [$c \in HasSubSite(itemP, j)$ and  $cj \in RootsOf(itemCh)$]
or [$c \in HasChild(itemP, j)$ and $cj \in InternOf(itemCh)$].
\end{verse}
The above mentioned observation extends it and states that:
\begin{verse}
For \underline{every} $c \in HasSubSite(itemP, j)$ and \underline{every}
$cj \in RootsOf(itemCh)$ holds $Viable?(c, cj, j)$ is true. 
And, for every $c \in HasChild(itemP, j)$, there is
\underline{at most one} $cj \in InternOf(itemCh)$ such that $Viable?(c, cj, j)$ is true.
\end{verse}
For both cases it is possible to conduct the computation for every $c \in itemP$ in
$O(1)$ rather than $O(|\An|)$ (i.e. $|\An(itemCh|)$). In the first case, 
compute only once the set $RootsOf(itemCh)$ and its maximum probability and pass this
to {\em every}\/ \linebreak         $c \in HasSubSite(itemP, j)$. 
In the second case, an off line  precompilation of 
the place of the $j$th child, for $j \in \{1, 2\}$, of every $c \in HasChild(itemP, j)$ in
$\An(R)$, for every $R$ enables finding it in $O(1)$; then the maximum probability of the
$j$th child is passed to $c$ for computing its maximum probability in $O(1)$.

More formally, the overbraced expression in figure~\ref{AlgPS} in each of the 
last two cases of algorithm~\ref{AlgPS} is rewritten. Let these two expressions 
be represented by the more general expression
\[\STACK{\sumMPD}
        {\alpha,~c_{l}\in \An(itemCh,m,q)} P(c)(c_{l}, l)\mul Pp(c_{l}, itemCh, m, q). \]
Substitute for this expression the following,
where $item, i$ and $j$ are as defined by algorithm~\ref{AlgPS}:
{\small
\[
\left\{
\begin{array}{lll}
\bIf~(c\in HasSubSite(item, l)) & : &
           PF(c)\mul \\ & &
           \STACK{\sumMPD}{\alpha,~c_{l} \in RootsOf(itemCh)} Pp(c_{l},itemCh,m,q), \\
\bIf~(c\in HasChild(item, l)) & : &
             PF(c) \mul \\
            & & \bIf~{\bf (~}\exists c_{l} \in InternOf(itemCh):~Parent?(c,c_{l},l){\bf~)}\\
              & & \bThen~\STACK{\sumMPD}{\alpha} Pp(c_{l},itemCh,m,q)\\
              & & \bElse~~~0. \\
\end{array}
\right.
\]
}
Both cases in this specification must be implemented as explained above 
in order to achieve linearity in $|\An|$ during the second phase of the
algorithm. The time complexity of the algorithm (two phases) is now proportional
to \mbox{$(|\RulesN| + |\An|) \mul n^{3}$}. As mentioned earlier, $\frac{|\RulesN|}{|\An|}$
is very small for DOP~STSGs. Therefore, the time complexity actually approaches
$O(|\An| \mul n^{3})$ as $\frac{|\RulesN|}{|\An|}$ approaches zero. For most DOP STSG
the actual time complexity indeed approaches the linear in STSG size.
This optimization does not result in any change in the space complexity, which remains
$O((|R|+|\An|) \mul n^{2})$.

It is interesting to observe the effect of this optimization in practice.
To this end, I conducted a  preliminary experiment\footnote{
This experiment was conducted with the first implementation
of the MPD algorithm. As most first implementations, it was also suboptimal due to choice of
simple but inefficient data-structures; thus, only the ratios between the CPU-times of
the various versions are of interest here, rather than the absolute CPU-times listed 
in the table. 
It is worth noting that subsequent versions improved the figures for all versions 
substantially (see the experiments in chapter~\ref{CHARSImpExp}).
} comparing various versions of the
algorithm: a linear version in STSG size, a square version and a version
that searches for the child of each address using Binary-Search on ordered
arrays. The results are listed in Table~\ref{Simaantable1}.
The experiments reported in table~\ref{Simaantable1} used the ATIS domain 
Penn Tree-bank version 0.5 (Penn TreeBank Project, LDC) without modifications. 
They were carried out, on a 
{\bf Sun Sparc station 10} with 64 MB RAM, parsing ATIS word-sequences.
The three versions were compared for execution-time by varying STSG sizes
on the same test set of 76 sentences, randomly selected. 
The sizes of the STSGs were varied by varying the allowed maximum depth
of elementary-trees and by projecting only from part of the tree-bank in some
other cases (with a maximum of 750 training trees). 
Average cpu-time  includes parse-forest generation, i.e. both phases.
It is clear that there is a substantial difference in growth of execution-time 
between the three versions as grammar size grows.
Note also that this experiment exemplifies, to some extent, the ratio between $|R|$ and $|A|$ 
for ATIS DOP STSGs. These observations hold also for experiments with DOP on other
domains, reported in chapter~\ref{CHARSImpExp}.
%
%
%
{\bf
\begin{table}[tb]
\vspace*{1em}
\hrule
\vspace*{0.1em}
\begin{small}
\begin{center}
\begin{tabular}{cccc|ccc}
{\footnotesize\bf num.\ of} & $|$\Rules$|$ & $|\An|$ & {\footnotesize\bf Avg.Sen.} & 
\multicolumn{3}{c}{\footnotesize Average CPU-secs.} \\
{\footnotesize\bf elem.\ trees} &  &  & {\footnotesize Length} & 
{\footnotesize linear} & {\footnotesize Bin.Search} & {\footnotesize Square} \\
\hline
74450 & 870 & 436831 & 9.5 &  445 & 993 & 9143 \\
26612 & 870 & 381018 & 9.5 &  281 & $-$ &  $-$ \\
19094 & 870 & 240619 & 9.5 &  197 & $-$ & 2458 \\
19854 & 870 &  74719 & 9.5 &  131 & 223  & 346 \\
\end{tabular}
\end{center}
\end{small}
\caption{Disambiguation times for various STSG sizes}
\vspace*{0.3em}
\hrule
\label{Simaantable1}
\end{table}
}
\subsection{Extension for disambiguating word-graphs}
\label{SecOptAlg4WGs}
The two-phase algorithm for parsing and disambiguating sentences under STSGs
can be easily extended for parsing and disambiguating word-graphs as those
produced by speech recognizers. As mentioned in chapter~\ref{CHDOPinML}
a word-graph is a Stochastic Finite State Machine (FSM). For notation and
definitions on word-graphs, the reader is referred to chapter~\ref{CHDOPinML}.
\DEFINE{Assumptions:}
{
 In this work we assume FSMs and SFSMs that contain no cycles, 
 i.e. paths $s_{x} \cdots s_{x} $ \lmdir $w_{1}^{n}$, for $n \geq 1$. 
 Due to the preceding assumption we may assume, without loss of generality,
 that the set of states $\Sigma$, of SFSM \SFSM, is equal to the set of numbers 
       \mbox{$\{0,1,\cdots (|\Sigma|-1)\}$}
       such that: 1)~each transition 
       $\tuple{s1,s2,w}$ fulfills \mbox{$s1 < s2$}, 2)~the start-state $S$ is~$0$,
       and 3)~the target state is~$(|\Sigma|-1)$.
}
Before discussing how to parse and disambiguate word-graphs under DOP STSGs,
it is clarifying to note that a sentence $w_{1}^{n}$ is also a simple word-graph.
The CKY algorithm exploits this by numbering a position between the words $w_{i}$ 
and $w_{i+1}$, for all $1\leq i\leq n$, by the number $i$. It also numbers the
position before the first word by $0$. These are the states of a word-graph with
the transitions $\tuple{i-1,i,w_{i}}$, for all \mbox{$1\leq i\leq n$}. Note that in a sentence,
every transition is from a state $i$ to the state $i+1$. In general word-graphs
this is not the case; transitions can be from any state $i$ to any state $j > i$.

Parsing an FSM with a CFG is known in the literature as the problem of computing
the intersection of the two machines; studies of algorithms for conducting this 
intersection are discussed in e.g.~\cite{GJvanNoord}. 
Below I discuss how to extend the two-phase MPD computation algorithm under STSGs for 
word-graphs, i.e. SFSMs. Firstly I discuss how to extend the CKY parser of the first phase 
for this task in a similar fashion to preceding work. After that, I discuss how to
compute the MPD of an SFSM under an STSG, in the second phase of the present 
algorithm.

Assume again that we are given the STSG \stsg = \STSG~ in CNF, and let 
\cfg = \CFG denote the CFG underlying it. Moreover, let $WG$ denote the wordgraph 
\SFSM ~where $\Sigma = $\mbox{$\{0,\cdots M\}$}, $M \geq 1$, 
$Q = \{w_{i} | 1\leq i\leq X\}$, and $S$ denotes~$0$ and $F$ denotes~$M$. 
For the rest of this section, we assume that \mbox{$Q\subseteq $\VT} holds. 
\subsubsection{First phase}
Recall that when the CKY algorithm parses the sentence $w_{1}^{n}$ under a
CFG, it maintains the following invariant for every entry $[i,j]$ in its table: 
\begin{verse}
  \ITEM{A}{\alpha}{\beta}$\in [i,j]$ if and only if (iff) 
  \mbox{$\RuleM{A}{\alpha\beta}$ \lmdir $w_{i}^{j}\beta$}\/ holds in the CFG at hand.
\end{verse}
Both parts of the CKY, {\bf Deduce} and {\bf Init}, maintain this. In particular, the
{\bf Deduce} part also relies in its inference on this invariant when combining a 
partial-derivation
of $w_{i}^{k}$ together with a partial-derivation of $w_{k}^{j}$ into a partial-derivation 
of $w_{i}^{j}$. 

For a word-graph with a set of states \mbox{$\{0,\cdots M\}$}, $M \geq 1$, 
the CKY employs a table as used for a sentence of length $M$. 
The only change in the CKY algorithm is in the {\bf Init} part, which now adds the
items \ITEM{A}{w_{l}}{}, for all $\RuleM{A}{w_{l}} \in \RulesN$, 
to entry \entry{i}{j} if there is a transition $\tuple{i,j,w_{l}}$ in the word-graph at hand.
Figure~\ref{FigCKYWGs} shows the CKY algorithm extended for word-graphs.
One can easily see that this extension of {\bf Init} still fulfills the CKY invariant.
A useful property of word-graphs is that if there is a path from state~$i$ to state~$k$
and a path from state~$k$ to state~$j$, then there is a path from state~$i$ to state~$j$.
Therefore, it is easy to prove that the above modest extension of the CKY is sufficient 
to maintain the invariant of the CKY algorithm for parsing word-graphs. The CKY table 
now contains a parse-forest consisting of all possible parses of every 
sequence of transitions, i.e. all paths, in the input word-graph. 
%
\begin{figure}
\hrule
\begin{tabbing}
{\bf Init:} \=                                           \\
            \> $\forall$ \= ~$\tuple{i,j,w}~ \in~ T$    \\
            \>           \> $\forall$ \= ~$\RuleM{A}{w} \in$ \Rules      \\
            \>           \>           \> ~~~add \ITEM{A}{w}{} to $[i,j]$ \\
            \>           \> $\forall$ \= ~$\ITEMN{B}{\beta}{} \in [i, j]$  \\
            \>           \>           \> $\forall~ \RuleM{A}{B C} \in$ \Rules \\
            \>          \>            \> ~~~add \ITEM{A}{B}{C} to $[i, j]$  \\
{\bf Deduce:}\=                  					\\
             \>$\forall$ \= ~$0 \leq i < M$ \AND~ $\forall$~ $i < j \leq M$			\\
             \>          \> ~~~$\forall$\= ~$i < k < j$  \AND~
                                            $\forall~ \ITEMN{A}{B}{C}\in [i, i+k]$ 		\\
             \>          \>             \>~~\bIf~ $\exists$ \ITEM{C}{\alpha}{} $\in [i+k, j]$\\
             \>          \>             \>~~\bThen~ add \ITEM{A}{BC} to $[i, j]$ \\
             \>          \> ~~~$\forall$\= ~$\ITEMN{B}{\beta}{} \in [i, j]$ \AND~
                                            $\forall~ \RuleM{A}{B C} \in$ \Rules \\
             \>          \>             \> ~~add \ITEM{A}{B}{C} to $[i, j]$  \\
\end{tabbing}
\hrule
\caption{CKY algorithm for parsing word-graphs under CFGs in CNF}
\label{FigCKYWGs}
\end{figure}

The time-complexity of this extended CKY for word-graphs is a function of the number
of states $M$. Similar to the original CKY, the time-complexity is 
$O(|\RulesN|\mul M^{3})$ and the space-complexity is $O(|\RulesN|\mul M^{2})$. 
Note that the number of transitions between any two states does not affect this
time-complexity. The number of transitions does not play a role in the parsing 
steps taken by the CKY {\bf Deduce} and it only introduces  an additional contant 
of time and space complexity during CKY {\bf Init}. But even this constant is accounted 
for the worst case complexity analysis given above (the $O()$ notation). 
\subsubsection{Second phase}
%
\DEFINE{Intersection derivation:}
{An intersection-derivation (or simply i-derivation) of \stsg~ with $WG$
 is a pair $(D_{stsg},D_{WG})$,
where $D_{stsg}$ is a \stsg~ derivation of the sentence $w_{1}^{n}$, and
$D_{WG}$ is a $WG$ derivation of $w_{1}^{n}$, i.e. 
$D_{WG}$ = \mbox{$0,k_{1},\cdots,k_{m},M$\lmdir $w_{1}^{n}$},
where $\forall 1\leq i\leq m$: $k_{i}\in \Sigma$.
}
\DEFINE{Probability of i-derivation:}
{The probability of the i-derivation $(D_{stsg},D_{WG})$ is equal to the multiplication 
of the probability of $D_{stsg}$ with the probability of $D_{WG}$. 
}
\DEFINE{MPiD:}
{The Most Probable i-Derivation (MPiD) of a word-graph $WG$ under the  STSG 
 \stsg~ is defined as the pair $(D_{stsg},D_{WG})$ with the maximum probability.
}
For computing the MPiD of $WG$ under \stsg~ we note that the second phase of the present
algorithm does not discriminate between a word-graph and a sentence; all it sees
is a parse-forest in a CKY table. Therefore, algorithm~\ref{AlgPS} can be
applied for computing the MPiD of an input word-graph without transition probabilities,
i.e. an FSM. For incorporating the probabilities of the transition of an SFSM
word-graph, the algorithm is extended in a very simple manner: for every item
\ITEM{A}{w_{l}}{} in entry \entry{i}{j} the term for $Pp$ is multiplied with
the transition probability $P(\tuple{i,j,w_{l}})$. Figure~\ref{SIMAANAlgWG} exhibits
the specification of this algorithm. A proof of the correctness of this 
algorithm is very simple and will be skipped here.
%
\begin{figure}[thb]
\hrule
\[
\begin{array}{rll}
& & \\
P_{mpd}(w_{0}^{n}) \hspf & \hspf = & 
\sumMPD_{\tiny\hspf\TABS
{\ITEM{S}{\alpha}{}$\in \entryN{0}{n}$} 
{$c\in \An(\ITEMN{S}{\alpha}{},~0,~n)$}
{and $Root?(c)$}{}}
Pp(c,~\ITEMN{S}{\alpha}{},~0,~n)\\
\mbox{ } & & \\
Pp(c,item,i,j) &\hspf =\hspf  &  \hspf \bCase item~\bOf \\
\mbox{ } & & \\
\hspfth
\ITEMN{A}{w_{l}}{} \hspft &\hspfth :\hspfth & \hspfth P(\tuple{i,j,w_{l}})\mul PF(c), \\
\\
\hspfth
\ITEMN{A}{B}{C} \hspft &\hspfth :\hspfth &  \hspfth
  \overbrace{
       \STACK{\sumMPD}{\alpha}
       \STACK{\sumMPD}{c1\in \An(\ITEMN{B}{\alpha}{},i,j)}
        P(c)(c1, 1)\mul
        Pp(c1, \ITEMN{B}{\alpha}{}, i, j)}, \\
\\
\hspfth
\ITEMNM{A}{BC}{} \hspft &\hspfth :\hspfth& \hspfth
\sumMPD_{i < k < j}~Pp(c, \ITEMN{A}{B}{C}, i, k)~\mul
\\
& & 
        \overbrace{
       \sumMPD_{\alpha}
            \sumMPD_{c2\in \An(\ITEMN{C}{\alpha}{}, k, j)}
P(c)(c2, 2)\mul Pp(c2, \ITEMN{C}{\alpha}{}, k, j)}
\end{array}
\]
\caption{Algorithm for computing the MPiD of word-graph $WG$}
\label{SIMAANAlgWG}
\vspace{1em}
\hrule
\end{figure}

The algorithm in figure~\ref{SIMAANAlgWG} can be adapted for computing the total
probability of a given word-graph under the given STSG. This can be achieved
by exchanging every $\sumMPD$ with a $\sum$. 

Similar to the original algorithm, the time-complexity when applying the optimization
(which is still valid) is $O(|\An|\mul M^{3})$ and the space-complexity 
is $O(|\An|\mul M^{2})$. The total times-complexity for both phases of the algorithm 
is $O((|R|+|\An|)\mul M^{3})$  and the space-complexity is $O((|R|+|\An|)\mul M^{2})$
%
%
%
%
\section{Useful heuristics for a smaller STSG}
\label{SecHeuristics}

Our practice has shown that as soon as the domains and the applications 
take practical forms, the interesting DOP STSGs acquire non-manageable sizes. In some
cases, these models could not be acquired or executed, even on sizeable machines, 
due to the huge number of elementary-trees (typically exceeding a hundered million 
for small domains). And in the few cases where they could be acquired and executed,
various problems were encountered: sparse data effects, extremely slow 
execution-times and extremely small probabilities. A possible solution is to
search for approximations that are manageable.

In analogy to n-gram models, Bod~\cite{RENSDES} suggested to infer DOP STSGs with 
elementary-trees of some maximal depth (i.e. number of edges in longest path). 
In many cases this proved quite useful, e.g. maximum depth four turned out to be
a good approximation in various situations. However, since limiting subtree-depth
to small values (typically depth one or two) simply implies sacrificing accuracy,
one ends up employing ``mid-range" values (e.g.~four, five or six). For mid-range values
the number of subtrees remains extremely large (depth four already exceeds 
a couple of million for small domains). Therefore, usually limiting subtree-depth 
alone is not effective enough\cite{BonnemaScha99}. Next I discuss shortly other similar 
heuristics that are at least as effective as the upper bound on depth and can be 
applied in conjunction with it.

An interesting aspect of the MPD in DOP is a general tendency 
for preferring  shorter derivations involving more probable trees. 
Shorter derivations imply a smaller number of substitutions. 
In general, one can imagine that there is an upper bound 
on the number of substitutions in most probable derivations of 
sentences of a certain domain under some DOP model. 

In practice, this knowledge can be exploited  in two different ways.
Firstly, during training, off line, a smaller size DOP STSG can be projected
from the tree-bank.  This is operationalized through setting an upper bound 
on the number of substitution-sites a DOP STSG's elementary-tree is allowed 
to have; for example, a maximum of two substitution-sites per elementary-tree.
This heuristic has been exploited in many experiments and turned out to reduce
the size of the DOP STSG up to {\em two orders of magnitude}\/ without loss of accuracy
or coverage e.g.~\cite{MyRANLP95,MyRANLPBook,BonnemaEtAl96,DOPNWORep,BonnemaEtAl97}.
And secondly, during the computation of the MPD, an upper 
bound on the number of substitutions can be exploited for pruning derivations
exceeding that upper bound. The latter pruning heuristic has not been applied yet
in our system. In any event, the two heuristics are complementary to each other and
can be applied simultaneously.

Other heuristics in the same spirit turned out to be useful: an upper-bound
on the number of terminals per elementary-tree and an upper bound on the number
of consecutive terminals per elementary-tree. In total, there are currently
four upper bounds in use: on depth (denoted $d$), on the number of 
substitution-sites (denoted $n$), on the number of terminals (denoted $l$) and on the
number of adjacent terminals (denoted $L$). 
These upper bounds are combined together in conjunction during learning a DOP STSG.
In most experiments reported in this thesis these heuristics are employed
in some form or another. 
\section{Conclusion}
\label{SecConcs}
In this chapter we presented various deterministic polynomial-time algorithms for parsing
and disambiguation under the DOP model. We also presented optimizations of these algorithms 
that make them particularly suitable to deal efficiently with the DOP STSGs, which are typically
extremely large; in particular, these algorithms have time-complexity linear in STSG size 
at almost no extra memory cost. We also discussed various useful heuristics for projecting 
smaller and more feasible DOP STSGs from large tree-banks. 

The majority of the algorithms in this chapter were originally presented 
in~\cite{MyNEMLAP94,MyRANLP95} and later in an improved version in~\cite{MyRANLPBook}. 
Prior to the presentation of these algorithms, preceding work on DOP was unaware 
of the possibility of {\em deterministic polynomial-time}\/ parsing and disambiguation 
under STSGs in general and under DOP in particular.

The present algorithms have been implemented in an environment for
training DOP STSGs and parsing and disambiguation of sentences and word-graphs 
(i.e. SFSMs), dubbed the Data Oriented Parsing and Disambiguation System (DOPDIS).
%
Since their implementation in DOPDIS, the present algorithms have enabled
intensive studies of the behavior of the DOP model, a model that
was considered unattainable by many. The speedup that these algorithms achieve 
(in comparison to preceding Monte-Carlo~\cite{RENSDES}) is in the order of magnitude 
of hundreds of times (due to the optimized deterministic polynomial nature of the 
algorithms rather than to any implementation detail). This has made DOPDIS a very attractive 
experimentation-tool for other work involving DOP 
e.g.~\cite{BonnemaScr,BonnemaEtAl96,DOPNWORep,BonnemaEtAl97}.
Currently, DOPDIS serves as the kernel of the Speech-Understanding Environment of 
the Probabilistic Natural Language Processing~\cite{DOPNWORep} in the OVIS system,
that is being developed in the Priority Programme Language and Speech Technology of the 
Netherlands Organization for Scientific Research (NWO).

\newcommand{\REFTHISSEC}[1]{(\ref{SecImpDetLearn}.#1)}
\chapter{Implementation and\ empirical~testing}
\label{CHARSImpExp}
This chapter discusses the details of the current implementation of the ARS learning 
and parsing algorithms (chapter~\ref{CHARS}) and exhibits an empirical study of its 
application to specializing DOP for two domains: the Dutch OVIS domain 
(train time-table information) and the American ATIS domain (air-travel information).
These domains are represented respectively by the Amsterdam OVIS tree-bank (syntactic
and semantic annotation) and the SRI-ATIS tree-bank\footnote{I am grateful to SRI International 
Cambridge~(UK) for allowing me to conduct experiments on their tree-bank. The SRI-ATIS
tree-bank differs considerably from the ATIS tree-bank of the Penn Treebank Project.}
(syntactic annotation). 

The experiments on the OVIS domain compare the behavior of various
DOP models and Specialized DOP models that are trained on the OVIS tree-bank.
In some of the experiments the models are trained only on the syntactic annotation 
of the tree-bank, and in the other experiments they are trained on the 
syntactic-semantic annotation. 
In each case, the experiments observe the effect of varying some training-parameters 
of the models (e.g specialization algorithm, subtree depth, training tree-bank size)
on their behavior; only one parameter is allowed to vary in each experiment, while the 
rest of the parameters are fixed. A similar but less extensive set of experiments 
compares the models on the SRI-ATIS tree-bank. To the best of my knowledge, the
present experiments are the most extensive experiments ever that test the DOP model on 
large tree-banks using cross-validation testing.

The structure of this chapter is as follows. Section~\ref{SecImpDet} discusses the 
implementation details of the learning and parsing algorithms. 
Section~\ref{CHIMPintro} introduces the goals of the experiments and their general setting,
and the measures that are used for evaluating the systems.
Section~\ref{SecOVISExps} exhibits experiments on the OVIS domain for parsing word-strings 
and word-graphs.
Section~\ref{SecATISExps} exhibits experiments on the ATIS domain for parsing word-strings. 
And finally 
section~\ref{CHIMPconcs} discusses the achievements and the results of the experiments 
and summarizes the general conclusions on the applicability and utility of ARS to 
specializing DOP for these domains.
\section{Implementation details}
\label{SecImpDet}
This section discusses the details of the implementations of 
the parsing and learning algorithms of chapters~\ref{CHARS} and~\ref{CHOptAlg4DOP} 
as used in the present experiments.
\subsection{ARS learning algorithms}
\label{SecImpDetLearn}
In order to implement the ARS learning algorithms of the preceding chapter,
it is necessary to take into consideration various issues 
such as data-sparseness, and time- and memory-costs of learning on existing 
tree-banks. Next we discuss how these issues are dealt with in the current
implementation.
\subsubsection{\REFTHISSEC{A}~~Data-sparseness effects:}
In order to reduce the effects of data-sparseness we incorporate the following solutions 
into the implementation:
\begin{description}
\item [Sparseness of lexical atoms:]
      Because sequences of actual words (lexical atoms) of the language often
      have low frequencies, the current implementation allows learning only SSFs that 
      are sequences of grammar symbols that are {\em not words}\/ of the language (i.e. are not
      terminals). This implies that the sequences of symbols that the learning algorithm
      considers consist of either PoSTag-symbols or higher level phrasal symbols.
      This is clearly a severe limitation of the current implementation since lexical information 
      is essential in disambiguation of syntax and semantics. Without a central role for
      lexical information in ambiguity reduction specialization, one can not expect 
      extensive ambiguity reduction to take place. Therefore, the current implementation 
      of the learning algorithms is severely suboptimal. 
      Note that this limitation is not so much a theoretical one as a practical choice
      dictated in part by data-sparseness and in part by the limited annotations of the 
      available tree-banks\footnote{These tree-banks do not include a  
      lexicon that describes the words of the language by e.g. feature structures.}. 
      Some suggestions on how to lexicalize the learning algorithms in practice
      (in the light of sparse-data problems) were discussed in section~\ref{CHARSConcs}.
\item  [Stop condition:]
      In the theoretical version of the sequential covering scheme, the iterations continue
      until the tree-bank is totally reduced into the roots of the original trees. However, 
      since in real tree-banks the SSFs that are closer to the roots of the trees are much less 
      frequent than SSFs that are closer to the leaves, the last iterations of the algorithm 
      operate on less frequent SSFs. This implies that the Constituency Probabilities of the
      SSFs become poorer and their ambiguity-sets become more prone to incompleteness.
      By setting a threshold on the frequency of the sequences of symbols that are considered 
      during learning it is possible to avoid these problems. Moreover, this threshold can
      serve as a limit on the size of the learned grammar. 
      Therefore, a threshold $\Phi$ is set on the frequency of SSFs; SSFs are considered 
      {\em Viable}\/ iff they abide by the requirements of the algorithm given in 
      figure~\ref{AlgScheme} and {\em also have a frequency that exceeds the threshold}.
      In general, after the learning algorithm stops there will remain a tree-bank of 
      partial-trees that is not represented in the learned grammar.
      After the last iteration all these partial-trees are then included in the learned grammar,
      thereby expanding the learned grammar to represent the whole tree-bank.
      For specializing DOP, the root nodes of all trees of the tree-bank are also marked 
      as cut nodes after the last iteration.

      This frequency threshold is supplemented by a ``coverage upper-bound" in the spirit
      of the one employed in~\cite{SamuelssonThesis}. In this case, however, the coverage 
      upper-bound is set in a direct manner on the percentage of nodes in the tree-bank 
      trees that is reduced in all iterations. In other words, as long as 
      the tree-bank contains a percentage of nodes (w.r.t.\ the original tree-bank) that 
      exceeds a priori selected percentage (one minus the upper-bound) and there are SSFs 
      to learn, the learning algorithm continues.
      Not all implementations benefit from this kind of a coverage upper-bound stop condition.
      In those implementations and experiments that did benefit from it,
      this fact will be mentioned and the value of the coverage upper-bound will be specified.
\end{description}
\subsubsection{\REFTHISSEC{B}~~Reducing time and memory costs of learning}
Off-line learning does not have to meet {\em real-time}\/ requirements. 
However, practical memory and time limitations on the learning system do exist: 
computers have physical limitations and our patience is not without limit. 
To enable acceptable time and memory costs of learning, the current 
implementation incorporates the following approximations:
\begin{description}
\item [Frequencies:] 
      The computation of the frequencies (total frequency and frequency as an SSF)
      of a sequence of symbols from the tree-bank trees is a time-consuming task; 
      in fact, the number of all sequences that can be extracted from the trees of a
      tree-bank prohibits exhaustive training on realistic tree-banks. Therefore, we approximate
      the frequencies of sequences of symbols in the tree-bank by assuming that
      {\em sequences that occur lower (closer to the leaves) in the trees rarely 
      occur again higher (closer to the root) in the trees}.  For implementing this
      in the sequential covering scheme, the frequency of a sequence of symbols is 
      computed only with respect to the current iteration. In other words, 
      at each iteration of the algorithm the frequencies are initiated at zero and
      the current frequencies are computed by extracting sequences of symbols from
      the frontiers of the partial-trees in the current tree-bank. 
\item [Length threshold:]
      The length of the sequences of symbols that are considered during learning can
      be limited by a predefined threshold without jeopardizing the quality of the 
      learned grammar; the threshold can be set at a length that is expected to be larger
      than most learned SSFs. In most of the present experiments the 
      ``magic" number~8 was selected to be the threshold on the length of 
      sequences of symbols that are considered during learning.
\end{description}
\subsubsection{\REFTHISSEC{C}~~Extensions to the learning algorithms}
In some experiments we also employed a different definition of the target-concept:
\begin{description}
\item [Equivalence classes for SSFs:]
A common problem in most tree-banks is repetition of categories under the same mother
node. For example, in the UPenn tree-bank there is the shallow multi-modifiers construction:
consecutive $PP$s that are the children of the same node in tree-bank trees constitute sequences 
of any length one can think of.
Usually this causes the number of grammar rules to explode and prohibits the recognition of 
constructions that are mere ``number modifications" of each other. The same problem occurs
in noun-phrases consisting of compound nouns. It is usually felt that these impoverished 
constructions tend to prohibit successful learning.

The solution proposed for the present algorithm is a simple and limited one; its only
merit is the fact that it is an approximation to the intuitive ideal solution.
It deals only with cases of consecutive repetitions of single symbols within the
borders of SSFs such as in \mbox{``$NP~PP~\cdots~PP$"}. More formally,
we redefine the target-concept of the learning algorithms from individual SSFs to 
equivalence classes of SSFs. To this end we define first the notion of a bracketed
SSF:
\DEFINE{Bracketed SSF:}
{
 A {\em bracketed SSF}\/ is obtained from a partial-tree $t$ by removing the labels 
 of its non-leaf nodes leaving behind a partial-tree with labels only on the leaf nodes.
}
\DEFINE{Initial bracketed-SSF:}
{An {\em Initial bracketed-SSF}\/ is obtained from a bracketed SSF by removing
 all nodes that dominate non-leaf nodes, leaving an ordered non-connected sequence of 
 partial-trees or individual-nodes.
}
An initial bracketed SSF is an ordered sequence of bracketed sequences of grammar symbols.
For example, for the partial-tree $s(np,vp(v,np(dt,n)))$ -~for a sentence such as ``np ate 
the banana"~- the initial bracketed SSF is \mbox{$np~v~(dt~n)$}.
An initial bracketed SSF $S$ is represented by an ordered sequence of brackets 
\mbox{$B_{1}\cdots B_{n}$}, where $B_{i}$ is the $i$-th  element of $S$ (which is
either a single symbol or a bracketed sequence of symbols in $S$).
Thus, the sequence  \mbox{$np~v~(dt~n)$} can be represented by the sequence of 
brackets \mbox{$(np) (v) (dt~n)$}.

Let $RP(B_{i})$ represent the function that removes consecutive repetitions of symbols 
from a bracket $B_{i}$. The equivalence classes of SSFs are obtained by partitioning the space 
of bracketed SSFs in the tree-bank by the following relation of one-symbol-repetition ($OSR$):
\center{
    $\tuple{B1_{1}\cdots B1_{n}, B2_{1}\cdots B2_{m}}\in OSR$~~ iff \\
    $m=n$ and for all \mbox{$1\leq i\leq n$}: $RP(B1_{i})$ is identical to $RP(B2_{i})$.
}
\end{description}
Not all implementations and experiments employ this definition. In those experiments where
the implementation is based on this extension, this will be stated explicitly.
\subsection{Implementation detail of the parsers}
\label{SecImpPDet}
The implementation of the various parsing and disambiguation algorithms 
(i.e. partial-parser TSG, DOP STSG, SDOP STSG and ISDOP STSG)
involved in the experiments is based on the DOPDIS implementation of chapter~\ref{CHOptAlg4DOP}.
Two implementation issues, however, must be addressed for the present experiments:
\begin{description}
\item [Unknown words:] 
   For enabling  the parsing and disambiguation of sentences containing words that are 
   unknown to the parser, we employ a simplified version of the Add-One 
   method~\cite{ADDONE} as follows: 
   \begin{enumerate}
    \item
     The elementary-tree sets of the specialized TSG (partial-parser), DOP STSG and the 
     SDOP STSG are extended with a set of elementary-trees 
    \[
    ~~\SETM{\RuleM{POS}{UNKNOWN}}{POS~is~a~PoSTag~of~the~whole~tree-bank}
    \]
    where the symbol UNKNOWN is a new terminal symbol that is used to denote every word that
    is not a terminal of the (S)TSG. A word in the input sentence that is not a terminal of 
    the (S)TSG is replaced by the terminal UNKNOWN and then the resulting sentence is parsed.
    \item Every elementary-tree \Rule{POS}{UNKNOWN} receives a probability in the DOP STSG
       and the SDOP STSG by the Add-One method as follows: when projecting the DOP or SDOP
       STSG from the training tree-bank the frequency of this elementary-tree is set to be
       equal to one and the frequencies of all other elementary-trees obtained from the training
       tree-bank are set to one more than their actual frequency. This probability assignment
       is equivalent to an Add-One method that assumes that there is exactly one unknown word
       in the domain; this is of course a wrong assumption that results in assigning too small
       probabilities to unknown words. The only reason for making this assumption is the 
       simplicity of its implementation. 
   \end{enumerate}
   In the DOP and SDOP models that allow the computation of semantics, unknown words do not
   have semantics and thus do not allow the computation of a semantic formula for the whole input 
   sentence.
\item [Projection parameters:]
   As explained in section~\ref{SecHeuristics}, to obtain manageable DOP models,
   a DOP STSG is projected from the training tree-bank under constraints on the form of 
   the subtrees that constitute its elementary-trees; these constraints are expressed as 
   upper-bounds on: the depth (d), the number of substitution-sites (n), the 
   number of terminals (l) and the number of consecutive terminals (L) of the subtree. 
   {\em In the experiments reported in this thesis these constraints
   apply to all subtrees but the subtrees of depth~1}, i.e. subtrees of depth~1 are all 
   allowed to be elementary-trees of the projected STSG whatever the constraints are.

   In the sequel we represent the four upper-bounds by the short notation 
   $d{\bf d}n{\bf n}l{\bf l}L{\bf L}$. For example, $d4n2l7L3$ denotes 
   a DOP STSG obtained from a tree-bank such that every elementary-tree has at most depth~4, 
   and a frontier containing at most 2~substitution-sites and 7~terminals; 
   moreover, the length of any consecutive sequence of terminals on the frontier of 
   that elementary-tree is limited to 3~terminals. 

   The same constraints are used in projecting SDOP models. Note however that for SDOP models
   obtaining the subtrees takes place only at the marked nodes; the elementary-trees of the 
   specialized TSG are atomic units and thus have depth~1. The elementary-trees of the SDOP
   model are combinations of the elementary-trees of the specialized TSG. Their depth is
   therefore computed such that the atomicity of the elementary-trees of the specialized
   TSG is not violated. To highlight this issues the projection parameters for the SDOP 
   models are shortly represented by
   $D{\bf D}n{\bf n}l{\bf l}L{\bf L}$ (note~$D$ rather than~$d$ for depth of a subtree).

   Since in the present experiments all projection parameters except for the upper-bound 
   on the depth will usually be fixed, the DOP STSG obtained under a depth upper-bound that is
   equal to an integer~$i$ will be represented by the short notation DOP$^{i}$. 
   A similar notation is used for the SDOP models e.g. SDOP$^{2}$ denotes a SDOP STSG which
    has elementary-trees of depth equal or less than~2. For ISDOP models 
   (section~\ref{SecSpecDOP}), there are two depth upper-bounds for respectively the SDOP 
   model and the DOP model. Therefore, the notation ISDOP$_{i}^{j}$ will be used to denote 
   an ISDOP model that integrates an SDOP$^{i}$ with a DOP$^{j}$ model.
\end{description}
%
\section{Empirical evaluation: preface}
\label{CHIMPintro}
In sections~\ref{SecOVISExps} and~\ref{SecATISExps} we present experiments both on the OVIS 
and the ATIS domains.
Before we exhibit the results of the experiments, it is necessary to state the goals of the
experiments and the limitations of the current implementation, and to list the definitions
of the evaluation measures that are used. This section discusses exactly these issues.
%
\subsection{Goals, expectations and limitations}
The main goal of the experiments in this chapter is to explore the merits and scope
of application of ARS for specializing DOP to certain domains. Essential in this respect
is to observe the influence of the ARS on the {\em tree-language coverage}\/ 
(see chapter~\ref{CHARS}) of the specialized grammars in comparison to the original
grammars. The experiments should clearly show that ARS specialization conserves the 
tree-language coverage. Moreover, the specialization of DOP models should not result in 
unexpectedly worse {\em accuracy}\/ results than the original DOP models; 
{\em in fact, when the amount 
of training material is sufficient we expect that the ARS should not result in any 
worsening of precision at all}. And finally, we expect from specialization a significant 
improvement in time and space consumption. Of course, the extent of 
this improvement strongly depends on the ability of the current implementation of the
learning algorithms to reduce ambiguity. 

The ARS algorithm that we test
here
is the GRF algorithm of section~\ref{SecSpecAlgs}.
Although the Entropy Minimization algorithm can be expected to be superior to the GRF 
algorithm\footnote{The Entropy Minimization algorithm is expected to result 
in less ambiguous, faster and smaller specialized grammars and SDOP models.}, 
the GRF algorithm has the advantage of much faster learning. Fast learning enables thorough 
experimentation with variations of the GRF algorithm on two tree-banks and on two tasks 
(sentence and word-graph parsing and disambiguation). Of course, being an ARS algorithm,
and despite of the fact that it might not be optimal, the experiments with the GRF algorithm 
must show that it is possible to specialize grammars without loss of tree-language coverage
and with some gain in processing time and space. 

The goals of and the expectations from the experiments are of course subject to the 
limitations of the available tree-banks, the hardware and the current implementation of 
the learning algorithm. For DOP, the available hardware (SGI Indigo2 640~MB~RAM) allows
extensive experimentation on tree-banks such as the OVIS~\cite{DOPNWORep} and 
SRI-ATIS~\cite{CarterACL97} tree-banks (respectively 10000~and 13335~annotated utterances). 
These two tree-banks represent the kind of domains that have limited diversity in language use and
are therefore interesting for the task of grammar specialization.
In fact, these domains represent the typical kind of training and test material that was 
used in earlier studies of Broad-Coverage Grammar (BCG) 
specialization~\cite{SamuelssonRayner91,Samuelsson94,SrinivasJoshi95,Neumann94}.
Of course, the conclusions of any experiments on this kind of limited 
domains do not generalize to other kinds of domains.

The main limitation of the current implementation of the ARS learning algorithms 
is that they are not lexicalized, i.e. lexicon-entries do not play any role in
the learning algorithms. This limitation means that the ambiguity reduction will 
be less spectacular than one might wish or expect. Without lexicalization it is very 
hard to control the combinatorial explosion of the number of analyses. As mentioned 
earlier, this limitation is merely a byproduct of data-sparseness and limited tree-bank
annotations.
Other clear limitations of the current implementations are the sequential covering
scheme's suboptimal search strategy and the many approximations that are necessary for
achieving a system that can run on the available hardware.
%
\subsection{The systems that are compared}
To observe the effects of ARS it is necessary to compare the specialized models (resulting
from ARS training) to the original (or base-line) models.
Preceding work on BCG specialization compares the specialized grammars 
(result of EBL training) to base-line systems that are based on
full-fledged BCGs e.g. the CLE system in~\cite{SamuelssonRayner91,Samuelsson94} and 
the XTAG system in~\cite{SrinivasJoshi95}. Since in our case we are interested mainly 
in the specialization of DOP models and the effects of specialization on the total process
of parsing and disambiguation (and since we have no access to any BCG-based system), 
the comparison in this work will be mainly between the specialized DOP models and the original 
DOP models. More specifically, in the present experiments the comparison is between three parsing 
and disambiguation models: DOP, SDOP and ISDOP. The DOP model serves as the base-line 
model and the SDOP and ISDOP models are the results of the ARS training.

In addition to this comparison, we also observe whether (and to what extent) ARS training
conserves the tree-language coverage; this is achieved by comparing the specialized grammar (a TSG)
directly to {\em the CFG that underlies the tree-bank (as the base-line parser)}. 
Note that usually the grammar that underlies a tree-bank is much smaller than any BCG and 
is based on an annotation scheme that is manually tailored to a certain extent for the domain.
This implies that when comparing the specialized grammar to the CFG underlying a tree-bank,
the net gain (e.g. speed-up) from specialization can be expected to be less spectacular than 
when comparing it against a real BCG. \\

In evaluating the results of specialization it is necessary to take care that the systems 
that are being compared do not feature extra optimizations that are not related to the
specialization. Therefore,
in the present comparison {\em all compared systems are based on the same parsing and 
disambiguation implementations}\/ that are described in chapter~\ref{CHOptAlg4DOP}. The
effects of the optimizations of chapter~\ref{CHOptAlg4DOP} are therefore included in
the {\em results of all systems}. Thus, any speed-up that is observed is entirely due to the
specialization using ARS\footnote{In most preceding work on grammar specialization, the 
reported speed-up is the result of two or more factors that include EBL training. 
In fact, usually the larger portion of speed-up is due to parser-optimizations and heuristics
- e.g. best-first - that are not related to EBL.}. 
The net speed-up which results from both ARS and the other optimizations that this thesis 
presents is therefore much larger than the speed-up observed in the present 
experiments\footnote{In fact, the speed-up which the optimization of 
chapter~\ref{CHOptAlg4DOP} achieves is much larger than that achieved by ARS. This speed-up
depends on the size of the DOP model. For the OVIS and SRI-ATIS domains, this speed-up can
vary between~10-100 times. When the comparison is against the Monte-Carlo 
algorithm~\cite{RENSMonteCarlo}, the speed-up is even larger - in some cases it is
approximately~500 times.}.\\

It is important here to note that none of the parsing and disambiguation systems involved in 
the present experiments employ any preprocessing tools e.g. Part-of-Speech 
Taggers. Generally speaking, the form of the input to these systems is a word-graph 
(an SFSM as defined in chapter~\ref{CHDOPinML}) of actual words of the language; of course,
a simple sequence of words (i.e.  a sentence) is a special case of a word-graph.

The output of each of DOP STSG, SDOP STSG and ISDOP STSG is the parse generated by the 
Most Probable Derivation (MPD)
(or MPiD in the case of an actual SFSM as chapter~\ref{CHOptAlg4DOP} explains) which the
STSG assigns to the input.
Along with the parse, when the training material contains semantic information, the output
also contains semantic information. When testing the tree-language coverage of a TSG or the CFG
underlying the tree-bank, the test will consist of checking whether the parse-space which
the TSG/CFG assigns to the input contains the right parse-tree (as found in the tree-bank)
or not.

\subsection{Evaluation measures}
%
%
In an experiment, a system is trained on one portion of the available tree-bank,
the training-set, and then tested on another portion, the test-set. The partitioning
into train/test sets is achieved by a random generator. Some experiments are conducted
on various independent partitions into train/test sets and the means and standard deviations
of the measurements on the various partitions are calculated and compared.
For the task of analyzing (i.e. parsing and disambiguating) sentences, the test-set is 
considered a sequence of pairs $\tuple{sentence,~test$-$parse}$, where 
the $sentence$ is fed as input to the system that is being evaluated and the $test$-$parse$ 
is considered the correct parse for that sentence (i.e. the gold-standard).
And for the task of analyzing word-graphs, the test-set is considered a sequence of 
triples \mbox{$\tuple{sentence, word$-$graph, test$-$parse}$}, where the $word$-$graph$ is the input to
the system, the $sentence$ is the correct sequence of words and the $test$-$parse$ is the correct 
parse for the sentence\footnote{Note that here we do not demand that the word-graph necessarily 
contain the correct sentence. This takes into consideration the realistic possibility that 
the word-graph that is output by a speech-recognizer does not contain the correct utterance.}.

The output of a parsing and disambiguation system for an input (sentence or word-graph) 
is a parse-tree which we refer to here with the term the {\em output-parse}.  
To compare the various systems, various measurements are done on the output of each
system. The measurements on the output of a system (parsing and disambiguation) are conducted
at two points: 1)~after parsing but before disambiguation referred to with ``parser quality
measures", and 2)~after parsing and disambiguation referred to with ``overall-system measures".
The measurements after the parsing stage (the spanning of a parse-space for the input) enable 
the evaluation of the string-language and the tree-language of the parser. And the measurements 
after the disambiguation stage enable the evaluation of the system as a whole. Together, the
two kinds of measures enable also the evaluation of the quality of the disambiguator (i.e. the 
quality of the probability distributions). 

Some notation is necessary for the definition of the evaluation criteria:
\begin{itemize}
\item The size of the test-set $TEST$, denoted $|TEST|$, is the number of trees in $TEST$.
\item A non-terminal node-label in a parse-tree is a pair \mbox{$\tuple{syntaxL, semanticL}$} where 
      $syntaxL$ is a syntactic category and $semanticL$ is a semantic category.
      If the tree-bank does not have any semantic categories then every non-terminal 
      node-label is a scalar $syntaxL$; the scalar $syntaxL$ can be viewed as equivalent
      to a pair \mbox{$\tuple{syntaxL,SEML}$} where $SEML$ is a nil semantic category.
\item For any tree $T$: $Frontier(T)$ denotes the sentence (ordered sequence of terminals)
      on the frontier of $T$,
      $NoLex(T)$ denotes the tree resulting from $T$ after removing all
      terminal nodes, $Syntax(T)$ denotes the tree resulting from $NoLex(T)$ after
      substituting instead of every node-label $\tuple{syntaxL,semanticL}$ the pair\linebreak
      $\tuple{syntaxL,SEML}$, $Semantic(T)$ denotes the tree resulting from $NoLex(T)$ after
      substituting instead of every node-label $\tuple{syntaxL,semanticL}$ the pair\linebreak
      $\tuple{SYNL,semanticL}$ where $SYML$ is a nil syntactic category.
\item For every natural number $n\leq |TEST|$:
      $I_{n}$ and $T_{n}$ denote respectively the $n$th input sentence/word-graph 
      and the $n$th test-parse (the correct parse for $I_{n}$) in $TEST$,
      and $O_{n}$ denotes the output-parse for input $I_{n}$ (output by the system 
      that is being evaluated). 
      The nil-tree $NIL$ is output by the system whenever the system fails to compute a
      tree for the input. $NIL$ does not contain any nodes at all.
\item For every tree $T$ (not equal to $NIL$): 
      the terminals of $T$ are enumerated from left to right by 
      consecutive natural numbers starting from one. For every non-terminal node in $T$:
      if the node has label $N$ and covers a (non-empty) sequence of terminals enumerated 
      $i,\cdots,~j$ then
      that node is represented by the triple $\tuple{N,i-1,j}$ referred to as a {\em labeled-bracket}; 
      the pair $\tuple{i-1,j}$ is called the {\em non-labeled bracket}\/ of node $\tuple{N,i-1,j}$. 
      We use $LaBr(T)$ to denote the set of labeled-brackets that correspond to the 
      non-terminal nodes of tree $T$, and $NonLaBr(T)$ to denote the set of non-labeled 
      brackets that correspond to the non-terminal nodes of~$T$. The tree $T$ can be 
      represented as a pair \mbox{$\tuple{Frontier(T), LaBr(T)}$}.
      The nil-tree $NIL$ is represented by the pair $\tuple{\epsilon,\emptyset}$.
\item For any two entities A and B of the same type\footnote{
      The meaning of the symbol ``==" depends on the type of A and B. For entities of the
      type {\em trees}\/ $A==B$ denotes the proposition that A is identical to B. For
      $A$ and $B$ that are (or compute to) integers, $A==B$ denotes simply integer equivalence.
      }:
     \begin{eqnarray*}
     EQ(A, B) & = & \left\{
     \begin{tabular}{cr}
     $1$ & if $A == B$,\\
     $0$ & otherwise.\\
     \end{tabular}
     \right.
     \end{eqnarray*}
\end{itemize}
Assuming that the system that is being evaluated consists of a parser and a disambiguator,
the evaluation measures can be divided into two sections: parser quality measures and 
overall-system measures. 
The overall-system measures can be divided into three sections:
exact-match measures, bracketing measures and other measures.
For computing the overall-system measures we employ the evaluation-environment
$graph2form$~\cite{BonnemaNWORep}. 

Next we list the definitions of the measures. In these definitions, 
$\sum_{i}{}$ ranges over $1\leq i\leq |TEST|$, where $TEST$ is the test-set:
\begin{enumerate}
\item {\bf Parser quality measures:}
\begin{description}
\item [Recognized] is the percentage of test-set sentences/word-graphs that is assigned a
                   parse by the system. This measures the coverage of the string-language 
                   of the parser.
\item [TLC] or Tree-Language Coverage, is the percentage of test-set sentences/word-graphs
              for which the test-parse is found in the parse-space that the parser assigns
              to the input. This measures the coverage of the tree-language of the parser
              and the quality of the correspondence between the string-language and 
              tree-language of the parser.
\end{description}
\item {\bf Exact match measures:}
\begin{description}
\item [Exact match] is equal to the 
         percentage $\frac{\sum_{i}{EQ(O_{i}, T_{i})}}{|TEST|\mul Recognized}$,
         Note the term {\em Recognized} in the formula, i.e.
         only the inputs that are assigned a tree that is not equal to $NIL$ are 
         counted in this term.
\item [Sem. exact match] or semantic exact match, is equal to
         \[ \frac{\sum_{i}{EQ( Semantic(O_{i}), Semantic(T_{i}) )}}{|TEST|\mul Recognized} \]
\item [Syn. exact match] or syntactic exact match, is equal to
         \[ \frac{\sum_{i}{EQ( Syntax(O_{i}), Syntax(T_{i}) )}}{|TEST|\mul Recognized} \]
\end{description}
\item {\bf Bracketing measures:} PARSEVAL~\cite{PARSEVAL} measures.
\begin{description}
\item [Labeled syntactic:]
      Let $SynLBO_{i}$ denote the set $LabBracket(Syntax(O_{i}))$ and
                 $SynLBT_{i}$ denote the set $LabBracket(Syntax(T_{i}))$:
  \begin{description}
  \item [Syn. LBR] or Syntactic Labeled Bracketing Recall; 
       \[ \frac{\sum_{i}{|SynLBO_{i}\cap SynLBT_{i}|}}{\sum_{i}{|SynLBT_{i}|}} \]
  \item [Syn. LBP] or Syntactic Labeled Bracketing Precision,
       \[ \frac{\sum_{i}{|SynLBO_{i}\cap SynLBT_{i}|}}{\sum_{i}{|SynLBO_{i}|}} \]
  \end{description}
\item [Labeled semantic:]
      Let $SemLBO_{i}$ denote the set $LabBracket(Semantic(O_{i}))$ and
                 $SemLBT_{i}$ denote the set $LabBracket(Semantic(T_{i}))$:
  \begin{description}
  \item [Sem. LBR] or Semantic Labeled Bracketing Recall, 
       \[ \frac{\sum_{i}{|SemLBO_{i}\cap SemLBT_{i}|}}{\sum_{i}{|SemLBT_{i}|}} \]
  \item [Sem. LBP] or Semantic Labeled Bracketing Precision,
       \[ \frac{\sum_{i}{|SemLBO_{i}\cap SemLBT_{i}|}}{\sum_{i}{|SemLBO_{i}|}} \]
  \end{description}
\item [Non-labeled:] Bracketing measures on non-labeled trees:
  \begin{description}
   \item [BR] or Non-Labeled Bracketing Recall,
       \[ \frac{\sum_{i}{|NonLaBr(O_{i}) \cap NonLaBr(T_{i})|}}
               {\sum_{i}{|NonLaBr(T_{i})|}} \]
   \item [BP] or Non-Labeled Bracketing Precision,
       \[ \frac{\sum_{i}{|NonLaBr(O_{i}) \cap NonLaBr(T_{i})|}}
               {\sum_{i}{|NonLaBr(O_{i})|}} \]
  \end{description}
\item [Crossing-measures:]
  Two non-labeled brackets $\tuple{h,j}$  and $\tuple{k,l}$ are called {\em crossing}\/ iff
  \mbox{$h < k < j < l$} or \mbox{$k < h < l < j$}. For every two sets of non-labeled
  brackets $U$ and $V$, $Cross(U,V)$ denotes the subset of brackets from $U$ that cross 
  with a least one bracket in $V$.
  \begin{description}
    \item [NCB recall] or Non-Crossing Brackets Recall, 
       \[ \frac{\sum_{i}{| NonLaBr(O_{i}) | -
                         | Cross( NonLaBr(O_{i}), NonLaBr(T_{i}) ) |}}
               {\sum_{i}{|NonLaBr(T_{i})|}} \]
    \item [NCB Precision] or Non-Crossing Brackets Precision,
       \[ \frac{\sum_{i}{| NonLaBr(O_{i}) | -
                         | Cross( NonLaBr(O_{i}), NonLaBr(T_{i}) ) |}}
               {\sum_{i}{|NonLaBr(O_{i})|}} \]
    \item [0-Crossing] or percentage of Zero-Crossing sentences,
           \[ \frac{\sum_{i} EQ( 0, Cross(O_{i}, T_{i}) )}{|TEST| * Recognized} \]
  \end{description}
\item [Input characteristics:]
      Some empirical information on the test-set:
  \begin{description}
   \item [\#of sens/WGs] or number of sentences/Word-Graphs in test-set $TEST$, 
          i.e. $|TEST|$. This measure is necessary since in some experiments the 
          results are reported only for some portion of the test-set. For example
          when parsing sentences the reported results are only for sentences that 
          contain at least two words. And when parsing word-graphs, in some tables
          the reported results concern only word-graphs that contain the correct sentence.
   \item [Sen. Length] or average Sentence Length, is the mean length of the sentences
         in the test-set $TEST$.
   \item [\#states in WGs] or the average number of states in the word-graphs in $TEST$.
   \item [\#trans in WGs] or the average number of transitions in the word-graphs in $TEST$.
  \end{description}
\item [CPU (secs.)] is the Average CPU-time that the parsing and disambiguation system
       consumes in order to process the input in test-set $TEST$. The CPU-time is reported
       in seconds and is the time consumed by the CPU as reported by the UNIX system on
       an Indigo2 machine (one R10000~2.5 processor, 640MB~RAM, IRIX64~6.2).
\end{description}
\end{enumerate}
In many experiments we report the mean and standard deviation of every measure computed
from the results of multiple partitions of the tree-bank into test/train sets.
In those cases, the standard deviation figure is reported between brackets to the right 
of the mean, i.e. \mbox{{\bf $mean$ ($std$)}}.
For convenience, all percentage figures in the tables are truncated at two
digits after the decimal point; then the rightmost digit (e.g. 2 in 67.62\%) was rounded 
to the closest of the three digits 0, 5 or~10 (the latter implies incrementing the 
neighboring digit). 
The standard deviations as well as all other figures that are not percentages (e.g. CPU-times)
are not rounded in this manner.
%
%

\newcommand{\REFTAB}[1]{\ref{#1} (page~\pageref{#1})}
\newcommand{\REFSUBSUBSEC}[2]{\ref{#1}.#2}
%
%
%
\begin{table}
\center{
\begin{tabular}{|c||c|c|c|c|}
\hline
Annotation  &\multicolumn{2}{|c|}{\#grammar symbols} & \multicolumn{2}{|c|}{\#grammar rules}  \\
Annotation  & non-terminals & terminals & lexical & non-lexical \\
\hline
Syntax+semantics & 433              &   816        &  1014            &  1708 \\
Syntax           & 43               &  816         &  921             &  433  \\
\hline
\end{tabular}
}
\caption{The OVIS tree-bank in numbers}
\label{FigOVISinNUms}
\end{table}

\begin{figure}[hbt]
\epsfxsize=15cm
\epsfbox{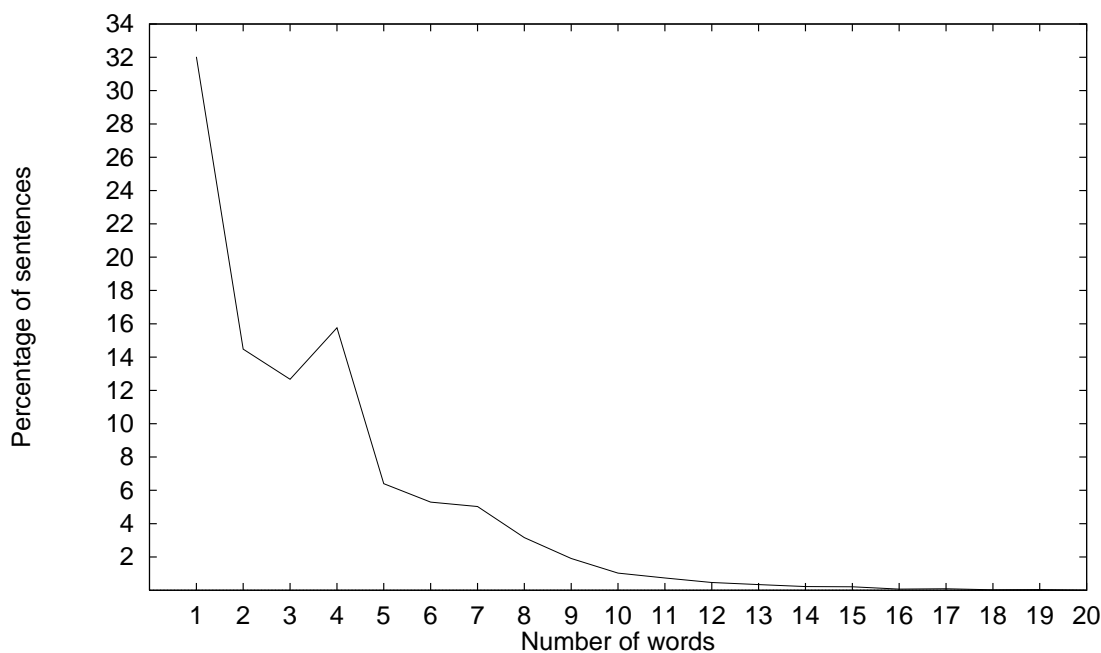}
\caption{OVIS tree-bank: percentage of sentences as a function of the number of words per sentence}
\label{FigSenLToWords}
\end{figure}

\section{Experiments on OVIS tree-bank}
\label{SecOVISExps}
The Amsterdam OVIS tree-bank\footnote{OpenbaarVervoer Informatie Systeem (OVIS) stands for
Public Transport Information System.} contains~10000 syntactic and semantic trees.
Each of the trees is the syntactic and semantic analysis of a transcribed user utterance.
The user utterances are typically answers to questions asked by the system in a dialogue 
that has the goal of filling the slots in a predefined form that represents a travel plan.
The slots in the form typically specify travel information e.g. a travel destination,
point of departure, day and time of departure or arrival. But also other kinds of 
information pertaining to the state of the dialogue, e.g. denial of earlier specified values. 

For detailed information concerning the syntactic and semantic annotation scheme of the
OVIS tree-bank we refer the reader to~\cite{DOPNWORep}. Here we describe the tree-bank only
briefly. The syntactic annotation of the OVIS tree-bank, although phrase-structure, does not 
completely conform to any existing linguistic theory. It does not contain traces of movements
or cyclic constructions and there is no partitioning of the non-terminal symbols into PoSTags 
and higher level phrasal symbols. The semantic annotation decorates the syntactic one at 
the nodes of the trees. It is based on the update-formalism~\cite{VeldVZanten}, 
which is a typed language specific for the OVIS domain. The semantic expression associated 
with an OVIS utterance is an {\em update-expression}\/ from the update-formalism; the
update-expression has a type in that formalism. In the OVIS tree-bank, the nodes of the 
syntactic trees are annotated with {\em semantic types}.
If the correct semantic expression of a node in a tree in the tree-bank can be computed from 
the semantics of its daughter nodes in the update-formalism, the node is decorated with the 
{\em type}\/ of that semantic expression. Following the philosophy of DOP on semantics 
(see~\cite{Scha90,vdBergEtAl,BonnemaEtAl97}), if the correct semantic-expression of a node
in a tree cannot be computed compositionally, then the node is directly decorated with the 
type of the correct semantic expression; if there is no semantic expression (in the formalism) 
that expresses an intuitively plausible meaning for the constituent dominated by the node,
the node is left unannotated semantically. 

Apart from the decoration of the nodes with semantic types, a rewrite system is associated 
with the trees in the tree-bank. In this rewrite-system, a ``ground" expression from the 
update-formalism is associated with every terminal (word of the language). And a {\em function}\/
is associated with every pair \mbox{$\tuple{rule, type}$} in the 
tree-bank, where \mbox{$type$} is the type decorating the node which constitutes the
left-hand side of \mbox{$rule$}, which is a syntactic rule in a tree in the tree-bank.
The function associated with the rule under a node in the tree-bank enables computing 
the semantic expression of the node from the 
semantic-expressions of its daughter nodes. 

Currently, the semantic and syntactic annotations are treated by the DOP model as 
one annotation in which the labels of the nodes in the trees are 
a juxtaposition of the syntactic and semantic labels.
This means that the DOP models that are projected from a syntactically and semantically 
annotated tree-bank are also STSGs. Although this results in many more non-terminal 
symbols (and thus also DOP model parameters), \cite{BonnemaScr,BonnemaEtAl96} show that the
resulting syntactic+semantic DOP models are better than the mere syntactic DOP models.
Here, we follow this practice of combining semantics with syntax into one simple formalism.\\

It is worth noting that a large portion of the OVIS tree-bank was 
annotated semi-automatically using the DOPDIS system (described in chapter~\ref{CHOptAlg4DOP}).
The annotation was conducted in cycles of training DOPDIS on the existing annotated 
material and then using it for semi-automatically annotating new material.

Along with the OVIS tree-bank, there is also an OVIS corpus of the actually spoken 
utterances and word-graphs that were hypothesized by a speech-recognizer for these 
spoken utterances. A spoken utterance and a word-graph in the corpus are associated 
with a transcribed utterance and its analysis in the tree-bank.\\
%
%
%
Table~\REFTAB{FigOVISinNUms} and figure~\REFTAB{FigSenLToWords} summarize the OVIS tree-bank
characteristic numbers and show the graphs of percentage of sentences to number of words.
The average sentence length for all sentences is~3.43 words. However, the results that
we report are only for sentences that contain at least two words; the numbers of those
sentences is~6797 and their average length is~4.57 words.
\subsection{Early experiments: January 1997}
\label{SecOVISOldExps}
In earlier work~\cite{MyNWORep96,MyACLWkSh97,MyECMLWkSh97} we reported experiments 
testing an early version of the GRF~algorithm (section~\ref{SecSpecAlgs}) on portions 
of the OVIS and the ATIS~\cite{HemphilEtAl90} domains.
Apart from the difference in the tree-banks involved, these experiments differ from 
the present experiment in some essential aspects that can be summarized as follows:
   \begin{description}
     \item [Competitors definition:]
       Rather than defining the set of competitors of an SSF to contain the competitors from
       all trees of the tree-bank, in this early implementation the set of competitors of an
       SSF contained only the competitors from the same tree as the SSF. This implies that an
       SSF may have a different set of competitors in a different tree of the tree-bank.
       As a consequence, an SSF might be learned only from some of the tree-bank trees that
       contain it, thereby jeopardizing the tree-language coverage ARS requirement.
     \item [Local context effects:]
       An SSF was learned only in those contexts in which its GRF exceeded its competitors in
       the same contexts. Since the context was not used during parsing, this also contributed
       to jeopardizing the tree-language coverage ARS requirement.
     \item [No Ambiguity-Sets Completion:]
       The learning algorithms did not employ any mechanism for completing the ambiguity-sets 
       as explained in section~\ref{CompoAmbS}.
     \item [Specialized DOP:] In these early experiments, the result of the learning algorithm
       was a tree-bank of trees containing marked nodes; some trees that did not fully reduce
       during learning contain unmarked nodes that are not dominated by any marked nodes.
       Rather than marking the roots of these trees (as done in the present algorithms) 
       {\em all}\/ these unmarked nodes were simply also marked as cut nodes. Therefore the
       resulting SDOP models differ from the current SDOP models in that the projection of
       subtrees took place also at these nodes.
   \end{description}
The following observations were made on basis of these early experiments: 
 \begin{itemize}
  \item In parsing OVIS and ATIS sentences the specialized TSG lost some tree-language coverage
        when compared to the CFG underlying the tree-bank. However, this loss of 
        tree-language coverage hardly affected the accuracy results of SDOP as compared to 
        DOP~\cite{MyACLWkSh97,MyECMLWkSh97}.
  \item In parsing OVIS word-graphs the SDOP models exhibited hardly any loss of precision
        or recall but enabled an average speed-up of~10 times in comparison with the original
        DOP models. The speed-up was larger on larger word-graphs. However, in these experiments
        many (about 36\%) word-graphs did not contain the right sentence (since the 
        speech-recognizer was still in its early training stages). This meant that on these
        word-graphs both DOP as well as SDOP scored zero recall and zero precision. Since these
        word-graphs were typically the hardest, neither SDOP nor DOP had the chance
        to improve on the other. See~\cite{MyNWORep96} for the details of these extensive
        experiments on parsing and disambiguation of word-graphs that are output by a 
        speech-recognizer. 
 \end{itemize}
The conclusion of these early experiments was clear: the speed-up that was achieved is
substantial and the SDOP models had coverage or accuracy that were comparable to the DOP
models. However, the loss of tree-language coverage in the specialized TSG was alarming and 
needed a remedy. 
After inspecting the implementation detail of the systems it turned out that the main
reason for the loss of tree-language coverage is simply that the implementation did not 
try to conserve the tree-language coverage in the first place; the dependency of the definition 
of the competitors of an SSFs on the current parse-tree and the dependency of the GRF function
on local context implied that the same SSF could be learned in one context but not learned
in many others.
This clearly undermines the tree-language coverage of the specialized grammar since it
makes our assumption concerning the completeness of the ambiguity-sets 
not justified. 
\subsection{Recent experiments on OVIS}
\label{SecOVISNewExps}
Theoretically speaking, the new versions of the ARS learning algorithms (as discussed in
chapter~\ref{CHARS}) are equipped with better mechanisms that enable conserving the 
tree-language coverage (as explained in sections~\ref{SecSufComp} and~\ref{SecQofCDisc}).
%
%
In the rest of this section we exhibit a new series of experiments that tests the
GRF algorithm of chapter~\ref{CHARS}, implemented as explained in 
section~\ref{SecImpDet}, on the OVIS domain. The present experiments 
are partitioned into three subseries: 
1)~a subseries that trains the models on {\em full-annotation}\/ (syntactic-semantic) and tests
  them on {\em utterances},
2)~a subseries that trains the models on {\em syntactic annotation}\/ only and tests them 
  on utterances, and
3)~a subseries that trains the models on full-annotation and tests
   them on {\em word-graphs}.


The present experiments observe the effect of various training parameters on the 
results of training the DOP / SDOP / ISDOP models. 
These training parameters are: the training tree-bank size,
the upper-bound on the depth of elementary-trees that the DOP/SDOP/ISDOP STSGs contain,
and the definition of the target-concept (either as an individual SSF or a generalization
of it into equivalence classes as specified in section~\ref{SecImpDet}). 
Other experiments test for the mean and standard deviations of each of the three models 
on different random partitions of the tree-bank into independent test and training sets.

Some of the training parameters were fixed in all experiments that are reported in the rest 
of this section: the upper-bound on the length of learned SSFs was set to~8 grammar symbols,
the threshold on the frequency of sequences of symbols was set on~5, the threshold on
the Constituency Probability of SSFs was set on~0.95, all DOP/SDOP/ISDOP STSG were
projected under the projection parameters~n2l7L3 (see section~\ref{SecImpPDet}). 

Since in the OVIS domain many of the utterances (approx. 33\%) consist of only one word 
(e.g. ``yes" or ``no"), the figures reported below compare the results of the parsers and 
disambiguators {\em only on utterances that contain at least two words}. This avoids trivial
input that might diminish the differences between the results of the various systems.
%
%
%

{
\begin{table}[htb]
\begin{center}
\begin{tabular}{|c||c|c|c||c|c|c|}
\hline
 & \multicolumn{3}{|c|}{B2000} & \multicolumn{3}{c|}{B4000} \\
 & DOP$^{4}$  &  ISDOP$_{2}^{4}$  &  SDOP$^{2}$  &  DOP$^{4}$  &  ISDOP$_{2}^{4}$  &  SDOP$^{2}$ \\
\hline
{\sl Recognized} &  89.10 & 88.10 & 80.15 & 91.90 & 91.60 & 85.00 \\
{\sl TLC} & 77.79  & 77.79  & 77.79  & 83.67 & 83.67 & 83.67 \\
Exact match &  83.65 & 82.95 & 86.05 & 87.35 & 86.85 & 89.80 \\
Syn. ex. match &  93.40 & 93.80 & 95.80 & 95.20 & 95.00 & 96.70 \\
Sem. ex. match &  84.50 & 84.15 & 86.95 & 88.00 & 87.65 & 90.15 \\
NCB recall &  84.75 & 82.85 & 70.60 & 88.25 & 87.65 & 77.45 \\
NCB prec. &  99.40 & 99.50 & 99.75 & 99.60 & 99.60 & 99.80 \\
0-Crossing &  96.85 & 97.35 & 98.55 & 97.30 & 97.75 & 98.80 \\
Syn. LBR &  83.70 & 81.70 & 70.00 & 87.60 & 86.90 & 77.05 \\
Syn. LBP &  98.20 & 98.15 & 98.90 & 98.85 & 98.75 & 99.25 \\
Sem. LBR &  81.50 & 79.70 & 68.50 & 86.05 & 85.20 & 75.85 \\
Sem. LBP &  95.60 & 95.75 & 96.80 & 97.10 & 96.85 & 97.70 \\
CPU (sec.) &  0.57 & 0.25 & 0.19 & 0.99 & 0.33 & 0.28 \\
Sen. length & \multicolumn{3}{|c|}{  4.40 } & \multicolumn{3}{|c|}{ 4.45 } \\
\# of sens & \multicolumn{6}{|c|}{  680.00 } \\
\hline
%
\hline
 & \multicolumn{3}{|c|}{B6000} & \multicolumn{3}{c|}{B9000} \\
 & DOP$^{4}$  &  ISDOP$_{2}^{4}$  &  SDOP$^{2}$  &  DOP$^{4}$  &  ISDOP$_{2}^{4}$  &  SDOP$^{2}$ \\
\hline
{\sl Recognized} &  94.10 & 93.40 & 88.25 & 95.15 & 94.40 & 91.30 \\
{\sl TLC} & 86.32 & 86.32 & 86.32 & 89.11 & 89.11 & 89.11 \\
Exact match &  88.75 & 88.35 & 90.50 & 89.50 & 88.65 & 90.00 \\
Syn. ex. match &  96.10 & 95.30 & 96.50 & 95.50 & 94.40 & 95.95 \\
Sem. ex. match &  89.20 & 89.00 & 90.85 & 89.95 & 89.25 & 90.65 \\
NCB recall &  90.70 & 89.40 & 81.60 & 92.15 & 90.80 & 86.55 \\
NCB prec. &  99.65 & 99.40 & 99.65 & 99.60 & 99.60 & 99.80 \\
0-Crossing &  98.45 & 97.30 & 98.15 & 97.55 & 97.35 & 98.40 \\
Syn. LBR &  90.00 & 88.60 & 81.10 & 91.55 & 90.00 & 86.05 \\
Syn. LBP &  98.90 & 98.50 & 99.00 & 98.95 & 98.65 & 99.20 \\
Sem. LBR &  88.55 & 87.20 & 80.00 & 90.35 & 88.55 & 84.95 \\
Sem. LBP &  97.30 & 96.95 & 97.65 & 97.60 & 97.05 & 97.95 \\
CPU (sec.) &  1.70 & 0.48 & 0.43 & 2.43 & 0.68 & 0.65 \\
Sen. length &  \multicolumn{3}{|c|}{ 4.45 } &  \multicolumn{3}{|c|}{ 4.50}\\
\# of sens &  \multicolumn{6}{|c|}{ 680.00 } \\
\hline
\end{tabular}
\end{center}
\caption{Varying training TB-size: results for sentence length $> 1$}
\label{FigTBsizeTable}
\end{table}
}

\begin{table}[htb]
\center{
\begin{tabular}{|c||c|c|c|||c|c|c|}
\hline
 & \multicolumn{3}{|c|||}{Number of elementary-trees} &
   \multicolumn{3}{|c|}{Number of internal-nodes}\\
 & \multicolumn{3}{|c|||}{(not including the lexicon)} &
   \multicolumn{3}{|c|}{(not including the lexicon)}\\
TB-size & Spec. TSG & SDOP$^{2}$ & DOP$^{4}$ & Spec. TSG & SDOP$^{2}$ & DOP$^{4}$\\
\hline
2000 & 549 & 4830 & 20215 & 2331 & 24284 & 108211 \\
4000 & 842  & 7725  & 33822  & 3288     & 37805    & 185041    \\
6000 & 1091 & 10174 & 46227 & 4069    & 49028    & 256908    \\
9000 & 1355 & 12824 & 61425 & 4883    & 60798    & 346479    \\
\hline
\end{tabular}
}
\caption{Grammar size as a function of tree-bank size}
\label{FigTBsizeGSize}
\end{table}

\begin{figure}[hbt]
\center{
\setlength{\unitlength}{0.240900pt}
\ifx\plotpoint\undefined\newsavebox{\plotpoint}\fi
\sbox{\plotpoint}{\rule[-0.200pt]{0.400pt}{0.400pt}}%
\begin{picture}(1500,900)(0,0)
\font\gnuplot=cmr10 at 10pt
\gnuplot
\sbox{\plotpoint}{\rule[-0.200pt]{0.400pt}{0.400pt}}%
\put(220.0,113.0){\rule[-0.200pt]{0.400pt}{184.048pt}}
\put(220.0,113.0){\rule[-0.200pt]{4.818pt}{0.400pt}}
\put(198,113){\makebox(0,0)[r]{$70$}}
\put(1416.0,113.0){\rule[-0.200pt]{4.818pt}{0.400pt}}
\put(220.0,240.0){\rule[-0.200pt]{4.818pt}{0.400pt}}
\put(198,240){\makebox(0,0)[r]{$75$}}
\put(1416.0,240.0){\rule[-0.200pt]{4.818pt}{0.400pt}}
\put(220.0,368.0){\rule[-0.200pt]{4.818pt}{0.400pt}}
\put(198,368){\makebox(0,0)[r]{$80$}}
\put(1416.0,368.0){\rule[-0.200pt]{4.818pt}{0.400pt}}
\put(220.0,495.0){\rule[-0.200pt]{4.818pt}{0.400pt}}
\put(198,495){\makebox(0,0)[r]{$85$}}
\put(1416.0,495.0){\rule[-0.200pt]{4.818pt}{0.400pt}}
\put(220.0,622.0){\rule[-0.200pt]{4.818pt}{0.400pt}}
\put(198,622){\makebox(0,0)[r]{$90$}}
\put(1416.0,622.0){\rule[-0.200pt]{4.818pt}{0.400pt}}
\put(220.0,750.0){\rule[-0.200pt]{4.818pt}{0.400pt}}
\put(198,750){\makebox(0,0)[r]{$95$}}
\put(1416.0,750.0){\rule[-0.200pt]{4.818pt}{0.400pt}}
\put(220.0,877.0){\rule[-0.200pt]{4.818pt}{0.400pt}}
\put(198,877){\makebox(0,0)[r]{$100$}}
\put(1416.0,877.0){\rule[-0.200pt]{4.818pt}{0.400pt}}
\put(321.0,113.0){\rule[-0.200pt]{0.400pt}{4.818pt}}
\put(321,68){\makebox(0,0){$1$}}
\put(321.0,857.0){\rule[-0.200pt]{0.400pt}{4.818pt}}
\put(423.0,113.0){\rule[-0.200pt]{0.400pt}{4.818pt}}
\put(423,68){\makebox(0,0){$2$}}
\put(423.0,857.0){\rule[-0.200pt]{0.400pt}{4.818pt}}
\put(524.0,113.0){\rule[-0.200pt]{0.400pt}{4.818pt}}
\put(524,68){\makebox(0,0){$3$}}
\put(524.0,857.0){\rule[-0.200pt]{0.400pt}{4.818pt}}
\put(625.0,113.0){\rule[-0.200pt]{0.400pt}{4.818pt}}
\put(625,68){\makebox(0,0){$4$}}
\put(625.0,857.0){\rule[-0.200pt]{0.400pt}{4.818pt}}
\put(727.0,113.0){\rule[-0.200pt]{0.400pt}{4.818pt}}
\put(727,68){\makebox(0,0){$5$}}
\put(727.0,857.0){\rule[-0.200pt]{0.400pt}{4.818pt}}
\put(828.0,113.0){\rule[-0.200pt]{0.400pt}{4.818pt}}
\put(828,68){\makebox(0,0){$6$}}
\put(828.0,857.0){\rule[-0.200pt]{0.400pt}{4.818pt}}
\put(929.0,113.0){\rule[-0.200pt]{0.400pt}{4.818pt}}
\put(929,68){\makebox(0,0){$7$}}
\put(929.0,857.0){\rule[-0.200pt]{0.400pt}{4.818pt}}
\put(1031.0,113.0){\rule[-0.200pt]{0.400pt}{4.818pt}}
\put(1031,68){\makebox(0,0){$8$}}
\put(1031.0,857.0){\rule[-0.200pt]{0.400pt}{4.818pt}}
\put(1132.0,113.0){\rule[-0.200pt]{0.400pt}{4.818pt}}
\put(1132,68){\makebox(0,0){$9$}}
\put(1132.0,857.0){\rule[-0.200pt]{0.400pt}{4.818pt}}
\put(1233.0,113.0){\rule[-0.200pt]{0.400pt}{4.818pt}}
\put(1233,68){\makebox(0,0){$10$}}
\put(1233.0,857.0){\rule[-0.200pt]{0.400pt}{4.818pt}}
\put(1335.0,113.0){\rule[-0.200pt]{0.400pt}{4.818pt}}
\put(1335,68){\makebox(0,0){$11$}}
\put(1335.0,857.0){\rule[-0.200pt]{0.400pt}{4.818pt}}
\put(1436.0,113.0){\rule[-0.200pt]{0.400pt}{4.818pt}}
\put(1436,68){\makebox(0,0){$12$}}
\put(1436.0,857.0){\rule[-0.200pt]{0.400pt}{4.818pt}}
\put(220.0,113.0){\rule[-0.200pt]{292.934pt}{0.400pt}}
\put(1436.0,113.0){\rule[-0.200pt]{0.400pt}{184.048pt}}
\put(220.0,877.0){\rule[-0.200pt]{292.934pt}{0.400pt}}
\put(25,800){\makebox(0,0){{\small\bf \% Recognized}}}
\put(916,-22){\makebox(0,0){{\bf Training-tree-bank size in $10^3$ units}}}
\put(220.0,113.0){\rule[-0.200pt]{0.400pt}{184.048pt}}
\put(1306,812){\makebox(0,0)[r]{DOP$^{4}$}}
\put(1328.0,812.0){\rule[-0.200pt]{15.899pt}{0.400pt}}
\put(423,599){\usebox{\plotpoint}}
\multiput(423.00,599.58)(1.407,0.499){141}{\rule{1.222pt}{0.120pt}}
\multiput(423.00,598.17)(199.463,72.000){2}{\rule{0.611pt}{0.400pt}}
\multiput(625.00,671.58)(1.820,0.499){109}{\rule{1.550pt}{0.120pt}}
\multiput(625.00,670.17)(199.783,56.000){2}{\rule{0.775pt}{0.400pt}}
\multiput(828.00,727.58)(5.920,0.497){49}{\rule{4.777pt}{0.120pt}}
\multiput(828.00,726.17)(294.085,26.000){2}{\rule{2.388pt}{0.400pt}}
\put(423,599){\raisebox{-.8pt}{\makebox(0,0){$\Diamond$}}}
\put(625,671){\raisebox{-.8pt}{\makebox(0,0){$\Diamond$}}}
\put(828,727){\raisebox{-.8pt}{\makebox(0,0){$\Diamond$}}}
\put(1132,753){\raisebox{-.8pt}{\makebox(0,0){$\Diamond$}}}
\sbox{\plotpoint}{\rule[-0.400pt]{0.800pt}{0.800pt}}%
\put(1306,767){\makebox(0,0)[r]{ISDOP$^{2}$}}
\put(1328.0,767.0){\rule[-0.400pt]{15.899pt}{0.800pt}}
\put(423,574){\usebox{\plotpoint}}
\multiput(423.00,575.41)(1.139,0.501){171}{\rule{2.016pt}{0.121pt}}
\multiput(423.00,572.34)(197.816,89.000){2}{\rule{1.008pt}{0.800pt}}
\multiput(625.00,664.41)(2.232,0.502){85}{\rule{3.730pt}{0.121pt}}
\multiput(625.00,661.34)(195.257,46.000){2}{\rule{1.865pt}{0.800pt}}
\multiput(828.00,710.41)(6.257,0.504){43}{\rule{9.928pt}{0.121pt}}
\multiput(828.00,707.34)(283.394,25.000){2}{\rule{4.964pt}{0.800pt}}
\sbox{\plotpoint}{\rule[-0.500pt]{1.000pt}{1.000pt}}%
\put(423,574){\makebox(0,0){$+$}}
\put(625,663){\makebox(0,0){$+$}}
\put(828,709){\makebox(0,0){$+$}}
\put(1132,734){\makebox(0,0){$+$}}
\put(1306,722){\makebox(0,0)[r]{SDOP$_{2}^{4}$}}
\multiput(1328,722)(41.511,0.000){2}{\usebox{\plotpoint}}
\put(1394,722){\usebox{\plotpoint}}
\put(423,371){\usebox{\plotpoint}}
\multiput(423,371)(35.377,21.717){6}{\usebox{\plotpoint}}
\multiput(625,495)(38.423,15.710){5}{\usebox{\plotpoint}}
\multiput(828,578)(40.240,10.192){8}{\usebox{\plotpoint}}
\put(1132,655){\usebox{\plotpoint}}
\put(423,371){\raisebox{-.8pt}{\makebox(0,0){$\Box$}}}
\put(625,495){\raisebox{-.8pt}{\makebox(0,0){$\Box$}}}
\put(828,578){\raisebox{-.8pt}{\makebox(0,0){$\Box$}}}
\put(1132,655){\raisebox{-.8pt}{\makebox(0,0){$\Box$}}}
\end{picture}
}
\caption{\%Recognized sentences as a function of tree-bank size}
\label{FigTBsizeCOV}
\end{figure}
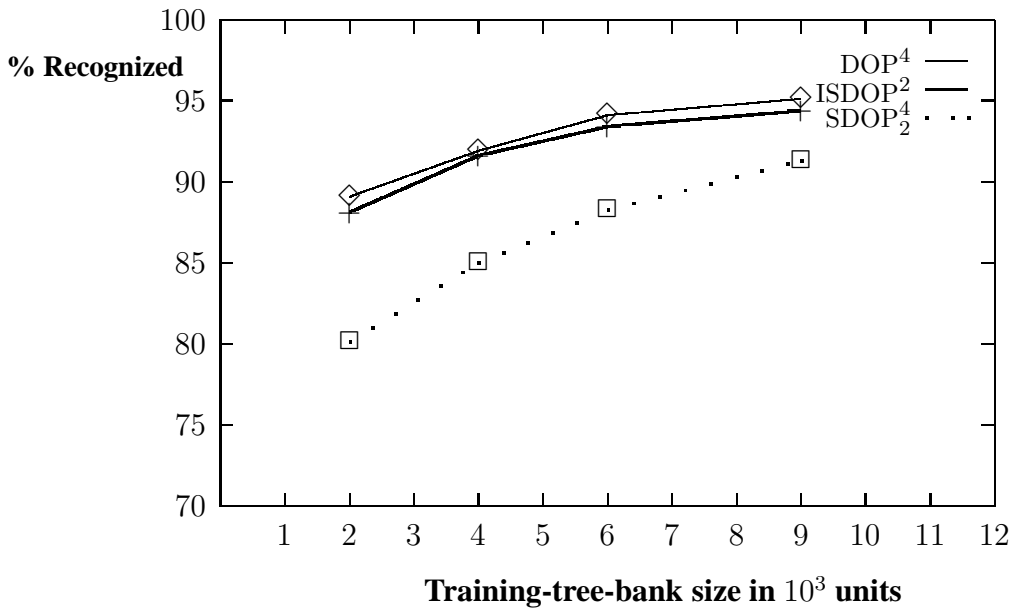
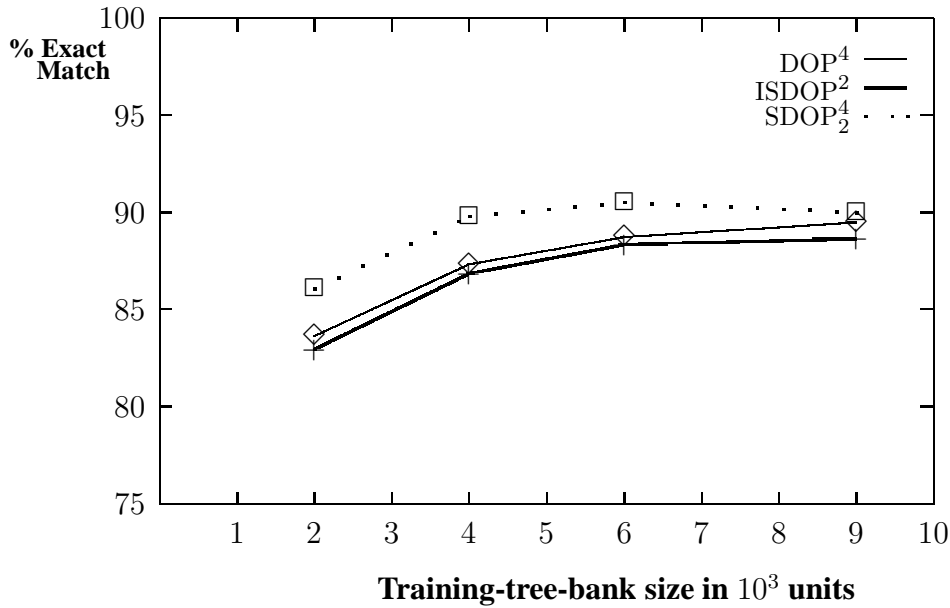
\begin{figure}[hbt]
\center{
\setlength{\unitlength}{0.240900pt}
\ifx\plotpoint\undefined\newsavebox{\plotpoint}\fi
\sbox{\plotpoint}{\rule[-0.200pt]{0.400pt}{0.400pt}}%
\begin{picture}(1500,900)(0,0)
\font\gnuplot=cmr10 at 10pt
\gnuplot
\sbox{\plotpoint}{\rule[-0.200pt]{0.400pt}{0.400pt}}%
\put(220.0,113.0){\rule[-0.200pt]{0.400pt}{184.048pt}}
\put(220.0,113.0){\rule[-0.200pt]{4.818pt}{0.400pt}}
\put(198,113){\makebox(0,0)[r]{$75$}}
\put(1416.0,113.0){\rule[-0.200pt]{4.818pt}{0.400pt}}
\put(220.0,266.0){\rule[-0.200pt]{4.818pt}{0.400pt}}
\put(198,266){\makebox(0,0)[r]{$80$}}
\put(1416.0,266.0){\rule[-0.200pt]{4.818pt}{0.400pt}}
\put(220.0,419.0){\rule[-0.200pt]{4.818pt}{0.400pt}}
\put(198,419){\makebox(0,0)[r]{$85$}}
\put(1416.0,419.0){\rule[-0.200pt]{4.818pt}{0.400pt}}
\put(220.0,571.0){\rule[-0.200pt]{4.818pt}{0.400pt}}
\put(198,571){\makebox(0,0)[r]{$90$}}
\put(1416.0,571.0){\rule[-0.200pt]{4.818pt}{0.400pt}}
\put(220.0,724.0){\rule[-0.200pt]{4.818pt}{0.400pt}}
\put(198,724){\makebox(0,0)[r]{$95$}}
\put(1416.0,724.0){\rule[-0.200pt]{4.818pt}{0.400pt}}
\put(220.0,877.0){\rule[-0.200pt]{4.818pt}{0.400pt}}
\put(198,877){\makebox(0,0)[r]{$100$}}
\put(1416.0,877.0){\rule[-0.200pt]{4.818pt}{0.400pt}}
\put(342.0,113.0){\rule[-0.200pt]{0.400pt}{4.818pt}}
\put(342,68){\makebox(0,0){$1$}}
\put(342.0,857.0){\rule[-0.200pt]{0.400pt}{4.818pt}}
\put(463.0,113.0){\rule[-0.200pt]{0.400pt}{4.818pt}}
\put(463,68){\makebox(0,0){$2$}}
\put(463.0,857.0){\rule[-0.200pt]{0.400pt}{4.818pt}}
\put(585.0,113.0){\rule[-0.200pt]{0.400pt}{4.818pt}}
\put(585,68){\makebox(0,0){$3$}}
\put(585.0,857.0){\rule[-0.200pt]{0.400pt}{4.818pt}}
\put(706.0,113.0){\rule[-0.200pt]{0.400pt}{4.818pt}}
\put(706,68){\makebox(0,0){$4$}}
\put(706.0,857.0){\rule[-0.200pt]{0.400pt}{4.818pt}}
\put(828.0,113.0){\rule[-0.200pt]{0.400pt}{4.818pt}}
\put(828,68){\makebox(0,0){$5$}}
\put(828.0,857.0){\rule[-0.200pt]{0.400pt}{4.818pt}}
\put(950.0,113.0){\rule[-0.200pt]{0.400pt}{4.818pt}}
\put(950,68){\makebox(0,0){$6$}}
\put(950.0,857.0){\rule[-0.200pt]{0.400pt}{4.818pt}}
\put(1071.0,113.0){\rule[-0.200pt]{0.400pt}{4.818pt}}
\put(1071,68){\makebox(0,0){$7$}}
\put(1071.0,857.0){\rule[-0.200pt]{0.400pt}{4.818pt}}
\put(1193.0,113.0){\rule[-0.200pt]{0.400pt}{4.818pt}}
\put(1193,68){\makebox(0,0){$8$}}
\put(1193.0,857.0){\rule[-0.200pt]{0.400pt}{4.818pt}}
\put(1314.0,113.0){\rule[-0.200pt]{0.400pt}{4.818pt}}
\put(1314,68){\makebox(0,0){$9$}}
\put(1314.0,857.0){\rule[-0.200pt]{0.400pt}{4.818pt}}
\put(1436.0,113.0){\rule[-0.200pt]{0.400pt}{4.818pt}}
\put(1436,68){\makebox(0,0){$10$}}
\put(1436.0,857.0){\rule[-0.200pt]{0.400pt}{4.818pt}}
\put(220.0,113.0){\rule[-0.200pt]{292.934pt}{0.400pt}}
\put(1436.0,113.0){\rule[-0.200pt]{0.400pt}{184.048pt}}
\put(220.0,877.0){\rule[-0.200pt]{292.934pt}{0.400pt}}
\put(60,830){\makebox(0,0){{\footnotesize\bf \% Exact}}}
\put(65,795){\makebox(0,0){{\footnotesize\bf ~~~~Match}}}
\put(938,-22){\makebox(0,0){{\bf Training-tree-bank size in $10^3$ units}}}
\put(220.0,113.0){\rule[-0.200pt]{0.400pt}{184.048pt}}
\put(1306,812){\makebox(0,0)[r]{DOP$^{4}$}}
\put(1328.0,812.0){\rule[-0.200pt]{15.899pt}{0.400pt}}
\put(463,377){\usebox{\plotpoint}}
\multiput(463.00,377.58)(1.077,0.499){223}{\rule{0.960pt}{0.120pt}}
\multiput(463.00,376.17)(241.007,113.000){2}{\rule{0.480pt}{0.400pt}}
\multiput(706.00,490.58)(2.856,0.498){83}{\rule{2.370pt}{0.120pt}}
\multiput(706.00,489.17)(239.081,43.000){2}{\rule{1.185pt}{0.400pt}}
\multiput(950.00,533.58)(8.031,0.496){43}{\rule{6.430pt}{0.120pt}}
\multiput(950.00,532.17)(350.653,23.000){2}{\rule{3.215pt}{0.400pt}}
\put(463,377){\raisebox{-.8pt}{\makebox(0,0){$\Diamond$}}}
\put(706,490){\raisebox{-.8pt}{\makebox(0,0){$\Diamond$}}}
\put(950,533){\raisebox{-.8pt}{\makebox(0,0){$\Diamond$}}}
\put(1314,556){\raisebox{-.8pt}{\makebox(0,0){$\Diamond$}}}
\sbox{\plotpoint}{\rule[-0.400pt]{0.800pt}{0.800pt}}%
\put(1306,767){\makebox(0,0)[r]{ISDOP$^{2}$}}
\put(1328.0,767.0){\rule[-0.400pt]{15.899pt}{0.800pt}}
\put(463,356){\usebox{\plotpoint}}
\multiput(463.00,357.41)(1.023,0.501){231}{\rule{1.834pt}{0.121pt}}
\multiput(463.00,354.34)(239.194,119.000){2}{\rule{0.917pt}{0.800pt}}
\multiput(706.00,476.41)(2.685,0.502){85}{\rule{4.443pt}{0.121pt}}
\multiput(706.00,473.34)(234.777,46.000){2}{\rule{2.222pt}{0.800pt}}
\multiput(950.00,522.40)(22.886,0.516){11}{\rule{32.556pt}{0.124pt}}
\multiput(950.00,519.34)(296.429,9.000){2}{\rule{16.278pt}{0.800pt}}
\sbox{\plotpoint}{\rule[-0.500pt]{1.000pt}{1.000pt}}%
\put(463,356){\makebox(0,0){$+$}}
\put(706,475){\makebox(0,0){$+$}}
\put(950,521){\makebox(0,0){$+$}}
\put(1314,530){\makebox(0,0){$+$}}
\put(1306,722){\makebox(0,0)[r]{SDOP$_{2}^{4}$}}
\multiput(1328,722)(41.511,0.000){2}{\usebox{\plotpoint}}
\put(1394,722){\usebox{\plotpoint}}
\put(463,451){\usebox{\plotpoint}}
\multiput(463,451)(37.581,17.631){7}{\usebox{\plotpoint}}
\multiput(706,565)(41.343,3.728){6}{\usebox{\plotpoint}}
\multiput(950,587)(41.471,-1.823){9}{\usebox{\plotpoint}}
\put(1314,571){\usebox{\plotpoint}}
\put(463,451){\raisebox{-.8pt}{\makebox(0,0){$\Box$}}}
\put(706,565){\raisebox{-.8pt}{\makebox(0,0){$\Box$}}}
\put(950,587){\raisebox{-.8pt}{\makebox(0,0){$\Box$}}}
\put(1314,571){\raisebox{-.8pt}{\makebox(0,0){$\Box$}}}
\end{picture}
}
\caption{Exact match as a function of tree-bank size}
\label{FigTBsizeEM}
\end{figure}
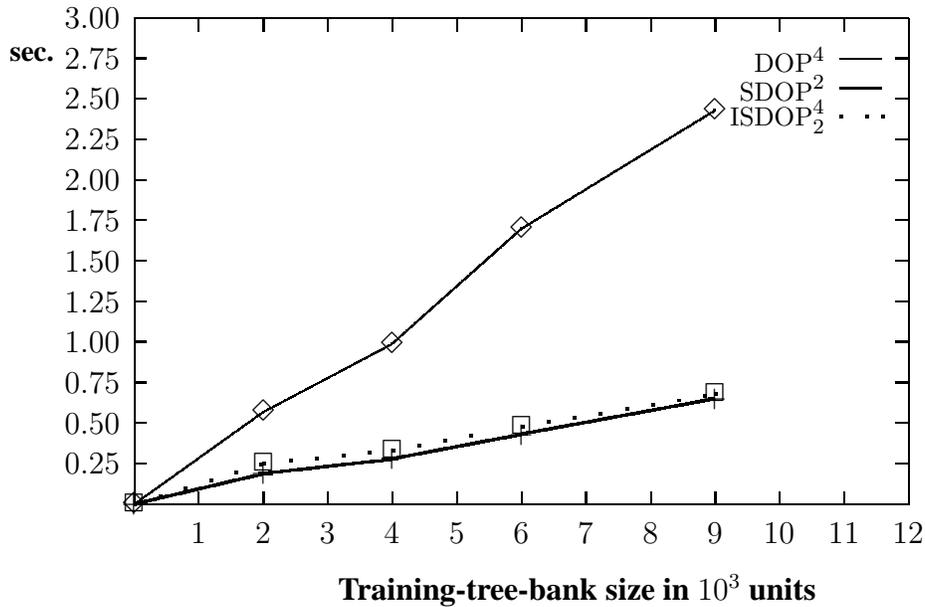
\begin{figure}[htb]
\center{
\setlength{\unitlength}{0.240900pt}
\ifx\plotpoint\undefined\newsavebox{\plotpoint}\fi
\sbox{\plotpoint}{\rule[-0.200pt]{0.400pt}{0.400pt}}%
\begin{picture}(1500,900)(0,0)
\font\gnuplot=cmr10 at 10pt
\gnuplot
\sbox{\plotpoint}{\rule[-0.200pt]{0.400pt}{0.400pt}}%
\put(220.0,113.0){\rule[-0.200pt]{292.934pt}{0.400pt}}
\put(220.0,113.0){\rule[-0.200pt]{0.400pt}{184.048pt}}
\put(220.0,177.0){\rule[-0.200pt]{4.818pt}{0.400pt}}
\put(198,177){\makebox(0,0)[r]{$0.25$}}
\put(1416.0,177.0){\rule[-0.200pt]{4.818pt}{0.400pt}}
\put(220.0,240.0){\rule[-0.200pt]{4.818pt}{0.400pt}}
\put(198,240){\makebox(0,0)[r]{$0.50$}}
\put(1416.0,240.0){\rule[-0.200pt]{4.818pt}{0.400pt}}
\put(220.0,304.0){\rule[-0.200pt]{4.818pt}{0.400pt}}
\put(198,304){\makebox(0,0)[r]{$0.75$}}
\put(1416.0,304.0){\rule[-0.200pt]{4.818pt}{0.400pt}}
\put(220.0,368.0){\rule[-0.200pt]{4.818pt}{0.400pt}}
\put(198,368){\makebox(0,0)[r]{$1.00$}}
\put(1416.0,368.0){\rule[-0.200pt]{4.818pt}{0.400pt}}
\put(220.0,431.0){\rule[-0.200pt]{4.818pt}{0.400pt}}
\put(198,431){\makebox(0,0)[r]{$1.25$}}
\put(1416.0,431.0){\rule[-0.200pt]{4.818pt}{0.400pt}}
\put(220.0,495.0){\rule[-0.200pt]{4.818pt}{0.400pt}}
\put(198,495){\makebox(0,0)[r]{$1.50$}}
\put(1416.0,495.0){\rule[-0.200pt]{4.818pt}{0.400pt}}
\put(220.0,559.0){\rule[-0.200pt]{4.818pt}{0.400pt}}
\put(198,559){\makebox(0,0)[r]{$1.75$}}
\put(1416.0,559.0){\rule[-0.200pt]{4.818pt}{0.400pt}}
\put(220.0,622.0){\rule[-0.200pt]{4.818pt}{0.400pt}}
\put(198,622){\makebox(0,0)[r]{$2.00$}}
\put(1416.0,622.0){\rule[-0.200pt]{4.818pt}{0.400pt}}
\put(220.0,686.0){\rule[-0.200pt]{4.818pt}{0.400pt}}
\put(198,686){\makebox(0,0)[r]{$2.25$}}
\put(1416.0,686.0){\rule[-0.200pt]{4.818pt}{0.400pt}}
\put(220.0,750.0){\rule[-0.200pt]{4.818pt}{0.400pt}}
\put(198,750){\makebox(0,0)[r]{$2.50$}}
\put(1416.0,750.0){\rule[-0.200pt]{4.818pt}{0.400pt}}
\put(220.0,813.0){\rule[-0.200pt]{4.818pt}{0.400pt}}
\put(198,813){\makebox(0,0)[r]{$2.75$}}
\put(1416.0,813.0){\rule[-0.200pt]{4.818pt}{0.400pt}}
\put(220.0,877.0){\rule[-0.200pt]{4.818pt}{0.400pt}}
\put(198,877){\makebox(0,0)[r]{$3.00$}}
\put(1416.0,877.0){\rule[-0.200pt]{4.818pt}{0.400pt}}
\put(321.0,113.0){\rule[-0.200pt]{0.400pt}{4.818pt}}
\put(321,68){\makebox(0,0){$1$}}
\put(321.0,857.0){\rule[-0.200pt]{0.400pt}{4.818pt}}
\put(423.0,113.0){\rule[-0.200pt]{0.400pt}{4.818pt}}
\put(423,68){\makebox(0,0){$2$}}
\put(423.0,857.0){\rule[-0.200pt]{0.400pt}{4.818pt}}
\put(524.0,113.0){\rule[-0.200pt]{0.400pt}{4.818pt}}
\put(524,68){\makebox(0,0){$3$}}
\put(524.0,857.0){\rule[-0.200pt]{0.400pt}{4.818pt}}
\put(625.0,113.0){\rule[-0.200pt]{0.400pt}{4.818pt}}
\put(625,68){\makebox(0,0){$4$}}
\put(625.0,857.0){\rule[-0.200pt]{0.400pt}{4.818pt}}
\put(727.0,113.0){\rule[-0.200pt]{0.400pt}{4.818pt}}
\put(727,68){\makebox(0,0){$5$}}
\put(727.0,857.0){\rule[-0.200pt]{0.400pt}{4.818pt}}
\put(828.0,113.0){\rule[-0.200pt]{0.400pt}{4.818pt}}
\put(828,68){\makebox(0,0){$6$}}
\put(828.0,857.0){\rule[-0.200pt]{0.400pt}{4.818pt}}
\put(929.0,113.0){\rule[-0.200pt]{0.400pt}{4.818pt}}
\put(929,68){\makebox(0,0){$7$}}
\put(929.0,857.0){\rule[-0.200pt]{0.400pt}{4.818pt}}
\put(1031.0,113.0){\rule[-0.200pt]{0.400pt}{4.818pt}}
\put(1031,68){\makebox(0,0){$8$}}
\put(1031.0,857.0){\rule[-0.200pt]{0.400pt}{4.818pt}}
\put(1132.0,113.0){\rule[-0.200pt]{0.400pt}{4.818pt}}
\put(1132,68){\makebox(0,0){$9$}}
\put(1132.0,857.0){\rule[-0.200pt]{0.400pt}{4.818pt}}
\put(1233.0,113.0){\rule[-0.200pt]{0.400pt}{4.818pt}}
\put(1233,68){\makebox(0,0){$10$}}
\put(1233.0,857.0){\rule[-0.200pt]{0.400pt}{4.818pt}}
\put(1335.0,113.0){\rule[-0.200pt]{0.400pt}{4.818pt}}
\put(1335,68){\makebox(0,0){$11$}}
\put(1335.0,857.0){\rule[-0.200pt]{0.400pt}{4.818pt}}
\put(1436.0,113.0){\rule[-0.200pt]{0.400pt}{4.818pt}}
\put(1436,68){\makebox(0,0){$12$}}
\put(1436.0,857.0){\rule[-0.200pt]{0.400pt}{4.818pt}}
\put(220.0,113.0){\rule[-0.200pt]{292.934pt}{0.400pt}}
\put(1436.0,113.0){\rule[-0.200pt]{0.400pt}{184.048pt}}
\put(220.0,877.0){\rule[-0.200pt]{292.934pt}{0.400pt}}
\put(60,820){\makebox(0,0){{\small\bf sec.}}}
\put(916,-22){\makebox(0,0){{\bf Training-tree-bank size in $10^3$ units}}}
\put(220.0,113.0){\rule[-0.200pt]{0.400pt}{184.048pt}}
\put(1306,812){\makebox(0,0)[r]{DOP$^{4}$}}
\put(1328.0,812.0){\rule[-0.200pt]{15.899pt}{0.400pt}}
\put(220,113){\usebox{\plotpoint}}
\multiput(220.00,113.58)(0.700,0.499){287}{\rule{0.660pt}{0.120pt}}
\multiput(220.00,112.17)(201.630,145.000){2}{\rule{0.330pt}{0.400pt}}
\multiput(423.00,258.58)(0.945,0.499){211}{\rule{0.855pt}{0.120pt}}
\multiput(423.00,257.17)(200.225,107.000){2}{\rule{0.428pt}{0.400pt}}
\multiput(625.00,365.58)(0.561,0.500){359}{\rule{0.549pt}{0.120pt}}
\multiput(625.00,364.17)(201.861,181.000){2}{\rule{0.274pt}{0.400pt}}
\multiput(828.00,546.58)(0.818,0.500){369}{\rule{0.754pt}{0.120pt}}
\multiput(828.00,545.17)(302.436,186.000){2}{\rule{0.377pt}{0.400pt}}
\put(220,113){\raisebox{-.8pt}{\makebox(0,0){$\Diamond$}}}
\put(423,258){\raisebox{-.8pt}{\makebox(0,0){$\Diamond$}}}
\put(625,365){\raisebox{-.8pt}{\makebox(0,0){$\Diamond$}}}
\put(828,546){\raisebox{-.8pt}{\makebox(0,0){$\Diamond$}}}
\put(1132,732){\raisebox{-.8pt}{\makebox(0,0){$\Diamond$}}}
\sbox{\plotpoint}{\rule[-0.400pt]{0.800pt}{0.800pt}}%
\put(1306,767){\makebox(0,0)[r]{SDOP$^{2}$}}
\put(1328.0,767.0){\rule[-0.400pt]{15.899pt}{0.800pt}}
\put(220,113){\usebox{\plotpoint}}
\multiput(220.00,114.41)(2.138,0.502){89}{\rule{3.583pt}{0.121pt}}
\multiput(220.00,111.34)(195.563,48.000){2}{\rule{1.792pt}{0.800pt}}
\multiput(423.00,162.41)(4.526,0.505){39}{\rule{7.226pt}{0.122pt}}
\multiput(423.00,159.34)(187.002,23.000){2}{\rule{3.613pt}{0.800pt}}
\multiput(625.00,185.41)(2.641,0.503){71}{\rule{4.364pt}{0.121pt}}
\multiput(625.00,182.34)(193.942,39.000){2}{\rule{2.182pt}{0.800pt}}
\multiput(828.00,224.41)(2.742,0.502){105}{\rule{4.543pt}{0.121pt}}
\multiput(828.00,221.34)(294.571,56.000){2}{\rule{2.271pt}{0.800pt}}
\sbox{\plotpoint}{\rule[-0.500pt]{1.000pt}{1.000pt}}%
\put(220,113){\makebox(0,0){$+$}}
\put(423,161){\makebox(0,0){$+$}}
\put(625,184){\makebox(0,0){$+$}}
\put(828,223){\makebox(0,0){$+$}}
\put(1132,279){\makebox(0,0){$+$}}
\put(1306,722){\makebox(0,0)[r]{ISDOP$_{2}^{4}$}}
\multiput(1328,722)(41.511,0.000){2}{\usebox{\plotpoint}}
\put(1394,722){\usebox{\plotpoint}}
\put(220,113){\usebox{\plotpoint}}
\multiput(220,113)(39.590,12.482){6}{\usebox{\plotpoint}}
\multiput(423,177)(41.309,4.090){5}{\usebox{\plotpoint}}
\multiput(625,197)(40.802,7.638){4}{\usebox{\plotpoint}}
\multiput(828,235)(40.939,6.868){8}{\usebox{\plotpoint}}
\put(1132,286){\usebox{\plotpoint}}
\put(220,113){\raisebox{-.8pt}{\makebox(0,0){$\Box$}}}
\put(423,177){\raisebox{-.8pt}{\makebox(0,0){$\Box$}}}
\put(625,197){\raisebox{-.8pt}{\makebox(0,0){$\Box$}}}
\put(828,235){\raisebox{-.8pt}{\makebox(0,0){$\Box$}}}
\put(1132,286){\raisebox{-.8pt}{\makebox(0,0){$\Box$}}}
\end{picture}
}
\caption{CPU-time as a function of tree-bank size}
\label{FigTBsizeCPU}
\end{figure}
\begin{table}
\center{
\begin{tabular}{|c||c|c|c|c|}
\hline
Annotation  &\multicolumn{2}{|c|}{\#grammar symbols} & \multicolumn{2}{|c|}{\#grammar rules}  \\
Annotation  & non-terminals & terminals & lexical & non-lexical \\
\hline
Syntax           & 32               &  911         &  1005        &  278 \\
\hline
\end{tabular}
}
\caption{The T-SRI-ATIS tree-bank in numbers}
\label{FigATISinNums}
\end{table}

\begin{table}
\center{
\begin{tabular}{|c|c|c|c||c|c|c|c|}
\hline
\multicolumn{8}{|c|}{Spec. TSG: 1523 elementary-trees, 2640 internal-nodes}\\
\hline
\hline
\multicolumn{4}{|c||}{Number of elementary-trees} & \multicolumn{4}{|c|}{Number of internal-nodes} \\
{\bf DOP$^{1}$} & {\bf DOP$^{2}$} & {\bf DOP$^{3}$} & {\bf DOP$^{4}$} & {\bf DOP$^{1}$} & {\bf DOP$^{2}$} & {\bf DOP$^{3}$} & {\bf DOP$^{4}$} \\
\hline
397  & 5920 & 24364 & 55178 & 1507 & 17795 & 162321 & 325431 \\
\hline
{\bf SDOP$^{1}$} & {\bf SDOP$^{2}$} & {\bf SDOP$^{3}$} & {\bf SDOP$^{4}$} & {\bf SDOP$^{1}$} & {\bf SDOP$^{2}$} & {\bf SDOP$^{3}$} & {\bf SDOP$^{4}$} \\
\hline
530 & 10464 & 33301 & 51817 & 2421 & 47545 & 210533 & 374373 \\
\hline
\end{tabular}
} 
\caption{Sizes of models to subtree-depth upper-bound (UB) trained on OVIS-syntax}
\label{FigSynDepthSize}
\end{table}
%

\subsection{Experiments using full annotation on utterances}
\label{ExpOVISFULL}
In this section the experiments were conducted on the OVIS tree-bank with its full
annotation (syntactic and semantic). Four sets of experiments were conducted. In each
set of experiments the value of one training parameter is varied and the rest of the 
parameters are fixed.
%
\subsubsection{A. Varying training-set size}
In this set of experiments the~10000 trees of the OVIS tree-bank were initially split into 
a tree-bank of~9000 trees (tree-bank B9000) and another tree-bank of~1000 trees (tree-bank A)
using a random generator.
The~1000 trees of tree-bank A were set aside to be used as the test-set in the experiments.
From tree-bank~B9000 (9000 trees) another four smaller tree-banks of sizes~6000, 4000 
and 2000~trees (denoted respectively tree-banks B6000, B4000 and B2000) were obtained by a 
random generator taking care that B2000$\subset$B4000$\subset$B6000$\subset$B9000.

In four independent experiments, each of tree-banks B2000 through B9000 was used as training 
material for projecting a DOP model and for ARS training and then projecting the SDOP and ISDOP 
models. Then the resulting DOP, SDOP and ISDOP models (twelve in total) were run independently
on the~1000 sentences of tree-bank~A. The results of parsing and disambiguation for each parser
were matched against the trees in tree-bank~A. 

The upper-bound on subtree-depth of the models was set on: DOP$^{4}$ and SDOP$^{2}$ (i.e. for
DOP models the upper-bound was set on~4 and for SDOP models on~2). These upper-bounds were chosen
for the following reasons: for DOP models depth~4 exhibited the best accuracy results for DOP
models and for SDOP models depth~2 gave the smallest models that had comparable accuracy results
(although not the best results the SDOP model can achieve as we will see in subsequent sections).

Table~\REFTAB{FigTBsizeTable} shows the sizes of the specialized TSG (i.e. partial-parser) and 
the DOP STSG and SDOP STSG models; 
the table shows the number of learned elementary-trees and also the size of the
TSG (measured as the total number of internal nodes of the elementary-trees of the TSG). 
The number of
elementary-trees of the specialized TSG is~1355 at B9000 but the rate of growth of this
TSG is decreasing as the training tree-bank sizes increases (adding~3000 trees to B6000 results 
in an increase of~1.24 times, while adding 2000~trees to B4000 and B2000 results in an increase
of~1.30 and~1.54 respectively). A similar growth-rate can be seen for the number of internal 
nodes figures. For the CFG underlying the training tree-bank the number of rules was:
788 (B2000), 1093 (B4000), 1368 (B6000) and 1687 (B9000).

Crucial here is to note the sizes of the SDOP STSGs against the DOP STSGs: 
at B9000 the size of the SDOP STSG is only~17\% of that of the corresponding DOP STSG. 
This is more than~5.7 times reduction in memory consumption.

The Tree-Language Coverage (TLC) of the specialized TSG is exactly equal to that of the CFG
underlying the training tree-bank at all tree-bank sizes. The ambiguity 
(measured as the number of active nodes in the parse-space) of the specialized TSG
is smaller than that of the CFG underlying the tree-bank: approx.~1.2 times at B2000
and~1.6 times at B9000. The modesty of these ambiguity reduction figures is the result of
the suboptimality of the non-lexicalized GRF algorithm when equipped with a 
sequential-covering scheme. 
%

%
Table~\REFTAB{FigTBsizeTable} exhibits and compares the results of the twelve
parsers on various measures. A general observation can be made here: while the ISDOP models
recognize (almost) as many sentences as the DOP models do, SDOP models recognize less.
However, the gap between the models narrows down from 9\% to 3\% as the tree-bank size increases.
In contrast to recognition power results, exact-match results show that the SDOP models score 
better results all the way. The ISDOP models score exact-match results that increasingly improve 
but remain slightly behind the DOP results (approx.~0.5-0.8\%). A similar observation 
applies to the (labeled) bracketing recall and precision results. This (slight) loss of 
accuracy in ISDOP models is not due to loss of tree-language coverage as the table shows.
This implies that it is due to a slight degradation in the quality of the probability 
distributions of the SDOP STSGs as the training tree-bank size increases. 
Thus, the SDOP models' accuracy improvement is mainly due to their limited recognition 
power rather than an improved disambiguation power. This observation is particularly
subject to the reservation that the  present experiments are conducted on one particular
partitioning to training/test material. Mean and standard deviation results on five different 
and independent partitions are discussed below and can be considered more stable. 

Concerning the speed of processing, the SDOP and ISDOP models are faster than the DOP 
models at all times: the speed-up increases from approx.~2-2.5 times at B2000 to 
approx. 4~times at B9000. 

Some of the results are exposed in a graphical manner in 
figures~\ref{FigTBsizeEM},~\ref{FigTBsizeCOV} and~\ref{FigTBsizeCPU} that show 
respectively the change in exact-match, Recognized and CPU-time as a function of tree-bank size.
%
%
\begin{figure}[hbt]
\epsfxsize=13.5cm
\center{
\epsfbox{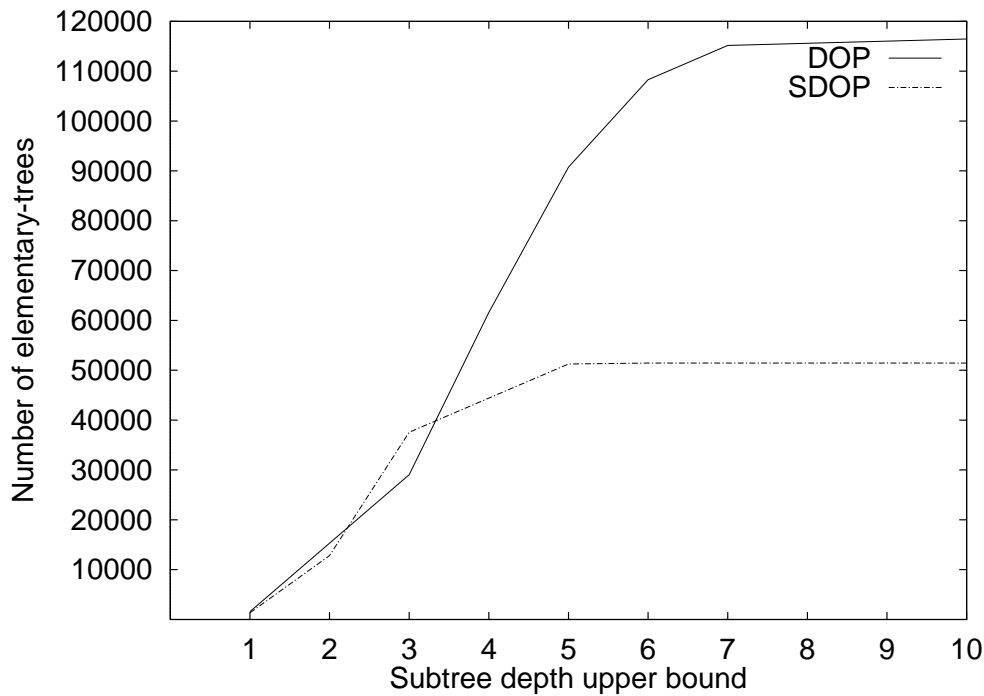}
}
\caption{Number of elementary-trees as a function of subtree depth}
\label{FigDepthSize}
\end{figure}

\begin{figure}[hbt]
\epsfxsize=13.5cm
\center{
\epsfbox{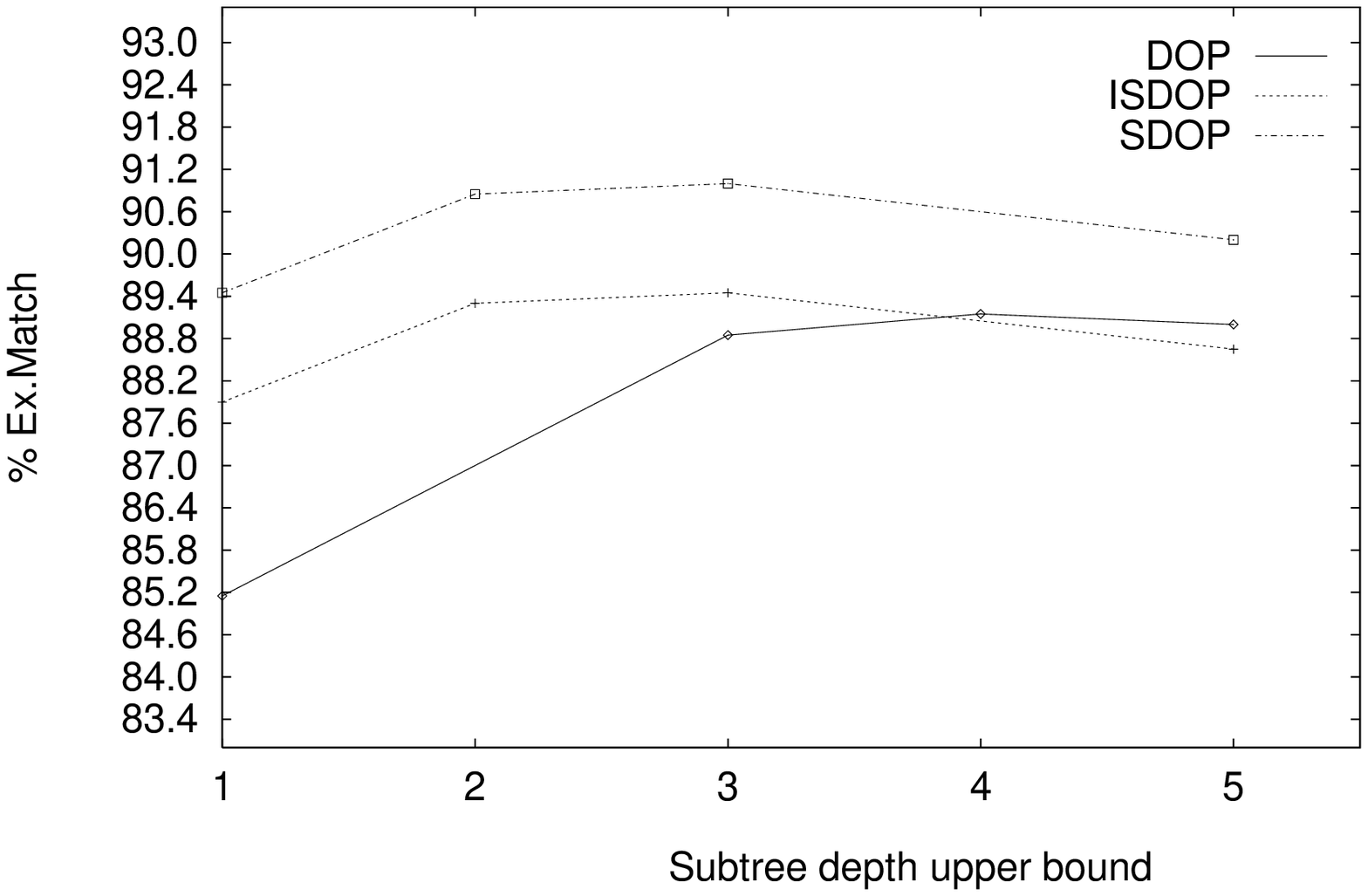}
}
\caption{Exact match as a function of subtree depth}
\label{FigDepthEM}
\end{figure}

\begin{figure}[htb]
\epsfxsize=13.5cm
\center{
\epsfbox{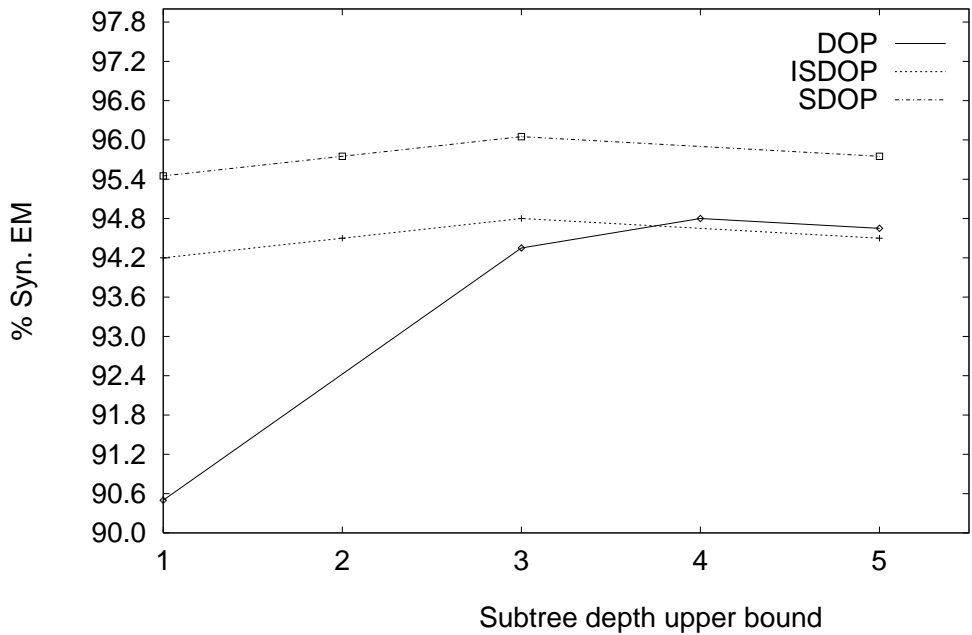}
}
\caption{Syntactic exact match as a function of subtree depth}
\label{FigDepthSynEM}
\end{figure}

\begin{figure}[hbt]
\epsfxsize=13.5cm
\center{
\epsfbox{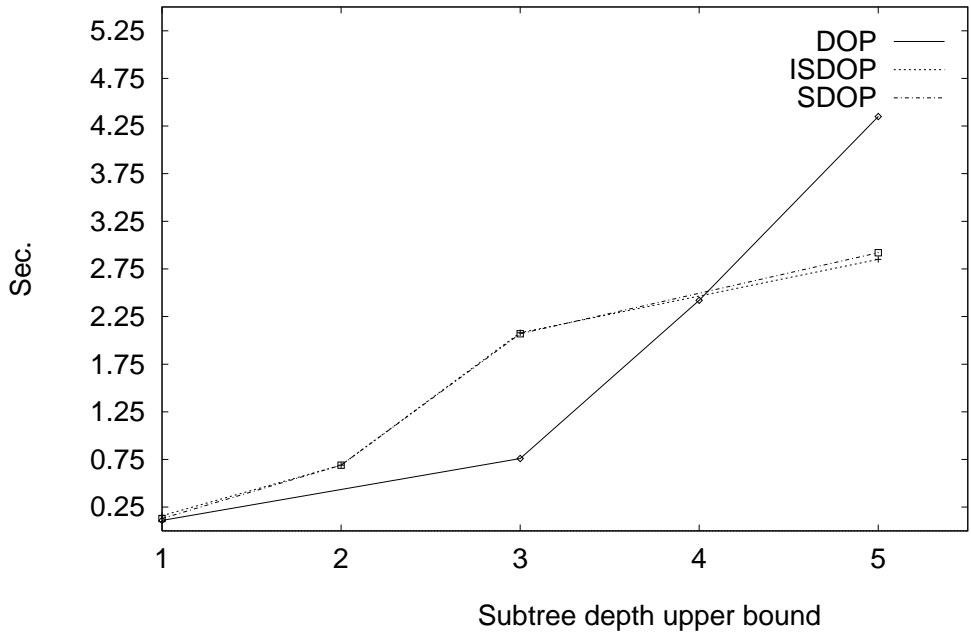}
}
\caption{CPU-time as a function of subtree depth}
\label{FigDepthCPU}
\end{figure}

\input{./D-CH_GramSpec/D-ExpResults/Figd4D2meansCPU}
{
\begin{table}[htb]
\begin{center}
\begin{tabular}{|c|c|c|c|c|}
\hline
Measures & DOP$^{1}$  &  DOP$^{3}$ &  DOP$^{4}$  &  DOP$^{5}$ \\
\hline
{\sl Recognized} &  95.20 & 95.20 & 95.20 & 95.20 \\
Exact match &  85.15 & 88.85 & 89.15 & 89.00 \\
Syn. ex. match &  90.50 & 94.35 & 94.80 & 94.65 \\
Sem. ex. match &  85.95 & 89.30 & 89.60 & 89.45 \\
NCB recall &  91.45 & 92.15 & 92.20 & 92.20 \\
NCB prec. &  98.90 & 99.55 & 99.60 & 99.60 \\
0-Crossing &  94.20 & 97.55 & 97.85 & 97.85 \\
Syn. LBR &  90.65 & 91.50 & 91.50 & 91.50 \\
Syn. LBP &  98.05 & 98.80 & 98.85 & 98.85 \\
Sem. LBR &  88.90 & 90.00 & 90.00 & 90.00 \\
Sem. LBP &  96.15 & 97.15 & 97.25 & 97.20 \\
CPU (sec.) &  0.11 & 0.76 & 2.42 & 4.35 \\
\hline
%
\hline
\hline
Measures & ISDOP$_{1}^{1}$  &  ISDOP$_{2}^{4}$  &  ISDOP$_{3}^{5}$  &  ISDOP$_{5}^{5}$ \\
\hline
{\sl Recognized} &  95.05 & 95.05 & 95.05 & 95.05 \\
Exact match &  87.90 & 89.30 & 89.45 & 88.65 \\
Syn. ex. match &  94.20 & 94.50 & 94.80 & 94.50 \\
Sem. ex. match &  88.65 & 89.90 & 90.05 & 89.30 \\
NCB recall &  91.70 & 91.80 & 91.80 & 91.80 \\
NCB prec. &  99.40 & 99.45 & 99.45 & 99.50 \\
0-Crossing &  96.65 & 97.10 & 97.40 & 97.40 \\
Syn. LBR &  91.05 & 91.05 & 91.05 & 91.05 \\
Syn. LBP &  98.65 & 98.65 & 98.65 & 98.65 \\
Sem. LBR &  89.30 & 89.45 & 89.50 & 89.40 \\
Sem. LBP &  96.80 & 96.90 & 97.00 & 96.90 \\
CPU (sec.) &  0.16 & 0.69 & 2.08 & 2.85 \\
\hline
\hline
%
\hline
Measures & SDOP$^{1}$  &  SDOP$^{2}$  &  SDOP$^{3}$  &  SDOP$^{5}$ \\
\hline
{\sl Recognized} &  92.30 & 92.30 & 92.30 & 92.30 \\
Exact match &  89.45 & 90.85 & 91.00 & 90.20 \\
Syn. ex. match &  95.45 & 95.75 & 96.05 & 95.75 \\
Sem. ex. match &  90.05 & 91.00 & 91.50 & 90.70 \\
NCB recall &  88.10 & 88.60 & 88.10 & 88.15 \\
NCB prec. &  99.60 & 99.60 & 99.60 & 99.65 \\
0-Crossing &  97.65 & 97.80 & 98.10 & 98.10 \\
Syn. LBR &  87.60 & 88.10 & 87.55 & 87.55 \\
Syn. LBP &  99.05 & 99.05 & 99.00 & 98.95 \\
Sem. LBR &  86.20 & 86.80 & 86.35 & 86.25 \\
Sem. LBP &  97.45 & 97.60 & 97.60 & 97.50 \\
CPU (sec.) &  0.13 & 0.69 & 2.07 & 2.92 \\
\hline
\hline
\end{tabular}
\end{center}
\caption{Varying subtree-depth upperbound: sentence length $> 1$}
\label{TabDepths}
\end{table}
}

{
\begin{table}[hbt]
\begin{center}
\begin{tabular}{|c|c|c|c|}
\hline
Measures & DOP$^{4}$  &  ISDOP$_{2}^{4}$  &  SDOP$^{2}$ \\
\hline
{\sl Recognized} &  95.25 (0.99) & 94.85 (1.42) & 91.45 (0.99) \\
{\sl TLC} & 89.60 (1.56) & 89.60 (1.56) & 89.60 (1.56) \\
Exact match &  89.35 (0.56) & 89.45 (1.55) & 91.30 (1.39) \\
Syn. ex. match &  94.90 (0.79) & 94.80 (0.79) & 96.30 (0.61) \\
Sem. ex. match &  89.90 (0.30) & 90.00 (1.30) & 91.50 (1.39) \\
NCB recall &  92.15 (1.51) & 91.30 (2.25) & 86.75 (1.55) \\
NCB prec. &  99.40 (0.21) & 99.35 (0.10) & 99.70 (0.05) \\
0-Crossing &  97.10 (0.72) & 96.90 (0.41) & 98.05 (0.34) \\
Syn. LBR &  91.45 (1.51) & 90.60 (2.06) & 86.35 (1.43) \\
Syn. LBP &  98.65 (0.29) & 98.55 (0.22) & 99.25 (0.15) \\
Sem. LBR &  90.05 (1.57) & 89.10 (2.03) & 85.20 (1.31) \\
Sem. LBP &  97.15 (0.36) & 96.90 (0.26) & 97.95 (0.26) \\
CPU (sec.) &  2.55 (0.23)  & 0.74 (0.09)  & 0.70 (0.03)  \\
\# of sens &  \multicolumn{3}{|c|}{685.75 (8.42) } \\
Sen. length & \multicolumn{3}{|c|}{ 4.50 (0.03) }\\
\hline
\end{tabular}
\end{center}
\begin{center}
\begin{tabular}{|c|c|c|c|}
\hline
Measures & DOP$^{1}$  &  ISDOP$_{1}^{1}$  &  SDOP$^{1}$ \\
\hline
{\sl Recognized} &  95.25 (0.99) & 94.75 (1.24) & 91.40 (0.86) \\
{\sl TLC} & 89.60 (1.56) & 89.60 (1.56) & 89.60 (1.56) \\
Exact match &  86.55 (0.97) & 88.05 (1.18) & 90.00 (1.17) \\
Syn. ex. match &  91.65 (0.72) & 93.75 (0.65) & 95.45 (0.50) \\
Sem. ex. match &  87.25 (0.83) & 88.70 (0.96) & 90.50 (1.16) \\
NCB recall &  91.65 (1.56) & 91.10 (1.90) & 86.05 (1.41) \\
NCB prec. &  99.00 (0.13) & 99.25 (0.17) & 99.60 (0.03) \\
0-Crossing &  94.85 (0.65) & 96.40 (0.39) & 97.75 (0.21) \\
Syn. LBR &  90.85 (1.45) & 90.30 (1.75) & 85.55 (1.36) \\
Syn. LBP &  98.10 (0.24) & 98.40 (0.27) & 99.05 (0.12) \\
Sem. LBR &  89.15 (1.51) & 88.70 (1.68) & 84.30 (1.24) \\
Sem. LBP &  96.30 (0.42) & 96.65 (0.29) & 97.60 (0.22) \\
CPU (sec.) &  0.12 (0.00)  & 0.17 (0.01)  & 0.14 (0.01)  \\
\# of sens & \multicolumn{3}{|c|}{ 684.60 (7.73) } \\
Sen. length & \multicolumn{3}{|c|}{ 4.50 (0.03)  } \\
\hline
\end{tabular}
\end{center}
\caption{Means and stds of 5 partitions: sentence length $> 1$}
\label{FigMeansTable}
\end{table}
}

{
\begin{table}[htb]
\begin{center}
\begin{tabular}{|c|c|c|c|c|}
\hline
Measures & DOP$^{1}$  &  DOP$^{2}$  &  DOP$^{3}$  &  DOP$^{4}$ \\
\hline
{\sl Recognized} &  99.70 & 99.70 & 99.70 & 99.70 \\
{\sl TLC} & 97.40 & 97.40 & 97.40 & 97.40 \\
Syn. ex. match &  84.20 & 89.30 & 92.30 & 92.40 \\
NCB recall &  97.70 & 98.20 & 98.80 & 98.80 \\
NCB prec. &  98.10 & 98.80 & 99.30 & 99.30 \\
0-Crossing &  89.60 & 93.60 & 95.90 & 96.30 \\
Syn. LBR &  96.00 & 97.10 & 98.00 & 97.90 \\
Syn. LBP &  96.50 & 97.60 & 98.50 & 98.30 \\
CPU (sec.) &  0.11 & 0.31 & 2.57 & 9.27 \\
\# of sens & \multicolumn{4}{|c|}{ 687.00 } \\
Sen. length & \multicolumn{4}{|c|}{ 4.60  } \\
\hline
%
Measures & ISDOP$_{1}^{1}$  &  ISDOP$_{2}^{4}$  &  ISDOP$_{3}^{3}$  &  ISDOP$_{4}^{4}$ \\
\hline
{\sl Recognized} &  99.60 & 99.60 & 99.60 & 99.60 \\
{\sl TLC} & 97.40 & 97.40 & 97.40 & 97.40 \\
Syn. ex. match &  87.40 & 90.50 & 92.00 & 91.70 \\
NCB recall &  98.10 & 98.30 & 98.40 & 98.50 \\
NCB prec. &  98.70 & 99.00 & 99.00 & 99.10 \\
0-Crossing &  93.00 & 94.90 & 95.60 & 95.80 \\
Syn. LBR &  96.50 & 97.10 & 97.50 & 97.40 \\
Syn. LBP &  97.10 & 97.70 & 98.10 & 98.00 \\
CPU (sec.) &  0.13 & 1.26 & 5.72 & 9.81 \\
\# of sens & \multicolumn{4}{|c|}{ 687.00 } \\
Sen. length & \multicolumn{4}{|c|}{ 4.60 } \\
\hline
%
\hline
Measures & SDOP$^{1}$  &  SDOP$^{2}$  &  SDOP$^{3}$  &  SDOP$^{4}$ \\
\hline
{\sl Recognized} &  99.10 & 99.10 & 99.10 & 99.10 \\
{\sl TLC} & 97.40 & 97.40 & 97.40 & 97.40 \\
Syn. ex. match &  87.50 & 90.60 & 92.10 & 91.80 \\
NCB recall &  97.60 & 97.90 & 98.00 & 98.00 \\
NCB prec. &  98.70 & 99.00 & 99.00 & 99.00 \\
0-Crossing &  92.90 & 94.90 & 95.60 & 95.70 \\
Syn. LBR &  96.10 & 96.70 & 97.10 & 97.00 \\
Syn. LBP &  97.10 & 97.80 & 98.10 & 98.00 \\
CPU (sec.) &  0.13 & 1.24 & 5.65 & 9.77 \\
\# of sens & \multicolumn{4}{|c|}{ 687.00 } \\
Sen. length & \multicolumn{4}{|c|}{ 4.60 } \\
\hline
\end{tabular}
\end{center}
\caption{Training on OVIS-syntax: results for sentence length $> 1$}
\label{FigSyntaxTable}
\end{table}
}

{
\begin{table}[htb]
\begin{center}
\begin{tabular}{|c|c|c||c|c|}
\hline
      & \multicolumn{4}{|c|}{target concept: equivalence classes SSF}\\
Measures & ISDOP$_{2}^{4}$  &  SDOP$^{2}$  &  ISDOP$_{3}^{3}$  &  SDOP$^{3}$ \\
\hline
{\sl Recognized} &  94.90 (1.17) & 91.30 (1.00) & 94.90 (1.17) & 91.20 (0.87) \\
{\sl TLC} & 89.60 (1.56) & 89.60 (1.56) & 89.60 (1.56) & 89.60 (1.56) \\
Exact match &  89.00 (1.35) & 91.00 (1.25) & 89.30 (1.37) & 91.30 (1.26) \\
Syn. ex. match &  94.50 (0.48) & 96.10 (0.44) & 94.80 (0.75) & 96.40 (0.53) \\
Sem. ex. match &  89.60 (1.16) & 91.40 (1.25) & 89.90 (1.11) & 91.80 (1.21) \\
NCB recall &  91.50 (1.84) & 85.90 (1.41) & 91.50 (1.79) & 85.70 (1.18) \\
NCB prec. &  99.40 (0.13) & 99.70 (0.05) & 99.30 (0.18) & 99.70 (0.05) \\
0-Crossing &  96.80 (0.40) & 98.00 (0.29) & 96.90 (0.37) & 98.20 (0.25) \\
Syn. LBR &  90.80 (1.72) & 85.40 (1.30) & 90.80 (1.67) & 85.30 (1.08) \\
Syn. LBP &  98.60 (0.17) & 99.10 (0.12) & 98.60 (0.27) & 99.20 (0.13) \\
Sem. LBR &  89.20 (1.70) & 84.30 (1.22) & 89.20 (1.66) & 84.20 (1.04) \\
Sem. LBP &  96.90 (0.20) & 97.80 (0.17) & 96.90 (0.27) & 97.80 (0.19) \\
CPU (sec.) &  0.63 (0.04)  & 0.58 (0.03)  & 1.93 (0.12)  & 2.02 (0.15)  \\
\# of sens & \multicolumn{4}{|c|}{ 684.60 (7.73) } \\
Sen. length &  \multicolumn{4}{|c|}{4.50 (0.03) } \\
\hline
\end{tabular}
\end{center}
\caption{Experiments with the definition of the target-concept 
         as equivalence classes of SSFs that are equivalent up to
         consecutive repetition of symbols. The results are for
         sentence \mbox{length~$>1$}}
\label{FigSIMSensd4D2}
\end{table}
}

{
\begin{table}[htb]
\begin{center}
\begin{tabular}{|c|c|c|c|}
\hline
Measures & DOP$^{4}$  &  ISDOP$_{2}^{4}$  &  SDOP$^{2}$ \\
\hline
Recognized &  99.60 (0.13) & 99.40 (0.15) & 98.30 (0.60) \\
{\sl TLC} & 97.40 (1.73) & 97.40 (1.73) & 97.40 (1.73) \\
Syn. ex. match &  92.25 (0.69) & 90.75 (1.04) & 91.20 (0.94) \\
NCB recall &  98.65 (0.33) & 98.25 (0.31) & 96.70 (0.82) \\
NCB prec. &  99.15 (0.17) & 99.00 (0.23) & 99.05 (0.24) \\
0-Crossing &  95.85 (0.33) & 95.15 (0.77) & 95.45 (0.76) \\
Syn. LBR &  97.55 (0.31) & 96.95 (0.21) & 95.50 (0.77) \\
Syn. LBP &  98.05 (0.25) & 97.70 (0.40) & 97.85 (0.42) \\
CPU (sec.) &  10.11 (0.87)  & 1.56 (0.17)  & 1.47 (0.15)  \\
\# of sens & \multicolumn{3}{|c|}{ 684.60 (7.73) } \\
Sen. length & \multicolumn{3}{|c|}{ 4.60 (0.05) } \\
\hline
\end{tabular}
\end{center}
\caption{ Syntactic OVIS: Results for sentence length $> 1$}
\label{FigSyntaxD2d4Table}
\end{table}
}

{
\begin{table}[htb]
\begin{center}
\begin{tabular}{|c|c|c|c|}
\hline
\multicolumn{4}{|c|}{Results on all word-graphs}\\
Measures & DOP$^{4}$  &  ISDOP$_{2}^{4}$  &  SDOP$^{2}$ \\
\hline
{\sl Recognized} &  99.30 (.30) & 99.10 (.35) & 98.10 (.60) \\
Exact match &  72.60 (.90) & 72.70 (1.40) & 73.20 (1.55) \\
Syn. ex. match &  80.70 (1.10) & 80.80 (1.70) & 81.40 (1.90) \\
Sem. ex. match &  75.10 (.50) & 75.20 (1.40) & 75.70 (1.55) \\
NCB recall &  90.70 (.70) & 90.00 (.55) & 87.90 (1.10) \\
NCB prec. &  93.30 (.80) & 93.10 (.60) & 93.60 (.70) \\
0-Crossing &  87.50 (1.00) & 87.30 (.90) & 87.80 (1.10) \\
Syn. LBR &  80.95 (1.50) & 80.20 (1.20) & 78.80 (1.10) \\
Syn. LBP &  83.30 (1.80) & 82.90 (1.60) & 83.90 (1.90) \\
Sem. LBR &  78.90 (1.60) & 78.20 (1.40) & 76.90 (1.30) \\
Sem. LBP &  81.20 (1.90) & 80.90 (1.80) & 81.80 (2.10) \\
CPU (sec.) &  33.10 (7.00) & 8.60 (2.00) & 8.20 (1.30) \\
\# of WGs &  \multicolumn{3}{|c|}{1000.00 (0) } \\
\#states in WGs &\multicolumn{3}{|c|}{  8.40 (.10) } \\
\#trans in WGs & \multicolumn{3}{|c|}{ 30.50 (2.00) } \\
Corr. sen. in WG & \multicolumn{3}{|c|}{ 89.80 (.55) } \\
\hline
%
\hline
\multicolumn{4}{|c|}{Results only on word-graphs containing the correct sentence}\\
Measures & DOP$^{4}$  &  ISDOP$_{2}^{4}$  &  SDOP$^{2}$ \\
\hline
{\sl Recognized} &  99.40 (.20) & 99.30 (.30) & 98.60 (.50) \\
Exact match &  80.80 (.70) & 80.90 (1.25) & 81.20 (1.30) \\
Syn. ex. match &  87.80 (.80) & 87.80 (1.40) & 88.20 (1.50) \\
Sem. ex. match &  83.10 (.70) & 83.00 (1.30) & 83.20 (1.40) \\
NCB recall &  94.50 (.50) & 93.90 (.40) & 92.50 (1.20) \\
NCB prec. &  96.40 (.70) & 96.20 (.75) & 96.40 (.70) \\
0-Crossing &  92.85 (.70) & 92.70 (.90) & 93.00 (1.05) \\
Syn. LBR &  88.70 (1.00) & 88.10 (1.00) & 86.90 (1.20) \\
Syn. LBP &  90.45 (1.50) & 90.30 (1.60) & 90.60 (1.60) \\
Sem. LBR &  87.00 (1.10) & 86.50 (1.20) & 85.30 (1.30) \\
Sem. LBP &  88.75 (1.60) & 88.60 (1.80) & 88.90 (1.90) \\
CPU (sec.) &  17.40 (3.70) & 4.10 (.80) & 4.05 (.85) \\
\# of WGs & \multicolumn{3}{|c|}{ 897.80 (5.50) } \\
\#states in WGs &\multicolumn{3}{|c|}{  7.30 (.10)  } \\
\#trans in WGs & \multicolumn{3}{|c|}{ 21.50 (1.10) } \\
Corr. sen. in WG & \multicolumn{3}{|c|}{ 100.00 (0) } \\
\hline
\end{tabular}
\end{center}
\caption{Results on OVIS wordgraphs: DOP$^{4}$ vs. ISDOP$_{2}^{4}$ and SDOP$^{2}$}
\label{WGsd4D2NoSim}
\end{table}
}
{
\begin{table}[htb]
\begin{center}
\begin{tabular}{|c|c|c|c|}
\hline
\multicolumn{4}{|c|}{Results on all word-graphs}\\
Measures & DOP$^{1}$  &  ISDOP$_{1}^{1}$  &  SDOP$^{1}$ \\
\hline
{\sl Recognized} &  99.40 (0.28) & 99.10 (0.38) & 98.20 (0.58) \\
Exact match &  69.50 (0.95) & 70.60 (1.36) & 71.10 (1.49) \\
Syn. ex. match &  77.90 (0.64) & 79.20 (1.43) & 79.70 (1.59) \\
Sem. ex. match &  71.80 (0.71) & 73.10 (1.29) & 73.60 (1.45) \\
NCB recall &  89.20 (0.64) & 89.10 (0.91) & 87.00 (1.13) \\
NCB prec. &  92.40 (0.92) & 92.60 (0.82) & 93.00 (0.78) \\
0-Crossing &  85.10 (0.96) & 86.10 (1.12) & 86.60 (1.15) \\
Syn. LBR &  78.90 (1.25) & 78.90 (1.61) & 77.50 (1.46) \\
Syn. LBP &  81.70 (1.78) & 82.00 (1.85) & 82.80 (1.94) \\
Sem. LBR &  76.50 (1.33) & 76.60 (1.66) & 75.30 (1.55) \\
Sem. LBP &  79.20 (1.85) & 79.60 (1.95) & 80.40 (2.06) \\
CPU (sec.) &  1.40 (0.22)  & 1.57 (0.23)  & 1.47 (0.20)  \\
\# of WGs & \multicolumn{3}{|c|}{ 1000.00 (0.00) } \\
\#states in WGs & \multicolumn{3}{|c|}{ 8.40 (0.11) } \\
\#trans in WGs & \multicolumn{3}{|c|}{ 30.50 (1.96) } \\
Corr. sen. in WG &\multicolumn{3}{|c|}{  89.80 (0.55) } \\
\hline
%
\hline
\multicolumn{4}{|c|}{Results only on word-graphs containing the correct sentence}\\
Measures & DOP$^{1}$  &  ISDOP$_{1}^{1}$  &  SDOP$^{1}$ \\
\hline
{\sl Recognized} &  99.40 (0.22) & 99.30 (0.30) & 98.70 (0.48) \\
Exact match &  77.40 (0.77) & 78.50 (1.18) & 78.80 (1.21) \\
Syn. ex. match &  84.90 (0.36) & 86.10 (1.05) & 86.40 (1.14) \\
Sem. ex. match &  79.40 (0.82) & 80.70 (1.29) & 81.00 (1.34) \\
NCB recall &  92.50 (0.68) & 92.80 (0.66) & 91.30 (1.28) \\
NCB prec. &  95.20 (0.81) & 95.50 (0.87) & 95.60 (0.81) \\
0-Crossing &  90.30 (0.80) & 91.40 (1.05) & 91.60 (1.02) \\
Syn. LBR &  86.10 (1.09) & 86.40 (1.29) & 85.10 (1.54) \\
Syn. LBP &  88.60 (1.49) & 89.00 (1.65) & 89.20 (1.59) \\
Sem. LBR &  84.00 (1.06) & 84.50 (1.44) & 83.20 (1.59) \\
Sem. LBP &  86.50 (1.42) & 87.00 (1.82) & 87.30 (1.78) \\
CPU (sec.) &  0.77 (0.25)  & 0.82 (0.24)  & 0.81 (0.24)  \\
\# of WGs & \multicolumn{3}{|c|}{ 897.80 (5.54) } \\
\#states in WGs &\multicolumn{3}{|c|}{  7.30 (0.08) } \\
\#trans in WGs & \multicolumn{3}{|c|}{ 21.50 (1.10) } \\
Corr. sen. in WG & \multicolumn{3}{|c|}{ 100.00 (0.00) } \\
\hline
\end{tabular}
\end{center}
\caption{Results on OVIS word-graphs: DOP$^{1}$ vs. ISDOP$_{1}^{1}$ and SDOP$^{1}$}
\label{WGsd1D1NoSim}
\end{table}
}

\subsubsection{B. Varying subtree depth upper-bound}
%
%
%
In general, deeper elementary-trees can be expected to capture more dependencies 
than shallower elementary-trees. This is a central quality of DOP models. 
It is interesting here to observe 
the effect of varying subtree depth on the three models that are being compared.
To this end, in a set of experiments, one random partition of the OVIS tree-bank into a test-set
of~1000 trees and a training-set of~9000 was used to test the effect of allowing the
projection of deeper elementary-trees in DOP STSGs, SDOP STSGs and ISDOP models.
The subtree depth for DOP, SDOP and ISDOP is defined as discussed in section~\ref{SecImpPDet}.

The training-set was used to project DOP, SDOP and ISDOP models.
DOP STSGs were projected with upper-bounds on subtree depth equal to~1, 3, 4, and~5.
Note that when the upper-bound is equal to~1, this results in a DOP STSG that is in fact
a SCFG (a maximum-likelihood probability assignment to the CFG underlying the training 
tree-bank). For SDOP STSGs the upper-bounds on subtree depth were equal to~1, 2, 3 and~5.
The ISDOP models were constructed from four different combinations of these SDOP and DOP models
(ISDOP$_{1}^{1}$, ISDOP$_{2}^{4}$, ISDOP$_{3}^{5}$ and ISDOP$_{5}^{5}$). 

Figure~\ref{FigDepthSize} shows the growth of number of elementary-trees as a function of
depth upper-bound for DOP as well as SDOP models. Although SDOP models are larger at lower
values for the depth upper-bound, the meaning of that value is different for the two models.
The most important fact is that the SDOP models stabilize already at depth upper-bound value~6,
whereas the DOP models are still growing even at depth upper-bound value~10~! Moreover, the
size of the largest SDOP model is less than half the largest DOP model.

Each of the twelve systems, trained only on the training-set, was run on the sentences of the
test-set (1000~sentences). The resulting parse-trees were then compared to the correct 
test-set trees. Table~\REFTAB{TabDepths} shows the results of these systems. The number of
sentences that consist of at least two words was~687 sentences and the reported results 
concern those sentences only. Note that the recognition power is not affected by the depth 
upper-bound in any of the systems. This is
because all systems allowed all subtrees of depth~1 to be elementary-trees. The
tree-language coverage for all twelve systems on the test-set was the same:~88.8\%; 
this implies that 
specialization did not result in any loss of tree-language coverage. Again the ambiguity reduction 
for the SDOP models in comparison with the DOP models was modest: approx. 1.5~times.

Figures~\ref{FigDepthEM}, \ref{FigDepthSynEM} and~\ref{FigDepthCPU} summarize graphically
the three most important measures: exact-match, syntactic exact-match and CPU-time.
In general, all models exhibit an increase of exact-match accuracy as depth increases with
the exception of a slight degradation at DOP$^{5}$, SDOP$^{5}$ and ISDOP$_{5}^{5}$. 
The degradation
that SDOP and ISDOP models exhibit is larger than the degradation of the DOP model.
An explanation of the degradation in the DOP model might be that including larger subtrees
implies many more subtrees and sparse-data effects. However, the degradation in the SDOP and
ISDOP models implies  another possible factor since the number of elementary-trees in these
models is much smaller and is comparable to smaller DOP models. It seems that all models tend
to assign too much of the probability mass to extremely large elementary-trees. 
This explains the sharper degradation in SDOP and ISDOP models: 
SDOP models do not include as many small elementary-trees as DOP models and thus tend to
assign a larger probability mass to extremely large elementary-trees.
This is an interesting observation that seems to explain our earlier observation of a 
slight degradation of the probability distributions of SDOP models as compared to the 
DOP models.

Also worth noting here is that while the ISDOP models have a recognition power that is 
comparable to the DOP models, their exact-match accuracy figures improve on
those of the DOP models. The best ISDOP models (ISDOP$_{3}^{5}$) are slightly more accurate 
than the best DOP models (DOP$^{4}$). Given the results of DOP$^{4}$ and SDOP$^{3}$, we 
conclude that the combination ISDOP$_{3}^{4}$ should be even more accurate. The SDOP models
again improve in accuracy on the DOP and ISDOP models at the small cost of 2.7\% loss of 
recognition power.

Apart from the exact match figures, bracketing accuracy figures in general tend to be better
for the DOP models than for the the ISDOP and SDOP models (even when the latter have better
exact match figures). Bracketing results tend to be a better indicator of the robustness of 
the accuracy of a system than exact match results. 
In the light of the improvement in exact match results,
the bracketing results might suggest that the SDOP models' are slightly more overfitted than 
the DOP models. 

A very interesting aspect of the effect of subtree depth on the models can be seen 
graphically in figures~\ref{FigDepthEM} and~ \ref{FigDepthSynEM}: the SDOP and ISDOP models
already start with much higher accuracy figures at a subtree depth upper-bound that is
equal to~1. This means that much smaller and much faster models already achieve an accuracy
comparable to deeper DOP models. For example, ISDOP$_{1}^{1}$ already achieves syntactic
exact-match 94.20\% while DOP$^{1}$ (a SCFG) remains far behind with 90.50\%; note that the
CPU times of both remain of the same order of magnitude though (0.16 and 0.11 seconds 
respectively). In fact ISDOP$_{1}^{1}$ has syntactic exact match that is comparable to
DOP$^{3}$ (although the exact-match figures do differ) while being approx. 4.8~times faster.
And ISDOP$_{2}^{4}$ is already much better than DOP$^{4}$ and DOP$^{5}$ while being
3.5 and 6.3~times faster respectively. The CPU graph in figure~\ref{FigDepthCPU} seems to 
indicate that increasing the subtree-depth upper-bound for the ISDOP and SDOP models does
not increase CPU-time as fast as for the DOP models. In fact, ISDOP$_{5}^{5}$ already
covers most possible subtrees that can be projected from the training-set and thus has
CPU-time consumption that is close to the maximum for ISDOP models. This is certainly not the
case for DOP models since there are a lot more subtrees that can be 
projected for larger values of the depth upper-bound; in fact the number of DOP subtrees 
and the size of DOP STSG increases rapidly as subtree depth increases\footnote{DOP models 
of depths larger than~5 tend to consume huge diskspace; Currently, the diskspace that is 
available to us is very limited. To avoid problems in the file-system we decided not to 
train or run such huge models.} 

We conclude from these experiments that ISDOP and SDOP models achieve as good accuracy results
as DOP models but ISDOP and SDOP models achieve these results much faster and at much smaller 
grammar-size. All models show a tendency towards suffering from sparse-data problems.
Due to an awkward bias in the DOP model towards preferring large elementary-trees to frequent
ones (only recently discovered~\cite{BonnemaScha99}), the probability distributions of the 
ISDOP and SDOP models tend to get worse than the DOP models as the subtree-depth upper-bound 
increases. 
%
%
\subsubsection{C. Varying train/test partitions: means and stds}

The experiments in this section report the means and standard-deviations (stds) of the parsing
and disambiguation results of each system (DOP, SDOP and ISDOP) on five independent partitions
of the OVIS tree-bank into training-set (9000 trees) and test-set (1000 trees). For every
partition, the three systems were trained only on the training-set and then tested on the 
test-set. Each of DOP and SDOP was trained with two different upper-bounds on subtree-depth
and the ISDOP system combined the DOP and SDOP systems; the resulting systems are DOP$^{4}$,
SDOP$^{2}$, ISDOP$_{2}^{4}$, DOP$^{1}$, SDOP$^{1}$ and ISDOP$_{1}^{1}$ for every partition
(i.e. thirty different parsing and disambiguation systems). 

After training on each of the five training-sets independently, the specialized TSGs consisted
of a mean number of elementary-trees of 1380.20 with std 29.39. These numbers do not include
the lexicon (i.e. rules that assign to every word a PoSTag).

Table~\REFTAB{FigMeansTable} lists the means and stds for each of the six systems.
At this point it is interesting to conduct two comparisons. Firstly,
compare the CPU-time and memory-usage for systems that exhibit comparable recognition 
power and accuracy results (i.e. DOP$^{4}$, SDOP$^{2}$ and ISDOP$_{2}^{4}$). And 
secondly, compare the accuracy results of the fastest systems (e.g. DOP$^{1}$, 
SDOP$^{1}$ and ISDOP$_{1}^{1}$).

In the first comparison, we see that the means of exact-match 
for ISDOP$_{2}^{4}$ and for SDOP$^{2}$ 
are respectively approx.\ 0.10\% and 1.95\% better than those for DOP$^{4}$. 
This comes relatively cheap since the cost in recognition-power is only 0.40\% for 
ISDOP and 3.80\% for SDOP. The mean speed-up is 3.65~times for SDOP and about 3.45~times 
for ISDOP. The other accuracy results (recall and precision) show again that the specialized 
systems have slightly less robust accuracy figures. The explanation given earlier holds 
here also (all systems have
exactly the same tree-language coverage results): the probability distributions of the
specialized systems are slightly worse than those of the DOP systems. The standard deviations
confirm this observation: the specialized systems are slightly less ``stable" than the 
DOP systems.


In the second comparison, concerning the fastest systems, we see that the three systems 
have CPU-times that can be considered comparable since they are so small; 
DOP$^{1}$ is only about 1.2 and~1.4 times faster than SDOP$^{1}$ and ISDOP$^{1}$ respectively.
However, both specialized systems achieve better exact match results (1.5\% more for ISDOP$^{1}$
and 3.45\% more for SDOP$^{1}$) at a small cost of recognition 
power (0.40\% and 3.80\% respectively).
Similarly, the other accuracy results concerning, syntactic match and semantic match 
exhibit similar improvements for the specialized systems; this is not fully true for the
(labeled) bracketing precision and recall, however.

Figure~\ref{Figd4D2meansCPU} shows the percentage of sentences that each system processes
as a function to a time-threshold. About 99.00\% of the sentences is processed 
within 3.55~secs by SDOP$^2$, 4.08~secs by ISDOP$_{2}^{4}$ and 14.65~secs by DOP$^{4}$.
Similarly, about 90\% is processed within 1.40, 1.45, 5.46 seconds respectively.
Let $X$ denote a time-threshold in seconds. For $X=1$ SDOP processes 81.40\%,
ISDOP processes 80.70\% and DOP processes 72.90\%. For $X=2$ the figures are 
95.80\%, 95.40\% and 79.60\% respectively. Clearly, the SDOP and ISDOP systems are much
faster than the DOP systems.
%
%
\subsubsection{D. Equivalence class as target-concept}
In this set of experiments we used the same five partitions as in the experiments concerning
varying the train/test partitions earlier in this section. We employ here also almost the 
same parameters and thresholds. The only difference lies in the definition of the 
target-concept: rather than using the notion of an SSF as target concept, here
we use the SSF equivalence-class definition of~subsection~\REFTHISSEC{C}. 
For convenience, we refer to the simple target-concept experiments with the SSF-based
experiments and to the present experiments with the EqC-based experiments.

After training it turned out that ``packing" together those SSFs that differ only in 
repetition of categories 
results in learning much less elementary-trees in the specialized TSG (on the five
partitions a mean of 642.80 with std 8.53). This is about half the number of elementary-trees
learned by the algorithm with the SSF-based definition; there are less elementary-trees of  
depth one (i.e. CFG rules) (about 71\%) and also fewer deep 
elementary-trees (about 28\%). The reason for having many less deep elementary-trees 
is that the number of training tree-bank trees that {\em did reduce totally}\/ by the 
iterations of the learning algorithm is much larger than in the SSF-based experiments.
Another reason for having many less elementary-trees is that shorter SSFs that had smaller 
GRF value than longer competitors in the SSF-based experiment, now join together with 
longer SSFs into the same equivalence class and thus are able more often to have a larger 
GRF value than their competitors; this results in learning shallower elementary-trees that
occur very often.

The results of the EqC-based experiments with ISDOP$_{2}^{4}$, SDOP$^{2}$ are listed in 
table~\REFTAB{FigSIMSensd4D2} and should be compared to those of table~\REFTAB{FigMeansTable}.
The results of the EqC-based definition of a target-concept show a clear deterioration 
of accuracy in comparison with the SSF-based experiments. Despite of this,
the EqC-based SDOP model (shortly SDOP$_{EqC}$) has higher bracketing recall results. 
Moreover, both the ISDOP$_{EqC}$ and SDOP$_{EqC}$ models are faster than those 
of the SSF-based experiments (about~1.2 times) and than the 
DOP$^{4}$ model (about~4.4 times). This might be due to learning a smaller specialized TSG.

From the table we also can see that the tree-language coverage remained exactly equal to
that of the DOP models and to that of the specialized models that are based on the 
simple definition. 
This implies that the deterioration of the accuracy results is again due to inferior 
probability distributions. One possible clarification for this is that the partition into 
equivalence classes of SSFs as done here is either too simplistic or the whole idea of 
partitioning the set of SSFs into equivalence classes is a bad one. It is not possible to 
determine this on the basis of the present experiments. Another plausible clarification
is that the awkward bias found in DOP STSGs~\cite{BonnemaScha99}, is magnified in SDOP
and ISDOP. 
%
%
%
\subsection{Experiments using syntax-annotation on utterances}
\label{ExpOVISSYNTAX}
In this series of experiments, the OVIS tree-bank was stripped of the semantic annotation
leaving syntactic trees only. The experiments are divided into subseries of experiments
reported in the following subsections.
\subsubsection{A. Varying subtree-depth upper-bound}
In this experiment we employed the same partition into train/test and the same training 
parameters as in the corresponding experiment on full annotation (experiment~B in 
section~\ref{ExpOVISFULL}). Table~\REFTAB{FigSynDepthSize} lists the sizes of the 
specialized~TSG and the sizes of the SDOP and DOP STSGs as a function to the
upper-bound on depth of subtrees (limited by depth~4). The specialized~TSG trained on syntax
has only~133 elementary-trees more than the syntactic DOP$^{1}$ (i.e. SCFG) but these 
elementary-trees are divided into only~240 (non-lexical) depth one elementary-trees 
and~290 deeper elementary-trees. 
Although the sizes of SDOP models seem larger than those of the DOP models for the same
value of the depth upper-bound, one must keep in mind that the same value has a different
meaning in the two cases as explained earlier. And as we have seen in table~\REFTAB{FigDepthSize},
the SDOP models reach much faster a much smaller maximum size than the DOP models.
The same applied in this case also.

The results of this experiment are listed in table~\REFTAB{FigSyntaxTable}.
Similar to earlier experiments, we see that the syntactic exact match for 
SDOP$^{1}$ and ISDOP$^{1}$ are 3.20\% better than the DOP$^{1}$ model while 
being slightly slower and slightly larger. But, in contrast to the earlier experiments,
the best DOP model (depth~4) is about 0.40\% better than the best ISDOP 
model (depth~3). Generally, however, both DOP models as well as SDOP and ISDOP improve
as the depth upper-bound is increased. An exception to this is the increase from value~3
to value~4 for SDOP and ISDOP; the explanation for this is clearly a worse probability
distribution due to having many large elementary-trees.

It is very hard to compare the models here since their accuracy results are not 
directly comparable: for example ISDOP$_{2}^{4}$'s syntactic 
exact-match falls between DOP$^{2}$ (1.20\% better) and DOP$^{3}$ (1.80\% worse).
But it is safe to state, however, that the specialization effort here is less successful 
than it is on the OVIS tree-bank with full annotation. 
\subsubsection{B. Varying train/test partitions: means and stds}
The same five partitions into train/test sets as in section~\ref{ExpOVISFULL} (subsection~C)
and the same training parameters are employed in an experiment on syntax only.
We also compare the DOP$^{4}$ against SDOP$^{2}$ and ISDOP$_{2}^{4}$ as in the earlier
experiment.

In clear contrast to the similar experiment on full annotation, the DOP
model has higher syntactic exact-match than ISDOP (1.50\%) and than SDOP (1.05\%).
However, the specialized models are about 6-7 times faster. The differences between
the bracketing accuracy results for the models are much smaller.
The exact-match measure is more stringent than the bracketing measures and exposes
a possible weakness in the specialized models compared to the DOP models. 
However, the speed of the specialized models might be attractive in application
where bracketing accuracy is a sufficient criterion on the output.

Mainly due to increase in recognition power of all three models (between 98.30\% and
99.70\%) compared to the full annotation experiment (between 91.40\% and 95.25\%),
the syntactic exact-match for the syntax-based systems is much less than that of
the systems that are based on full annotation. Clearly the semantic annotation reduces
both the recognition power and the CPU-time of the learned systems. Specialization seems
to profit from this more than the regular DOP models. A possible reason for this is that
the semantic annotation results in discriminating between SSFs that should be considered
different in any case. Another possible reason is that the probability distributions of 
the specialized models that are trained on the full annotation suffer less badly than 
those trained only on syntax. The real reasons for this situation are still not totally 
clear and this needs further investigation. 

\subsection{Experiments using full annotation on word-graphs}
\label{SecOVISWGs}
The five partitions into training/test sets of the experiment of
section~\mbox{\ref{SecOVISNewExps}.C} are used here also for an experiment on
parsing and disambiguation of the word-graphs that correspond to the 
utterances in the test-sets. Of course, every training-set (test-set) is used for
training (resp. testing) independently from the other partitions.

The result of parsing and disambiguation of a word-graph is an output parse-tree
that is compared to the test-parse on various measures including a test for equality of
the utterances on the frontiers both trees. The selection of the output parse-tree
is based on the Most Probable Intersection Derivation (MPiD) as defined in 
section~\ref{SecOptAlg4WGs}. In short, an i-derivation (intersection derivation) is a 
combination of an STSG derivation and an SFSM derivation for the same sentence (that must
be in the intersection of the string-languages of both machines);
the probability of an i-derivation is the multiplication of the probability\footnote{
The input word-graphs have probabilities that are obtained from the speech-recognizers
likelihood by a normalization heuristic due to Bonnema~\cite{BonnemaNWORep}. The only
rationale behind this heuristic is, in fact, that it combines better with the DOP probabilities
than raw speech-recognizer likelihoods. The issue of ``scaling" the likelihoods of
the speech-recognizer in a well-founded way is still under study. In any case,
Bonnema's heuristic divides the likelihood of every transition of length $m$ time-units 
by a normalization factor that is computed from the word-graph. The normalization factor 
is the likelihood of a time-unit in any transition in the word-graph to the power $m$. 
The likelihood of a time-unit in a transition of length $n$ is the n-th root of the
likelihood of the transition. 
} of the
STSG derivation with the probability of the SFSM derivation. The MPiD, for a given STSG
and a given SFSM, is the most probable of all i-derivations of any sentence that is in 
the intersection between the string-language of the SFSM and the string-language of the STSG.
The parsing and disambiguation algorithms that compute the MPiD are described in detail
in section~\ref{SecOptAlg4WGs}. 

We employ the same training parameters as the experiment of
section~\mbox{\ref{SecOVISNewExps}.C}. Table~\REFTAB{WGsd4D2NoSim}  
and table~\REFTAB{WGsd1D1NoSim} exhibit the mean and
std results of six systems each trained and tested on the five partitions into train/test
sets. The results for each system are reported once on all word-graphs and once only
on those word-graphs that contain (i.e. accept) the correct utterance.

The exact-match results of ISDOP$_{2}^{4}$ and SDOP$^{2}$, 
table~\REFTAB{WGsd4D2NoSim}, are slightly better than those of DOP$^{4}$; the
recognition power of the three systems is comparable though. The main difference between
the systems is in CPU-time: the specialized systems are about four times faster 
(and consume half the space). As the experiments of section~\mbox{\ref{SecOVISNewExps}.C} show,
except for exact-match, in general, the other accuracy measures of DOP$^{4}$ are 
slightly better that those of ISDOP$_{2}^{4}$; SDOP$^{2}$ has lower recognition power
and recall results, but improved precision results, in comparison with the other two systems.

Table~\REFTAB{WGsd1D1NoSim} exhibits the results of
DOP$^{1}$, SDOP$^{1}$ and ISDOP$_{1}^{1}$. The specialized systems SDOP$^{1}$ and 
ISDOP$_{1}^{1}$ are clearly better than  DOP$^{1}$ (i.e. SCFG) in all accuracy  measures, 
while their CPU-times do not differ much. It is clear that specializing the SCFG is
most rewarding. Compared to the ``deeper" systems (DOP$^{4}$, ISDOP$_{2}^{4}$ and SDOP$^{2}$),
however, all three have inferior accuracy results; their CPU-times figures are much smaller though.
%
\begin{figure}[h]
\epsfxsize=13.5cm
\center{
\epsfbox{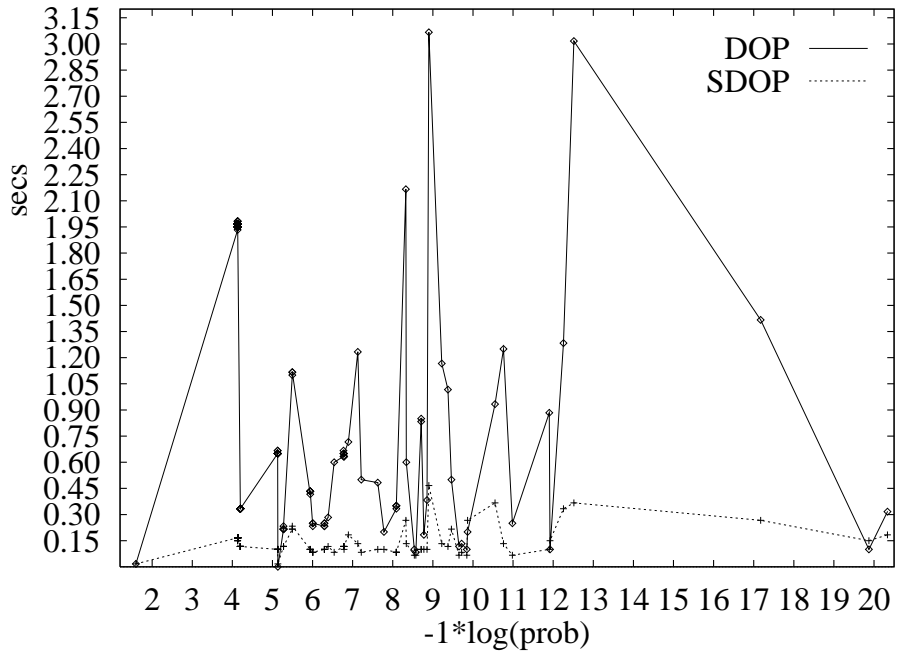} 
}
\caption{CPU-time as a function of input probability (input length 4 words)}
\label{FigFreqSpeedA}
\end{figure}

\begin{figure}[h]
\epsfxsize=13.5cm
\center{
\epsfbox{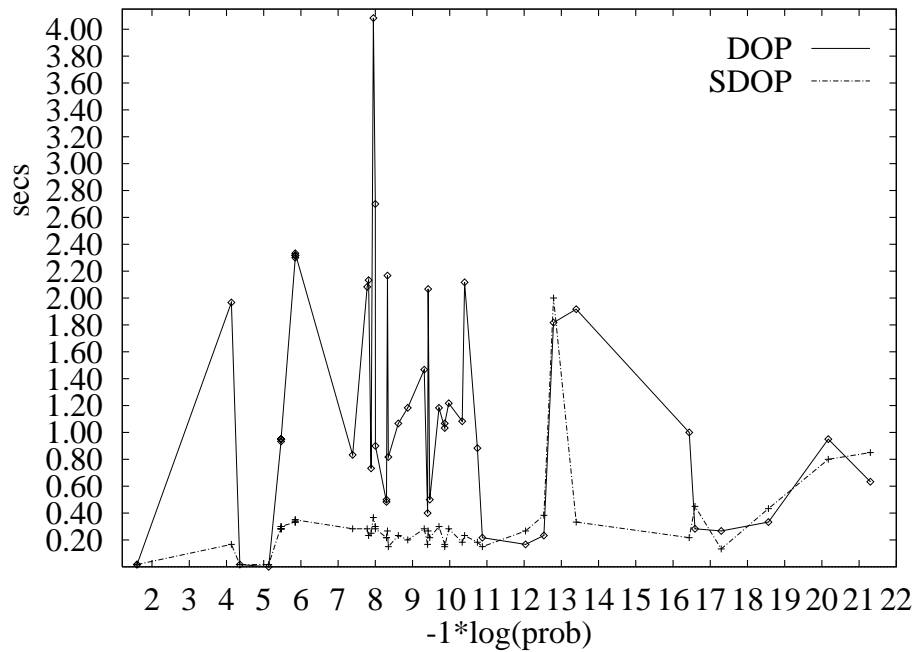}
}
\caption{CPU-time as a function of input probability (input length 6 words)}
\label{FigFreqSpeedB}
\end{figure}

\subsection{More frequent input processed faster}
The goal in this subsection is to observe the CPU-time of both DOP and ISDOP as a function of 
the relative frequency of the input. Ideally, for observing whether a system processes more 
frequent input faster, the only parameter that should be allowed to change is the frequency 
of the input; neither the grammar nor the input should change. This implies that it is 
necessary to conduct an experiment on two different tree-banks that share the same underlying
grammar and contain exactly the same utterances but differ only as to the frequencies of the 
utterances and the other linguistic phenomena. Since this experiment is very hard to 
arrange in order for the results to be meaningful, 
we will compare here the CPU-times of the systems 
on utterances of different frequencies from the same tree-bank.
This means that other factors than frequency may interfere in the results, e.g.
the ``degree of ambiguity" of an utterance and its length. We try to minimize the effect of
such factors.

Here we compare the efficiency-behavior of DOP$^{4}$ model to that of ISDOP$_{2}^{4}$ model.
Both systems are borrowed from the experiments reported in
subsection~\REFSUBSUBSEC{ExpOVISFULL}{B}.
Since the specialization algorithm is not lexicalized, actual words should not play 
a role in the comparison. Therefore, both systems were adapted to parse and evaluate 
PoSTag sequences rather than word-sequences. Moreover, a suitable test-set is extracted 
from the original test-set: it consists of the correct sentential PoSTag sequences that 
are associated with the word-sequences of the original test-set.

As expected, it is hard to estimate the relative frequencies of sentential PoSTag sequences 
due to data sparseness.
Luckily, we are not interested directly in the relative frequencies of sentential PoSTag 
sequences but in the {\em shape of the probability distribution that they constitute over 
the various sentential PoSTag sequences}. A reasonable approximation of the shape of this 
distribution is DOP's probability distribution over these sequences. After all, DOP's
probability distribution over these sequences is derived from their relative frequencies.
However, DOP's probabilities depend to a certain extent on the length of the input.
To avoid this effect, we decided to conduct comparisons only on inputs of the same length.

In figures~\ref{FigFreqSpeedA} and~\ref{FigFreqSpeedB} we plot the CPU-time as a 
function of the $-log(prob)$, where
$prob$ is the DOP$^{4}$ probability of the test PoSTag sequence. In figure~\ref{FigFreqSpeedA}
we plot this for sentence length~4 (164 events) and in figure~\ref{FigFreqSpeedB} for sentence 
length~6 (62 events). We selected these two plots because they are representative of the various
other possible plots for other sentence lengths. It is clear that the ISDOP$_{2}^{4}$ is
much faster than DOP$^{4}$ (while they are comparable in precision and coverage as 
subsubsections~\REFSUBSUBSEC{ExpOVISFULL}{B} and~\REFSUBSUBSEC{ExpOVISFULL}{C} show).
However, generally speaking SDOP$_{2}^{4}$ processes {\em more probable input slightly faster 
than it processes less probable input}\footnote{Note that because the x-axis is $-log(prob)$,
the points at the left-side of the x-axis represent higher probability values than points 
at the right-side.}, 
while DOP$^{4}$ seems to behave in an unpredictable manner in this respect. 
The slight fluctuations in the SDOP$_{2}^{4}$ plot may be attributed to two factors.
Firstly, these are often input sequences that are parsed and disambiguated by the Par+DOP 
model because they are not in the string-language of the SDOP model. And secondly, these 
sequences may constitute more ambiguous input. In any case, this modest experiment clearly
shows that the behavior of the specialized model with respect to the frequency of the input
is much more attractive than that of the original DOP model.
%
\subsection{OVIS: summary of results and conclusions}
%
To summarize the experiments on parsing and disambiguation of OVIS 
sentences and word-graphs: 
\begin{itemize}
\item On full annotation, specialization results in at least equally accurate systems 
      that are significantly smaller and faster than DOP systems. 
      However, the specialized systems seem to have slightly more overfitted probability 
      distributions than the DOP systems.
      A hypothesis on why this overfitting takes place in the specialized systems is that
      these systems are based on STSGs that consist of many larger subtrees and fewer
      smaller subtrees; the larger subtrees seem to claim too much of the probability mass 
      (because there are fewer smaller subtrees than in the DOP models). This is due to
      a strange bias in the DOP model (which carries over to SDOP models).
\item Both DOP and the specialized models tend to improve as the size of 
      the training-set increases.
\item The specialized models achieve better accuracy results than DOP models when the
      subtree depth is limited to shallow subtrees. In general, as subtree-depth upper-bound 
      increases, all systems improve and become almost as accurate. However, from a certain 
      value for the depth upper-bound, the systems start to suffer from sparse-data problems
      or/and overfitting.
\item In particular, specializing DOP$^{1}$ (i.e. an $SCFG$) 
      results in specialized systems SDOP$^{1}$ and ISDOP$_{1}^{1}$ that are equally fast 
      but that have much improved accuracy figures.
\item Extending the definition of the target-concept to equivalence classes of SSFs
      results in accuracy results that are slightly inferior to those of the DOP model 
      and the specialized models that are based on the original definition. The speed-up,
      however, improves slightly.
\item On the syntactic OVIS tree-bank specialization results in worse exact-match than the
      DOP models. Bracketing measures, however, are comparable to the DOP models.
      The specialized models are also much faster. The ISDOP$_{1}^{1}$ and SDOP$^{1}$ models 
      improve in all measures on the DOP$^{1}$ model (i.e. SCFG).
\item In general, more frequent input is parsed and disambiguated faster by the specialized 
      models, whereas DOP tends to show unpredictable behavior in this respect.
\end{itemize}
We conclude here that specializing DOP by the current implementation of the GRF algorithm 
to the OVIS domain
did not jeopardize the tree-language coverage (but it also reduced the ambiguity only by 
a limited amount due to the fact that it is not lexicalized). The specialized
DOP models are faster, smaller and as accurate as (if not better than) the regular DOP models.
In particular, two specializations are most successful. Firstly, the specialization of 
the SCFG (i.e. DOP$^{1}$) is quite successful: the specialized model is almost as fast 
but improves the accuracy results significantly. And secondly, the specialization of the
deepest DOP models results in much smaller and faster but equally accurate models.
%
%

%
%
\section{Experiments on SRI-ATIS tree-bank}
\label{SecATISExps}
In this section we report experiments on syntactically annotated utterances from the 
SRI International ATIS tree-bank. The utterances of the tree-bank originate from the ATIS 
(Air Travel Inquiry System;~\cite{HemphilEtAl90}) domain. For the present experiments,
we have access to~13335 utterances that are annotated syntactically (we refer to this 
tree-bank here as the SRI-ATIS tree-bank). The annotation scheme originates from the
linguistic grammar that underlies the Core Language Engine (CLE) system~\cite{Alshawi92}.
The annotation process is described in~\cite{CarterACL97}, it is a semi-automatic process
with a human in the annotation loop; the CLE system, supplemented with various 
features, is used for suggesting analyses for the utterances that are being annotated.
Among these features, there is a {\em preference mechanism}\/ that can be trained on
the annotated part of the material and enables partial disambiguation of the space of analyses.
Another feature is a set of heuristics that enable rapid manual selection of the correct 
analysis by employing a compact and smart representation of the analyses.

The rest of this section is structured as follows. Subsection~\ref{SubSecPrep}
discusses the detail of preparations that are necessary for conducting the experiments.
Subsection~\ref{SubSecATISFirst} reports a set of experiments testing DOP and the 
specialized models on the ATIS tree-bank.
And subsection~\ref{SubSecATISConc} summarizes the results of the experiments on the ATIS domain.
\subsection{Necessary preparations}
\label{SubSecPrep}
To conduct the present experiments on the SRI-ATIS tree-bank two issues had to be
addressed. Firstly, the tree-bank has an underlying grammar that is cyclic. 
And secondly, the tree-bank contains traces of movements, i.e. epsilons.
Since the DOPDIS system can not deal with any of those, we had to devise solutions.
The tree-bank is transformed by a simple algorithm into a tree-bank that has an acyclic 
underlying grammar.
And the DOP projection mechanism is adapted to allow epsilon traces in elementary-trees
but the epsilons are always internal to other elementary-trees. This section 
describes in detail both solutions. 
\subsubsection{Removing cycles}
Since our parsers and disambiguators assume acyclic grammars ,
some measures had to be taken in order to remove the cycles from the grammar that underlies
the SRI-ATIS tree-bank. We decided to apply a series
of simple transformations {\em to the trees of the tree-bank}\/ (rather than directly to 
the grammar underlying the tree-bank). The transformations are applied in turn to every tree
of the tree-bank. The transformations that we developed do not change the tree-bank trees 
too much; a guideline was that only minimal changes that can be justified linguistically 
or that are really necessary should be used. We employed three transformation that were 
applied in a pipeline to every tree in the tree-bank. The transformations, in their order of 
application are:
\begin{description}
\item [Bamboo-trees:] If the tree $t$ in the tree-bank has a partial-tree $Bt$ that 
           involves only unary productions (often referred to with the term ``Bamboo" 
           partial-trees), 
           all nodes in $Bt$ are removed except for the bottom and the top nodes that are
           now connected directly. This reduces the number of cycles effectively. The intuition
           behind this is that the many internal nodes in a Bamboo partial-tree, if necessary
           at all, should be really internal and not visible\footnote{Of course, by removing 
           these internal nodes we loose the internal structure of Bamboo-trees. For 
           conserving the internal structure of a Bamboo-tree, one could represent its internal 
           nodes by a single node labeled by a new non-terminal, constructed by juxtaposing 
           the labels of the internal nodes in some order (e.g. lowest node to highest node 
           in the tree).}. They should not serve as substitution sites or as left-hand sides
           of grammar rules. The choice of removing the internal nodes was the simplest 
           sensible solution. 
\item [NP in VP:] If the tree $t$ has an internal node labeled NP (noun-phrase)
         that derives only a VP (verb-phrase), the VP is renamed into an INFVP, i.e.
         infinitive VP. This avoids the common cycle \mbox{$VP\lmd NP\lmd VP$}.
         The intuition here is that all these VPs  are in fact infinitive VPs rather
         than usual VPs.
\item [Clear cycles:] 
       If in tree $t$ there is a node $N_{t}$ labeled $XP$ that governs a partial-tree that
       has an internal node $N_{l}$ labeled also $XP$ and if all branches along the path between
       $N_{t}$ and $N_{l}$ result only in empty-strings ($\epsilon$), then the whole 
       partial-tree between $N_{t}$ and $N_{l}$ is removed and $N_{l}$ becomes a direct
       child of the node that was parent of $N_{t}$ (exactly in the same place instead of 
       $N_{t}$). Virtually all such $XP$s in the SRI-ATIS tree-bank are in fact VPs. 
       Again the intuition is that the removed partial-tree between the $XP$s should be 
       also considered internal. This transformation removes only a small part of traces 
       of movement in the tree-bank, many others remain in tact. 
\end{description}
%
%
The resulting tree-bank has an underlying grammar that is acyclic. As we shall see, the 
coverage and the tree-language coverage of that grammar remains very high. In the sequel, 
the name ``T-SRI-ATIS tree-bank" refers to the tree-bank obtained from the original
SRI-ATIS tree-bank after these transformations.

Table~\ref{FigATISinNums} (page~\pageref{FigATISinNums}) summarizes some of the 
characteristic numbers of the T-SRI-ATIS tree-bank. The mean sentence length 
is approx.~8.2 words and the number of trees is~13335 trees.
\subsubsection{Training and parsing in the presence of epsilons}
Let the term {\em empty-frontier partial-tree}\/ refer to a partial-tree that has a 
frontier that consists of a sequence of epsilons only (i.e. no other symbols).
The DOPDIS system does not allow epsilon rules or empty-frontier elementary-trees
because we think that such creatures {\em should not be left on their own}\/ 
in the first place. In our view, for projecting STSGs, all empty-frontier partial-trees
in the tree-bank should be treated as {\em internal to other partial-trees that
subsume them in the tree-bank trees}. To achieve this effect, it is necessary to adapt 
the DOP (and SDOP) projection mechanism. The new mechanism projects all subtrees of the
tree-bank trees except for the {\em empty-frontier subtrees}\/ (i.e. the mechanism weeds out 
the empty-frontier subtrees). Crucial here is that all empty-frontier partial-trees in 
the tree-bank trees are now internal to the elementary-trees of the STSG. We stress this 
fact again: the resulting STSG generates also the empty-frontier partial-trees but always
as part of other partial-trees that are not empty-frontier.
For calculating the depth of a subtree, the mechanism assigns depth 
zero to empty-frontier partial-trees. The probabilities of the subtrees are calculated as usual 
from the relative frequencies of the non-empty-frontier subtrees only.
\subsection{Experiments on T-SRI-ATIS}
\label{SubSecATISFirst}
This section exhibits the first series of experiments that we conducted on the T-SRI-ATIS 
tree-bank. It is important to stress here that the present experiments differ completely
from earlier experiments~\cite{RENSDES,GoodmanDES,MyRANLP95} with DOP on the ATIS tree-bank 
of the Penn Treebank Project; the latter tree-bank contains only approx.\ 750~trees (vs.\
13335~trees in the SRI-ATIS) that exhibit much less variation of linguistic phenomena than 
the SRI-ATIS tree-bank. Moreover, the two tree-banks are annotated differently.

The specialization algorithm that is used here is the GRF algorithm implemented as 
described in section~\ref{SecImpDet} with the equivalence classes of SSFs as the target
concept. 
Some of the training parameters were fixed in all experiments that are reported in the rest 
of this subsection: the upper-bound on the length of learned SSFs was set to~8 grammar symbols,
the threshold on the frequency of sequences of symbols was set on~10, the threshold on
the Constituency Probability of SSFs was set on~0.87, and all DOP/SDOP/ISDOP STSGs were
projected under the parameters~n2l4L3 (as explained in section~\ref{SecImpPDet}) unless
stated otherwise.
%
\begin{figure}[htb]
\epsfxsize=13cm
\epsfbox{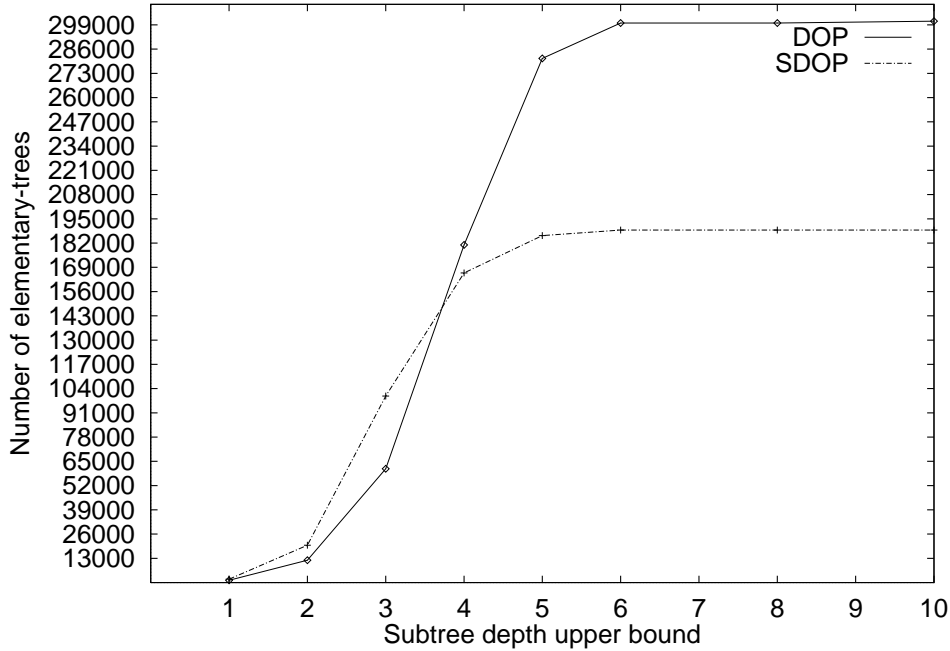}
\caption{T-SRI-ATIS: Number of subtrees to depth upper-bound}
\label{FigDepthUBSize}
\end{figure}

{
\begin{table}[htb]
\begin{center}
\begin{tabular}{|c|c|c|c|c|}
\hline
Measures & DOP$^{1}$  &  DOP$^{2}$  &  DOP$^{3}$ & DOP$^{4}$\\
\hline
{\sl Recognized} &  99.90 & 99.90 & 99.90 & 99.90 \\
{\sl TLC} & 99.50 & 99.50 & 99.50 & 99.50 \\
Syn. ex. match &  46.00 & 69.70 & 79.10 & 82.70 \\
NCB recall &  92.00 & 96.60 & 97.50 &  98.30 \\
NCB prec. &  94.20 & 97.40 & 97.90  &  98.40 \\
0-Crossing &  62.60 & 79.20 & 84.00 &  87.60 \\
Syn. LBR &  88.80 & 95.40 & 96.70 &  97.30 \\
Syn. LBP &  90.80 & 96.20 & 97.20 &  97.40 \\
CPU (sec.) &  2.49 & 16.29 & 146.73 &  710.58 \\
\hline
%
Measures & SDOP$^{1}$  &  SDOP$^{2}$  &  SDOP$^{3}$  & SDOP$^{6}$\\
\hline
{\sl Recognized} &  99.90 & 99.90 & 99.90  &  99.90 \\
{\sl TLC} & 99.50 & 99.50 & 99.50 & 99.50 \\
Syn. ex. match &  50.80 & 71.20 & 79.80  &  83.60 \\
NCB recall &  93.50 & 96.60 & 97.70 &  98.20 \\
NCB prec. &  95.10 & 97.20 & 98.00 &  98.30 \\
0-Crossing &  65.90 & 80.50 & 84.60 &  87.30 \\
Syn. LBR &  90.60 & 95.40 & 96.90 &  97.30 \\
Syn. LBP &  92.10 & 96.00 & 97.20 &  97.40 \\
CPU (sec.) &  10.34 & 55.67 & 363.68 &  942.74 \\
\hline
\# of sens &   \multicolumn{4}{|c|}{1000.00 } \\
Sen. length & \multicolumn{4}{|c|}{ 8.20 } \\
\hline
\end{tabular}
\end{center}
\caption{ATIS experiments: DOP and SDOP models of various depths}
\label{TabATISdepth}
\end{table}
}

{
\begin{table}[htb]
\begin{center}
\begin{tabular}{|c|c|c||c|c|}
\hline
Measures & DOP$^{1}$ &  DOP$^{2}$ & SDOP$^{1}$ & SDOP$^{2}$\\ 
\hline
{\sl Recognized} &  99.97 (0.05)  & 99.97 (0.05)  &  99.93 (0.05)  & 99.93 (0.05)  \\
{\sl TLC} & 99.50 (2.16) & 99.50 (2.16)  & 99.50 (2.16) & 99.50 (2.16)  \\
Syn. ex. match &  46.50 (1.08) & 70.50 (1.46)   &  50.00 (1.46) & 70.80 (0.69)  \\
NCB recall &  92.50 (0.31) & 96.80 (0.19)   &  93.50 (0.24) & 96.80 (0.19)  \\
NCB prec. &  94.50 (0.25) & 97.60 (0.16)  &  95.00 (0.24) & 97.50 (0.20)  \\
0-Crossing &  63.20 (0.56) & 80.50 (1.53)   &  65.70 (0.56) & 80.60 (0.71)  \\
Syn. LBR &  89.20 (0.37) & 95.60 (0.25)   &  90.50 (0.38) & 95.60 (0.14)  \\
Syn. LBP &  91.20 (0.25) & 96.40 (0.25)   &  91.90 (0.42) & 96.20 (0.15)  \\
CPU (sec.) &  2.48 (0.07)  & 16.19 (0.34)  &  10.70 (0.56)  & 57.89 (3.00)   \\
\# of sens & \multicolumn{4}{|c|}{1000.00 (0.00) }  \\
Sen. length & \multicolumn{4}{|c|}{8.20 (0.04) } \\
\hline
%
\hline
\end{tabular}
\end{center}
\caption{ATIS: Means and STDs}
\label{TableMeansSTDs}
\end{table}
}

\subsubsection{A. Varying subtree depth}
A training-set of 12335~trees and a test-set of 1000~trees were obtained by partitioning
the T-SRI-ATIS tree-bank randomly. Both DOP and SDOP models with various depth upper-bound 
values were trained on the training-set and tested on the test-set. It is noteworthy that
the present experiments are extremely time-consuming: for upper-bound values larger than three,
the models become huge and very slow, e.g. it takes more than 10~days for DOP$^{4}$ to parse 
and disambiguate the test-set (1000~sentences).

Figure~\ref{FigDepthUBSize} exhibits the number of elementary-trees as a function of
subtree-depth upper-bound for the models DOP and SDOP (under the projection parameters
n2l4L3). The size of the SDOP models becomes smaller than the DOP models only from 
subtree-depth upper-bound equal to four.
The DOP model without any upper-bound on depth, i.e. the UTSG of the tree-bank,
consists of twice as many elementary-trees as the corresponding SDOP model; this means that the
SDOP model results in approx. 40-50\% reduction of memory-use. However, both models are 
extremely large and, currently, this reduction does not result in useful speed-up.
However, reducing the number of subtrees can be instrumental in dealing with sparse-data
effects.

Table~\REFTAB{TabATISdepth} shows the results for depth upper-bound values smaller than four.
A striking figure is that the percentage of recognized sentences of the SDOP is exactly equal
to that of the DOP model. This means that the ISDOP models are equal to the SDOP models in 
this case. After studying the specialized tree-bank trees it turned out that after reducing
a big portion of the tree-bank trees, the learning algorithm went on learning grammar rules
until it reduced most of the trees totally. This happened because the stop condition of
the algorithm relies on a frequency threshold that is very hard to set at suitable values.
A better stop-condition would be to set an upper-bound on how much of the tree-bank should 
be covered by the training algorithm, e.g. check that the ratio between the number of nodes 
that remained after this iteration and the number of nodes in the original tree-bank trees 
does not drop under a certain a priori set minimum value.

The SDOP system turned out to have a parse-space that is on average about 1.2~times smaller than
the DOP systems (i.e. we compare the parse-space of the Specialized TSG and the CFG underlying
the training tree-bank). Clearly, this is a very small reduction in ambiguity. This is a 
bit surprising because one expects more reduction of ambiguity on the longer ATIS
sentences than on the shorter OVIS sentences. This ``surprise" is exactly a consequence of the 
severe practical limitations of the current implementation.
In any case, although these SDOP models have a string-language and tree-language that are 
comparable to the DOP models, it remains interesting here to observe how these SDOP models
compare to the DOP models, i.e. whether they conserve tree-language coverage, 
have good probability distributions and constitute any improvement on the DOP models.

As table~\REFTAB{TabATISdepth} shows, the shallow SDOP models (subtree depth limited to three)
are still larger and therefore also slower than their DOP counterparts. The $SDOP^{1}$ and
the DOP$^{1}$ models are the fastest in their families and the least accurate. However, 
SDOP$^{1}$ already improves on DOP$^{1}$ by some 4.78\% syntactic exact match. The differences
in exact match between the SDOP and DOP models become smaller but are still substantial. 
The SDOP models are also some 4 times slower for SDOP$^{1}$ and about 2.5 times for SDOP$^{3}$.
It is very interesting that by specializing the DOP model (the one that consists of all
tree-bank subtrees without any depth upper-bound) we obtained a series of limited-depth SDOP 
models that are more accurate but also slower. In fact the SDOP models fit exactly in
between the DOP models in matters of accuracy and CPU-time: SDOP$^{i}$ fits between DOP$^{i}$
and DOP$^{i+1}$. Given that the size of the SDOP model becomes smaller than the DOP model
only from depth upper-bound value four, we also expect speed-up to be observed only at those 
values. This means that this speed-up is currently not really useful.
%
%
%
\subsubsection{B. Varying training/test partitions: means and stds}
Four independent partitions into test (1000 trees each) and training sets (12335 trees each)
were used here for training and testing the DOP and the SDOP models. The training parameters
remain the same as in the preceding subsection (A). The means and the stds are exhibited
in table~\REFTAB{TableMeansSTDs}. The same situation as earlier occurs here also. The SDOP 
models and the DOP models are not directly comparable; the SDOP models fit in between the 
DOP models, where SDOP$^{i}$ fits in between DOP$^{i}$ and DOP$^{i+1}$. The SDOP models are
more accurate but slower than the DOP models.

\subsubsection{Discussion}
It is on the one hand disappointing that specialization by the GRF did not result in
useful speed-up. On the other hand, three facts must be kept in mind. Firstly, the results 
show that the
specialized models form an alternative to the DOP models since they fit in between the
DOP models; especially SDOP$^{1}$ offers a more attractive PCFG model than the DOP$^{1}$ model.
Secondly, this experiment has shown that on the ATIS tree-bank the GRF algorithm results in 
speed-up only at very large DOP models that are impractical. And thirdly, and most importantly,
the seriously suboptimal current implementation (GRF algorithm without lexicalization) 
did not result in ambiguity reduction. It turns out that without significant ambiguity-reduction,
the GRF learning algorithm does not provide useful speed-up. It does reduce the number 
of subtrees significantly, however, a thing that can help avoid sparse-data effects.

The question at this point is why does the speed-up come only at large depth 
values (four or deeper), while in the OVIS case the speed-up was already at SDOP$^{2}$. 
First of all, the fact that the SDOP models have the same string-language and tree-language
as the DOP models already gives a clue. Contrary to the OVIS case, here the learning algorithm 
went on learning (43~iterations) until it reduced many of the tree-bank trees totally. From a 
certain iteration on (20th-22nd), the algorithm learned only CFG rules. The stop condition, 
a frequency lower-bound for SSFs, is too weak to control such behavior. Clearly, it is necessary 
to employ a stronger stop condition with direct monitoring of the coverage of the learned grammar.
Secondly, the ATIS trees are larger than the OVIS trees and there is more variation. In the 
OVIS case, many trees were learned as is in one iteration. This means that these trees did not
have cut nodes except their root nodes (and the PoSTag level nodes); and this implies that there
are many less subtrees in the SDOP models than the DOP models already at small depth upper-bounds.
The language use in OVIS is constrained by a dialogue protocol, a thing that makes it much
simpler than the language use in the ATIS domain.
Thirdly, the SDOP$^{2}$ in OVIS achieves the same accuracy results as the DOP$^{4}$,
whereas in the ATIS case, although SDOP$^{2}$ improves on DOP$^{2}$ in accuracy, it still 
does not compare to deeper DOP models. Clearly, subtree depth is a more important factor in 
accurate parsing of the ATIS domain than it is in parsing the OVIS domain. In OVIS, DOP$^{1}$
achieves syntactic exact match that is only~4\% less than the best model $DOP^{4}$; in ATIS,
the difference is about~35\%~! It is very hard to bridge such a gap by the shallower SDOP 
models. 
%
%
\subsection{ATIS: summary of results and conclusions}
\label{SubSecATISConc}
The results of the ATIS experiments can be summarized as follows:
\begin{itemize}
\item The SDOP model that implements the idea of the DOP model as containing
      {\em all}\/ subtrees of the tree-bank is substantially smaller than the original
      DOP model. When using a subtree depth upper-bound the SDOP models become smaller
      only from value~3. And the SDOP models can be faster than their DOP counterparts 
      only from that value.
\item The SDOP models have recognition power that is equal to the DOP models.
      The accuracy is also comparable. 
\item The series of SDOP models with the various depth upper-bound values constitutes a
      series of models that are hard to compare to their DOP counterparts. In fact the
      SDOP and DOP models can be merged into one series of models since $SDOP^{i}$ fits
      in between DOP$^{i}$ and DOP$^{i+1}$ in matter of accuracy and speed.
\item The model SDOP$^{1}$ is an SCFG that achieves much better accuracy than the DOP$^{1}$
      model at some limited cost of time and space.
\end{itemize}
In conclusion, the ATIS experiments have exposed a weakness in the current implementation
of the specialization algorithm: it does not reduce ambiguity substantially to allow speed-up.
However, the fact that the SDOP models have accuracy results that improve on their DOP 
counterparts implies that the specialized models have good probability
distributions and tree-language coverage. It might be very attractive to set the parameters
at such values that the SDOP models have less recognition power than the DOP models: this 
might mean some speed-up. However, we believe that restricting the string-language can be 
better done by ambiguity-reduction, i.e. restriction of the tree-language.

\section{Concluding remarks}
\label{CHIMPconcs}
In this chapter we discussed the details of a first implementation of the specialization
algorithms of chapter~\ref{CHARS}. This implementation is severely limited due to the
currently available hardware. It employs the GRF measure, is not lexicalized and embodies 
many approximations that were dictated by the need to implement a system that runs within
acceptable time and space limits.

The chapter presented also an extensive experiments on the OVIS and ATIS domains.
We experimented with {\sl approx.~150 DOP and SDOP systems}, each trained on~10000-12335
trees and tested on~1000 sentences or word-graphs.
The experiments reported in this chapter are the first that 
train and test the DOP model on such large tree-banks extensively
using the cross-validation method. It is very hard to compare the results of the DOP
models or the SDOP models to other corpus-based models because (to the best of our knowledge)
there are no such systems that report experiments on these two tree-banks\footnote{
Most other work uses larger benchmarks, e.g. UPenn's Wall Street Journal, which are still 
too large to use for serious DOP experiments given the available hardware.}.
But the chapter presents the classical comparison between DOP and the Maximum-Likelihood SCFG
(DOP model of depth-one subtrees) underlying the training tree-bank (or so called 
the ``tree-bank grammar"~\cite{Charniak96}). The DOP and SDOP models (of subtree depth 
upper-bound larger than one) clearly improve on the SCFG models in both domains. 
In the ATIS domain, the improvement is drastic (about~35\% extra tree exact match improvement).
In any case, the exact-match figures of the DOP models on these two differently annotated 
tree-banks (for two different languages and domains) exhibit the potential of 
DOP as a model of natural language disambiguation.

The experiments also show that the specialized DOP models (SDOP and ISDOP) are in general 
as accurate as the DOP models and have a comparable tree-language coverage. Due to the severely 
suboptimal current implementation, the ambiguity-reduction is small (1.2-1.6 times) and much 
less than one might expect.   

On the OVIS domain, the SDOP and ISDOP models are as accurate as but smaller 
and faster than the DOP models. The experiments on parsing and disambiguation of utterances 
and word-graphs from that domain show clearly that the SDOP models are more attractive than
the DOP models.  In contrast to the DOP models, the SDOP and ISDOP models have the property
that the processing times for more frequent input are smaller.

On the ATIS domain, the experiments show that the SDOP models constitute a different sequence of
models than the DOP models. While the DOP model with a given subtree-depth upper-bound has 
accuracy results that is smaller than the SDOP model with the same value of subtree-depth 
upper-bound, the DOP model has smaller processing times.
Therefore, it is very hard to directly compare the two 
kinds of models and it is even harder to conclude sharply that the one is better or faster 
than the other. Worth mentioning, however, is that the SDOP model that contains all subtrees 
of the tree-bank is much smaller than its DOP counterpart.

It is appropriate here to stress that both the SDOP and the DOP systems are based on
the same optimized parsing and disambiguation algorithms of chapter~\ref{CHOptAlg4DOP}.
Therefore, the speed-up that the optimizations and heuristics of chapter~\ref{CHOptAlg4DOP}
provide is already included in the CPU-times and the sizes of both the DOP and the SDOP models. 
However, it is safe to say that these optimizations and heuristics
have often been a major factor in making the experiments not only considerably more efficient 
but also feasible at all\footnote{ For example, in most cases, we could not project DOP models 
of some subtree-depth upper-bound three or more if we did not limit the number of 
substitution-sites to two per subtree. And in many other cases, it was clear that without
the optimization of the parsing algorithm to time-complexity linear is grammar size,
running a 5-fold cross-validation DOP experiment could consume many more weeks.}.

Clearly, the conclusions from the present experiments remain limited to the domains that
were used. It is very hard to generalize these conclusions to other domains. Moreover,
due to the severely suboptimal current implementation, the original question pertaining 
to the utility of the ARS specialization algorithms in offline ambiguity-reduction 
is here only partially answered. The answer to this question depends on whether the 
theoretical ideas of chapter~\ref{CHARS} can be implemented in an algorithm that does not 
embody strong assumptions but remains tractable. 

Future work in this direction should address the limitations of the current implementation.
A future implementation can be better off if it adopts the entropy-minimization formulae, 
does not employ the simple sequential covering scheme but a more refined one, and exploits 
lexical information to reduce ambiguity. A possible track of research here is to use lexical
information both during learning and parsing. 
During learning the lexical information can be incorporated in the measures of ambiguity 
of SSFs (or equivalence classes of SSFs). Moreover, when the lexical infomation is represented
in feature-structures, an instantiated feature-structure is associated with every subtree in 
the ambiguity-set of an SSF that is learned from the tree-bank. During parsing, instead 
of constructing all the instantiated feature-structures of a learned SSF by unification and 
inheritance, the feature-structures in the ambiguity set of that SSF are simply checked on 
whether they fit with the lexical input or not. This is expected to play a major role in 
speed-up of parsing since it enables the immidiate pruning of ambiguity-sets.
On another track, it might be worth the effort to develop fast methods for partial-parsing 
the specialized grammar based on some clever indexing schemes or on the Cascade of Finite 
State Transducers discussed shortly in chapter~\ref{CHARS}. Such ideas have proved to be
essential in earlier applications of EBL for specializing 
grammars~\cite{SamuelssonRayner91,Samuelsson94,SrinivasJoshi95,Neumann94}.
%

\chapter{General conclusions}
\label{CHDiscussion} 
This thesis concentrated on the efficiency and the complexity of natural language 
disambiguation, especially under the Data Oriented Parsing model. On the one hand, it 
established that some problems of probabilistic disambiguation belong to classes that are
considered intractable. And on the other hand, it provided various solutions for improving 
the efficiency of probabilistic disambiguation. Some of these solutions constitute optimized 
extensions of existing parsing and disambiguation algorithms, and others try to tackle the 
efficiency problem more fundamentally through specializing the models to limited domain 
language use. A central hypothesis that underlies the latter kind of solutions is that existing 
performance models can be more efficient if they are made to account for the property that 
more frequent input is processed more efficiently. The thesis also provided an empirical study of 
instances of these solutions on collections of data that are associated with some  
applications. Two questions may be raised at this point.
Firstly, {\em in how far do the solutions that this thesis provides solve the efficiency 
problem of the DOP model~?}\  And secondly, {\em in how far do these
solutions enable modeling the property that more frequent input is processed more 
efficiently~?}\ Next we try to answer both questions.

We note first that the answer to the first question can be conclusive only in the light 
of some application that imposes clear time and space requirements. Therefore, it is suitable
to restate that question as a more specific question that can be answered on the basis of our 
actual experience in this thesis: {\em did our solutions enable substantially more efficient 
disambiguation under the DOP model on the domains employed in this thesis~?}\
Clearly, the answer to this question may entail general conclusions concerning the present methods 
only when the applications being addressed involve similar limited domains.

If we look back at the starting point of this work we can apply the following simple-minded 
quantitative reasoning about how far we are now from that point. About five years ago, 
before this work started, the average time for processing an ATIS-domain Part-of-Speech (PoSTag) 
sequence by a DOP model trained on~700 trees was about~12000 
seconds~\cite{RENSMonteCarlo,RENSDES}. 
Three years ago, the optimized algorithms of chapter~\ref{CHOptAlg4DOP} enabled a 
net~100-500 times speed-up on the same material~\cite{MyRANLP95}. 
The efficiency improvement that these algorithms enable can also be seen in the relatively 
acceptable time and space results of the DOP model on OVIS domain utterances 
(section~\ref{SecOVISExps}). Obviously, these algorithms have improved the efficiency of 
the DOP model substantially and have made it possible to experiment with the DOP model on 
larger tree-banks more extensively.
However, we cannot conclude that these optimizations have completely solved the efficiency 
problem for DOP on all similar domains. The experiments that are reported in 
section~\ref{SecATISExps} expose the magnitude of the efficiency problem of the DOP model 
even when using these optimized algorithms.
For the DOP models that are employed in these experiments, trained on~15 times as many trees as
the experiments mentioned above, the average time for processing a sentence by the optimized
algorithms is~100-1000 seconds\footnote{Note that this concerns processing actual word-sequences 
which are more ambiguous than PoSTag sequences.}. Although our optimized algorithms enable 
a substantial speed-up when compared to those described in~\cite{RENSMonteCarlo}, alas both
kinds of algorithms are currently not yet useful for practical applications in the ATIS domain.
Of course, we can still apply pruning techniques to improve the efficiency of these algorithms,
but we feel that pruning techniques will not solve the problem fundamentally.
This clearly indicates that the ultimate goal of making the DOP model 
efficient enough to be useful in real-world applications will not be achieved 
{\em solely by optimizations of traditional parsing techniques}. 
The question as to what kinds of ``non-traditional" techniques 
should make this goal attainable is discussed in chapter~\ref{CHARS}.

The main motivation behind chapter~\ref{CHARS} is exactly the efficiency bottleneck
that traditional methods of parsing and disambiguation face when applied to the DOP model.
Usually, the time-complexity and actual processing-times of these methods depend on 
general characteristics of the probabilistic grammar and the input sentence, e.g. size, length
and ambiguity.
The idea behind ambiguity-reduction specialization (ARS) (chapter~\ref{CHARS}) is that
it should be possible to make {\em processing time depend on general properties of the 
distribution of sentences in some domain of language use}, rather than only on 
properties of individual sentences. Therefore, the ARS framework aims at learning 
from a tree-bank that represents a limited domain, a {\em specialized}\/ probabilistic grammar 
that processes more frequent input more efficiently than comparable less frequent input.
For specializing a probabilistic grammar, the ARS framework focuses the specialization effort 
on the observation that language use in a limited domain is much less ambiguous than it is 
in less limited domains.
Specialization by ARS, therefore, aims at finding the least ambiguous grammar (in the Information
Theoretic sense) that represents the limited language use in the domain without resulting 
in extreme overfitting or loss of precision.

The ARS framework can be implemented in a whole range of possible ways, of which only one 
has been tested here: the cheapest one. In our opinion, there are many other implementations that 
can improve radically on this one. Despite of the inferiority of the current 
implementation, it has been possible to exemplify that the ARS framework can specialize 
the DOP model to a domain so that it enables efficiency improvement without loss of 
recognition-power or precision of disambiguation. In fact, this is a remarkable result 
considering how hard it is to improve efficiency without jeopardizing the disambiguation-precision
of a probabilistic model. Moreover, as we can see from the experiments on the OVIS domain
(chapter~\ref{CHARSImpExp}) the specialized DOP models are usually faster on more frequent
input. 

Nevertheless, the experiments of section~\ref{SecATISExps} also show that the current 
implementation of the ARS is not able to realize the theoretical promise of that framework 
on every limited domain of language use. 
The current implementation reduces ambiguity only to a small extent 
and does not always result in DOP models that are efficient enough to be useful in applications.
On the ATIS domain, the current implementation failed to achieve an efficiency improvement
that is comparable to its achievement on the OVIS domain. We think that this is mainly
due to the inferiority of the {\em current implementation}. Future work may be able to show 
how other implementations can improve the efficiency of DOP on domains like the ATIS.
However, another important factor should be kept in mind: it is unclear in how far the 
ATIS and the OVIS tree-banks that we employed for our experiments can be considered 
{\em statistical samples}\/ from their respective domains. This is of course a major issue 
in determining the behavior of the learning algorithms.

In summary, the conclusion of this work has two sides. On the one side, we see that this 
thesis offers the DOP model a much better computational position than before. The efficiency 
improvement enables more serious experimentation with the model on larger tree-banks and the 
study of complexity explains why disambiguation under the DOP model is so hard. 
On the other side, it is fair to say that the efficiency problem of the DOP model is not totally
solved by this thesis. It will remain an interesting subject of research where
substantial improvement can be achieved.

Unfortunately, many people consider the efficiency problem as a kind of 
``gory detail" research subject. This point of view has its roots in modeling techniques 
that aim solely at the correctness of computer systems that model tasks that are conceptually
clear. Indeed, this view is suitable for such computer systems. However, when modeling 
intelligent behavior, this point of view is truly mistaken. Efficiency is often considered 
a {\em hallmark of intelligent behavior}. There are many examples of this fact, 
but one such example is most suitable here: the game of Chess.
The main difference between an expert and a novice who knows the rules 
of the game is mainly the efficiency of the expert that has {\em specialized}\/ himself in 
Chess through experience. Not modeling the efficiency of the Chess expert in any way simply 
means studying the whole space of possibilities, which no intelligent Chess program does.

We envision that future research in natural language processing will address the 
efficiency problem more frequently than currently is the case and we hope that the present study 
exemplifies the range of interesting research directions pertaining to the efficiency of 
performance models. The subject of modeling and exploiting efficiency properties of human 
language processing is both interesting and rewarding. 



\bibliographystyle{apalike}
\bibliography{BIBLIOGRAPHY}
\samenvatting
\begin{quotation}
{\small\sl {\large\sl D}it proefschrift analyseert de computationele eigenschappen van
hedendaagse performance-modellen van menselijke taalverwerking, zoals
Data-Oriented Parsing (DOP)~\cite{Scha90,Scha92,RENSDES}. Het constateert
enkele belangrijke beperkingen en tekortkomingen, en doet voorstellen voor
verbeterde modellen en algorithmes, gebaseerd op technieken uit Explanation-Based
Learning. Experimenten met implementaties van deze algorithmes leveren
bemoedigende resultaten op .}
\end{quotation}
{\Large\cal H}et is algemeen bekend dat formele grammatica's van natuurlijke talen zeer
ambigu zijn. Vaak kennen deze grammatica's zeer veel analyses toe aan een uiting.
Het overgrote deel van deze analyses wordt door een mens echter helemaal niet waargenomen.
Desambigu\"{e}ring, het kiezen van die ene analyse die door een mens als meest
plausibel wordt beschouwd, vormt een van de belangrijkste doelstellingen
van de huidige {\em performance modellen van natuurlijke taal parsering}.
Veel van deze modellen implementeren desambigu\"{e}ring door gebruik te
maken van een {\em probabilistische grammatica}, die bestaat uit regels waaraan
toepassings-waarschijnlijkheden zijn toegekend. Deze waarschijnlijkheden
worden geschat op basis van een geannoteerd corpus (een {\em tree-bank}\/),
dat bestaat uit een grote, representatieve hoeveelheid uitingen die elk voorzien zijn van
een boomstruktuur die de juiste analyse van de uiting representeert.
De toepassings-waarschijnlijkheden van de regels in een dergelijke
probabilistische grammatica maken het mogelijk de verschillende analyses van een 
uiting te rangschikken op waarschijnlijkheid, zodat de analyse met de hoogste kans als de meest
plausibele analyse uitgekozen kan worden.

Het Data-Oriented Parsing (DOP)~\cite{Scha90,Scha92,RENSDES} model
onderscheidt zich van andere performance-modellen doordat de
"probabilistische grammatica" die gebruikt wordt een zeer redundant karakter heeft.
In dit model wordt een tree-bank die iemands taal-ervaring representeert,
in zijn geheel opgeslagen in het geheugen. Vervolgens dient dit geheugen
als databank voor het parseren van nieuwe uitingen door middel van {\em analogie}. 
In de thans bestaande realisaties van dit model, wordt een nieuwe input-zin geanalyseerd 
doordat er nagegaan wordt op welke manieren deze zin gegenereerd had kunnen worden door 
het combineren van ``parti\"{e}le analyses" (brokstukken van de bomen in de tree-bank).
De voorkomens-frequenties van de verschillende brokstukken in de databank
kunnen dan gebruikt worden om de
waarschijnlijkheden van de verschillende mogelijke analyses te berekenen.

Plausibele performance-modellen zijn erg ineffici\"{e}nt, en dat geldt in sterke 
mate voor DOP. Modellen die in zekere mate in staat zijn om input-zinnen suksesvol te
desambigu\"{e}ren op basis van de informatie in een tree-bank, lijken wat
betreft hun {\em  effici\"{e}ntie-eigenschappen} helemaal niet op het
menselijke taalverwerkings-vermogen. Het is evident dat effici\"{e}ntie in
menselijk gedrag in het algemeen en in taalkundig gedrag in het bijzonder, een essentieel
kenmerk is van intelligentie. Bovendien vormen ``echte'' applicaties,
waarin effici\"{e}ntie altijd belangrijk is, het natuurlijke biotoop van
performance-modellen.

Dit proefschrift betreft de computationele complexiteit en de effici\"{e}ntie
van probabilistische desambigu\"{e}rings-modellen in het
algemeen en van het DOP model in het bijzonder. Allereerst presenteren we
in een theoretisch geori\"{e}nteerd hoofdstuk een 
complexiteits analyse van probabilistische desambigu\"{e}ring
binnen het DOP model en soortgelijke modellen.
Deze analyse impliceert dat effici\"{e}nte desambigu\"{e}ring met zulke
modellen niet bereikt zal kunnen worden met behulp van uitsluitend conventionele
optimalisatie technieken.
Daarom wordt in de volgende hoofdstukken een nieuwe aanpak van het
ineffici\"{e}ntie-probleem ontwikkeld. Deze aanpak integreert twee verschillende
optimalisatie-methodes: een conventionele en een niet-conventionele.
De conventionele optimalisatie richt zich op het bereiken van effici\"{e}nte 
{\em deterministisch polynomiale-tijd}\/ desambigu\"{e}rings algorithmes voor DOP.
De niet-conventionele optimalisatie, die centraal staat in het proefschrift,
richt zich op het {\em specialiseren van performance modellen voor domeinen met 
een specifiek taalgebruik}\/ door middel van {\em leren}.
Beide manieren van aanpak worden in dit proefschrift toegepast op het DOP
model, en empirisch getoetst op bestaande, applicatie-gerichte, tree-banks.

De motivaties, methodes, en bijdragen van het proefschrift worden
hieronder met betrekking tot ieder van deze onderwerpen samengevat.

\paragraph{Computationele complexiteit:} De computationele
complexiteits studie gepresenteerd in  hoofdstuk~\ref{CHComplexity}, bevat
bewijzen dat verschillende problemen van probabilistische
desambigu\"{e}ring NP-hard zijn.  Dit betekent
dat ze niet opgelost kunnen worden m.b.v.\ deterministische polynomiale-tijd
algorithmes. Deze desambigu\"{e}rings-problemen worden hier beschouwd voor twee 
soorten grammatica's: het soort grammatica's dat door DOP wordt gebruikt,
genaamd Stochastic Tree-Substitution Grammars (STSG's), en de ``traditionele'' 
Stochastic Context-Free Grammars (SCFG's).
Voor STSG's wordt van de volgende problemen bewezen dat ze NP-hard zijn:
(1)~het berekenen van de meest waarschijnlijke parse (Most Probable Parse -
MPP) van een uiting, (2)~het berekenen van de MPP van een woord-graaf\footnote{Een woord-graaf
wordt als output opgeleverd door een spraakherkenner die gesproken uiting
analyseert. Het is een {\em Stochastic Finite
State Transducer} die de verschillende hypotheses van de spraakherkenner
(en hun rangschikking) effici\"{e}nt representeert.}, en (3)~Het berekenen van de meest
waarschijnlijke zin van een woord-graaf. We bewijzen tevens dat ook voor SCFGs het
berekenen van de meest waarschijnlijke zin van een woord-graaf NP-hard is.

\paragraph{Ge\"{o}ptimaliseerde algorithmes:}
Voorafgaande aan het werk dat in dit proefschrift wordt gepresenteerd
bestonden er slechts ineffici\"{e}nte non-deterministische {\em exponenti\"{e}le}\/
tijdscomplexiteit algorithmes voor het desambigu\"{e}ren onder DOP~\cite{RENSDES}. 
Deze situatie heeft vaak geresulteerd in onbetrouwbare en tijdrovende empirische 
experimenten. In dit proefschrift worden de eerste effici\"{e}nte {\em deterministisch
polynomiale-tijd}\/ algorithmes voor desambigu\"{e}ren onder het DOP model beschreven
(hoofdstuk~\ref{CHOptAlg4DOP}).
Deze algorithmes richten zich op het berekenen van de meest waarschijnlijke derivatie (Most
Probable Derivation -~MPD). Een belangrijke bijdrage aan de effici\"{e}ntie van
desambigu\"{e}ring onder DOP wordt geleverd door het beperken van de invloed van de meest
vertragende factor: de {\em grootte}\/ van een DOP STSG. Dit wordt bereikt door twee methodes
te combineren: (1)~een conventionele optimalisatie van de algorithmes,
zodat deze algorithmes een lineaire tijdscomplexiteit in de STSG grootte
hebben, en (2)~verschillende heuristieken die een DOP STSG reduceren tot een kleinere
doch meer accurate grammatica. Samen resulteren deze twee optimalisaties in
een versnelling van twee ordes van grootte, vergeleken met de algorithmes die gebruikt
werden voorafgaande aan dit werk. Bovendien, omdat de grootte van een DOP STSG kleiner 
is geworden, is het effect van het ``sparse-data'' probleem veel kleiner geworden dan 
oorspronkelijk het geval was.

\paragraph{Specialisatie door middel van ambigu\"{i}teits-reductie:}
Centraal in dit proefschrift staat een niet-conventionele optimalisatie
methode die performance modellen specialiseert voor specifieke
domeinen van taalgebruik (hoofdstuk~\ref{CHARS}).
In veel taalverwerkings toepassingen is het taalgebruik op een of andere manier beperkt.
Deze beperkingen worden bepaald door het systeem-ontwerp (bijvoorbeeld beperkte vrijheid 
in dialogen) en/of door de keuze van het domein van de applicatie, bijvoorbeeld openbaar 
vervoer informatie, ticket reserverings systemen en computer handleidingen.
Een interessante eigenschap van\linebreak   menselijk taalbegruik in specifieke domeinen is dat het
minder breed  en minder ambigu is dan het taalgebruik dat verondersteld wordt door
linguistische { \em Broad-Coverage Grammatica's} (BCGs).
Deze eigenschap van menselijk taalgebruik heeft
betrekking op {\em hele domeinen},  meer dan op {\em individuele}\/ uitingen.
Zulke eigenschappen kunnen worden gemeten als statistische {\em biases}\/ in
samples van geanalyseerde uitingen uit het domein. Wij menen de
ineffici\"{e}ntie van de huidige performance-modellen grotendeels te kunnen
verklaren uit het feit dat ze geen rekening houden met zulke
statistische biases in beperkte domeinen. Deze modellen maken gebruik van 
{\em tree-banks} die geannoteerd zijn onder linguistische BCG's, die juist gericht 
zijn op {\em niet}-beperkt taalgebruik. De desambigu\"{e}rings-algorithmes die door de 
huidige performance-modellen worden gebruikt, hebben daardoor een feitelijk tijdsverbruik dat
onafhankelijk is van de eigenschappen van het domein.  Het tijdsverbruik van deze algorithmes 
is alleen afhankelijk van de eigenschappen van individuele zinnen (b.v.\ zinslengte), en van 
de BCG (b.v.\ de ambiguiteit van de BCG). In dit proefschrift wordt een direkt verband gelegd
tussen deze situatie en het ontbreken, in de huidige performance modellen,
van een aantrekkelijke eigenschap van menselijke taalverwerking: {\em frequente en minder 
ambigu\"{e} uitingen worden door een mens effici\"{e}nter geanalyseerd}. Volgens dit
proefschrift kan deze eigenschap verkregen worden door het interpreteren van de statistische 
biases in beperkte domeinen binnen een {\em Informatie-Theoretisch}\/ raamwerk, dat
performance-modellen {\em specialiseert voor beperkte domeinen}.

Het proefschrift presenteert een raamwerk dat deze idee\"{e}n
implementeert, genaamd het  ``Ambiguity-Reduction Specialization (ARS) framework''.
Het ARS framework incorporeert de bovengenoemde effici\"{e}ntie eigenschappen in
performance modellen, door middel van een ``off-line'' leeralgorithme dat
gebruik maakt van een tree-bank. Het doel van dit leeralgorithme is het beperken van zowel de
herkennings-kracht als de ambigu\"{\i}teit van de linguistische BCG die voor de annotatie 
van de tree-bank werd gebruikt, zodat er gespecialiseerd wordt voor het domein.
Dit resulteert in een {\em gespecialiseerde grammatica}, en in een {\em
gespecialiseerde tree-bank}\/ geannoteerd onder deze grammatica.  Deze nieuwe tree-bank kan
dienen voor het verkrijgen van een kleinere en minder ambigu\"{e} probabilistische grammatica 
onder een bepaald performance-model. In het ARS framework wordt (voor het eerst) deze 
specialisatie-taak uitgedrukt in termen van {\em beperkte optimalisatie}. De algorithmes voor
de uitvoering van deze taak kunnen daardoor geformuleerd worden als
leeralgorithmes die gebaseerd zijn op beperkte optimalisatie. Er worden
twee verschillende specialisatie-algorithmes gepresenteerd. Het principi\"{e}lere algorithme 
is gebaseerd op de noties van {\em entropie}\/ en Shannon's {\em optimale codelengte}, het 
practischere algorithme is gebaseerd op intu\"{i}tive statistische maten. Tevens presenteert 
dit proefschrift een nieuw parseer-algorithme dat de gespecialiseerde grammatica en de 
oorspronkelijke BCG integreert op een complementaire manier, zodat de parser geen tijdverlies 
lijdt wanneer de gespecialiseerde grammatica faalt in het herkennen van de input.

\paragraph{Empirisch onderzoek:}
De boven genoemde leer- en parseeralgorithmes zijn\linebreak
ge\"{i}mplementeerd in computer programma's, en worden gebruikt in een
project van de Nederlanse organisatie voor Wetenschappelijke Onderzoek (NWO). Het proefschrift 
rapporteert (hoofdstuk~\ref{CHARSImpExp}) uitgebreide empirische experimenten die de boven 
besproken theoretische idee\"{e}n testen op twee tree-banks, 
OpenbaarVervoer Informatie Systeem (OVIS) en Air Travel Inquiry System (ATIS).
Deze tree-banks representeren twee domeinen, twee talen en twee
desambigueertaken: het desambigueren van uitingen en het desambigueren van
woord-grafen in een dialoogsysteem. In deze experimenten wordt het meer practische,
maar minder optimale leeralgorithme, toegepast op het specialiseren van het DOP model voor
gelimiteerde domeinen. De experimenten laten zien dat in beide domeinen de resulterende
gespecialiseerde DOP STSG's (genaamd SDOP STSGs) substantieel kleiner zijn dan de oorspronkelijke 
DOP STSG's. Bovendien, in \'{e}\'{e}n van de domeinen (OVIS) zijn, op beide desambigueertaken, 
de SDOP STSG's niet alleen minstens zo accuraat als de oorspronkelijke DOP STSG's, maar ook
veel effici\"{e}nter. In het andere domein (ATIS) zijn de SDOP STSG's ook effici\"{e}nter dan de
oorspronkelijke DOP STSG's, maar deze effici\"{e}ntie verbetering wordt bereikt slechts voor
DOP modellen die onbruikbaar zijn in de praktijk.

Tevens wordt de hypothese getoetst dat de gepresenteerde specialisatie-methode resulteert 
in effici\"{e}ntere parsering van frequente en minder ambigu\"{e} uitingen.
Ondanks het feit dat dit wordt getest in een sub-optimaal experiment op het
OVIS domein blijkt dat deze hypothese ondersteund wordt door de emipirische resultaten.
De parseertijd van de SDOP STSGs is kleiner voor frequente invoer, dit in tegenstelling tot de
parseertijd van DOP STSGs, die duidelijk onafhankelijk is van de frequentie van de invoer.\\

De conclusie heeft betrekking op beide onderzoeksonderwerpen die aan elkaar
worden gerelateerd in dit proefschrift: enerzijds de computationele en
effici\"{e}ntie-aspecten van het DOP model, en anderzijds het specialiseren van 
performance-modellen voor beperkte domeinen. De studie naar de computationele aspecten van 
het DOP model levert een complexiteits-analyse en een effici\"{e}nt algorithme op.
De empirische resultaten laten duidelijk zien dat het nieuwe algorithme een
aanzienlijke effici\"{e}ntie-verbetering oplevert. Deze resultaten maken
echter ook duidelijk dat de computationele aspecten en de  effici\"{e}ntie van het DOP -model
verdere onderzoek vereisen. De studie naar het specialiseren van performance modellen
voor gelimiteerde domeinen heeft nieuwe inzichten omtrent het modelleren
van effici\"{e}ntie-eigenschappen van menselijk taalverwerking opgeleverd.
Onze hypothese betreffende de relatie tussen statistische biases en deze eigenschappen blijkt 
ondersteund te worden door de empirische resultaten. Het zou echter voorbarig zijn te 
concluderen dat de gepresenteerde methode sucsesvol toepasbaar is op elk beperkt domein. 
De studie in dit proefschrift is immers beperkt gebleven tot sub-optimale implementaties die 
verschillende approximaties bevatten, als gevolg van beperkingen in de totnutoe beschikbare 
hardware. Het is daarom noodzakelijk om deze studie voort te zetten in toekomstig onderzoek.

\curriculum

The author was born in Haifa on the 12th of September 1964.
In~1984 he started his studies in Computer Science at the Technion (Israel
Institute of Technology) and obtained the~B.A.\ degree in~1988.
During~1988 and~1989 he worked both as a teacher at high school and 
as a software engineer in Haifa. 
In~1989 he moved to Amsterdam and followed a course in Dutch language.
Between 1990~and~1992 he studied Informatics (formal specification languages 
and programming science) at the University of Amsterdam (UvA) and obtained 
the M.A.\ (``doctoraal") degree cum laude. 
During the year~1993 he worked as a research-assistant (formal specification 
languages for real-time systems) at Delft University of Technology (TUD). 
From~1994 to~1996 he worked as a researcher (robust parsing) for the
Foundation for Language and Speech (STT) at Utrecht University. 
During~1996 he worked as a researcher (learning efficient parsing) in 
the Priority Programme Language and Speech Technology (TST) of 
the Netherlands Organization for Scientific Research (NWO). And in mid~1997 he 
received a joint grant from NWO~(TST) and~STT for writing the present dissertation.


\pagestyle{empty}

\noindent
{\em Titles in the ILLC Dissertation Series:}

\newcommand{\illcpublication}[3]{\item[ILLC #1: ]{\bf #2}\\{\em #3}}

\begin{list}{}{ \settowidth{\leftmargin}{ILL}
		\setlength{\rightmargin}{0in}
		\setlength{\labelwidth}{\leftmargin}
		\setlength{\labelsep}{0in}
}

\illcpublication{DS-1993-01}{Paul Dekker}{Transsentential Meditations; Ups and downs in dynamic semantics}
\illcpublication{DS-1993-02}{Harry Buhrman}{Resource Bounded Reductions}
\illcpublication{DS-1993-03}{Rineke Verbrugge}{Efficient Metamathematics}
\illcpublication{DS-1993-04}{Maarten de Rijke}{Extending Modal Logic}
\illcpublication{DS-1993-05}{Herman Hendriks}{Studied Flexibility}
\illcpublication{DS-1993-06}{John Tromp}{Aspects of Algorithms and Complexity}
\illcpublication{DS-1994-01}{Harold Schellinx}{The Noble Art of Linear Decorating}
\illcpublication{DS-1994-02}{Jan Willem Cornelis Koorn}{Generating Uniform User-Interfaces for Interactive Programming Environments}
\illcpublication{DS-1994-03}{Nicoline Johanna Drost}{Process Theory and Equation Solving}
\illcpublication{DS-1994-04}{Jan Jaspars}{Calculi for Constructive Communication, a Study of the Dynamics of Partial States}
\illcpublication{DS-1994-05}{Arie van Deursen}{Executable Language Definitions, Case Studies and Origin Tracking Techniques}
\illcpublication{DS-1994-06}{Domenico Zambella}{Chapters on Bounded Arithmetic \& on Provability Logic}
\illcpublication{DS-1994-07}{V. Yu. Shavrukov}{Adventures in Diagonalizable Algebras}
\illcpublication{DS-1994-08}{Makoto Kanazawa}{Learnable Classes of Categorial Grammars}
\illcpublication{DS-1994-09}{Wan Fokkink}{Clocks, Trees and Stars in Process Theory}
\illcpublication{DS-1994-10}{Zhisheng Huang}{Logics for Agents with Bounded Rationality}
\illcpublication{DS-1995-01}{Jacob Brunekreef}{On Modular Algebraic Protocol Specification}
\illcpublication{DS-1995-02}{Andreja Prijatelj}{Investigating Bounded Contraction}
\illcpublication{DS-1995-03}{Maarten Marx}{Algebraic Relativization and Arrow Logic}
\illcpublication{DS-1995-04}{Dejuan Wang}{Study on the Formal Semantics of Pictures}
\illcpublication{DS-1995-05}{Frank Tip}{Generation of Program Analysis Tools}
\illcpublication{DS-1995-06}{Jos van Wamel}{Verification Techniques for Elementary Data Types and  Retransmission Protocols}
\illcpublication{DS-1995-07}{Sandro Etalle}{Transformation and Analysis of (Constraint) Logic Programs}
\illcpublication{DS-1995-08}{Natasha Kurtonina}{Frames and Labels. A Modal Analysis of Categorial Inference}
\illcpublication{DS-1995-09}{G.J. Veltink}{Tools for PSF}
\illcpublication{DS-1995-10}{Giovanna Cepparello}{Studies in Dynamic Logic}
\illcpublication{DS-1995-11}{W.P.M. Meyer Viol}{Instantial Logic. An Investigation into Reasoning with Instances}
\illcpublication{DS-1995-12}{Szabolcs Mikul\'as}{Taming Logics}
\illcpublication{DS-1995-13}{Marianne Kalsbeek}{Meta-Logics for Logic Programming}
\illcpublication{DS-1995-14}{Rens Bod}{Enriching Linguistics with Statistics: Performance Models of Natural Language}
\illcpublication{DS-1995-15}{Marten Trautwein}{Computational Pitfalls in Tractable  Grammatical Formalisms}
\illcpublication{DS-1995-16}{Sophie Fischer}{The Solution Sets of Local Search Problems}
\illcpublication{DS-1995-17}{Michiel Leezenberg}{Contexts of Metaphor}
\illcpublication{DS-1995-18}{Willem Groeneveld}{Logical Investigations into Dynamic Semantics}
\illcpublication{DS-1995-19}{Erik Aarts}{Investigations in Logic, Language and Computation}
\illcpublication{DS-1995-20}{Natasha Alechina}{Modal Quantifiers}
\illcpublication{DS-1996-01}{Lex Hendriks}{Computations in Propositional Logic}
\illcpublication{DS-1996-02}{Angelo Montanari}{Metric and Layered Temporal Logic for Time Granularity}
\illcpublication{DS-1996-03}{Martin H. van den Berg}{Some Aspects of the Internal Structure of 
Discourse: the Dynamics of \linebreak Nominal Anaphora}
\illcpublication{DS-1996-04}{Jeroen Bruggeman}{Formalizing Organizational Ecology}
\illcpublication{DS-1997-01}{Ronald Cramer}{Modular Design of Secure yet Practical Cryptographic Protocols}
\illcpublication{DS-1997-02}{Nata\u{s}a Raki\'{c}}{Common Sense Time and Special Relativity}
\illcpublication{DS-1997-03}{Arthur Nieuwendijk}{On Logic. Inquiries into the Justification of Deduction}
\illcpublication{DS-1997-04}{Atocha Aliseda-LLera}{Seeking Explanations: Abduction in Logic, Philosophy of Science and Artificial Intelligence}
\illcpublication{DS-1997-05}{Harry Stein}{The Fiber and the Fabric: An Inquiry into Wittgenstein's Views on Rule-Following and Linguistic Normativity}
\illcpublication{DS-1997-06}{Leonie Bosveld - de Smet}{On Mass and Plural Quantification. The Case of French `des'/`du'-NP's.}
\illcpublication{DS-1998-01}{Sebastiaan A. Terwijn}{Computability and Measure}
\illcpublication{DS-1998-02}{Sjoerd D. Zwart}{Approach to the Truth: Verisimilitude and Truthlikeness}
\illcpublication{DS-1998-03}{Peter Grunwald}{The Minimum Description Length Principle and Reasoning under Uncertainty}
\illcpublication{DS-1998-04}{Giovanna d'Agostino}{Modal Logic and Non-Well-Founded Set Theory: Translation, Bisimulation, Interpolation}
\illcpublication{DS-1998-05}{Mehdi Dastani}{Languages of Perception}
\illcpublication{DS-1999-01}{Jelle Gerbrandy}{Bisimultations on Planet Kripke}
\illcpublication{DS-1999-02}{Khalil Sim\'{a}an}{Learning Efficient Disambiguation}

\end{list}

\end{document}